%% file: main.tex
\documentclass[11pt]{article}
\usepackage[margin=1in]{geometry}
\usepackage[T1]{fontenc}
\usepackage{lmodern}
\usepackage{graphicx}
\graphicspath{{./}}
\usepackage{pdflscape}
\newcounter{algorithm}
\usepackage{float}
\usepackage{amsmath,amssymb}
\usepackage{array}
\usepackage{booktabs}
\usepackage[numbers,sort&compress]{natbib}
\usepackage{xcolor}
\usepackage[colorlinks=true,allcolors=blue!55!black]{hyperref}
\usepackage{caption}
\makeatletter
\newcommand{\secbarrier}{\par\vspace{0pt}%
  \ifx\@deferlist\@empty\else\clearpage\fi}
\makeatother
\title{Through the Looking Glass:\\Directly Reading and Writing Transformers}
\author{Mark Oskin\\
Professor\\
School of Computer Science and Engineering\\
University of Washington\\
\texttt{mhoskin@uw.edu}}
\date{September 2026}

\begin{document}
\maketitle

\begin{abstract}
How many of a transformer's components decide a token? Counted the standard way --- feed-forward
units and attention channels, scored by the absolute value of what each contributes to the logit ---
one prediction rests on three thousand to two hundred thousand of them. But contributions are
signed, and across the eighteen models here the mass pushing away from the predicted token is a
median of seven times the mass carrying it. Divide by the net instead and the same attribution on
the same predictions returns dozens. On the model this paper is built on, fifty-three components
carry ninety percent of a prediction, thirteen it cannot survive losing, and eight suffice to produce
it with everything else at that position zeroed. Across twelve models trained by other people, $124$ million to seven billion parameters, the
sufficient set runs from two components to sixteen, and what a prediction draws on, followed all
the way back, is one to three percent of the model, a share that does not grow with size. The rest of the model is neither idle nor holding work for later tokens:
three quarters of what a layer adds to the residual is a fixed linear map of the state it received.

Everything here is read from the model's own parameters and activations, with nothing trained and
nothing fitted, and it names a component on both sides. What a component writes is named from the
predictions it drives, reaching close to half of every model, and the name is a share of the drive
that predicts what deleting the component costs. What it reads is decoded from its weights in the
frame of its own layer, at $58.9$ percent above chance over its eight strongest inputs and below
chance once enough of the list is asked for; sorting the remainder by which upstream component
supplies it separates it into grammatical categories the embedding cannot see.

A name can be acted on. An association the model does not hold installs into one spare unit, key and
value read from the weights, moving the target from rank $578$ to rank $1$ for a quarter of one
percent of held-out loss --- a fortieth of what a rank-one update costs to reach the top ten. Where
an edit lands is set by the architecture: a write before the second layer keeps a tenth of its
direction at the readout, one in the last third up to eight tenths. An attention head installed to mark a token and a unit installed two layers above it to read the
mark make an edit fire only where that token occurred earlier in the context, and a unit the model trained for
itself is driven from two layers upstream with $86$ percent of the effect passing through it.
Replacing the activation with an order-preserving one makes a unit's inputs readable at the
instrument's ceiling, flat across the list rather than decaying, at the price of turning an install
into a two-part edit.
\end{abstract}

\input{intro}
\input{lens}
\input{naming}
\input{editing}
\input{activation}
\input{othermodels}
\input{related}
\input{discussion}
\input{conclusion}

\appendix
\input{gallery}

\input{setmodel}
\input{pitfalls}
\input{appgraph}

\bibliographystyle{plainnat}
\bibliography{refs}

\end{document}

%% file: intro.tex
\section{Introduction}
\label{sec:intro}

A transformer produces a token. How, specifically, did it do that? Are there identifiable parts of
the model that matter to that particular prediction, and can they be traced --- a ``circuit'' used
\emph{for that specific token}? They can, and the circuit is small. A handful up to a few dozen
components do the work behind any one prediction, and most of what a transformer computes at a
position is a side effect of the architecture: moving data between layers and keeping its own
geometry organized.

That is not what the literature would lead one to expect. In the basis a transformer actually has,
individual feed-forward units and attention channels, published counts run to thousands or tens of
thousands of components for a single behavior \citep{marks2024sparsefeature}, and the accounting in
common use reproduces that here: measured the standard way, one prediction in the models below rests
on three thousand to two hundred thousand components. A number that size is not an explanation of
anything.

It is also wrong. Contributions to a logit are signed, so they cancel, and the mass pushing away from
the prediction is a median of seven times the net across the models here and twenty times it on the
model this paper is built on. The standard count divides by the absolute total, and so counts that
opposing mass as part of the answer. Divide by the net instead and the same attribution on the same
predictions returns dozens. The opposing mass is not a rival circuit the net hides. A model writes its
prediction into the token frame directly rather than assembling it elsewhere and turning it on, and
it does not use that frame as scratch space \citep{oskin2026offaxis}; so the net contribution is the
model writing the output, and the rest is the residual stream being carried between layers and kept
organized so that many things can share it \citep{elhage2022superposition}. Our lens separates that
management from the prediction, and what remains is a small legible circuit.

\label{sec:whatitmeans}
This paper concerns the legibility and editability of transformers. What exactly does that mean? The
intuitive versions are easy enough to state.

\begin{itemize}
\item \textbf{A component is legible} if what it does can be stated. Two things have to be readable:
what drives it, and what it produces. A name for either has to meet two conditions. It must be
\emph{coherent}: the tokens that drive the component share something, and so do the tokens it
promotes. And it must be \emph{verified} against the model's behavior rather than against whether the
list looks sensible.
What a component is credited with writing depends about equally on its weights and on when it fires,
so a column read on its own narrows the answer to a seventh of the vocabulary and no further
(Appendix~\ref{sec:geometry-tried}). What does supply them is behavior: the predictions a component
drives name what it writes, and the positions where it fires name what it reads. Where no token
covers the answer, a component can still be described by the circuit it fits within.

\item \textbf{A model is legible} if the connections between its components can be followed. When it
generates a token there must be a trace of why it produced \emph{that} one, and the trace must
explain the choice to someone reading it. The path must be traceable end to end and land in the token
embedding space where it meets the input and the output, even when what it passes through in between
is conceptual. Ideally a trace is small enough for a person to read; it must in any case be complete enough to
include what actually decided the token.

\item \textbf{A component is editable} if its behavior can be changed, the model respects the change,
and everything unrelated is left alone. Install an intended behavior into one named component,
measure how strongly the model takes it, and measure how much else moved. Neither number means
anything alone, since any edit is respected if pushed hard enough and any edit is harmless if pushed
gently. Editability is therefore a trade-off rather than a threshold.
\end{itemize}

Our aim was a lens derived from the transformer's own parameters, with no training or fitting of its
own. Good fitted explainers exist: a sparse autoencoder trained on a model's activations recovers
features a person can name, and we reproduce that result in Appendix~\ref{sec:baseline-pitfalls}. The
reason we did not build on one is that a fitted explainer is a second model, and understanding it
becomes a prerequisite to understanding the transformer it was meant to explain. Simple is good if it
works.

On the model this paper is built on, the lens returns three numbers answering three questions about
one prediction. Fifty-three components carry ninety percent of it. Thirteen it cannot survive losing:
removing them changes the answer, where removing the same number at random leaves it alone. And eight
suffice to produce it --- keep those eight at a position, zero every other feed-forward unit and
attention channel there, let the survivors recompute, and the model emits the same token. Eight
random components in their place never do, on any of $115$ predictions we tested.

The rest of the model is neither idle nor holding work for later tokens. Keep only the circuit at a
position and the cost at the token being predicted is a quarter of what an equally sized random set
costs, and on each of the next five hundred and twelve tokens the two cost the same. What the
remainder is doing is managing the workspace: three quarters of what a layer adds to the residual is
a fixed linear map of the state it received, applied whatever the model is about to say, and that
work is spread across every component rather than concentrated in a few. Cancellation is far more
severe arriving at a component than leaving the model --- about a dozen upstream components reach
ninety percent of a unit's incoming drive counted signed and many thousands counted absolute, a
factor of roughly a thousand against about twenty at the readout.

Most of this paper's effort goes into the gap between finding a component and reading it. A
component is named on both sides. What it writes is named from the predictions it drives, which
reaches close to half of every model, and the name is a share of the drive that predicts what
deleting the component costs. What it reads is named from its own weights, decoded in the frame of
its own layer rather than through the vocabulary: asked for a unit's eight strongest drivers, the
parameters recover the true set at $58.9$ percent, thirty-seven times chance, and that agreement is
confined to the head of the list. Somewhat over a quarter of what drives a unit falls on tokens the
embedding groups into a category; the remaining three quarters have no name in the vocabulary at
all, being concepts the model builds for itself, and sorting them by which upstream component
supplies them separates them into grammatical categories the embedding cannot see. The same reading
extends to attention, where a channel's value projection is evaluated at the position it attends to,
and the drop between what a layer's attention reads and what its feed-forward block reads is the
context that layer delivered.

Naming a component is one half of the problem; acting on it is the other. An association the model
did not hold installs into a single spare component, with the key and the value both derived from
the weights without fitting, moving the target from rank $578$ to rank $1$ for a quarter of one
percent of held-out loss --- a fortieth of what the rank-one update the editing literature
established costs to reach the top ten. Where such a write can be made turns out to be a property
of the architecture rather than of the component chosen. The bottom two layers erase the input token
and stash it off the token axis, so a write made there loses most of its direction by the readout.
The top layer answers in the token frame, which is the frame the readout reads and the only one a
write acts on directly, so the last hop of a circuit belongs in that frame and the earlier hops do
not --- though not in that layer, which consumes readout-frame content rather than carrying it,
and is the one place preservation falls instead of rising.
Between the two, the residual carries a write in the coordinates it was written in, and the baseline
uses about a quarter of its width, the same quarter at every depth, so a direction drawn at random
carries a signal from one installed component to another with nothing else in the model oriented to
read it.
That is enough to build a circuit across two layers from components that did not exist before, and
to drive a unit the model trained for itself from two layers upstream, with $86$ percent of the
effect passing through it.

Two further sections take the lens beyond the model it was built on.
Section~\ref{sec:mono} replaces the activation function with an order-preserving one, sigmoid or
softplus, which makes a unit's inputs readable at the
instrument's ceiling, flat across the whole list where the conventional unit decays below chance;
the price is that such a unit is never silent, so an association installed into it alone arrives
everywhere at once, and the install becomes a two-part edit. ReLU buys legibility with accuracy. Section~\ref{sec:stock} applies the lens unchanged to twelve models trained by other people,
from $124$ million to $7$ billion parameters across seven architecture families. There the
sufficient set runs from two components to sixteen with the largest models at its small end, and on gated feed-forward models nearly half the
components carrying a prediction fire negative, so a scope read without that sign lands on the wrong
tokens.

Section~\ref{sec:circuit} builds the lens and applies it to the baseline: the accounting, what is
necessary, what is sufficient, and what the rest of the model is doing. Section~\ref{sec:naming}
names components, on the write side from behavior and on the read side from weights, and follows the
part the vocabulary cannot name back to its sources. Section~\ref{sec:edit} acts on the names: a
gain, an install, where an edit can be placed, a circuit across two layers, a tap into a trained
unit, and a comparison against a rank-one editor. Sections~\ref{sec:mono}
and~\ref{sec:stock} are the two above. Section~\ref{sec:related} places the work,
Section~\ref{sec:discussion} takes up what the measurements ask of the architecture, and
Section~\ref{sec:conclusion} closes. Appendix~\ref{sec:gallery} is a gallery of traces drawn
mechanically rather than chosen, two per model, where a reader can see what a trace actually looks
like; Appendix~\ref{sec:setmodel} is the set-operator model of Section~\ref{sec:mono} and how it is
trained; Appendix~\ref{sec:baseline-pitfalls} is the measurements that failed, the conventions that flatter and are therefore not used here, and the ways each number here could have been made to look better by choosing a different null. Appendix~\ref{sec:appgraph} is the dependency graph: the edge
relation between components, the closure it builds, and why that closure is a necessity object
rather than a sufficient one.

%% file: lens.tex
\section{Reading a prediction: The Looking Glass}
\label{sec:circuit}

This section builds the instrument necessary to read a transformer. We begin with the baseline model
used for every measurement here, and with the further models used later in the paper. We then
introduce the lens, and show how it identifies the parts of a transformer doing real work for a
prediction. Next we ask which parts are \emph{necessary} for a correct prediction and which are
\emph{sufficient} to produce it, and then what the rest of the model is doing while that small set
carries the answer. We close with the limitations of the lens. 

\subsection{Models}
\label{sec:circuit-model}

The baseline is a model trained here on the GPT-2 small
architecture~\citep{radford2019language}: twelve layers of width $d=768$, twelve attention heads,
learned positional embeddings, LayerNorm, tied input and output embeddings, and no bias term
anywhere. It is trained at context length $T=2048$ rather than the released model's $1024$, and the
released GPT-2 appears separately among the models of Section~\ref{sec:stock}. Writing $x^{(l)}$ for the residual stream entering layer $l$,
\begin{equation}
  x^{(l+1)} \;=\; x^{(l)} \;+\; \mathrm{Attn}^{(l)}\!\big(\mathrm{LN}(x^{(l)})\big)
             \;+\; \mathrm{FFN}^{(l)}\!\big(\mathrm{LN}(h^{(l)})\big),
\end{equation}
with $h^{(l)}$ the stream after the attention write. The feed-forward block is the object this paper
is about:
\begin{equation}
    \mathrm{FFN}^{(l)}(z) \;=\; W_{\text{out}}^{(l)}\,\phi\big(W_{\text{in}}^{(l)} z\big),
  \qquad W_{\text{in}}^{(l)} \in \mathbb{R}^{m \times d},\; m = 4d = 3072 .
  \label{eq:ffn}
\end{equation}
The block decomposes into $m$ independent \emph{units}. Unit $u$ owns one row of the up-projection
and one column of the down-projection,
\begin{equation}
  w_u = W_{\text{in}}^{(l)}[u,:] \in \mathbb{R}^{d},
  \qquad
  c_u = W_{\text{out}}^{(l)}[:,u] \in \mathbb{R}^{d},
  \label{eq:unit}
\end{equation}
and contributes $a_u(z)\,c_u$ to the block output, where its activation is the scalar
\begin{equation}
    a_u(z) \;=\; \phi\big(w_u \cdot z\big).
  \label{eq:act}
\end{equation}
The read row $w_u$ decides which inputs drive the unit, and the write column $c_u$ decides what it
contributes once driven. Neither projection carries a bias, which is worth noting here because
Section~\ref{sec:edit} has to build one out of the layer's own geometry when an installed unit needs
an off state. A component in what follows is one feed-forward unit or one (head, channel) pair of
attention, in both cases a scalar multiplying a write column, so the two are counted on the same
footing. For the baseline,
\begin{equation}
  \phi(\zeta) \;=\; \mathrm{GELU}(\zeta) \;=\; \zeta\,\Phi(\zeta),
  \label{eq:gelu}
\end{equation}
with $\Phi$ the standard normal CDF. Two properties of Equation~\ref{eq:gelu} are recorded here
because they matter later: GELU is unbounded above, and it is not monotone, having a minimum near
$\zeta = -0.7517$ and rising on both sides of it.

\subsubsection{Training and seeds}
\label{sec:circuit-seeds}

Training is one epoch of OpenWebText at batch size 16, peak learning rate $3\times10^{-4}$ with 2000
warmup steps and cosine decay to a tenth of peak, for 272{,}687 steps. Nothing about the recipe is
unusual.

Eight seeds were trained, differing only in initialization and data order. Five converged and three
did not, by a margin wide enough that nothing turns on where the line is drawn: the failures sit at
perplexity 21.6, 24.4 and 39.5 against a converged band of 19.0 to 19.7.
Table~\ref{tab:baseline-seeds} shows all eight rather than the five, because broken models can be
deceptively legible, as Appendix~\ref{sec:baseline-pitfalls} shows. The seeds are therefore separated on quality
alone, fixed before any legibility number is computed.

\begin{table}[htbp]
\centering
\small
\caption{The eight baseline seeds. Only initialization and data order differ. The excluded seeds are shown rather than dropped because
Appendix~\ref{sec:baseline-pitfalls} needs them. Two columns are defined properly in Table~\ref{tab:contract} and appear here only to show
that the excluded seeds differ on more than perplexity: $\mathrm{ov}@8$ is how well a unit's parameters name the inputs that drive it, and
$\mathrm{PR}$ is the participation ratio of the early layers, a measure of how much representational
variety the model retained. A collapsed model reads as more legible, which is why the two are always
quoted together. $\mathrm{ov}@8$ here is taken under the ranked summary of Appendix~\ref{sec:pit-convention}, raw rather than chance-corrected.}
\label{tab:baseline-seeds}
\begin{tabular}{@{}llrrrr@{}}
\toprule
seed & status & val.\ ppl & LAMBADA & $\mathrm{ov}@8$ & $\mathrm{PR}$ (L0--4) \\
\midrule
1 & converged & 19.00 & 0.267 & 34.6 & 103.0 \\
2 & converged & 19.02 & 0.248 & 35.4 & 102.6 \\
3 & converged & 19.09 & 0.274 & 33.6 & 106.7 \\
4 & converged & 19.10 & 0.272 & 35.5 & 105.5 \\
5 & converged & 19.65 & 0.258 & 38.4 &  98.4 \\
\midrule
6 & excluded  & 21.62 & 0.209 & 48.7 &  45.7 \\
7 & excluded  & 24.39 & 0.214 & 40.7 &  41.3 \\
8 & excluded  & 39.51 & 0.152 & 49.2 &  30.3 \\
\bottomrule
\end{tabular}
\end{table}

We use LAMBADA~\citep{paperno2016lambada} to assess model quality in addition to perplexity. The five
converged seeds are of similar quality, spanning $0.248$ to $0.274$, a spread we read as seed and
evaluation noise; later comparisons are held against that range, so two models whose ranges overlap
are called equal.

\subsubsection{Additional models used}
\label{sec:circuit-models}

Two further groups of models appear from Section~\ref{sec:naming} onward.
The first is five more models trained here, listed in Table~\ref{tab:models}; each changes only
what the feed-forward unit does with its pre-activation, so any difference between them is
attributable to that change.

\begin{table}[htbp]
\centering\small
\caption{The models trained for this paper. The five others differ from the baseline only in the feed-forward unit, sharing its depth, width, head count, parameter budget, data and schedule. The case for choosing
an activation function on these grounds is Section~\ref{sec:mono}, where the set-operator row is taken up as well and Appendix~\ref{sec:setmodel} defines it; this table is a reference for the rows that appear in the tables that follow.}
\label{tab:models}
\input{T_models}
\end{table}

The second is twelve models trained by other people, none of them modified for this work: GPT-2,
OPT, SmolLM2, three of Qwen2.5, Gemma-3, OLMo-2, two of Llama-3.2, TinyLlama
and Mistral, spanning $124$ million to $7$ billion parameters across seven architecture families and
three activation functions. Section~\ref{sec:stock} is where they are the subject, and
Table~\ref{tab:stock} lists them with their sizes and activations.

One family is deliberately absent from that set: the GPT-NeoX line, of which the Pythia suite is
the part we tested. Those models compute their two sublayers in parallel from the same input,
$x + \mathrm{Attn}(\mathrm{LN}_1(x)) + \mathrm{FFN}(\mathrm{LN}_2(x))$, rather than letting the
feed-forward block read what attention has just written, and every reading in this paper assumes the
sequential arrangement. Extending the work to that design is left for later.

Results in Sections~\ref{sec:circuit} through~\ref{sec:edit} are measured on the baseline. The exceptions are noted where another model behaves differently.

\subsection{The lens}
\label{sec:circuit-lens}

The lens is built from three pieces: an accounting of the mass each component writes toward the
prediction, a decision about which of that mass counts as explaining it
(Section~\ref{sec:circuit-accounting}), and a frame in which
anything read below the readout is scored.

Write the logit
of token $t$ as an exact sum over components. Each component $u$ contributes its
activation $a_u$ times the projection of its write column $c_u$ (Equation~\ref{eq:unit}) onto the
readout direction,
\begin{equation}
  \kappa_u \;=\; a_u \,\big\langle c_u,\; \gamma \odot (U_t - \bar{U}) / \rho \big\rangle,
  \qquad
  \sum_u \kappa_u \;=\; N,
  \label{eq:contrib}
\end{equation}
with $\gamma$ the final-norm gain and $\rho$ the norm of the final residual state. The sum runs over
feed-forward units and attention channels, so $N$ is the net those components deliver. The token and
positional embeddings reach the readout along the residual path and are not among the terms; their
share is the floor that Section~\ref{sec:circuit-suf} measures and sets aside.

The logit lens and the tuned lens are \emph{projections}: they choose what to read a hidden state
through \citep{nostalgebraist2020logitlens,belrose2023tunedlens}. The projection here is the ordinary one. What this lens changes is the \emph{accounting} applied afterward, which of the terms in Equation~\ref{eq:contrib} are counted as explaining the prediction and which are set aside as the architecture's own traffic.

\paragraph{The embedding frame.} A transformer's residual stream does not arrive in token coordinates until very late, and models
mostly operate off the token axis. The prediction reaches that axis in the last few layers by being
written onto it, rather than by the accumulated content turning onto
it \citep{oskin2026offaxis}. Equation~\ref{eq:contrib} is read in the right basis by construction: it is evaluated \emph{at} the
readout, and the unembedding is the frame the readout is written in. Anything read earlier is not. Ask what a mid-stack component passes to the component that reads it.
Score the writer through the unembedding and the reader through its own layer's basis and the answer
is nothing, a correlation of $-0.02$ against a random control, which reads as \emph{the graph does
not link}. Score both sides in the reader's own frame and the same edges correlate $+0.28$ and beat a size-matched random control on $76$ percent of cases (Section~\ref{sec:naming-provenance}). The two halves had been scored in different bases, and a null produced by the wrong frame looks exactly like a null.

The lens we construct uses the \emph{layer-native token table}. Every reading taken between layers
depends on it. Fix a position $p$ and a set of contexts $\mathcal{C}$ drawn from held-out text. A context is one
sequence at the model's training length, and $p$ is a single position inside it --- the same position
in every context --- so a context contributes one measurement rather than one per token. For a
candidate token $t$, let $z^{(l)}(c,t)$ be the input to layer $l$'s feed-forward block at position
$p$ when $t$ is substituted there in context $c$. The table is the average of that state over contexts,
\begin{equation}
  \tilde{E}[l,t] \;=\; \frac{1}{|\mathcal{C}|}\sum_{c \in \mathcal{C}} z^{(l)}(c,t) \;\in\; \mathbb{R}^{d},
  \label{eq:etil}
\end{equation}
which is the layer's own picture of the token, in the layer's own frame. One row per token per layer,
computed once in a single pass, with nothing fitted and no per-input work.

That it is measured rather than fitted is the point. A tuned lens
\citep{belrose2023tunedlens} would also supply a per-depth frame, but it is trained, and then
understanding the explainer becomes part of understanding the model. $\tilde{E}$ is read off the
model's own forward passes, so a reading taken through it is a derivation rather than a second
model's opinion. Where this paper decodes anything below the readout --- what a component fires on,
what an edge carries, what a mid-stack write means to the layer that consumes it --- it is scored
against $\tilde{E}$ at that layer.

\subsection{Identifying components that matter}
\label{sec:circuit-accounting}

The ninety percent used throughout is a bar on contribution mass. A set clears it when its members
account for nine tenths of the signed total arriving at the readout. How much of a prediction such a
set actually carries is a separate measurement, made in
Section~\ref{sec:circuit-necessity}.

Ranking components by $\kappa_u$ and asking how many are needed to reach ninety percent of the total
requires choosing what the total is. Three choices are available:
\begin{equation}
  n_{90}^{\text{net}}: \textstyle\sum_{i \le n} \kappa_{(i)} \ge 0.9 N,
  \qquad
  n_{90}^{\text{pos}}: \ldots \ge 0.9 \textstyle\sum_u \max(\kappa_u, 0),
  \qquad
  n_{90}^{\text{abs}}: \ldots \ge 0.9 \textstyle\sum_u |\kappa_u| .
  \label{eq:denominators}
\end{equation}
Contributions cancel heavily, so the three denominators differ by orders of magnitude.
Table~\ref{tab:accounting} measures the consequence on the same predictions with the same
attribution: keeping the sign and dividing by the net returns tens of components, while discarding
the sign returns tens of thousands. The opposing mass is twenty times the net at the median on
this model, seven across the eighteen measured here, and it is spread thin: reaching ninety percent
of the prediction takes a quarter of every component in the model under the positives-only
denominator, and half of them under the absolute one. It is not a rival circuit that the net
denominator hides; Appendix~\ref{sec:pit-opposing} tests that against its floor.

\begin{table}[htbp]
\centering\small
\caption{Components needed to reach ninety percent of a prediction, under the three denominators of
Equation~\ref{eq:denominators}. Only the denominator changes. Median over 115 predictions, each the model's own top-1 token with
probability at least 0.3. The two rightmost conventions are the ones in common use; the same
comparison across twelve models trained by other people is Table~\ref{tab:stock}.}
\label{tab:accounting}
\input{T_accounting_base}
\end{table}

\begin{figure}[htbp]
\centering
\includegraphics[width=\textwidth]{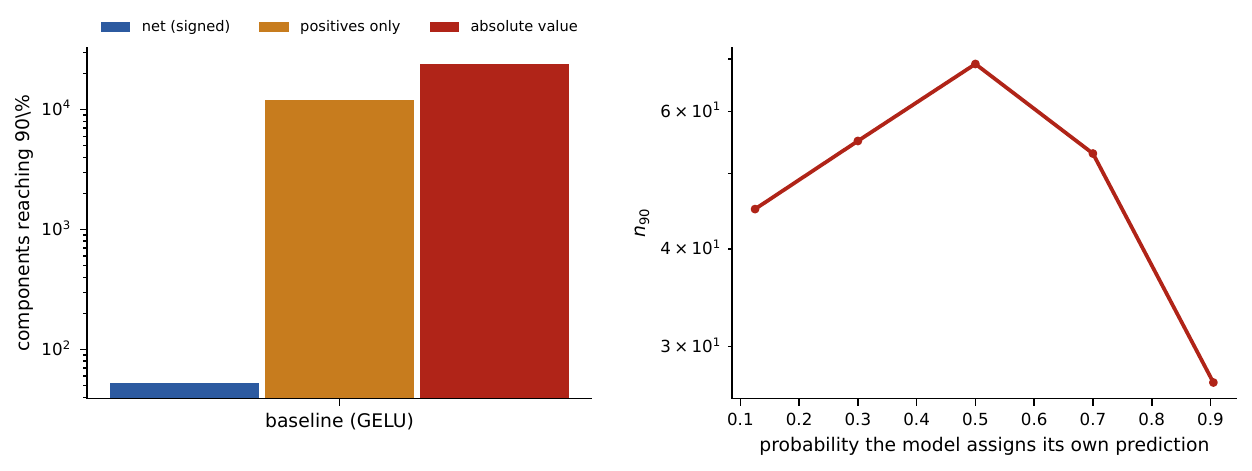}
\caption{Left: the same counts on a log axis. The choice of denominator moves the answer by two to
three orders of magnitude, which is larger than any difference between the models being compared.
Right: circuit size against how confident the prediction is. Confident predictions rest on fewer
components, so a circuit size quoted without the confidence of the predictions it was measured on is
not comparable across papers. Every other number in this section is measured at probability
$\ge 0.3$.}
\label{fig:accounting}
\end{figure}

The count is insensitive to where the bar is set. On the same ranking the median count at eighty
percent is $0.75$ to $0.82$ of the count at ninety, and at ninety-nine percent it is $1.16$ to
$1.25$, across the five models trained here it was run on. Nearly the whole usable range of the bar
moves the answer by less than a factor of two. The tail is enormous counted absolutely and nets to almost nothing, so raising the bar finds few components left to add.

\subsection{What part of a transformer is necessary}
\label{sec:circuit-necessity}

\begin{figure}[htbp]
\centering
\includegraphics[width=0.66\textwidth]{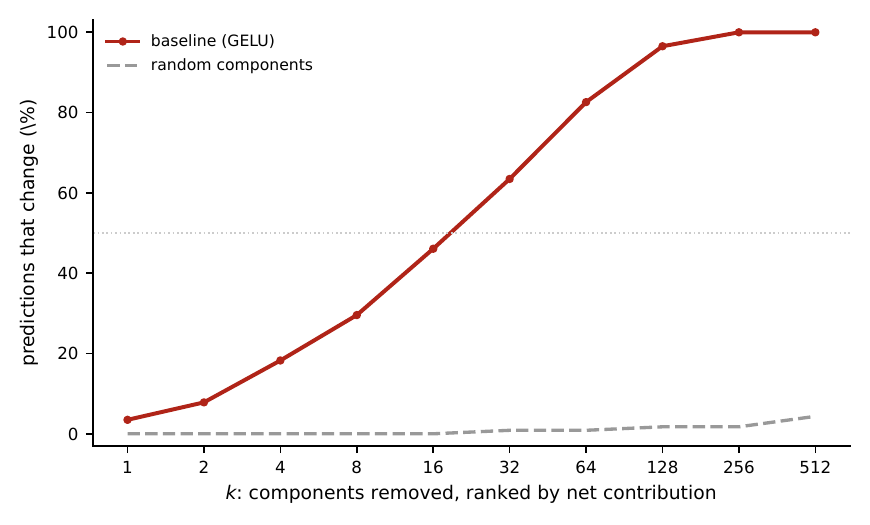}
\caption{Removing the $k$ highest-contributing components changes the prediction, where removing $k$ random components leaves it alone. Bands are bootstrap 95 percent intervals over predictions. The random
control stays at or below 5 percent across the whole range, while by $k=128$ the identified
components have changed nearly every prediction. The baseline is plotted here;
Figure~\ref{fig:flipk-stock} shows the same curves for the twelve models trained by other people.}
\label{fig:flipk}
\end{figure}

The accounting names a set. We call it the one-hop set, because every member is credited for writing
to the output directly rather than through anything else. Whether that set is the reason for the prediction is settled by removing components and re-running the model. Remove the top $k$ at the position being predicted; Figure~\ref{fig:flipk} reports how often the prediction
changes; the control removes the same \emph{number} of components, chosen at random.

Few components are needed in any model we measured, which is a statement about transformers rather
than about anything built here; Section~\ref{sec:stock} makes that case across twelve models
trained by other people.

Removing a set is not the same as adding up its contributions. Equation~\ref{eq:contrib} is linear,
so it predicts exactly what a removal should cost: the sum of the contributions taken away. Perform
the removal and the \emph{target logit} falls by only $42$ to $50$ percent of that
(Table~\ref{tab:linear}). Both quantities are logit mass, and neither is a rate at which the
prediction changes; that is Figure~\ref{fig:flipk}, and a removal can take half the predicted logit
away and still leave the argmax where it was. The shortfall is made up by the components left
behind, which compute something different once the removed ones are gone --- the self-repair
reported by \citet{mcgrath2023hydra} and \citet{rushing2024selfrepair}. Ablation is therefore a second measurement rather than arithmetic on
the first.

\begin{table}[htbp]
\centering\small
\caption{Removal cost against predicted cost, both measured in logits. Each entry is the realized
drop in the target logit divided by the sum of the contributions of the components removed, so
$1.00$ would mean the model absorbed the removal exactly as the linear attribution predicts and the
removal test would add nothing. About half the predicted loss is instead recovered by the components left behind. This is a statement about logit mass rather than accuracy, which is Figure~\ref{fig:flipk}.}
\label{tab:linear}
\input{T_linear}
\end{table}

\paragraph{What to measure when a component is removed.} Removing components shrinks the residual
stream. The final norm then divides by a smaller quantity, so every logit inflates together, and a
drop in the target's logit mixes two effects: what happened to that token, and a shift applied to the
whole vocabulary. Only the first is about the prediction, so damage has to be read off something a
uniform shift cannot move.

We use the \emph{margin}: the target logit minus its strongest competitor. A shift common to every
logit cancels from that difference, and the margin falls monotonically as more components are
removed, which any measure of damage has to do. It also joins two readings that could otherwise
disagree --- whether the prediction flipped, and how much damage was done --- because a flip
\emph{is} the margin reaching zero. The change in log probability is equally invariant, and is used
where a likelihood is the natural reading.

Anything read off the argmax was never exposed to this, because a constant added to every logit
cannot change which entry is largest: the flip curves, the minimality search, the sufficiency
numbers and $n_{90}$ itself are all unaffected. On this model the choice barely matters, and
Table~\ref{tab:margin} shows the raw drop and the margin disagreeing about the direction of the
effect in three percent of cases at the median $k$. It matters elsewhere. Measured in raw logits, the
ratio of realized to predicted effect comes out \emph{negative} on GPT-2, which is not a possible value.
Section~\ref{sec:stock-instrument} reports
raw drops disagreeing with the margin up to forty-five percent of the time and falling as $k$ grows.

\begin{table}[htbp]
\centering\small
\caption{The two quantities on the baseline. ``Sign disagreement'' is how often the raw logit drop and
the margin disagree about the direction of the effect, at the median and worst $k$; ``non-monotone
steps'' counts the times the median drop \emph{falls} as $k$ grows, which a measure of damage should
never do. Here they agree, with no non-monotone steps. Table~\ref{tab:margin-stock} is the same table on
models where the raw quantity fails.}
\label{tab:margin}
\input{T_margin}
\end{table}

\paragraph{What the removal costs elsewhere.} Any sufficiently large perturbation lowers a logit, so
the effect on the target has to be read against the collateral. Define specificity as the drop in the
target logit divided by the mean absolute change across the rest of the vocabulary.
Table~\ref{tab:specificity} compares three rankings.

\begin{table}[htbp]
\centering\small
\caption{Specificity: target-logit drop divided by mean absolute change elsewhere. Ranking by
absolute contribution selects large negative contributors, and removing them can \emph{raise} the
target logit, which is why that ranking scores worst. Ranking by activation magnitude selects
high-norm components whose removal disturbs the whole distribution. The net ranking is the most specific at every $k$, by a margin over the alternatives that widens as $k$ grows.}
\label{tab:specificity}
\input{T_specificity}
\end{table}

\paragraph{The required graph.}\label{sec:circuit-req}
The required graph is the smallest prefix of the one-hop set, ordered by contribution, whose removal
changes the prediction. Every required-graph number in this paper is found by scanning prefix
lengths upward and stopping at the first that flips.
Appendix~\ref{sec:pit-ablate} covers the flaws in the more attractive-looking alternatives. On some predictions no prefix flips at all, so the quantity is undefined there; the proportion
is reported with each table, and medians are taken over the predictions where it exists.

\subsection{What part of a transformer is sufficient}
\label{sec:circuit-suf}

Necessity asks what a prediction cannot survive losing. The opposite question is what it takes to
\emph{build} it: keep a set of components, zero every other feed-forward unit and attention
channel at that position, let the survivors recompute from what is left, and ask whether the model
still emits the same token. Nothing is spliced in. A component that is kept fires from the residual
that remains, which is the token embedding plus whatever the other survivors have written.

\paragraph{The choice of replacement value.} A transformer has no null value for a component's
output, so an ablation must write something in its place, and three conventions are in common use:
zero, the component's mean output over a sample of inputs, and a value resampled from a different
input. Nothing in the architecture prefers one. For the necessity measurement above the choice barely matters
--- removing the same components under all three changes the same predictions to within seven points
of each other, because a large signed contribution is gone whichever value is written where it was.
It matters a great deal for what follows, and zero is the choice made here; the alternatives are
tested below and in Appendix~\ref{sec:circuit-asym}.

Finding such a set is a search, and its shape follows from why a kept set fails: some member is
starved, its drivers among the components that were zeroed, and the repair is to put those drivers
back. Algorithm~\ref{alg:suf} gives the procedure --- seed with the required graph, repair starved
members from their suppliers until the model emits the token, then cut every member the set can
do without --- with no free parameters and a stopping rule the model supplies.

\begin{figure}[htbp]
\centering
\includegraphics[width=\textwidth]{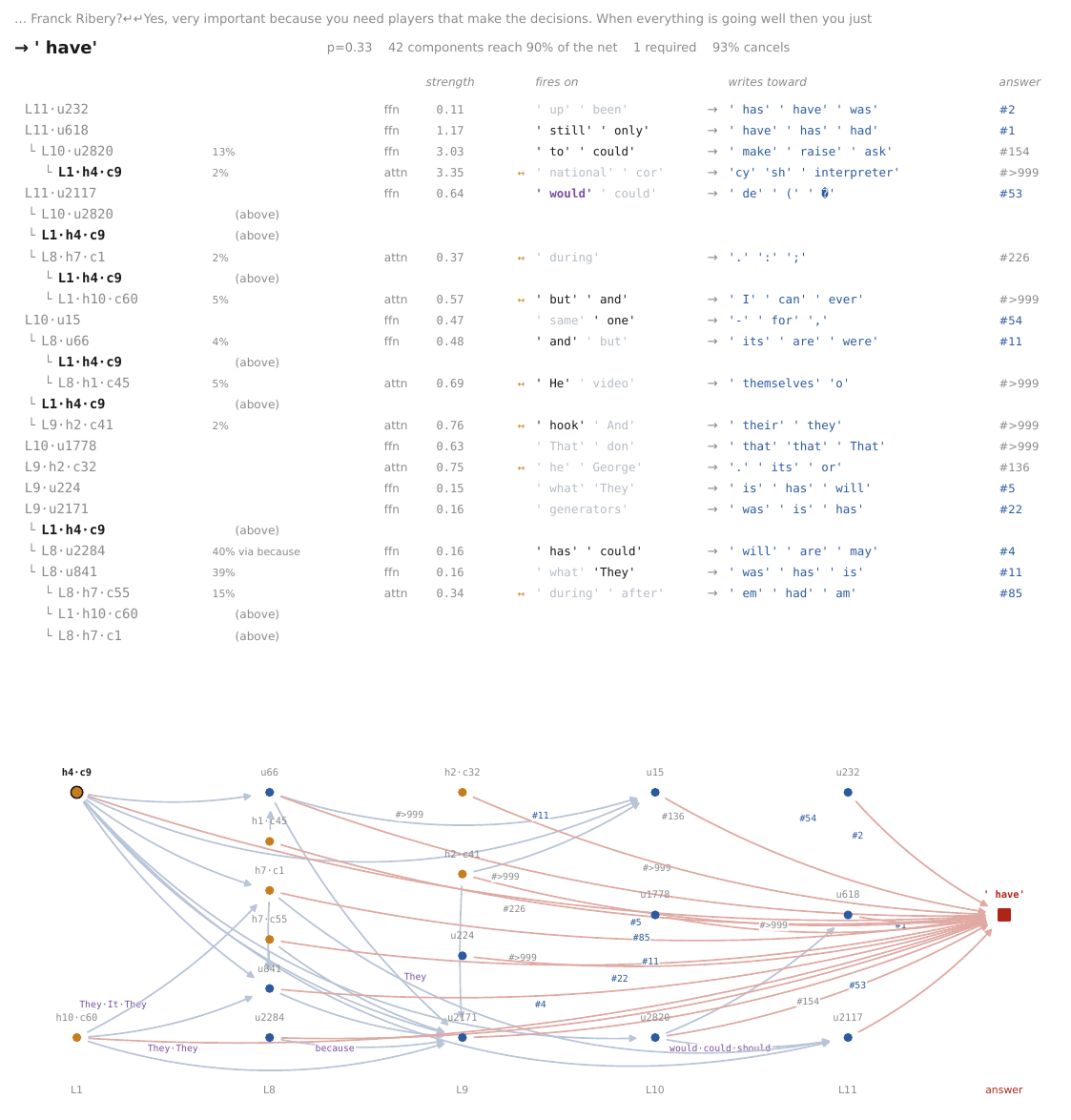}
\caption{One prediction's sufficient set. Keeping these eighteen components and zeroing every other
feed-forward unit and attention channel at this position leaves the model predicting the same token;
$n_{90}$ is $42$ of the model's $46{,}080$ components, and seven of the eighteen are attention
channels. Bold marks the one component whose \emph{removal} from the intact model changes the
prediction: an attention channel at layer one, feeding a unit at layer ten, feeding the unit at layer
eleven that promotes the predicted token first among the whole vocabulary. Rows are the members
nearest the readout first, with what each draws on indented beneath it and its share of that member's
incoming drive; the last column is where the predicted token sits in what the component promotes. In
the graph, a member that writes to the answer has an edge to it, labeled with the rank the predicted
token holds in what that member promotes. Appendix~\ref{sec:gallery} has the full key.}
\label{fig:example}
\end{figure}

Figure~\ref{fig:example} is one such set drawn out, with what each member fires on and what it
writes toward.

\begin{figure}[htbp]
\centering
\small
\begin{minipage}{0.94\textwidth}
\rule{\textwidth}{0.7pt}\vspace{0.3em}\\
\refstepcounter{algorithm}\label{alg:suf}\textbf{Algorithm~\thealgorithm}\; The smallest set that builds a prediction, at position $p$ for token $t$.
\vspace{0.2em}\\ \rule{\textwidth}{0.4pt}
\begin{tabular}{@{}r@{\hspace{0.9em}}l@{\hspace{1.1em}}p{0.67\textwidth}@{}}
& $m(S)$ & the margin of $t$ over its nearest rival when only $S$ is kept at $p$;
           the set is \emph{sufficient} when $m(S) > 0$ \\[0.45em]
& $\delta_u(S)$ & what component $u$ delivers to the readout under $S$, short of what it delivers
           with the model intact \\[0.45em]
1 & \textbf{seed}   & $S \gets$ the required graph: the shortest prefix of
                       Equation~\ref{eq:contrib} whose \emph{removal} changes the token, or the
                       one-hop set where no prefix does \\[0.2em]
2 & \textbf{repair} & while $m(S) \le 0$: take the $u \in S$ with the largest $\delta_u(S)$ and add
                       the components supplying it, shortest prefix first, until $u$ delivers again.
                       The prediction counts as a member of $S$, its suppliers being the one-hop
                       set \\[0.2em]
3 & \textbf{cut}    & score $d_u \gets m(S \setminus \{u\})$ for every $u \in S$; stop when no
                       single removal leaves a sufficient set; otherwise drop the largest group
                       off the cheap end of $d$ that $S$ survives, and repeat \\
\end{tabular}
\rule{\textwidth}{0.7pt}
\end{minipage}
\caption{Every line is a forward pass of the model scored against its own answer, with nothing fitted and nothing spliced in. Step 2 is what separates this from a search over the contribution ranking:
it uses the dependency structure to say \emph{which} components are missing rather than trying
prefixes until one works, and it stops when the model emits the token rather than when a budget runs
out. Step 3 scores many candidates in one pass.}
\end{figure}

\begin{table}[htbp]
\centering\small
\caption{The smallest set that reproduces a prediction, and what the same number of components does
when chosen differently. ``Converged'' is how many of the sampled predictions the procedure verified. A prediction is dropped
when the repair step of Algorithm~\ref{alg:suf} reaches its round limit without the kept set becoming
sufficient, which is a limit of the procedure rather than a fact about the model and, since the
repair has typically built hundreds of components by then, evidence that the criterion is not one
that volume satisfies.
``Floor'' is the share of predictions on which keeping \emph{nothing} at the position --- so that the
residual is the token embedding alone --- already returns the same token. No set kept on top of that
can be credited with producing it, and the search returns a set of size one on every one of them, so
those predictions are counted here and then set aside. The size and the two controls are medians over
what remains after both exclusions. The last column is
the median set as a fraction of the model's components.}
\label{tab:suf}
\input{T_recsuf_base}
\end{table}

Eight components reproduce the prediction, out of $46{,}080$ (Table~\ref{tab:suf}). That is under two
parts in ten thousand of the model, and it is not an artifact of the test. Zeroing everything else
also deletes the competition, so the floor has to be measured first and then excluded; on this model it is $12$ percent of
predictions. On the predictions that remain, eight components drawn at random reproduce the token on \emph{none} of
them --- not one of $115$ --- and the eight highest-contributing on $7$ percent, against the search's
hundred.

Two checks say a set is not right for the wrong reasons. Adding components does not satisfy the
criterion: where the repair never converges it has already built a mean of $489$ components, at
most $962$ or two percent of the model, and the token still does not come out. And the sets are
conditioned on their context rather than carrying their token --- kept at a position in a different
context whose intact answer is a different token, seven percent still emit the original and
eighty-nine percent emit neither it nor the new one.

The set does not depend on where the search starts, and it is not the only one. Starting the
search from a random draw of components the size of the required graph, instead of from the required
graph itself, and running the same repair and cut reaches a sufficient set on $85$ percent of
predictions against $89$, at the same median size of six on that draw, for about half as much work again. Where
both starts succeed, the set found from the random start and the set found from the required graph
overlap at a Jaccard of $0.31$, against $0.0004$ for two random sets of the same sizes, and roughly
half the pairs share at least half their members. A fifth share nothing. That is what an
overdetermined prediction should give, and the repair reports the same thing on its own on half the
predictions it succeeds on: the model holds more than one set of components that can carry the
answer alone.
Section~\ref{sec:compare-suf} reports the same measurement on the models trained elsewhere.

The search uses the answer: every step asks whether the
\emph{known} token survives, so what it establishes is that such a set exists. It is not a
procedure that predicts one in advance, which is worth pursuing and not attempted here. And the set is sufficient at one position, with the rest of the sequence
computing normally, so any attention channels among its members read ordinary values from the
positions around them. What the measurement licenses is that eight of the model's components need to be active
\emph{here} for this token, not that eight components are all that are needed for the entire
transformer to function.

The result does not depend on zeroing. Zero is one choice of what to write where a removed
component was, and the objection it invites is that a model with most of a position blanked is
nowhere it has ever been. Section~\ref{sec:circuit-bookkeeping} supplies a replacement that keeps it
on-distribution: the layer-to-layer map, which is what the removed components would have delivered if
they were doing only carriage. Run under both backgrounds, the prefix-and-prune search of Figure~\ref{fig:example}
returns a median of $18$ components against the map and $16$ against zeros. The pair is to be
read against each other rather than against the $8$ of Table~\ref{tab:suf}, which is a
different search on a draw excluding a different floor. The carriage (described below)
alone predicts the token on a quarter of predictions, and those are excluded here for the same reason
the token embedding's own successes are: a prediction the background alone already makes is not
evidence about any set. A small set suffices whether the space around it is emptied or filled with
the traffic the architecture would normally have put there.

Every size reported here is an upper bound returned by a heuristic search, leaving the minimal sufficient circuit open. What the
sizes establish is that sufficient circuits orders of magnitude smaller than the published counts can
be located at all. The cost of that search grows with the model. On the twelve-layer models it
converges in a few dozen forward passes. On a twenty-four-layer model most predictions still
converge, and a minority exhaust a budget an order of magnitude larger and return a bound rather than
an answer. Finding the circuits in a very large model is a search problem, and scaling that search is left for
later work.

\paragraph{What the surviving sets look like.} They are small graphs rather than bags, as
Figure~\ref{fig:example} shows: its members sit at five depths, several draw on one another, and the
seven attention channels among them are the only route by which anything outside the position reaches
it. Between two
thirds and three quarters of the members are feed-forward units and the rest are attention channels,
spanning about five layers, and most feed-forward members draw part of their drive from other
members --- though only about a tenth of it at the median, the rest arriving from components outside
the set entirely.

\paragraph{Do models say ``not this token'' in a circuit?} A prediction can be reached by pushing the
answer up or by pushing its rivals down, and the accounting so far counts only the first. Asking
about the second needs positions where the model faced two candidates. Pairs of positions that share
their last four tokens and end in different confident predictions, $A \rightarrow X$ and
$B \rightarrow Y$, are such positions, since what separates them is what attention carried in from
earlier context. There are a hundred per model, from held-out text at the training length. Where a
token wins, its net-denominator $n_{90}$ is tens of components; where it loses, it still carries a
circuit of the same order and keeps $58$ to $73$ percent of its push. Its promoters overlap across
the two contexts at a Jaccard of $0.28$ to $0.45$, and the winner's and the loser's promoters overlap
at $0.14$ to $0.33$ (Table~\ref{tab:pair}). The loser is not pushed down: its net
push stays positive on at least $95$ percent of pairs. It is out-pushed, by the winner's circuit
firing harder in the winner's context, by a median of three to five logits. We looked for circuits
that say ``not this,'' components whose removal would hand the prediction to the token they hold
down, and they are not there: the mass pushing the loser down is no different from any other
token's (Appendix~\ref{sec:pit-opposing}). Removing eight of the
winner's promoters moves the prediction on $30$ to $72$ percent of pairs; removing eight of the
loser's suppressors moves it to the loser on $8$ to $26$ percent, and would do the same for any
token. The baseline's other converged seeds fall within the ranges quoted, with one exception taken
up in Appendix~\ref{sec:baseline-pitfalls}.

\begin{table}[htbp]
\centering\small
\caption{Two answers at one position. On a hundred contrastive pairs per model, two positions
sharing their last four tokens and ending in different confident predictions: the net-denominator
$n_{90}$ of the winner where it wins and of the loser both where it loses and where, at the other
position, it wins; the share of the loser's winning-context push its top sixteen promoters still
deliver where it loses; and the Jaccard overlap of the loser's top sixteen promoters across the
two contexts, and of the winner's with the loser's at the position the winner takes. Medians over
pairs.}
\label{tab:pair}
\input{T_pair}
\end{table}

Three numbers now sit on top of one another, answering three questions about the same prediction.
The signed accounting says which components \emph{carry} it: fifty-three at the median, read off
the parameters and activations with nothing removed. Ablation says which of those it \emph{cannot
lose}: thirteen. And the search says which set \emph{builds} it: eight. Fewer than a sixth of the
components carrying ninety percent of the logit are needed to reproduce the token.

\subsection{What the rest of the model is doing}
\label{sec:circuit-bookkeeping}

A prediction resting on a few dozen components leaves an obvious question: what is everything else doing? ``Nothing'' is not credible in a trained network. Three candidates are tested in turn: staging
for later tokens, effort on rival answers, and the architecture's own overhead. If the
remainder were staging information for \emph{later} tokens, then removing everything at a
position except that position's circuit would spare the token being predicted and damage the ones
that follow, which attend back to it. Table~\ref{tab:bookkeep} tests exactly that and finds no such staging. Keeping only the circuit costs a quarter as much at the token itself as keeping an equally
sized random set, which is the necessity result restated. On the tokens that follow, it costs the
same as the random set, and it goes on costing the same however far out one looks: the median
per-prediction ratio of the two costs runs $0.98$ to $0.99$ from sixteen tokens to five hundred and
twelve, with no trend in distance (Table~\ref{tab:horizon}). Whatever the rest of the model is doing
at this position, it is not principally holding things for the future, and the window it is not
holding them for is the whole context. Section~\ref{sec:compare-standing} reports the same sweep on
nine further models.

\begin{table}[htbp]
\centering\small
\caption{Everything at a position is removed except the set named. Cost at this token is the fall in
the margin; cost on the next sixteen is the rise in their cross-entropy. Keeping the circuit beats keeping a random set for the token being predicted alone, matching it on the tokens after. The two later-token medians are taken separately, so their ratio differs from the median
per-prediction ratio quoted in the text.}
\label{tab:bookkeep}
\input{T_bookkeep}
\end{table}

\begin{table}[htbp]
\centering\small
\caption{The same test at six horizons, nested on one set of positions so the horizon is the only
thing varying. Each entry is the median over predictions of one prediction's circuit cost divided by
its random-set cost, so $1.00$ means the circuit is worth no more than an equally sized random set
for the tokens that follow. ``At it'' is the same ratio at the predicted token, where a low number
is the necessity result.}
\label{tab:horizon}
\input{T_horizon_base}
\end{table}

A second candidate is that the remainder is spent on near misses. Section~\ref{sec:circuit-suf}
found that where a token wins, the token that lost at that position still carries a circuit of tens
of components and keeps most of its push, and the same records say how much of the model's pushing
goes to candidates that do not win. At the position where a token wins with net push $N$, the token
that lost there carries $0.63$ to $0.84$ of $N$ across the six models of Table~\ref{tab:pair}, a
competitor drawn from the model's top fifty carries $0.43$ to $0.79$, and a random vocabulary token
carries none. Counted as positive mass rather than net, the runner-up draws on $0.94$ to $0.99$ of
what the winner draws on and a top-fifty competitor on $0.88$ to $0.98$. The model promotes its near
misses nearly as hard as its answer, each with a circuit of the winner's order: $26$ to $48$
components on the four smaller models, one to two hundred on Gemma-3 and Qwen. That is effort of
the same kind as the answer's, and it accounts for a few dozen components per candidate, not for
the tens of thousands whose pushes cancel.

The remaining candidate is that the remainder is the architecture's own overhead: moving information
between layers, and keeping a superposed residual stream organized so that many concepts can share
one vector space \citep{elhage2022superposition}. Work of that kind has to happen somewhere, and it happens \emph{everywhere}. It has no reason to point at the token being predicted, and shows up as an enormous signed mass, spread over tens of thousands of components, that nets to little at the readout. That overhead has a known shape. \citet{oskin2026offaxis} report a transformer's residual stream
turning as a near-rigid body from depth to depth, with the prediction arriving on the readout in the
last few layers by being written onto it. What follows is an independent check on that picture, made
with a different instrument and for a different purpose --- not the geometry of the stream, but how
much of a layer's work a fixed map of the incoming state can account for. It confirms the published
result and adds two things to it: a size for the reshaping, and a test of whether the map can stand
in for the components.

Write $x^{(\ell)}$ for the residual entering layer $\ell$
at the position being read, so that the layer's whole contribution there is
\begin{equation}
  \Delta^{(\ell)} \;=\; x^{(\ell+1)} - x^{(\ell)},
  \label{eq:carupd}
\end{equation}
everything the attention and feed-forward blocks write into the stream at that depth. Carriage means
the part of that update which is the same transformation applied to whatever happens to be there, so
it is what a \emph{fixed} map of the incoming state can account for. Fit that map over a sample of
positions $\mathcal{P}$ and score it on a disjoint sample $\mathcal{P}'$:
\begin{equation}
  A_\ell \;=\; \arg\min_{A} \sum_{p \in \mathcal{P}} \big\| x^{(\ell)}_p A - x^{(\ell+1)}_p \big\|^2 ,
  \qquad
  \hat{\Delta}^{(\ell)} \;=\; x^{(\ell)}\big(A_\ell - I\big),
  \label{eq:carmap}
\end{equation}
with a ridge term for conditioning. The share of the update the map accounts for is the part of
$\Delta$ lying \emph{along} $\hat{\Delta}$, which is bounded by one however large the prediction is:
\begin{equation}
  s_\ell \;=\; \Bigg( \frac{\sum_{p \in \mathcal{P}'} \big\langle \Delta_p,\; \hat{\Delta}_p /
  \|\hat{\Delta}_p\| \big\rangle^{2}}{\sum_{p \in \mathcal{P}'} \|\Delta_p\|^{2}} \Bigg)^{1/2} .
  \label{eq:carshare}
\end{equation}
A ratio of norms is not so bounded and will exceed one when the map's prediction is large and misaligned.
Restricting $A$ to a rotation gives the rigid variant, by orthogonal Procrustes on the same samples:
$Q_\ell = UV^{\!\top}$ where $X^{\!\top}Y = U\Sigma V^{\!\top}$. A rotation preserves every length and
angle, so $s_\ell(A_\ell) - s_\ell(Q_\ell)$ is the shear --- the part of the map that reshapes the
stream rather than turning it.

Measured this way, most of what a layer writes is carriage. A fixed map of the incoming state
accounts for $74$ percent of the update on the baseline and $79$ on the sigmoid model, with a
positive held-out $R^2$ at every depth, and a rotation alone accounts for the bulk of that (Table~\ref{tab:carriage}). That much
is the near-rigid body of \citet{oskin2026offaxis}, recovered here without appeal to any frame.

What is left over is where this measurement goes past the published one. The stream is reshaped
between depths and not merely reoriented, and the reshaping is concentrated at the entrance: the
first layer shears the stream three to five times as much as any layer after it. 

\begin{table}[htbp]
\centering\small
\caption{How much of a layer's update a fixed map of the incoming state accounts for, as a median
over layers. ``Rigid part'' restricts that map to a rotation, which preserves every length and angle;
the shear is what the full map explains beyond it, and is the median of the per-layer difference
rather than the difference of the two medians beside it, so those columns are not meant to subtract.
The shear is concentrated in the first layer in both models.}
\label{tab:carriage}
\input{T_carriage}
\end{table}

Whether the map can stand in for the components is tested by substitution. We run the model with a
layer's own output replaced by what the map predicts, and with the
actual writes of a chosen set $S$ added back on top:
\begin{equation}
  \tilde{x}^{(\ell+1)} \;=\; x^{(\ell)} A_\ell \;+\; \sum_{u \in S \cap \ell} a_u c_u ,
  \label{eq:carsub}
\end{equation}
applied at the position being read while the rest of the sequence computes normally, exactly as in
the ablations above. With $S$ empty and one layer substituted, the model still produces the
same token on a median of $92$ percent of predictions.

One part of what the accounting sets aside is not carriage but supply. A member of a circuit has
inputs of its own, written by components upstream, and those are causally necessary to it: remove a
node's attributed sources and its activation falls by two thirds, where removing as many random
upstream components leaves it unchanged (Appendix~\ref{sec:circuit-branch}). The one-hop accounting
cannot see them, because it credits a component only for what its own write lands on the readout
and never for what a later component reads from it. The sufficiency search of
Section~\ref{sec:circuit-suf} recovers them by repair, and the closure of Appendix~\ref{sec:appgraph}
recovers all of them, at one to three percent of the model. The cancellation that governs the readout
governs a component's input too, and more severely: reaching ninety percent of a node's incoming
drive takes about a dozen sources counted with sign and many thousands counted without, a ratio near
a thousand against about twenty at the readout, which is why the branching figures of
Appendix~\ref{sec:appgraph} are computed signed. The bar of ninety percent is a choice, and
Section~\ref{sec:circuit-accounting} tests it: at the readout the count moves by less than a factor
of two between eighty and ninety-nine percent, because the tail is enormous counted absolutely and
nets to almost nothing. A bar at one hundred percent, the first count at which the signed sum
reaches the net, is barely larger, and what it never does is draw in the thousands of sources that
cancel, since a signed sum reaches the net without them. Raising the bar buys a somewhat larger
closure, still a graph of overlapping backward trees at a few percent of the model, not the model.

\subsection{Limitations}
\label{sec:circuit-limits}

Five qualifications travel with every number in this section.

The first is that credit is direct. A component is scored by what its own write places on the
readout, so one that acts only by changing what a later component computes never enters the ranking,
however much the prediction turns on it. This reaches further than the accounting: the sufficient
sets of Section~\ref{sec:circuit-suf} are grown from candidates the same equation nominates, so a
component that acts only through others cannot enter one of those either, however much it matters.
The recursion of Appendix~\ref{sec:circuit-branch} recovers such components as sources, which is
much of why a closure is larger than the set it closes over.

The second is that the unit is a choice. An attention head is sixty-four channels, and counting the head
rather than the channel cuts $n_{90}$ by more than half while lifting attention's apparent share by
two thirds (Table~\ref{tab:granularity}). Channels are used throughout because they
put attention and feed-forward components on the same footing; the per-head figures are given so the
comparison with work that counts heads is available.

\begin{table}[htbp]
\centering\small
\caption{The one-hop accounting with attention counted per channel and per head. The result
survives the coarser unit, though the numbers do not transfer between the two conventions.}
\label{tab:granularity}
\input{T_granularity}
\end{table}

The third is that late layers are over-credited. A component is a neuron activation times its write
column, and late layers project more directly onto the readout, so direct contribution flatters them.
Comparing attributed mass against causal effect on this model, the later half of the stack is credited
with about twice the direct contribution of the earlier half while removing its top $k$ changes barely
more predictions --- at $k=16$, thirty-two percent against twenty-nine. The attribution over-credits
late layers by a factor of $0.53$ at $k=16$ and $0.65$ at $k=64$. Section~\ref{sec:stock} reports the
same measurement on models trained by other people, where it does not always come out this way.

The fourth is that the prediction is a choice. Circuit size falls with the model's confidence in its
own output (Figure~\ref{fig:accounting}, right), so these figures describe confident predictions, and
a number quoted at a different confidence floor describes a different population.

The fifth is that a collapsed model reads as maximally legible. The worst excluded baseline seed sits far
below its siblings on LAMBADA (Table~\ref{tab:baseline-seeds}) and reads as perfectly
legible: $n_{90}=1$, a single component flips the prediction, and attention accounts for all of it. Seeds that merely miss the quality bar read normally: across the eight seeds
of the sigmoid construction of Section~\ref{sec:mono}, the three that failed the screen span the same $n_{90}$, the same attention share
and the same depth as the five that passed. The quality screen therefore travels beside every figure here.

\secbarrier

%% file: T_models.tex
\setlength{\tabcolsep}{5pt}
\begin{tabular}{@{}ll>{\raggedright\arraybackslash}p{0.26\textwidth}>{\raggedright\arraybackslash}p{0.34\textwidth}@{}}
\toprule
model & activation & property & what it is for \\
\midrule
baseline (GELU) & GELU & not monotone & the conventional reference; every result in Sections~\ref{sec:circuit}--\ref{sec:edit} is measured on it \\
sigmoid & logistic & monotone, injective, bounded & order-preserving and bounded \\
softplus & softplus & monotone, injective, unbounded & separates boundedness from order-preservation \\
ReLU & ReLU & monotone, \emph{not} injective & separates order-preservation from information-preservation \\
unshaped sigmoid & logistic & monotone, no crispness term & isolates the shaping term from the activation \\
set operators & fuzzy set operations & bounded, two operands per unit & explicit logic in place of a nonlinearity \\
\bottomrule
\end{tabular}

%% file: T_accounting_base.tex
\begin{tabular}{lrrrr}
\toprule
model & net (signed) & positives only & absolute value & inflation \\
\midrule
baseline (GELU) & 53 & 11,974 & 23,971 & 452$\times$ \\
\bottomrule
\end{tabular}

%% file: T_linear.tex
\begin{tabular}{lrrr}
\toprule
model & $k=4$ & $k=16$ & $k=64$ \\
\midrule
baseline (GELU) & 0.42 & 0.50 & 0.47 \\
\bottomrule
\end{tabular}

%% file: T_margin.tex
\begin{tabular}{lrrrrr}
\toprule
& & \multicolumn{2}{c}{sign disagreement} & \multicolumn{2}{c}{non-monotone steps} \\
\cmidrule(lr){3-4}\cmidrule(lr){5-6}
model & target logit & median & worst & raw & margin \\
\midrule
baseline (GELU) & 12.3 & 3\% & 17\% & 0 & 0 \\
\bottomrule
\end{tabular}

%% file: T_specificity.tex
\begin{tabular}{lrrrr}
\toprule
model & $k$ & net-ranked (ours) & absolute-ranked & activation-ranked \\
\midrule
baseline (GELU) & 4 & \textbf{3.6} & 2.0 & 2.9 \\
 & 16 & \textbf{5.1} & 2.9 & 3.0 \\
 & 64 & \textbf{5.5} & 3.2 & 2.7 \\
\addlinespace
\bottomrule
\end{tabular}

%% file: T_recsuf_base.tex
\setlength{\tabcolsep}{4pt}
\begin{tabular}{lrrrrrrr}
\toprule
& & & & & \multicolumn{2}{c}{same size, other sets} & \\
\cmidrule(lr){6-7}
model & preds & floor & converged & size & top-$k$ & random & \% of model \\
\midrule
baseline (GELU) & 150 & 12\% & 115 & \textbf{8} & 7\% & 0\% & 0.017\% \\
\bottomrule
\end{tabular}

%% file: T_pair.tex
\footnotesize
\begin{tabular}{@{}lrrrrrr@{}}
\toprule
& \multicolumn{3}{c}{$n_{90}$, net denominator} & loser's push & \multicolumn{2}{c}{promoter overlap} \\
\cmidrule(lr){2-4}\cmidrule(lr){6-7}
model & winner & loser & loser where it wins & kept & loser, across contexts & winner vs loser \\
\midrule
baseline (GELU) & 74 & 39 & 89 & 60\% & 0.33 & 0.14 \\
sigmoid & 58 & 41 & 66 & 67\% & 0.39 & 0.19 \\
set operators & 39 & 22 & 44 & 73\% & 0.45 & 0.28 \\
GPT-2 124M & 80 & 60 & 108 & 67\% & 0.39 & 0.33 \\
Gemma-3 1B & 250 & 212 & 251 & 48\% & 0.28 & 0.19 \\
Qwen2.5 1.5B & 146 & 126 & 158 & 54\% & 0.39 & 0.19 \\
\bottomrule
\end{tabular}

%% file: T_bookkeep.tex
\begin{tabular}{lrrrrr}
\toprule
& & \multicolumn{2}{c}{cost at this token} & \multicolumn{2}{c}{cost on the next 16} \\
\cmidrule(lr){3-4}\cmidrule(lr){5-6}
model & circuit & keep circuit & keep random & keep circuit & keep random \\
\midrule
baseline (GELU) & 57 & 10.84 & 36.18 & 0.090 & 0.071 \\
\bottomrule
\end{tabular}

%% file: T_horizon_base.tex
\setlength{\tabcolsep}{4.5pt}
\begin{tabular}{lrrrrrrrr}
\toprule
& & & \multicolumn{6}{c}{tokens after the prediction} \\
\cmidrule(lr){4-9}
model & preds & at it & 16 & 32 & 64 & 128 & 256 & 512 \\
\midrule
baseline (GELU) & 206 & 0.25 & 0.98 & 0.98 & 0.99 & 0.99 & 0.99 & 0.98 \\
\bottomrule
\end{tabular}

%% file: T_carriage.tex
\begin{tabular}{lrrrrr}
\toprule
& & & & \multicolumn{2}{c}{shear by depth} \\
\cmidrule(lr){5-6}
model & map explains & rigid part & shear & first layer & the rest \\
\midrule
baseline (GELU) & 74\% & 63\% & 11 & 35 & 11 \\
sigmoid & 79\% & 66\% & 11 & 50 & 11 \\
\bottomrule
\end{tabular}

%% file: T_granularity.tex
\begin{tabular}{lrrrr}
\toprule
& \multicolumn{2}{c}{per channel} & \multicolumn{2}{c}{per head} \\
\cmidrule(lr){2-3}\cmidrule(lr){4-5}
model & $n_{90}$ & attention & $n_{90}$ & attention \\
\midrule
baseline (GELU) & 53 & 27\% & 24 & 45\% \\
\bottomrule
\end{tabular}

%% file: naming.tex
\section{What's in a Name?}
\label{sec:naming}

Section~\ref{sec:circuit} showed that a prediction rests on a few dozen components and that they can
be found without training anything. That does not say what any of them \emph{is}. This section asks the second question, of a unit and of a channel, and of what each
writes and what each reads. The two sides are named by different means: what a component writes is
read off the predictions it drives, and what it reads is decoded from its parameters in the frame of
its own layer.

\subsection{What a unit writes}\label{sec:naming-vocab}\label{sec:naming-write}

Section~\ref{sec:circuit}
already answers which components carry a prediction: the accounting set is the components whose direct
contribution reaches ninety percent of the net, and the shares that define it are the same shares used
here. Those components wrote that token, by construction rather than by inference, since they are
selected by their contribution to it. So a unit can be named without
reading its column weights at all. Take a prediction, take its strongest contributors, and repeat until the
model is covered.

Doing that costs little. A million and a half predictions take minutes, and each one
supplies a few dozen memberships.  What comes back has to be reported in three parts (Table~\ref{tab:naming}). A component is \emph{covered} when it appears among the
strongest contributors to any prediction at all; the loosest statement, reaching most of the model. It is \emph{characterized} when it appears there twenty times or more, which is when there
is enough to say what it writes. And its set of tokens is \emph{coherent} when the set holds together
against a frequency-matched null, judged in a distributional space built from the training corpus, so
the yardstick is the data the model was trained on rather than another model's geometry. Taken
together, close to half of the components in a model have a nameable write.

\begin{table}[htbp]
\centering\small
\caption{Naming components by the predictions they drive, on the baseline over $1.5$ million
predictions. \emph{Covered} is appearing among the top-$k$ contributors to any prediction,
\emph{characterized} is appearing there twenty times or more, and \emph{coherent} is the share of
those whose token set beats a frequency-matched null in a distributional space built from the training
corpus. Every share is of all $46{,}080$ components. Widening $k$ trades a little coherence for a
great deal of coverage, which is why the last row and not the first is the operating point.}
\label{tab:naming}
\input{T_naming}
\end{table}

Components write a closely related set of tokens rather than one exact token, with the relation within a set grammatical rather than semantic. A unit promotes negated auxiliaries, or first names,
or units of measurement, or the tokens that open a sentence --- categories a reader can state in a
phrase, and not the topical groupings that the word \emph{concept} usually suggests. Only a tenth of one percent of components write a single token.
Weighted by how much each token receives rather than counted, a third of components put half their
contribution on a single token, so the two ways of asking give different answers.  The breadth is visible on the page.
Every node in Appendix~\ref{sec:gallery} carries the list of tokens it writes toward, and the token
being predicted is one entry among several --- so a component that helps deliver an answer is seen
promoting it alongside its neighbors rather than singling it out.

A component supplies a stable share: across the contexts where a component and a token appear
together, its coefficient of variation has a median of $0.24$, and two thirds of pairs sit below
$0.3$. So the amount belongs to the component rather than to the context.  That number can be checked by deleting the component and watching what the prediction loses, which is
a different measurement from the attribution that produced it. The two agree at $r = 0.60$ and the
relation is monotone: components credited with under five percent of the drive cost about a third of a
logit when removed, those credited with five to fifteen percent cost four fifths of one, and those
above fifteen percent cost more than two. A name of the form \emph{this supplies a seventh of the
drive toward} `\texttt{ dogs}' is therefore a claim about the model.

\begin{figure}[htbp]
\centering
\includegraphics[width=\textwidth]{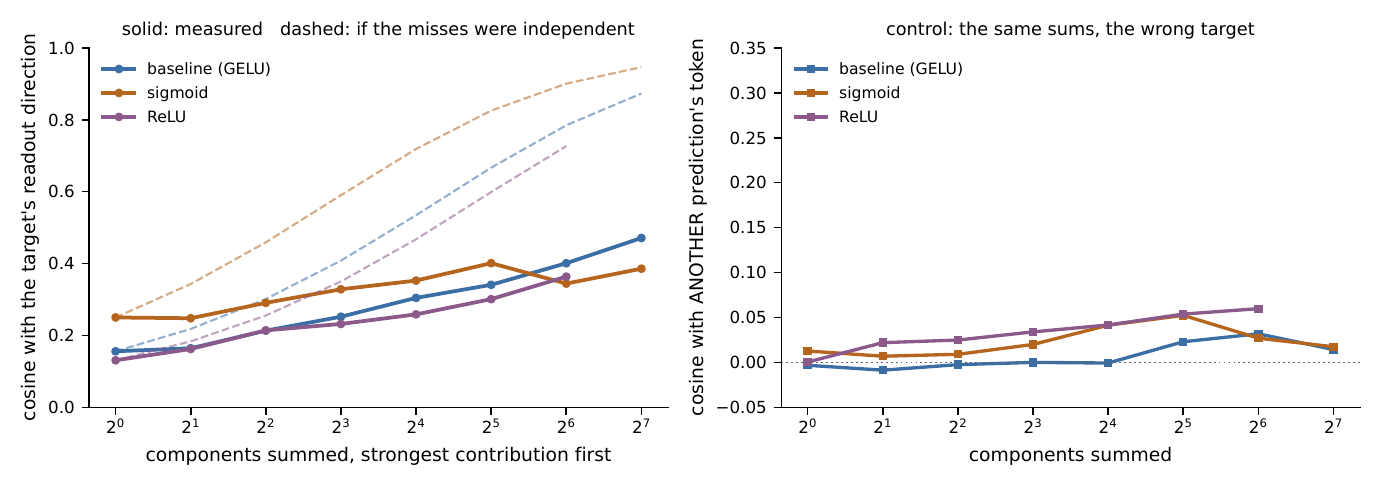}
\caption{How a prediction is assembled. The cosine here is taken against a \emph{known} target rather
than searched for over the vocabulary, which is what makes it answerable where the column measures of
Appendix~\ref{sec:geometry-tried} are not. Left: the cosine between the target's readout direction and
the running sum of what its components write, added strongest contribution first; the dashed line is
the climb independent misses would give, anchored on each model's own first member. Every model falls
below it, so the sum is carried by its largest terms rather than by their number. Right: the control, the same partial sums scored against a different prediction's token. Flat and near
zero for every model plotted, so what the left panel shows is aim rather than undirected drift toward
the token cone.}
\label{fig:sum}
\end{figure}

Components nameable this way are late in the model. In
the baseline's feed-forward blocks the characterized share runs at one or two percent through the
first layers, reaches a quarter by layer six, half by layer eight, and ninety-five percent at the
last; attention lags by about two layers and then catches up. Coherence does not move with depth, sitting between eighty and ninety-nine percent everywhere, so depth changes how many components can be named rather than how well. What does change is breadth: two tokens carry half a component's
contribution in the middle of the stack and nine do at the last layer, which is the same distributed
commitment seen in Figure~\ref{fig:sum}, arriving here as a property of the names.

\begin{figure}[htbp]
\centering
\includegraphics[width=\textwidth]{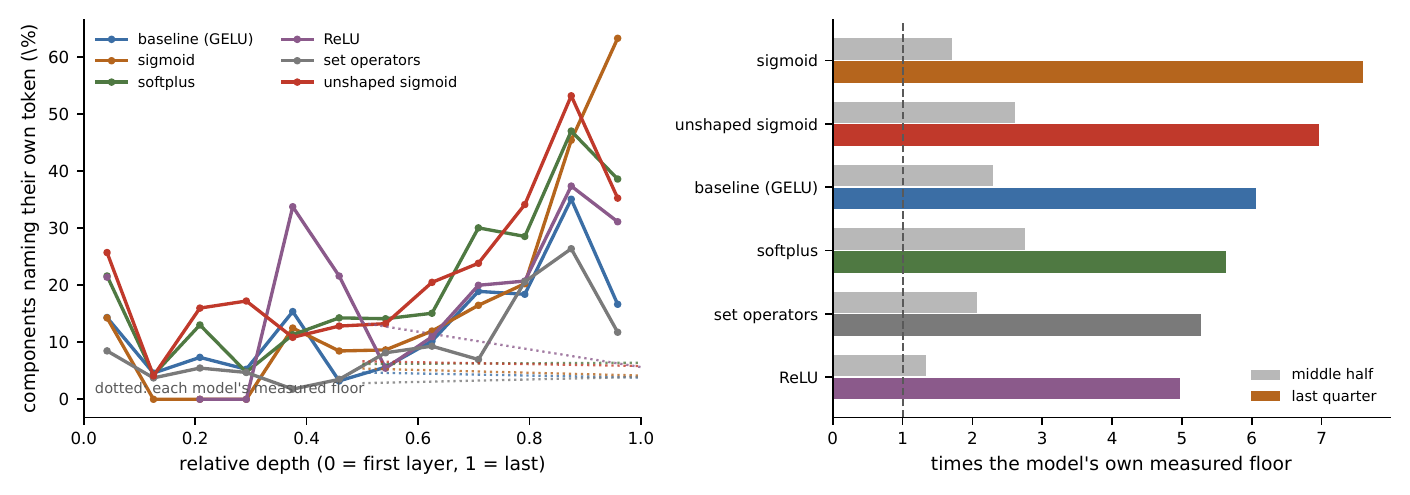}
\caption{A component \emph{names its own token} when the token the model predicts lands inside the
top hundred of its write column's ordering over the vocabulary, signed by how the component fired.
Left: the share by relative depth, one line per model, with each model's own measured floor dotted.
Right: the middle half of the stack and the last quarter, as multiples of that floor, which is the
only form in which models can be set beside one another. Rank exists only for components appearing in
a trace, so this is conditioned on a component mattering to some prediction.}
\label{fig:rankdepth}
\end{figure}

Three other measurements put the same event at the same depth. The residual first carries the
token the model will \emph{predict} at layer eight: the token \emph{at} the position starts at
$+0.45$ and is gone within a single layer, while the one it emits sits below zero through the
body of the stack. Write columns first clear chance there. And a component's own ranking of the token it helps
predict lifts there --- score a component by whether that token lands in the top hundred of what it
promotes, and the last quarter of the stack clears its own floor by five to eleven times against the
middle's one and a third to five and a half, higher in the last quarter than in the middle in every
model measured (Figure~\ref{fig:rankdepth}).\footnote{The floor of this measurement is set by the
readout's geometry --- tokens with large centered rows sit near the top of almost any column --- and runs
from half a percent in one model to thirteen in another, which is why the figure reports multiples of
each model's own floor. Appendix~\ref{sec:geometry-tried} takes the readout apart; the depth result
survives that correction, the floor halving and the margin over it widening.}

A name says which tokens a component writes toward, leaving the sum to pick one of them. Components write toward the result and rarely at it.  The predicted token
sits inside the top hundred of two thirds of the weighted contribution, and inside the top thousand of
ninety-three percent of it, so no member selects the answer on its own.
Individually each is an aimed near miss. Score a contributing component's own write, scaled
by how it fired, against the direction the readout would need: a typical member sits at $0.07$, which is far too small for a maximum over the vocabulary to register and still carries aim,
because the same
vectors score about zero against a different prediction's token. Adding them in order of contribution
takes the alignment to $0.45$, three times the best single member (Figure~\ref{fig:sum}). The climb is
slower than independent averaging would give, and not because the misses share a direction --- measured
directly they are close to orthogonal, at a mean pairwise cosine of $0.004$. It is that the members are
unequal: the strongest contribution is twice as well aimed as a typical one, $0.156$ against $0.071$,
so the sum is carried by its largest terms and each additional weak member buys less than the last.

\subsection{What a unit reads}\label{sec:naming-answer}

A unit's parameters name the
tokens that drive it, and the name can be checked against the model rather than against whether the
list looks sensible. Asked for a unit's eight strongest drivers, the baseline's parameters read
through the layer basis described below recover the true set at $58.9$ percent, thirty-seven times
chance --- the strongest agreement between a component's weights and its behavior anywhere in this
paper. How far down the list that holds is answered below as well.

A read row is a covector rather than a vector in the residual stream: it consumes states and returns a number, expressed in the frame of the layer that hosts it rather than the frame the output head reads. Projecting it
through the vocabulary, as the logit lens reads
activations~\citep{nostalgebraist2020logitlens}, asks it what it would mean somewhere it is not, and
recovers $7$ percent against a chance rate of $1.6$. The repair needs no fitting. The layer-native
token table $\tilde{E}$ of Equation~\ref{eq:etil} is the layer's own picture of each token, in the
layer's own frame, and a read row decodes against it directly, both sides centered across the
candidate vocabulary $V$:
\begin{equation}
  s_u(t) = \big(w_u - \bar{w}_u\big) \cdot \big(\tilde{E}[l,t] - \bar{\tilde{E}}[l]\big),
  \qquad
  \mathcal{P}_u^K = \operatorname*{arg\,top-}K_{\,t \in V} \; s_u(t).
  \label{eq:decode}
\end{equation}
The architecture's carriage cancels on both sides of that dot product, which is why no transport
between frames is required and why nothing has to be trained to supply one. With the same weights, the same tokens and the same unit, the basis is the whole of the difference between $7$ percent and $58.9$.

\begin{figure}[htbp]
\centering
\includegraphics[width=0.74\textwidth]{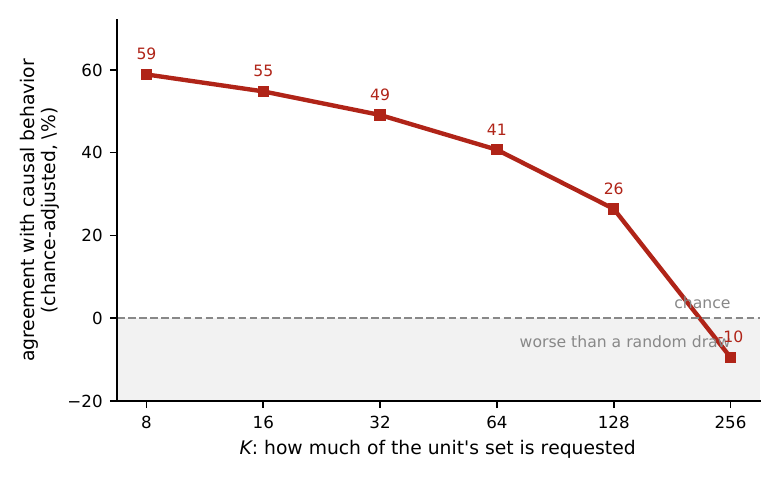}
\caption{Whether a baseline unit's parameters identify the set that drives it, against how much of
that set is requested. Chance-adjusted per Equation~\ref{eq:overlap}; the raw chance rate rises from
1.6 percent at $K=8$ to 50 percent at $K=256$. Agreement decays with $K$ and passes below chance,
where the parameters select tokens overlapping the true set less often than a random draw. Table from
16 contexts, causal behavior from 16 disjoint contexts, context length 2048. Agreement here is taken under the mean summary of Equation~\ref{eq:causal}, chance-corrected; Appendix~\ref{sec:pit-convention} says what the alternative would give.}
\label{fig:agreementk}
\end{figure}

What a unit
\emph{does} respond to is measured by substitution rather than inferred. On a disjoint set of contexts
$\mathcal{C}'$, so that the naming and the ground truth cannot share sampling noise,
\begin{equation}
  \bar{a}_u(t) = \frac{1}{|\mathcal{C}'|}\sum_{c \in \mathcal{C}'} a_u\big(z^{(l)}(c,t)\big),
  \qquad
  \mathcal{A}_u^K = \operatorname*{arg\,top-}K_{\,t \in V} \; \bar{a}_u(t).
  \label{eq:causal}
\end{equation}
and the two sets are compared, corrected for the rate at which a random list of the same length would
agree:
\begin{equation}
  \mathrm{ov}@K(u) = \frac{|\mathcal{P}_u^K \cap \mathcal{A}_u^K|}{K},
  \qquad
  \widehat{\mathrm{ov}}@K = \frac{\mathbb{E}_u[\mathcal{P}_u^K \cap \mathcal{A}_u^K]/K - K/|V|}{1 - K/|V|}.
  \label{eq:overlap}
\end{equation}
The candidate set $V$ is the most frequent tokens in the corpus, and its size sets the chance rate:
the sweep of Figure~\ref{fig:agreementk} uses $|V| = 512$, so chance at $K=8$ is $1.6$ percent and at
$K=256$ is $50$, while the unit-against-channel comparison of Table~\ref{tab:readname} uses $256$.
Since a chance-corrected overlap must reach zero once $K$ approaches $|V|$, the two are read against their own candidate sets rather than against each other. A name that survives this is the set that moves the unit.\footnote{Equation~\ref{eq:causal} averages activations and takes the top $K$. Ranking within each context and averaging the ranks instead is not interchangeable with it: on the baseline the choice is worth $18$ points of agreement, on a bounded unit $2$. The sweeps over $K$ and the ceiling comparison of Section~\ref{sec:mono-legibility} use the mean; the per-seed tables use the rank, and each says which. Appendix~\ref{sec:pit-convention} says why both are reported.} 

Four quantities recur from here on, each reported under the conditions in Table~\ref{tab:contract}. The sweep over $K$ is the one worth naming here: the
baseline's own agreement changes sign between $K=8$ and $K=256$, so a paper quoting only the
first number would report a conventional transformer as substantially legible.

\begin{table}[htbp]
\centering
\small
\caption{The measurements used throughout the paper, and the condition each is reported under. Each condition is a specific failure from Appendix~\ref{sec:baseline-pitfalls} that moved a number in the flattering direction.}
\label{tab:contract}
\begin{tabular}{@{}l@{\hspace{1.2em}}p{0.30\textwidth}p{0.44\textwidth}@{}}
\toprule
& what it measures & always reported \\
\midrule
$\widehat{\mathrm{ov}}@K$ & whether the parameters name the inputs that causally drive the component
  & swept over $K$, never at one cut; chance-corrected; table and ground truth from disjoint
    contexts; quoted with the number of contexts behind it \\
\addlinespace[0.35em]
$n_u$ & how large the set being named actually is
  & as excess over the unit's own median, so a baseline firing rate does not count as response \\
\addlinespace[0.35em]
$\mathrm{PR}$ & how much representational variety the model retained
  & split-half, to remove the estimation ridge; beside every legibility figure, because a collapsed
    model reads as more legible \\
\addlinespace[0.35em]
LAMBADA, ppl & quality
  & beside every legibility figure, per seed, at a matched step count \\
\bottomrule
\end{tabular}
\end{table}

\paragraph{How far down the list the name holds.}\label{sec:naming-shape} The $58.9$ percent is one point on a curve that falls (Figure~\ref{fig:agreementk}). By the time two
hundred and fifty-six tokens are asked for, the parameters select worse than a random draw. So a unit has a few tokens named at the head of its list, and where that head ends is set by the
activation function rather than by the read row running out of what it knows. The read row itself
orders the vocabulary well throughout, which Section~\ref{sec:naming-readout} demonstrates on
models differing only in that function.

A second instrument asks a different question and reaches further down. Rather than asking what a
unit reads, collect the positions in ordinary text where it fires hardest and describe it by the
tokens sitting there --- \emph{when it turns on}. That description needs a floor of its own,
because a unit's firing correlates with general context and any random token would name it a
little: the floor is what the same unit yields at two dozen to five dozen tokens back. Against it,
two units in three clear at forty positions and three in four at eighty. Neither instrument
substitutes for the other --- the parameters say what moves a unit and stop after a handful of
tokens; the firing context describes far more units and says nothing about cause.

\paragraph{How much of a model works on concepts.}\label{sec:naming-about}

That the parameters pick out the right tokens says nothing about whether those tokens have anything
to do with each other. Somewhat over a quarter of a unit's response falls on tokens forming a concept the token embedding can see and we would identify as related.
The question resists a yes or no, for a reason that belongs to the embedding rather than to the
units. Similarity in embedding space,

\begin{equation}
  \mathrm{coh}(S) = \frac{1}{|S|(|S|-1)} \sum_{t \neq t' \in S}
  \frac{\hat{E}[t] \cdot \hat{E}[t']}{\|\hat{E}[t]\|\,\|\hat{E}[t']\|},
  \qquad \hat{E}[t] = E[t] - \bar{E}.
  \label{eq:coh}
\end{equation}
geometrizes \emph{paradigmatic} classes, where members are mutually substitutable and cosine finds
them. A \emph{relational} class is different: a corner can be on a street, in a room, or of a bed, and the members belong to a common concept by co-occurring in a structure, which nothing in the training objective pushes together. A unit responding to such
a class is invisible to Equation~\ref{eq:coh} by construction.

We score each token in a unit's set by how
well it belongs to the others, and the cut is calibrated against a random set of the same size rather
than chosen in raw cosine units. What is reported is the share of the unit's response on either side
of that cut. On the baseline eleven of a unit's top forty tokens fall on the coherent side and carry
$28$ percent of its response, leaving twenty-nine tokens and the other $72$ percent to the remainder.

That is a much smaller fraction than Table~\ref{tab:naming} reports for the same models, because the two are asking different questions. Equation~\ref{eq:coh} asks whether a set is paradigmatic, and answers
in the embedding, where relational and grammatical classes are invisible by construction. The naming
asks whether a set holds together in the corpus, where a class of tokens that occur in the same places
does hold together whether or not any word covers it. Nine sets in ten pass the second test and one in
three passes the first, and the distance between those numbers is the size of what a token embedding
does not group together: negated auxiliaries, sentence-openers, units of measurement, the tokens that close a
parenthesis. Those are real categories with no name in the vocabulary.

One thing that fraction cannot separate. The decode centers both sides across the candidate
vocabulary, which removes what the architecture's carriage does to every token alike but not what it
does to one token in particular, so a unit whose work is moving a specific token's information from
one depth to another reads here as a unit that responds to that token. Since most of what a component
writes is carriage in the first place (Section~\ref{sec:circuit-bookkeeping}), both shares are upper
bounds on what is semantic: a component doing a little prediction and a great deal of transport is
indistinguishable here from one doing only the first.

Taken together a unit is nameable on both sides by different means, with the embedding the weak link in naming it further.

\subsection{Naming an attention channel}\label{sec:naming-attnwrite}

A channel is named the same way a unit is (Table~\ref{tab:naming}). A prediction's strongest
contributors include channels, so what a channel writes is named on the same terms and to the
same standard, and its names are of the same kind --- numbers, the marks that close a clause, the openings of sentences.
Counted apart at a matched budget the two kinds come out close: about a third of each is
characterized, and the sets that result are coherent within a few points of one another. The naming reaches channels as readily as units. Where the two kinds differ is in depth. A channel is characterized in almost
no cases through the first third of the stack --- under a twentieth of them, against an eighth of
feed-forward units at layer zero.  This rises to a third by layer six, and from layer eight onward runs at
three quarters to five sixths, which is where units already are.

\begin{figure}[htbp]
\centering
\includegraphics[width=0.94\textwidth]{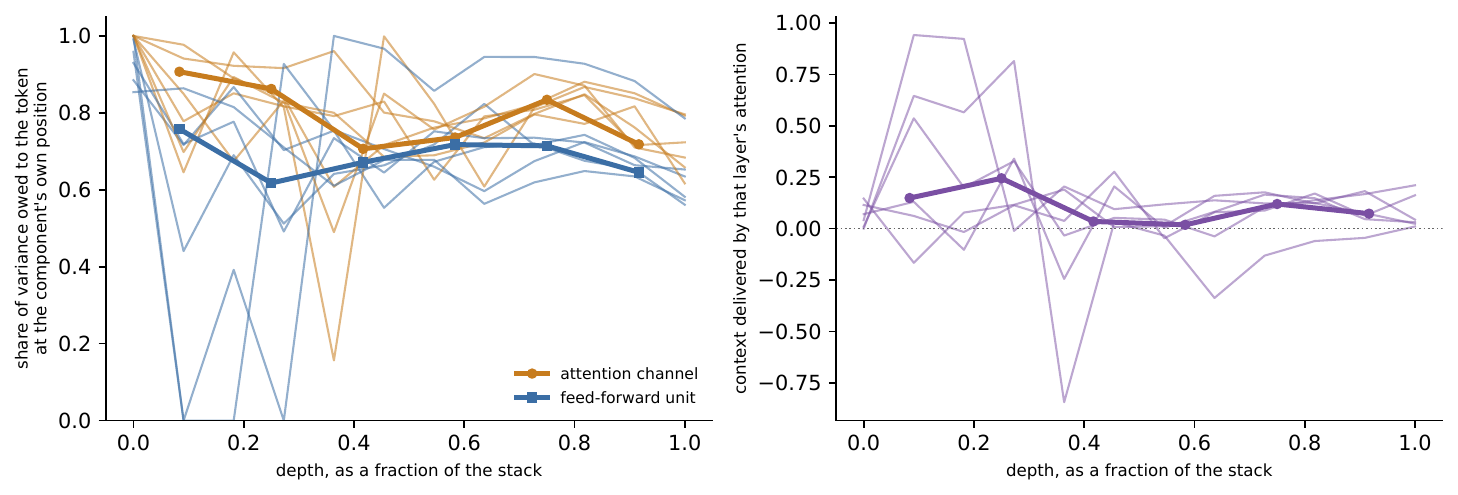}
\caption{Left: the read side of an attention channel and of a feed-forward unit, scored with the same
instrument in one pass per model. Right: both against depth. A channel's read side is measured from
its value projection, which is evaluated at the position being attended to and therefore moves when
the probe moves; a channel's \emph{output} at a position is assembled from other positions and does
not.}
\label{fig:attnread}
\end{figure}

\paragraph{What a channel collects.}\label{sec:naming-attn} Attention admits the same two questions a
feed-forward unit does. What a channel produces at a position is gathered from other positions, and
what it \emph{collects} from a position is local, which puts it within reach of the same techniques used to name a unit's inputs.  In fact, the instrument that names a unit's read side works better on a channel, without modification. A channel's value-projection row is its read row; decode it against the same layer-native
table and check it against the tokens the channel measurably picks up when one is substituted at the
position it reads. On the baseline the agreement is $0.844$ at eight tokens against the unit's
$0.598$, and where the unit's falls away as more of the list is asked for --- to $0.279$ by
sixty-four --- the channel's \emph{rises}, to $0.892$ (Figure~\ref{fig:attnread}).

That difference is the activation function.  A unit's ordering has to pass through its activation function on the way to its behavior, and a non-monotone one does not preserve it, where a channel's value path is linear and nothing stands between the row and what it collects.
Where the read side is concerned attention is the easier component to name. Described by when it
turns on instead, a channel does slightly less well than a unit: three in five at forty firing
positions against two units in three.

\begin{table}[htbp]
\centering\small
\caption{Whether a component's parameters name the tokens that drive it, against how much of the list
is asked for. Both kinds are scored by one instrument: a unit's read row and a channel's
value-projection row are each decoded against the layer's own token table
(Equation~\ref{eq:decode}) and checked against the set that causally moves the component, measured by
substitution on disjoint contexts and corrected for chance as in Equation~\ref{eq:overlap}. A channel
is flat or rising in every model because its value path is linear. A unit's ordering must survive its
activation function first, and what happens then depends on the function: the baseline falls below
chance by $128$ tokens, ReLU decays without crossing, and the sigmoid model holds almost flat. The
candidate set here is $256$ tokens, half the size used in Figure~\ref{fig:agreementk}, so the
baseline crosses at about two fifths of the candidate set in both --- $K\approx104$ of $256$ here and
$K\approx208$ of $512$ there. Where a unit's parameters stop naming its inputs is a share of the pool they are scored against rather than a fixed $K$. Agreement here is taken under the mean summary of Equation~\ref{eq:causal}, chance-corrected; Appendix~\ref{sec:pit-convention} says what the alternative would give.
This table is the one exception to the disjoint sets of Table~\ref{tab:contract}: its layer table
and its causal profile are accumulated in a single pass over the same contexts, which
Appendix~\ref{sec:baseline-pitfalls} measures at about a third of a point.}
\label{tab:readname}
\input{T_readname}
\end{table}

What a channel collects at the position it is reading is a separate question, and read that way
an attention channel reads tokens, and reads them at the bottom of the stack.
Substitute a token at one position, read the value projection there, and take the share of a
channel's variance owed to that token rather than to the context around it. At layer zero the quantity is $1.000$ on all six models. Above it the quantity falls, by an amount
that differs sharply between models, and the feed-forward units of the same layers sit lower at
almost every layer of every model, with ReLU (Table~\ref{tab:attn}) the exception.

A channel reads $\mathrm{ln}_1(x)$ and
the feed-forward unit above it reads $\mathrm{ln}_2(x + \mathrm{attn}(x))$, so the drop in token share
between them at one layer is the context that layer's attention delivered.

Two things about that quantity hold on every model measured. The first attention block delivers
little of it in absolute terms, between $0.00$ and $0.15$, so the token is present at the bottom of
the stack and moves little between positions --- though on the softplus and unshaped sigmoid models
that little is a large share of the small total those models deliver anywhere. And the layer that delivers most of it always sits in the first half of the stack. What does not hold everywhere
is the balance between the two ends: four of the six models deliver more context in their first third
than their last, and the softplus model and the unshaped sigmoid run the other way.  Why those two run the other way is left for later work.

\begin{table}[htbp]
\centering\small
\caption{What an attention channel collects at the position it is reading, and what its layer
delivers from elsewhere. Token share is the fraction of a channel's variance owed to the token at
that position rather than to its context; layer zero must return $1.000$, since the residual there is
the token, which is the check that the measurement works. ``Layers $>$ FFN'' counts the layers at
which the channel's token share exceeds that of the feed-forward unit above it, which reads the
residual after attention has written to it. The context a layer delivers is that difference, peaking in the first half of every stack.}
\label{tab:attn}
\input{T_attn}
\end{table}

All of this reaches the value side at a single position. What a channel finally contributes is that
value aggregated over the sequence by a pattern we do not explore further in this work.  Decoding
attention across the entire context is left to future work.

Finally, naming this way reaches only the components that write a prediction. A quarter of the
model contributes to predictions without ever being among the strongest contributors to one, and
those components are not idle: ranking contributions to the \emph{read rows} of the strongest
components rather than to the readout, a third of the leaders one level back are drawn from that
band. They drive the components that write, which no ranking against the readout can see.
Reaching them is what the graph built in Section~\ref{sec:circuit} is for.

\subsection{What the graph shows}
\label{sec:naming-graph}\label{sec:naming-provenance}

The three quarters of a unit's response that the embedding groups into nothing
(Section~\ref{sec:naming-about}) --- call them its \emph{anonymous} tokens --- have an origin even
where they have no name. Section~\ref{sec:circuit} supplies a graph over the same components, and
read through that graph those tokens are written by components upstream of the unit. Saying which
ones takes care (Table~\ref{tab:shapes}). Write $C$ for the unit being explained and $A$ for a component credited with
driving it. Of the three shapes that produce that credit, the one to locate is the \emph{direct} source. That
means ruling out \emph{siblings}, which read what $C$ reads and write nothing into it, and
\emph{mediated} sources, which reach $C$ only through something else.

\begin{table}[htbp]
\centering\small
\setlength{\tabcolsep}{5pt}
\begin{tabular}{@{}l l p{0.30\textwidth} p{0.28\textwidth}@{}}
\toprule
& shape & what it would mean & where it is settled \\
\midrule
direct & $A \rightarrow C$ & $A$ writes into the row $C$ reads with &
Equation~\ref{eq:provscore} \\
\addlinespace
siblings & $A \leftarrow X \rightarrow C$ & $A$ and $C$ read the same thing and never touch, so $A$
fires when $C$ fires while sending it nothing &
Equation~\ref{eq:damage}, with the same-layer null of Table~\ref{tab:causal} \\
\addlinespace
mediated & $A \rightarrow B \rightarrow C$ & $A$ reaches $C$, but only through $B$ &
handled via search (Section~\ref{sec:circuit-suf}) \\
\bottomrule
\end{tabular}
\caption{The three shapes that would put $A$ high in $C$'s ranking. Ranking cannot tell them apart,
because all three make $A$'s firing track $C$'s. Deletion kills the sibling case: with no edge between them, removing $A$ leaves $C$ untouched. Deletion cannot separate the mediated case, a real path, from a direct one, so the sufficiency search reaches it instead, because a set that needs $B$ has to keep $B$ to work.}
\label{tab:shapes}
\end{table}

What follows measures where a unit's \emph{feed-forward} drive comes from; a channel's value side is
taken up at the end of the section.

\paragraph{Ruling out siblings.} The anonymous tokens are not noise. Split sixteen contexts into two
groups of eight, average a unit's response over each group separately, and the two profiles agree at $0.648$, so the unit responds to those tokens the same way on text it has not seen before. Fitting one does not name it either --- every fitted estimator tried returns negative
held-out variance (Appendix~\ref{sec:naming-fitfail}).

No fitting is necessary, however.  The architecture that supplies the atoms also supplies an ordering among them. Take a unit $u$ at layer $\ell$ with read row $w_u$. Every
upstream component $v$ at layer $\ell_v$ has a write column $c_v$, and an activation on token $t$
which we write $a_v(t)$, in the sense of Equation~\ref{eq:act}. The drive $v$ contributes to $u$ is
then
\begin{equation}
  s_u(t) \;=\; \sum_{v \,:\, \ell_v < \ell} \omega_v \, a_v(t),
  \qquad
  \omega_v \;=\; \cos\!\big(\mathrm{carry}(c_v, \ell_v, \ell),\, w_u\big),
  \label{eq:provscore}
\end{equation}
with $\mathrm{carry}$ the composition of fitted layer maps of Section~\ref{sec:circuit-bookkeeping}
and no coefficient fitted anywhere. The weight $\omega_v$ is how much of what $v$ writes lands along the row $u$ reads with.

\begin{figure}[htbp]
\centering
\includegraphics[width=0.94\textwidth]{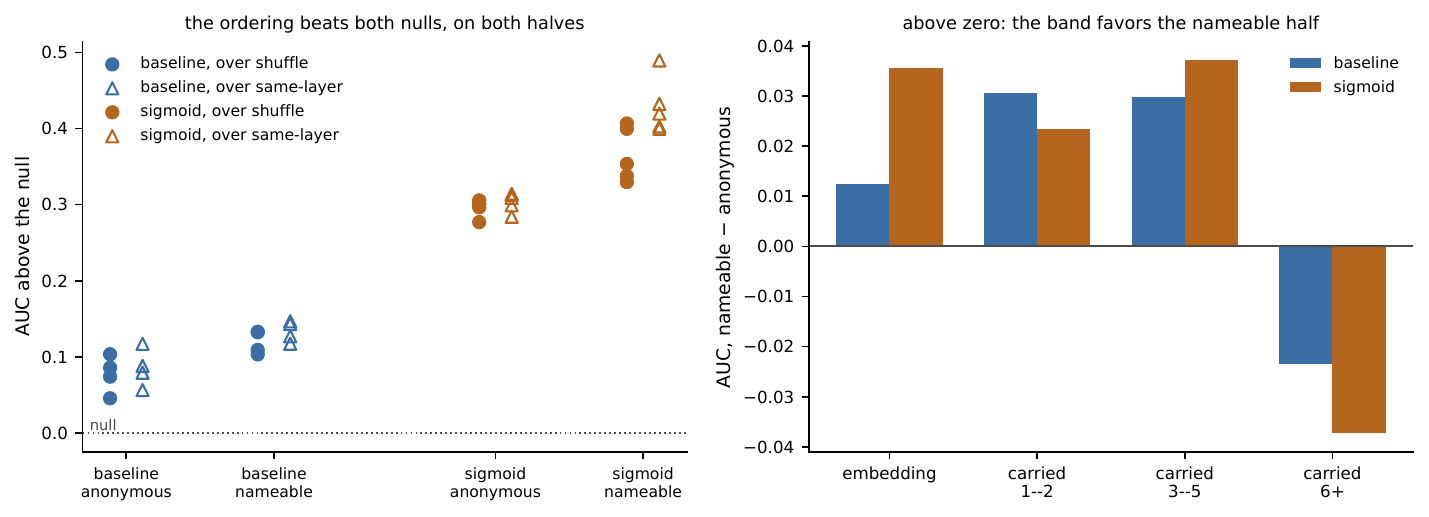}
\caption{Left: how far Equation~\ref{eq:provscore} ranks a unit's drivers above each of its two
nulls, one point per seed. Plotted as the excess, so the null is the dotted line and any point above
it is the ordering beating that null; filled circles are the excess over the shuffle, open triangles
over the same-layer control. Every group clears both, on the anonymous part as well as the nameable
one. Right: the same score split by how far each term was carried, as the gap between the two halves.
A bar above zero is a band that favors the nameable tokens and one below it favors the anonymous
tokens, so the sign change across the bands is the result: what arrives at the embedding is what a
static token table can read, and what has been carried furthest is what it cannot.}
\label{fig:prov}
\end{figure}

Ranked by $s_u$, the tokens that drive $u$ sit above the rest of the candidate pool, for
the anonymous part as well as the nameable one, in both families and at every seed that trained
(Figure~\ref{fig:prov}, left). Three nulls say the ordering is not an artifact of the machinery that
produced it: permuting which component holds which alignment, setting every weight to one, and
substituting components at the target's own layer that cannot have driven it. All three return chance
or close to it.

The sibling shape is not disposed of that way. Two components that read the same thing fire at the
same times, so a sibling of $u$ has a firing profile shaped like $u$'s --- and a sum of enough
siblings rebuilds $u$'s own profile, which then ``predicts'' what drives $u$ while none of them wrote
anything into it.

Turning that shape into a score takes a substitution. Weight each writer not by whether it wrote
into the row $u$ reads with, but by whether it \emph{resembles} $u$, meaning it reads what $u$ reads:
\begin{equation}
  \omega^{\text{sib}}_v \;=\; \cos\!\big(w_v,\, w_u\big),
  \label{eq:weights}
\end{equation}
read row against read row, with only the weight changed.
Equation~\ref{eq:provscore} is the direct shape scored and Equation~\ref{eq:weights} is the sibling
shape scored. On the baseline the two rank a unit's drivers about equally well, so no ranking can
choose between them.

Deleting components can, because the two shapes predict different damage. Ablate a set $A$ and
re-read the unit; what a score expects to lose, token by token, is that same sum restricted to what
was removed,
\begin{equation}
  \hat{d}_u(t) \;=\; \sum_{v \in A} \omega_v \, a_v(t) .
  \label{eq:damage}
\end{equation}
with $\omega_v$ either weight. With $|A| = 1$ the two differ by a positive scalar, so one deletion can never separate them. With $|A|$ larger
the weights become a mixture over profiles.
Ablating twenty-four components at random, so the choice favors neither shape, the direct weighting
tracks the damage and the sibling weighting stays flat, in both families.

That settles who writes the anonymous tokens. Where those writers sit explains why the vocabulary misses them. Split Equation~\ref{eq:provscore} by how far
each term was carried and the two halves of a unit's set separate: the embedding term favors the
\emph{nameable} tokens; terms carried six layers or more favor the anonymous ones
(Figure~\ref{fig:prov}, right).

This is the shape \citet{oskin2026offaxis} predicts. A term that enters at the
embedding and is read a few layers later has been turned only a few times and still lies near the
coordinates the token table is written in. A term written at layer one and read at layer nine has been
turned nine times and lies somewhere else. Decoding both through one static table shows the first
alone, and what a reader then sees is a nameable quarter beside an anonymous remainder --- one set of
tokens, read in two frames, with the table in only one of them.

\begin{table}[htbp]
\centering\small
\caption{One unit's causal top forty, separated by where each token's drive arrives from. Same units,
same tokens, and same measurement as Table~\ref{tab:gelu-units}, which lists them as a single list and
where no description fits; each token is assigned here to the source band supplying most of its drive,
standardized per band so the assignment is not simply the largest band. The parts are categories: capitalized sentence-openers, modals, do-support, prepositions and wh-words, third-person pronouns. Baseline model, $K=40$, context length 2048.}
\label{tab:gelu-units-rot}
\input{T_gelu_units_rot}
\end{table}

Sorted by where it came from, the remainder stops looking like a grab-bag. Assigning each of a unit's
tokens to the band supplying most of its drive splits the set into a median of four groups, and those
groups are categories: a nearest-centroid classifier fitted to half a unit's tokens recovers the
source of the other half well above a shuffled control, and part-of-speech purity within a group
exceeds a matched random partition.
Mean pairwise cosine, the measure Equation~\ref{eq:coh} uses to call a token nameable, barely moves on
the same partition. The groups are classes like auxiliaries and copulas,
degree adverbs, clause openers, numbers and time words, whose members are mutually substitutable in a
grammatical position rather than close in a distributional neighborhood.

So the remainder is written by components upstream of the unit, and it separates into categories
once the terms are sorted by where they came from.  The reason a reader misses it is that the basis it is read in is
the wrong one. Table~\ref{tab:gelu-units} is what one unit's set looks like with every term decoded
through the same table; Table~\ref{tab:gelu-units-rot} is the same set with the terms separated by
where they came from.

The same ordering reaches a channel. Equation~\ref{eq:provscore} with the unit's read row replaced
by the channel's value-projection row, and the gain of the norm feeding attention rather than the one
feeding the feed-forward block, traces what drives a channel to the components upstream of it.
Attention at layer $\ell$ reads the pre-attention residual, so summing strictly below $\ell$ excludes
its own layer's attention from its sources. Against the same two nulls the ordering holds on both families, and the controlled figure ---
the excess over same-layer channels --- separates the baseline from everything else across ten models
and seeds: the baseline's three seeds sit well below the five shaped models, which cluster tightly
together. A channel is about twice as traceable as a unit on the baseline and about as traceable on
the rest. The baseline is the hard case here as it is for naming.

Those two figures rest on different footings. A channel's drive
is a linear read of the residual and Equation~\ref{eq:provscore} is a linear sum, so the predictor
matches the target's functional form exactly; a feed-forward unit's response is measured after its
activation function, and the same linear score is fed through a nonlinearity. Attention scores
higher for that reason alone, before anything about legibility enters. The same effect raises its
same-layer null on the baseline, where channels reading the identical state in parallel correlate
with the target for free, which is why the excess over that null rather than the raw figure is the
number quoted.

An attention channel now has incoming edges, and few of them
clear the same two percent of incoming drive that a feed-forward node's sources clear: across the
baseline's gallery pages, most channels carry a read side and only a handful carry an edge into it. A
channel draws on the residual broadly rather than on a handful of writers, which is a different shape
of dependency from the one the feed-forward graph has.

A component, then, carries a name on each side. Close to half the model's components are named
by what they write, read off the predictions they drive, and the share each supplies belongs to the
component rather than to the context. The head of a unit's inputs is named from its weights, decoded
in the frame of its own layer, and where that head ends is set by the activation function. The
remainder --- three quarters of what a unit reads, which the embedding groups into nothing --- is
named by which components supply it, and sorted by where it came from, it is grammar. A channel is
named on the same terms on both sides, and more easily on the read side, because nothing stands
between its row and what it collects.

\begin{table}[htbp]
\centering\small
\caption{The provenance ordering and the five objections it was tested against. Rows above the
deletion block are areas under the ROC curve, where $0.5$ is chance; the sibling and deletion
blocks are differences, where $0$ means the two weightings are indistinguishable. The baseline is the hard case throughout, ruling out siblings only under deletion and passing the depth split only in its embedding half, where the sigmoid model passes every one by more. Context length
2048, 16 contexts, 512 candidate tokens.}
\label{tab:causal}
\input{T_causal}
\end{table}

\secbarrier

%% file: T_naming.tex
\setlength{\tabcolsep}{7pt}
\begin{tabular}{lrrrr}
\toprule
contributors taken & covered & characterized & coherent & named \\
\midrule
top 5 & 42.6\% & 32.1\% & 91.8\% & \textbf{29.5\%} \\
top 10 & 53.4\% & 43.0\% & 90.4\% & \textbf{38.9\%} \\
top 20 & 64.9\% & 54.3\% & 87.9\% & \textbf{47.7\%} \\
top $n_{90}$ & 85.9\% & 57.2\% & 84.3\% & \textbf{48.2\%} \\
\bottomrule
\end{tabular}

%% file: T_readname.tex
\setlength{\tabcolsep}{6pt}
\begin{tabular}{l lrrrrr}
\toprule
model & component & $K{=}8$ & $K{=}16$ & $K{=}32$ & $K{=}64$ & $K{=}128$ \\
\midrule
baseline (GELU) & channel & 0.844 & 0.862 & 0.881 & 0.892 & 0.898 \\
 & unit & 0.598 & 0.531 & 0.435 & 0.279 & \textbf{-0.116} \\
\addlinespace
sigmoid & channel & 0.651 & 0.670 & 0.689 & 0.688 & 0.674 \\
 & unit & 0.743 & 0.746 & 0.748 & 0.740 & 0.680 \\
\addlinespace
ReLU & channel & 0.522 & 0.533 & 0.546 & 0.545 & 0.544 \\
 & unit & 0.513 & 0.465 & 0.388 & 0.282 & 0.139 \\
\addlinespace
\bottomrule
\end{tabular}

%% file: T_attn.tex
\setlength{\tabcolsep}{5pt}
\begin{tabular}{lrrrrrr}
\toprule
& \multicolumn{3}{c}{token share of a channel's read} & \multicolumn{3}{c}{context its layer delivers} \\
\cmidrule(lr){2-4}\cmidrule(lr){5-7}
model & layer 0 & above & layers $>$ FFN & at layer 0 & peak layer & peak \\
\midrule
baseline (GELU) & 1.000 & 0.61--0.98 & 10 of 12 & 0.01 & 1 & 0.54 \\
sigmoid & 1.000 & 0.66--0.96 & 11 of 12 & 0.04 & 1 & 0.94 \\
softplus & 1.000 & 0.63--0.85 & 10 of 12 & 0.12 & 5 & 0.28 \\
ReLU & 1.000 & 0.16--1.00 & 6 of 12 & 0.00 & 3 & 0.81 \\
set operators & 1.000 & 0.49--0.90 & 10 of 12 & 0.07 & 3 & 0.34 \\
unshaped sigmoid & 1.000 & 0.62--0.89 & 11 of 12 & 0.15 & 4 & 0.19 \\
\bottomrule
\end{tabular}

%% file: T_gelu_units_rot.tex
\begin{tabular}{@{}llp{0.70\textwidth}@{}}
\toprule
unit & where from & the tokens it drives on \\
\midrule
A & from the embedding & \texttt{\ what}, \texttt{\ cor}, \texttt{\ where}, \texttt{\ much}, \texttt{\ Ad}, \texttt{\ out}, \texttt{\ at}, \texttt{\ But} \\
 & carried 1--2 layers & \texttt{\ now}, \texttt{\ of}, \texttt{\ on}, \texttt{\ such}, \texttt{\ children}, \texttt{\ who}, \texttt{\ them}, \texttt{\ or} \\
 & carried 3--5 layers & \texttt{\ about}, \texttt{\ here}, \texttt{\ name}, \texttt{\ within}, \texttt{\ how}, \texttt{\ when}, \texttt{\ after}, \texttt{\ part}, \texttt{ines}, \texttt{\ as}, \texttt{\ from}, \texttt{ers}, \texttt{\ before} \\
 & carried 6+ layers & \texttt{\ And}, \texttt{\ Zhang}, \texttt{\ Com}, \texttt{\ so}, \texttt{\ Mun}, \texttt{\ little}, \texttt{You}, \texttt{\ year}, \texttt{\ more}, \texttt{\ would}, \texttt{\ You} \\
\addlinespace
B & from the embedding & \texttt{\ al}, \texttt{\ war}, \texttt{\ intelligence}, \texttt{\ come}, \texttt{\ here}, \texttt{\ Zhang}, \texttt{\ help}, \texttt{request}, \texttt{\ off}, \texttt{\ life}, \texttt{\ now}, \texttt{\ take} \\
 & carried 1--2 layers & \texttt{\ again}, \texttt{\ does}, \texttt{\ Ad}, \texttt{\ did}, \texttt{'t}, \texttt{\ think}, \texttt{\ used}, \texttt{\ do}, \texttt{\ really}, \texttt{\ has}, \texttt{\ current}, \texttt{\ This} \\
 & carried 3--5 layers & \texttt{\ need}, \texttt{\ officials}, \texttt{\ have}, \texttt{\ and}, \texttt{\ or}, \texttt{\ how}, \texttt{\ way}, \texttt{\ never}, \texttt{\ want}, \texttt{'s} \\
 & carried 6+ layers & \texttt{\ death}, \texttt{\ Red}, \texttt{\ name}, \texttt{\ home}, \texttt{\ time}, \texttt{\ by} \\
\addlinespace
C & from the embedding & \texttt{\ Com}, \texttt{\ Pad}, \texttt{\ Ad}, \texttt{\ Zhang}, \texttt{\ He}, \texttt{\ We}, \texttt{\ It}, \texttt{\ As}, \texttt{\ The}, \texttt{\ In}, \texttt{\ Red}, \texttt{\ This}, \texttt{\ Po}, \texttt{\ For}, \texttt{com} \\
 & carried 1--2 layers & \texttt{\ something}, \texttt{\ an}, \texttt{\ issues}, \texttt{\ war}, \texttt{\ any}, \texttt{\ all}, \texttt{\ read} \\
 & carried 3--5 layers & \texttt{\ products}, \texttt{ution}, \texttt{\ same}, \texttt{\ even}, \texttt{\ they}, \texttt{\ one}, \texttt{\ she} \\
 & carried 6+ layers & \texttt{\ Mun}, \texttt{\ They}, \texttt{son}, \texttt{It}, \texttt{\ him}, \texttt{\ he}, \texttt{They}, \texttt{\ reported}, \texttt{ner} \\
\addlinespace
D & from the embedding & \texttt{\ things}, \texttt{\ This}, \texttt{\ there}, \texttt{\ community}, \texttt{\ family} \\
 & carried 1--2 layers & \texttt{\ can}, \texttt{\ may}, \texttt{\ is}, \texttt{\ make}, \texttt{\ could}, \texttt{\ using}, \texttt{\ should}, \texttt{\ the}, \texttt{\ how}, \texttt{context}, \texttt{\ about}, \texttt{\ it}, \texttt{\ are}, \texttt{\ they}, \texttt{\ by} \\
 & carried 3--5 layers & \texttt{\ see}, \texttt{\ does}, \texttt{\ did}, \texttt{\ to}, \texttt{\ then}, \texttt{\ where}, \texttt{\ this}, \texttt{\ these}, \texttt{\ do}, \texttt{\ into}, \texttt{\ will}, \texttt{\ that}, \texttt{\ was}, \texttt{\ which}, \texttt{\ around} \\
\bottomrule
\end{tabular}

%% file: T_causal.tex
\setlength{\tabcolsep}{5pt}
\begin{tabular}{lrr}
\toprule
& baseline (GELU) & sigmoid \\
\midrule
\multicolumn{3}{l}{\emph{Does the ordering work?}} \\
\quad ranks anonymous drivers above the pool & 0.584 & 0.831 \\
\quad ranks nameable drivers above the pool & 0.634 & 0.869 \\
\addlinespace
\multicolumn{3}{l}{\emph{Against three nulls}} \\
\quad permute which writer holds which alignment & chance & chance \\
\quad set every weight to one (no target in the score) & 0.509 & 0.543 \\
\quad substitute units at the target's own layer & chance & chance \\
\addlinespace
\multicolumn{3}{l}{\emph{Against the sibling shape, which the ranking cannot see}} \\
\quad weight by resemblance instead, anonymous & $+0.002$ & $+0.071$ \\
\quad weight by resemblance instead, nameable & $-0.137$ & $+0.135$ \\
\addlinespace
\multicolumn{3}{l}{\emph{Under deletion, which siblings cannot survive}} \\
\quad one writer removed: predicted vs.\ actual damage & $+0.148$ & $+0.231$ \\
\quad \quad the same, permuted across tokens & $+0.011$ & --- \\
\quad 24 writers removed: direct over sibling & $+0.059$ & $+0.335$ \\
\quad \quad sibling weighting alone & $+0.0005$ & $-0.061$ \\
\addlinespace
\multicolumn{3}{l}{\emph{Where the two halves arrive from}} \\
\quad embedding band favors the nameable half & $+0.015$ & $+0.115$ \\
\quad depth $\times$ half interaction & spans zero & $+0.150$ \\
\addlinespace
\multicolumn{3}{l}{\emph{How many seeds}} \\
\quad clear the shuffle on anonymous tokens & 4 of 5 & 5 of 5 \\
\quad clear the same-layer control & 5 of 5 & 5 of 5 \\
\bottomrule
\end{tabular}

%% file: editing.tex
\section{Editing a Model: Say what?}
\label{sec:edit}

Naming a component says what it is responsible for. This section writes to the components so
named, and the edits come in three kinds of increasing reach. The first turns a component up and
asks whether the tokens its name predicts are the ones that move. The second installs an
association the model does not hold into a spare component, key and value both read from the
parameters. The third drives a unit the model trained for itself, from two layers below it, with
nothing installed at all. Between the second and the third sit two questions every edit has to
answer --- how deep to write, and in which frame --- and a circuit built across two layers that
depends on both answers. The section closes by scoring the install against the standard rank-one
editor.

\subsection{Turning a component up}
\label{sec:edit-gain}

The simplest edit follows from a component being identified by what it writes and who reads it:
multiply a component's activation at the position being
predicted by a gain $\alpha$ and run the model again, with nothing else touched and nothing trained.
That is the smallest intervention that still has a predicted consequence --- it changes how loud a
component is without changing what it says, so the scope its parameters name is the same before and
after and the prediction is entirely about where the logits should move.

Three quantities are read together.  The \emph{own scope} is
the set the edited component's write column promotes. The \emph{other scopes} are the sets named by
the other components edited on the same predictions, which controls for an edit that simply raises
whatever it touches. And a size-matched \emph{random} draw from the vocabulary catches an edit that
moves the distribution wholesale. Any edit is respected if pushed hard enough and harmless if pushed
gently, so strength is a variable.  We sweep $\alpha \in \{2, 4, 8\}$, over $360$ edits spanning
eight depths of the baseline.

\begin{table}[htbp]
\centering\small
\caption{An edit installed in one component of the baseline, scored against the scope its own
parameters name and against the scopes of the other components edited in the same predictions, at
matched strength. ``Respects the name'' is the share of edits moving their own scope more than
another component's.}
\label{tab:edit-baseline}
\input{T_edit_baseline}
\end{table}

Table~\ref{tab:edit-baseline} reports the results. The edit moves the tokens the parameters named and leaves
the other components' targets alone in ninety-seven percent of cases, and the effect on the named
scope exceeds the effect elsewhere by more than an order of magnitude.

\begin{figure}[htbp]
\centering
\includegraphics[width=\textwidth]{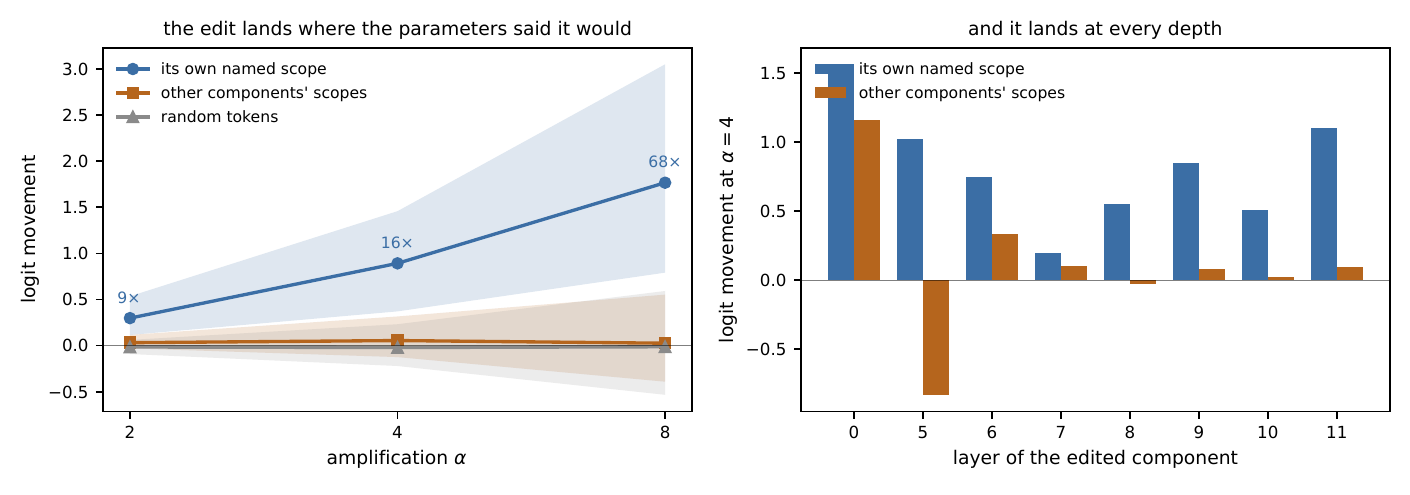}
\caption{Left: what a gain of $\alpha$ does to the three sets, median over $360$ edits with the
interquartile range shaded. The named scope grows roughly in proportion to $\alpha$ while the
collateral stays flat at zero, so the ratio between them \emph{widens} with strength --- ten times at
$\alpha=2$ and sixty-eight at $\alpha=8$. This is the answer to the objection the definition itself
raises: the edit is not merely respected because it was pushed hard, since pushing harder buys no
collateral damage. Right: the same edit by the depth of the component it was installed in, at
$\alpha=4$. It lands at every depth measured, including layer zero.}
\label{fig:edit}
\end{figure}

Figure~\ref{fig:edit} sweeps the strength.  Doubling $\alpha$
roughly doubles the movement on the named scope --- $0.30$, $0.89$, $1.77$ --- while what happens to
every other component's scope stays flat and near zero at all three settings. 

\subsection{Installing an association that was not there}
\label{sec:edit-install}

A gain is not what is usually meant by writing to a model. The model-editing literature installs
associations a model does not hold \citep{meng2022rome}, treating a feed-forward layer as a linear
associative memory and applying a rank-one update: a key for the subject, a value for the target. We
adopt that framing here.

Writing a target token's
unembedding direction into a single unit's output column, in closed form and with no gradient, is
already established --- by \citet{dai2022kn} through token embeddings and by \citet{hakimi2026ove} through
unembedding columns, the latter explicitly a per-unit version of the rank-one mechanism. What no
method we can find does is choose the \emph{site} from the weights. ROME localizes by causal tracing;
the knowledge-neuron line scores neurons by integrated gradients; the closest parameter-side method
still localizes with a logit-lens score against cached activations. Every one of them reads the site
off what the model \emph{does} on some input. Section~\ref{sec:naming} finds what a unit reads, which makes the site available directly.

That is worth having for a reason the same literature supplies. \citet{hase2023localization} report
that where causal tracing localizes a fact predicts almost nothing about where editing it succeeds ---
the tracing effect accounts for a fraction of a percent of the variance in edit success, against the
$94$ percent explained by the choice of layer alone, with the raw correlation slightly negative.
The causal tracing that picks the site, the
expensive step of the standard pipeline, is not doing the work
it appears to do.

Legibility supplies both halves of the update:

\begin{itemize}
\item the \textbf{key} is the layer-native row for token $A$, $\tilde{E}[l,A]$ of
Equation~\ref{eq:etil} centered over the candidate vocabulary --- the state the layer sees when $A$ is
the current token, measured once and fitted to nothing;
\item the \textbf{value} is the readout direction for token $B$, $U_B - \bar{U}$, so that a component
firing on that key lands on $B$'s logit.
\end{itemize}

They are written into a \emph{spare} component, the one whose activation over the corpus is smallest,
so nothing the model was using is destroyed and the cost is bounded by construction. The pair is
drawn at random from frequent tokens, and the control is built in: $B$'s rank where $A$ appears is
several hundred before the edit, so the model cannot be said to have been leaning toward it.

\begin{table}[htbp]
\centering\small
\caption{Installing ``after $A$, say $B$'' into one spare component, over $1{,}176$ trials across
seven token pairs, twelve depths and seven strengths, with the read gain swept separately in
Section~\ref{sec:mono-edit}. Each site is shown at the weakest strength that
works. The edit moves $B$ to the top of the distribution where $A$ occurs and leaves its rank
elsewhere alone; what separates the two sites is the price.}
\label{tab:put}
\input{T_put}
\end{table}

\begin{figure}[htbp]
\centering
\includegraphics[width=\textwidth]{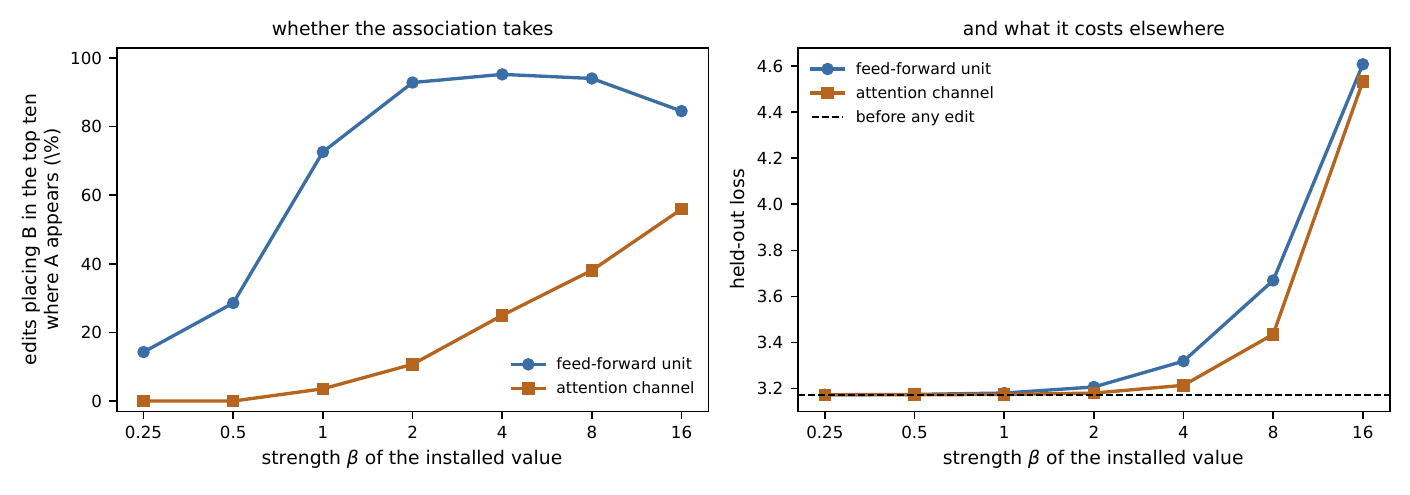}
\caption{Left: how often the edit takes, against how hard it is pushed. A feed-forward unit takes it
on nearly every trial from $\beta=2$ upward, and at every depth through layer eight; an attention
channel needs roughly eight times the strength and succeeds at scattered depths. Right: held-out loss
against the same strength, with the unedited model dashed. The unit's working range costs almost
nothing and the channel's does not.}
\label{fig:put}
\end{figure}

It works, on one component (Figure~\ref{fig:put}). A single spare feed-forward unit at $\beta=1$ moves $B$ from rank
$578$ to rank $1$ at the positions where $A$ occurs, leaves its rank elsewhere where it was ($778$ to
$782$), and costs a quarter of one percent of held-out loss (Table~\ref{tab:put}). Raising $\beta$
buys nothing after that: the association is already first, and the loss climbs, reaching half again
the unedited value by $\beta=16$. Neither key nor value was fitted, traced, or optimized.

\subsection{Editing attention}
\label{sec:edit-attn}

A simple edit to attention --- amplifying a channel and reading the result at the positions
where that channel is actually active --- moves the tokens it promotes on $88$ to $91$ percent of
trials, against $93$ to $94$ percent for a unit, with the collateral on other components'
scopes near zero in both cases; the channel's movement is about a quarter of the unit's at matched
strength. The sign matters here. What a channel gathers
flips sign with context, so its output column demotes exactly the tokens it promotes whenever the
gathered value is negative, and a scope read without that sign scores the wrong end of the list about
half the time --- taken unsigned the same measurement returns chance, $47$ to $49$ percent, with the
scope moving the wrong way.

Installing a new behavior into attention is harder.
An edit counts only if $B$ reaches the top ten where $A$ occurs, is \emph{not} promoted where it
does not, and itself costs under ten percent of held-out loss --- which is what separates an
installed association from a component shouting. On that criterion $10$ percent of channel edits
succeed against $32$ percent of unit edits.

The reason is the same fact that makes a channel's read side legible.
A channel's value is collected from whatever position it attends to, so what arrives at $A$ is an
attention-weighted average rather than the value itself. The rank of $B$ away from $A$ \emph{falls},
from $826$ to $1142$, so the channel demotes the target where it was not installed. What happens
instead is \emph{attenuation} at the place the edit was aimed. Measured over twelve token pairs and
five depths, an installed channel gathers $-0.06$ where $A$ is absent and $+0.38$ where $A$ is the
current token, a contrast a third the size of what a unit delivers. That is why the channel needs
$\beta = 8$ where the unit needs $\beta = 1$, and why it costs a tenth of held-out loss where the unit costs a quarter of a percent. The extra
strength makes up for the averaging, but at a cost.

Which head takes the edit matters. Heads differ in how much of what they deliver is
about the position they are at rather than the positions they looked at, and the difference is large:
put the same read row into one channel of every head at once and the correlation between what a head
gathers and the value at the current position runs from $-0.02$ to $0.75$. In a separate sweep over twelve
pairs, five depths and five strengths, installing in the quietest channel of the layer installs and
stays local on $1$ percent of trials; installing instead in the head that gathers the key most
sharply raises what arrives at the key from $0.30$ to $1.53$, brings the target from rank $574$ to
rank $14$, and takes that share to $8$ percent.

What does succeed is editing the query and key as well. Adding $\kappa\,(k_C \!\cdot\! z_q)(w \!\cdot\! z_p)$ to a head's
score makes it attend to
positions carrying $C$.  But $\kappa$ has no absolute meaning: the term competes with the head's
existing score, whose scale differs by orders of magnitude between heads.
A $\kappa$ chosen once is far too small to move some
heads and saturates others. With $\kappa$ swept
in multiples of the head's own score standard deviation there is a window, and inside it the
installed term and the existing one compose rather than compete: the existing score carries distance,
so their sum expresses \emph{recent and $C$} rather than either alone.
What that buys is a fact conditioned on a token the position does not carry; how much it buys
depends on where the reading component takes its read row, and Section~\ref{sec:place-circuit}
reports the result once the reader has been built.

What
attention writes at a position is a mixture over \emph{other} positions, so a value edit alone
arrives diluted at the reader. A feed-forward unit's output at a position depends on that
position alone, which is what lets an edit there be both strong and local. The model-editing
literature writes to feed-forward layers predominantly, and the work here suggests that is a
reasonable default rather than a necessary one. Attention takes an edit when the pattern is installed
alongside the value and the strength is set in the units the head itself has been trained to.

An attention channel is the more legible of the two on its read side and the worse of the two to
install into; Section~\ref{sec:mono-edit} finds the same on the activation arms.

\subsection{Where to place an edit}
\label{sec:place-depth}

What to write is a question about meaning, and Section~\ref{sec:naming} answers it: a component's
parameters say which tokens turn it on and which it promotes. Three questions about the architecture
remain --- how deep a write has to be made before it arrives, which of several available frames it
should be expressed in, and what has to be true for anything to read it --- and the answer to all
three starts from the recurrence. A transformer carries its state in a stream that every layer adds to:
\begin{equation}
h_{\ell+1} \;=\; h_\ell \;+\; A_\ell\!\left(\mathrm{LN}(h_\ell)\right) \;+\;
F_\ell\!\left(\mathrm{LN}(h_\ell)\right),
\label{eq:stream}
\end{equation}
with $A_\ell$ the attention block at depth $\ell$ and $F_\ell$ the feed-forward block. The addition
in Equation~\ref{eq:stream} is the whole of the architecture's contribution to an edit. Anything
written into $h$ at depth $\ell$ is still in $h$ at every later depth, in the same coordinates,
because nothing in the recurrence moves it. What changes with depth is how much company it has.
Every layer above adds its own contribution on top, and a write made early is a smaller and smaller
share of what the readout finally sees.

Writing a direction into the residual stream and observing the consequence is the mechanism behind
activation steering \citep{turner2023actadd, rimsky2024caa, li2023iti}. The measurement here asks
the prior question: of what is written, how much is still there at the end.

A unit direction $w$ is added to the stream at $10$ percent of positions immediately before layer
$X$, scaled to a tenth of the median residual norm at that depth. Writing $h_L$ for the state
arriving at the final norm and $\Delta = h_L^{\,\text{written}} - h_L^{\,\text{clean}}$ for the
change it causes, two quantities are read at the written positions:
\begin{equation}
\mathrm{survival}(X) \;=\; \cos\!\left(\Delta, w\right), \qquad
\mathrm{spill}(X) \;=\;
\frac{\operatorname{med}_{p \notin W} \lVert \Delta_p \rVert}
     {\operatorname{med}_{p \in W} \lVert \Delta_p \rVert},
\label{eq:survival}
\end{equation}
where $W$ is the set of written positions. Survival asks whether the write still points its own
way. Spill asks how much of it has been carried to positions it was never made at.\footnote{\citet{timkey2021rogue} show that a few dominant coordinates can control a cosine measure. Both quantities here are differences, in which a near-constant coordinate cancels before the cosine is taken, and masking the ten coordinates that carry most of the measured change moves the comparison of Figure~\ref{fig:norotation} by at most $0.012$.}

Four kinds of direction are written. A
\emph{private} direction is drawn at random, so nothing in the model was built to read it. A
\emph{layer-native} direction is a row of the token table $\tilde{E}[X,\cdot]$ of
Equation~\ref{eq:etil}, returned to residual coordinates by dividing out the layer's norm gain. A
\emph{working-state} direction is a leading principal direction of the residual at $X$ with the
readout subspace projected out. A \emph{readout} direction is $U_B - \bar{U}$ for a token $B$.

\begin{figure}[htbp]
\centering
\includegraphics[width=\textwidth]{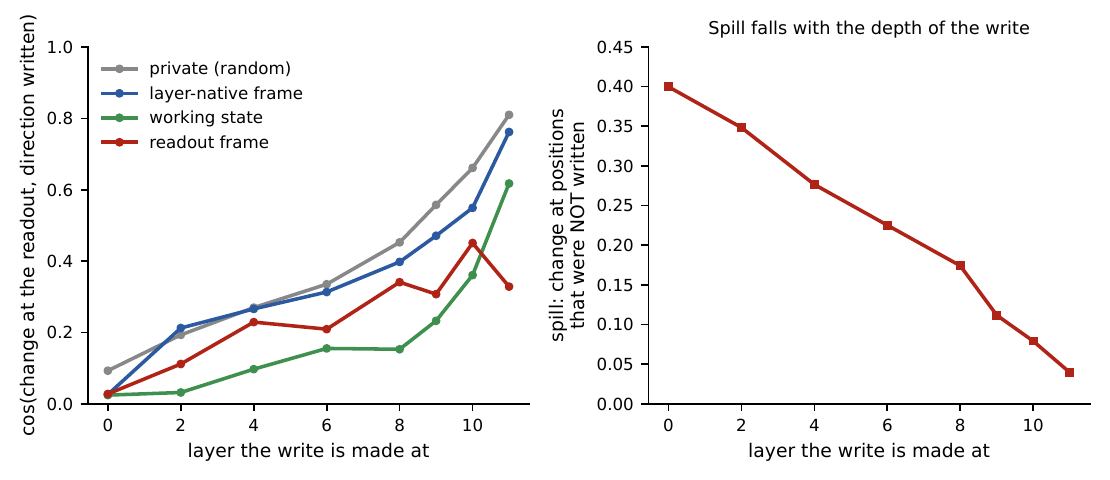}
\caption{Left: how much of a written direction still points its own way at the readout, against the
depth it was written at, for four kinds of direction. Right: spill, the change at positions the
write was never made at, relative to the change where it was. Written at $10$ percent of positions, at a tenth of the median residual norm.}
\label{fig:survival}
\end{figure}

Figure~\ref{fig:survival} reports the result. Depth dominates. A write made before layer two survives at
$0.11$ and a write made before layer eleven at $0.33$ to $0.81$, a spread far larger than any
difference between the four kinds of direction at fixed depth. Spill falls by a factor of ten over
the same range, from $0.40$ at layer zero to $0.04$ at layer eleven. An early write is buried under
what follows it and carried to positions it was never made at. A late write stays where it was put in
both senses, which is why every edit in this paper is made in the last third of the stack.

The private direction is the best preserved of the four at every depth, which is the key to reading
the rest of the panel. Preservation is the default behavior of Equation~\ref{eq:stream}, so the
gap between a family and the private baseline measures how much the model \emph{acted} on it. On
that reading the layer-native direction is barely touched, sitting level with the private baseline.
The working-state direction is the most transformed at seven of the eight depths measured, which is
the rotating stream of \citet{oskin2026offaxis} seen from the inside. The readout direction sits
between them, and its consequence is the only one of the four aimed anywhere in particular.

Writing $P$ for the projection onto the leading $64$
principal directions of the centered unembedding, the share of the response that is visible to the
readout is $\lVert P^{\!\top}\Delta \rVert / \lVert \Delta \rVert$, which for an isotropic direction
in $768$ dimensions is $0.289$. The three non-readout families sit at $0.23$ to $0.30$ at every
depth, which is to say at chance. The readout family rises from $0.27$ to $0.43$, with the excess
growing toward the layers where the prediction is committed. A write in the readout frame is the only one the model ultimately answers.

That is also why the constructions here stop one layer short of the top. Preservation rises with
depth for every family, with one exception: over the last step the readout frame is the only one
that gets \emph{worse}, by $0.122$ on the baseline and $0.086$ on the sigmoid model, where the
layer-native, working-state and private families all improve substantially over the same step. The
last layer reads readout-frame content and answers it, which consumes it, and does nothing of the
kind to the other three. Both models agree. A write placed there to be read by something above it
has nothing above it left, and a write placed there to survive is the one kind the commit layer
spends. Every construction below installs its reader at layer ten of twelve for that reason.

A division of labor follows from the two token frames behaving oppositely --- the layer-native frame
carried by the stream untouched, the readout frame the one every later component responds to --- and
it is the reason a circuit needs more than one component. A write in
the readout frame acts on the output directly, at the price of being the frame every later component
is listening to. A write in any other frame is inert at the output and reaches it only through
something that reads it. A circuit therefore carries its intermediate state in one frame and commits
its answer in the other.

\subsection{What the turn does to a write}
\label{sec:place-transport}

The frame turns from depth to depth, and there are two ways to measure that turn. Only one of them
describes what happens to content already in the stream.
Orthogonal Procrustes between consecutive layer-native token tables gives the map that describes how
each depth's own reading frame is oriented,
\begin{equation}
R_\ell \;=\; \arg\min_{R \in O(d)} \lVert X_\ell R - X_{\ell+1} \rVert_F
\;=\; U V^{\!\top}, \qquad X_\ell^{\!\top} X_{\ell+1} = U \Sigma V^{\!\top},
\label{eq:procrustes}
\end{equation}
and the same fit run between the residual \emph{states} rather than the tables gives a second map
$Q_\ell$. The two differ by a factor of four: the table turns by a median plane angle of $56$ degrees
per step where the state turns by $10$ to $18$. Content sitting in the stream is acted on by the
second of these, and a comparison against the first says nothing about it.

One step is unlike the rest. The state turns by $81$ degrees between layers zero and one, against $10$
to $18$ everywhere above, so the bottom of the stack is a different operation from the body of it.  The
model is stashing away the input token and building the initial workspace.  What follows does not
apply to those early one to two layers.

The difference
of two conditional means is a \emph{difference in differences}.
Capture the reader's input with the marker written and again
without it, and take the change at the marked positions only,
\begin{equation}
\Delta_{h \to r} \;=\;
\operatorname{E}_{p \in W}\!\left[x_r^{\,\text{written}}\right] -
\operatorname{E}_{p \in W}\!\left[x_r^{\,\text{clean}}\right].
\label{eq:did}
\end{equation}
Equation~\ref{eq:did} is compared against the direction as written and against the same direction
carried through the composed state map $\prod_{\ell=h}^{r-1} Q_\ell$. Because that turn is small the
two candidates are themselves close, so a third quantity is needed to isolate what the turn
contributes on its own: remove the written component from both and ask how much of what is left lies
along the turned increment.

\begin{figure}[htbp]
\centering
\includegraphics[width=0.72\textwidth]{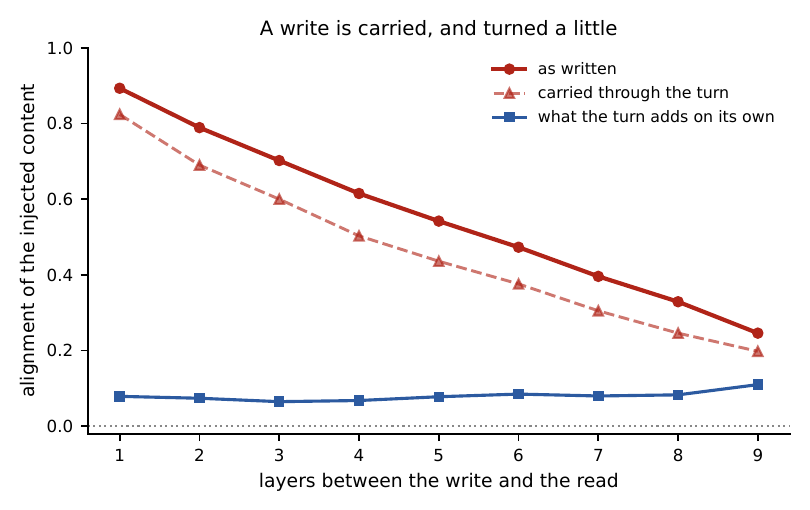}
\caption{A direction written into the residual before layer three, followed to every later depth.
Solid is its alignment with itself, dashed with the same direction carried through the composed
state-level turn, and the lower curve is what the turn contributes once the direct part is removed
from both. Medians over probe directions at ten percent of positions.}
\label{fig:norotation}
\end{figure}

Figure~\ref{fig:norotation} reports the result. A write made before layer three is aligned with itself
at $0.70$ three layers later and with its turned image at $0.60$, and the turn's own contribution is
$0.065$. The ordering holds at every gap: the write is more itself than it is a turned copy of
itself, and the turn adds a flat $0.065$ to $0.11$ on top. Both components are present, as the
recurrence requires, and the direct one is the larger.

What grows with distance is the model's response to the write rather than a turned copy of it. Since
the alignment with the write falls from $0.89$ at one layer to $0.25$ at nine, the share of the
change orthogonal to the write rises from $0.45$ to $0.97$ over the same range. By the far end almost
all of what a reader sees is the response, and no map fitted here predicts its direction.

Two rules follow, and both are used in everything below. A read row is measured at its destination,
as the conditional mean difference of the reader's own input, rather than assumed from the writer's
intent or carried there analytically: the write arrives diluted and wrapped in a response that
nothing predicts, so the only reliable statement about the reader's input is a measurement of it. And
the turn is a correction worth a tenth of the signal rather than the thing to correct for, so a read
row carried to its destination through the turn is at best a tenth better than one that ignores the
turn.  On a thirty-six-layer stack the same ordering holds from layer seven through thirty-one, and the
turned image leads only above a single anomalous twenty-degree step near the top, on one model of
unusual shape.

\subsection{A circuit across two layers}
\label{sec:place-circuit}

An edit that spans layers has to carry its intermediate state in some frame, and the stream between
writer and reader is a communication channel. There are two ways to use it. The first is to write in
the model's own geometry so that components already present read the
result. The price is that the write has to be expressed in a frame that must be estimated, and the
estimate is what the accuracy of the edit then rests on. The second way is available whenever both
ends of the circuit are supplied, and it is the one the construction below uses: carry the
intermediate state on a direction drawn at random, which is a coordinate system chosen rather than
found and therefore known exactly.

A channel of that kind is available because a trained transformer operates in a fraction of its
width, and in the same fraction more or less at every depth. The residual's effective dimensionality, measured as
a participation ratio of the covariance spectrum, is roughly $170$ to $195$ through the middle of the
stack once the small number of very high variance coordinates of \citet{sun2024massive} is set aside.
Pooling the layers a wire has to cross barely raises it: over layers eight to eleven the union
measures $183$ dimensions where the largest single layer in that span measures about $175$, and four
disjoint subspaces of that size would need some six hundred. The layers share one working subspace
rather than each occupying its own, so a direction drawn once is untouched at every depth it passes
through, and about five hundred and eighty of the model's seven hundred and sixty-eight dimensions
are free for the purpose.

\begin{figure}[htbp]
\centering
\includegraphics[width=0.58\textwidth]{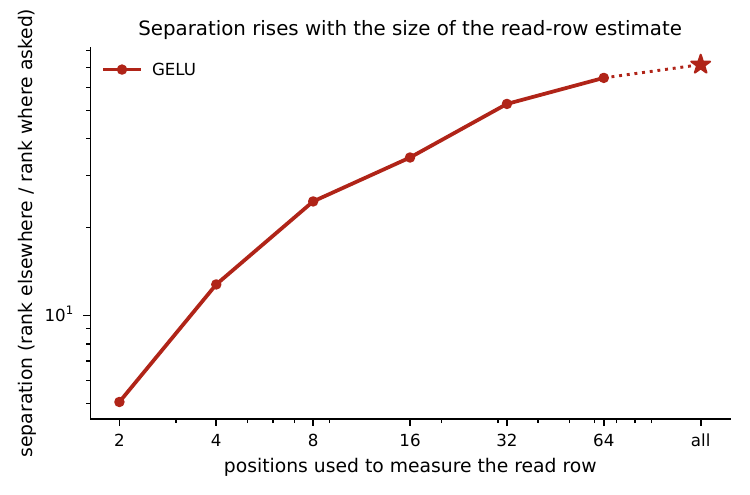}
\caption{Separation achieved by the cross-layer circuit, against the number of positions used to form
the measured read row, with every other setting held fixed.}
\label{fig:estimation}
\end{figure}

Being private buys knowledge of the carrier rather than separation from the model. A random direction
still places about $\sqrt{175/768} \approx 0.48$ of its length inside the space the model uses, and
that overlap costs little: holding it and the norm fixed and varying only how many active directions
carry it, from all of them down to one, leaves held-out loss flat, with the cost instead growing
roughly quadratically in the size of the write. Nor does the carrier's frame change what the circuit
achieves. Repeating the construction with the marker drawn from the readout frame, from the layer's
own token frame, and from the working state leaves the target at median rank $12$ and the separation
at $71.6$ in every case, because the reader's row is measured where it will be used and a measured
conditional mean isolates whatever was written there. What a private channel saves is the estimate,
which is what the construction turns out to be limited by (Figure~\ref{fig:estimation}).

Circuits of this shape occur in trained models without being put there. \citet{neo2024context}
report attention heads that recognize a context and activate a downstream token-predicting unit
accordingly, which is the found version of what follows. Both halves of the division of labor appear
in a single construction. An attention head at depth $h$ is installed to fire where a chosen token
occurs and to write a private direction. A feed-forward unit at depth $r > h$ is installed to read that direction
and to write a target token's readout direction. Neither component existed before, and the model
supplies only the stream between them. 

The unit's read row is the conditional mean of Equation~\ref{eq:did}, measured at its own layer,
and a component installed this way needs a silent state at the positions the edit is not aimed at. These
models carry no bias anywhere, so one is built from the near-constant direction of the read space.
That direction is closely related to the massive activations of \citet{sun2024massive}: a small
number of coordinates that are large and nearly input-independent, and that serve as an implicit
bias the architecture never declares. With $\bar{x}$ the mean
input to the reading layer, $u = \bar{x} / (\bar{x} \cdot \bar{x})$ satisfies $u \cdot x \approx 1$
across positions, and subtracting $t\,u$ from a read row subtracts $t$ from its pre-activation:
\begin{equation}
\hat{r} \;=\; r \;-\; t\,u, \qquad
t \;=\; Q_{q}\!\left(\left\{\,r \cdot x_p \;:\; p \notin W \right\}\right),
\label{eq:threshold}
\end{equation}
with the threshold taken as a quantile of the row's own projection over positions the edit is not
aimed at, so it is measured rather than tuned against the target.

\begin{figure}[htbp]
\centering
\includegraphics[width=\textwidth]{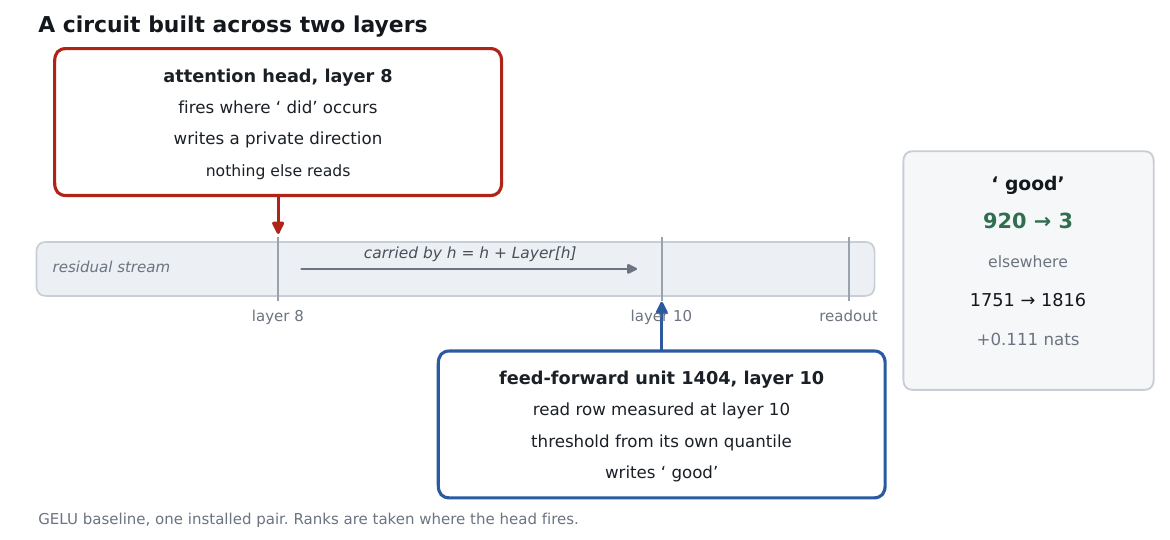}
\caption{One installed pair, end to end. The head marks the positions, the stream carries the marker
two layers without turning it, and the unit answers there. Ranks are for the target token at the
positions the head fires on and at all other positions.}
\label{fig:circuitex}
\end{figure}

Figure~\ref{fig:circuitex} shows an instance. A head at layer eight fires where \texttt{'\,did'}
occurs; a unit at layer ten answers \texttt{'\,good'}, which moves from rank $920$ to rank $3$ where
the head fires while its rank elsewhere goes from $1751$ to $1816$. The cost is $0.111$ nats of
held-out loss.

The same construction answers a condition carried from elsewhere in the sequence, which is the
installation deferred from Section~\ref{sec:edit-attn}. Over $24$ triples and $2{,}400$ trials on
the baseline, with the correction of Section~\ref{sec:mono-comp} applied, the target reaches median rank $29$ where $C$ has occurred against $4{,}970$
elsewhere, a separation of $171$. On $96$ percent of triples some setting brings the target into the
top ten there, and on $54$ percent it does so inside the ten percent loss budget. Every triple that
clears the budget also leaves the target's rank elsewhere alone, and the cheapest setting that does
so costs $0.029$ nats.

Reporting a construction like this at a single operating point hides the shape of what it buys, so
the frontier is the object to report: for each loss budget, the best rank any setting achieves
within it. Over the full cross of write strength, read gain, saturation and read-row calibration,
the baseline brings the target into the top ten for $0.121$ nats, and for $0.179$ once the
specificity criterion is imposed (Figure~\ref{fig:circuitcost}). Section~\ref{sec:mono-edit} reports the same frontier for the
activation arms.

If the circuit is limited by the model, the amount of data used to measure the read row should barely
matter; if by the estimate, a great deal.

Figure~\ref{fig:estimation} answers it. Separation rises from $5.1$ at two positions to $71.6$ at the
full sample, monotonically and without turning over at the largest sample available. The binding constraint on a circuit installed this way
is the accuracy of one's estimate of the destination frame, which is a measurement that can be improved rather than a
property of the model that cannot.

\subsection{A reader is what makes off-axis content visible}
\label{sec:place-reader}

The claim that a write outside the readout frame is inert until something reads it can be put to a
single test, with both halves and their controls in one measurement. A direction is written at layer
eight at $10$ percent of positions. An unrelated target token is then scored three ways: with the
write alone, with a reading unit at layer ten alone, and with both.

\begin{figure}[htbp]
\centering
\includegraphics[width=0.78\textwidth]{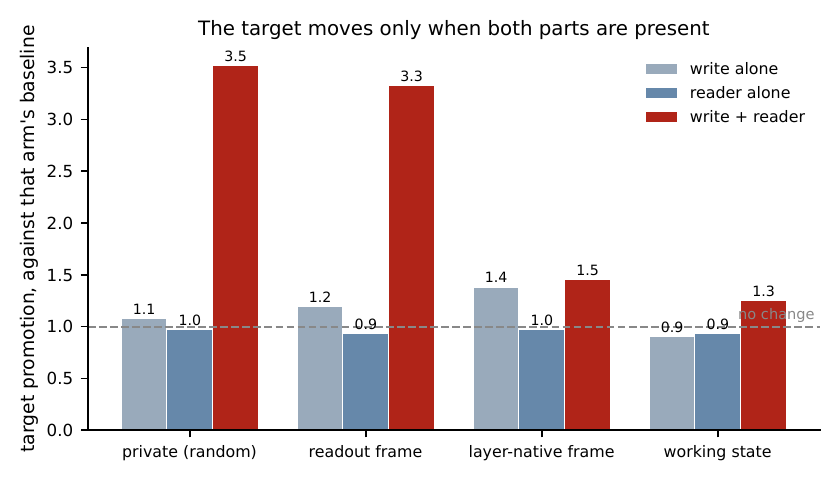}
\caption{Promotion of an unrelated target token at the written positions, relative to each arm's own
unedited baseline, for four kinds of written direction. The reading unit at layer ten has a measured,
thresholded read row and writes the target's readout direction.}
\label{fig:reader}
\end{figure}

Figure~\ref{fig:reader} depicts the result. The reading unit alone leaves the target where it was, at $1.0$
for every kind of write, which is the threshold of Equation~\ref{eq:threshold} doing its work. The
write alone moves the target by $1.1$ times. The two together move it by $3.5$, for $0.13$ nats. The
largest effect belongs to the private direction, and the two model-derived directions give the
least, which is consistent with the reader having to find a frame the model is also acting on.

\subsection{Driving a component the model already has}
\label{sec:place-tap}

Everything to this point installs both ends of a circuit. A unit the model trained can be driven the
same way, with nothing installed, through the row it already reads with. A unit's pre-activation at depth $\ell$ is $a_u = W_a[u] \cdot \left(g \odot \mathrm{LN}(h)\right)$,
with $g$ the layer's norm gain. To first order in a perturbation of the stream,
\begin{equation}
\frac{\partial a_u}{\partial h} \;\propto\; \Pi_{\perp \mathbf{1}}\!\left(W_a[u] \odot g\right),
\label{eq:drive}
\end{equation}
the read row multiplied elementwise by the gain and projected off the constant direction, since
layer normalization removes the mean and a component along $\mathbf{1}$ adds to $\lVert h \rVert$
without reaching the unit.\footnote{Both details matter in practice. Dividing by the gain rather
than multiplying, or retaining the constant component, produces a direction that suppresses the unit
it was meant to excite.} Units are selected by the margin between the top two tokens their write
column decodes to, which picks out components that answer with a single token.

Whether the write reaches the target through the unit is a mediation question. Writing $r_0$ and
$r$ for the target's rank before and after the write in the intact model, and $r_0^{\text{abl}}$ and
$r^{\text{abl}}$ for the same pair with the unit's write column zeroed,
\begin{equation}
m \;=\; 1 \;-\;
\frac{r_0^{\text{abl}} - r^{\text{abl}}}{r_0 - r}
\label{eq:mediation}
\end{equation}
is the share of the promotion that the tapped unit carries. Comparing the ablated arm against the
intact model's baseline rather than its own inflates $m$ without bound, because ablating a unit that
normally promotes the target moves that baseline.

\begin{figure}[htbp]
\centering
\includegraphics[width=\textwidth]{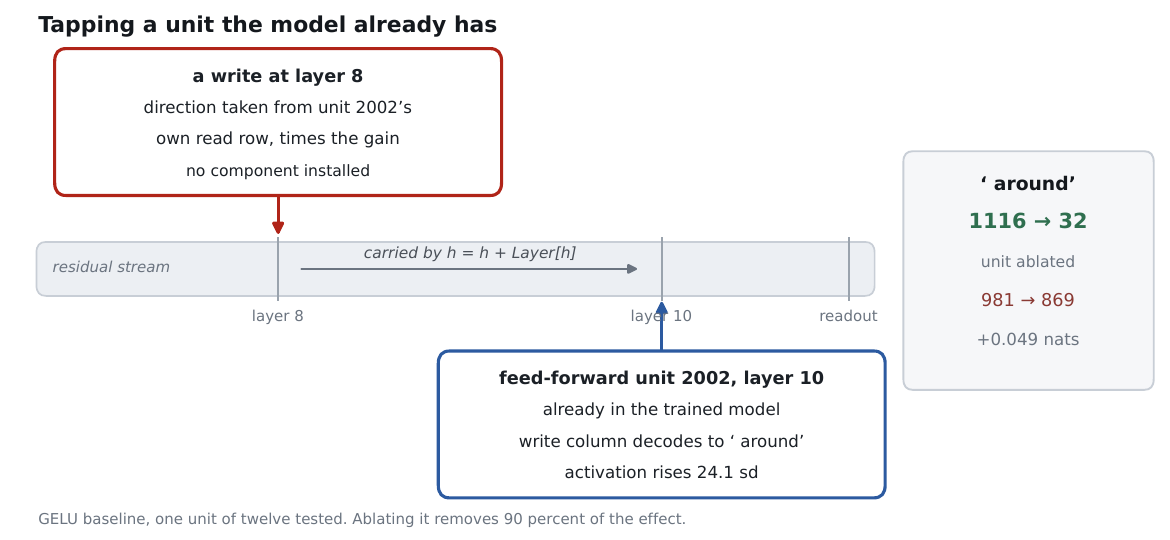}
\caption{An instance of the tap. The write at layer eight uses the layer-ten unit's own read row as
its direction, and no component is installed anywhere. Ablating that single unit removes ninety
percent of the target's promotion.}
\label{fig:tapex}
\end{figure}

\begin{figure}[htbp]
\centering
\includegraphics[width=0.58\textwidth]{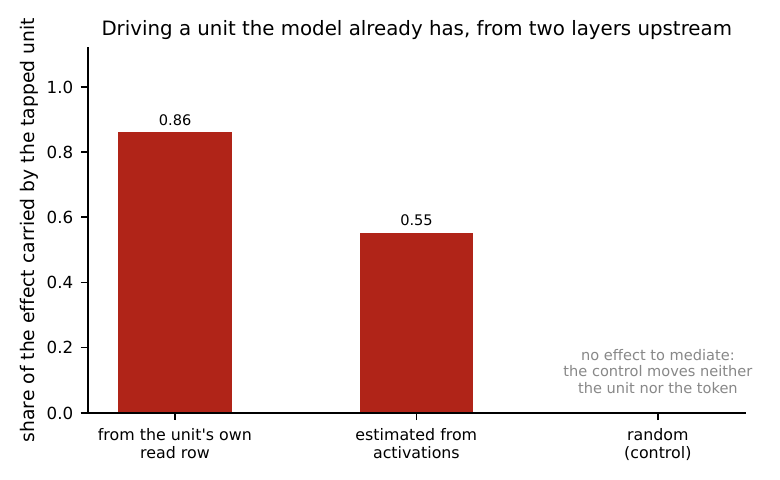}
\caption{Share of the target's promotion carried by the tapped unit, by how the driving direction was
obtained, over twelve units per model at four write strengths. The random control moves neither the
unit nor the target, so it has no effect to mediate.}
\label{fig:tap}
\end{figure}

Figures~\ref{fig:tapex} and \ref{fig:tap} report the result. A write at layer eight raises a
layer-ten unit's activation by $24.1$ standard deviations and moves the token that unit writes from
rank $1116$ to rank $32$, for $0.049$ nats. Ablating that one unit takes the same write from $981$ to
$869$, so ninety percent of the promotion travels through it. Over twelve units the figure is $0.86$.

What makes this work is driving the unit in the frame it already reads in. A direction taken from
the weights through Equation~\ref{eq:drive} gives $0.86$, where one estimated from the unit's
activations gives $0.55$. A native unit's read row is available
exactly, in $W_a$, and estimating it from activations only adds noise.

\subsection{Scoring against a rank-one editor}
\label{sec:edit-rome}

We implemented the rank-one update of \citet{meng2022rome} and
scored it under the criterion used above. Writing $k_*$ for the key, $v_*$ for the value and
$C = \operatorname{E}\!\left[k k^{\!\top}\right]$ for the second moment of the layer's keys over a
corpus sample, that update is
\begin{equation}
\hat{W} \;=\; W \;+\;
\frac{\left(v_* - W k_*\right)\left(C^{-1}k_*\right)^{\!\top}}
     {\left(C^{-1}k_*\right)^{\!\top} k_*}\,,
\label{eq:rome}
\end{equation}
which spreads the write across the layer along $C^{-1}k_*$ rather than loading it into one unit. The
value $v_*$ is obtained by gradient descent on the target's likelihood.

Equation~\ref{eq:rome} has two ingredients, and separating them is informative. The covariance term
is what preserves specificity: replacing $C^{-1}$ by the identity brings the target to rank five on
the baseline while promoting it seventy times as far everywhere else. The gradient is the expensive
one. Substituting a closed-form value, the target's own readout direction, in place of the optimized
$v_*$ brings the target into the top ten for $0.068$ nats, where the gradient version reaches rank
one for $0.611$. The part of the method that carries its specificity is the fast part to compute.

\begin{figure}[htbp]
\centering
\includegraphics[width=0.66\textwidth]{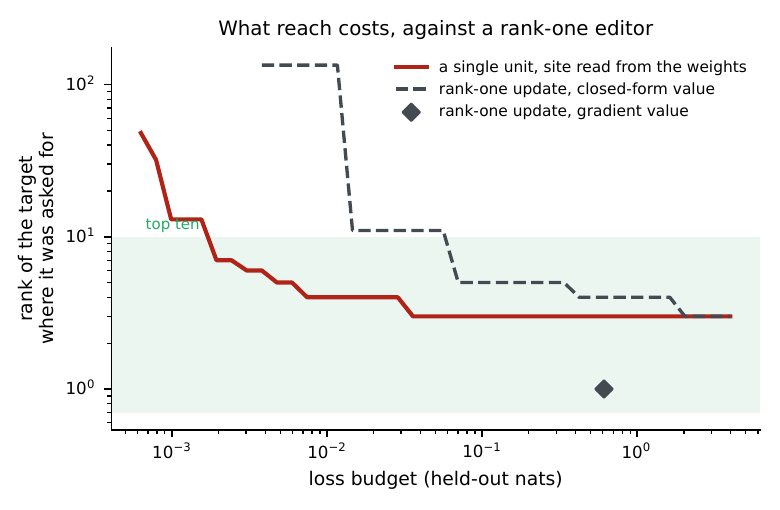}
\caption{Reach against held-out loss for the install of Section~\ref{sec:edit-install} and for the
rank-one update of Equation~\ref{eq:rome}, both at layer ten and both swept over their own strength.
The dashed curve substitutes a closed-form value for the gradient-optimized one; the marker is the
method with its gradient value. Specificity is unconstrained in these panels and reported in the
text.}
\label{fig:rome}
\end{figure}

Figure~\ref{fig:rome} places the two on the same axes, with each swept over its own strength so the
comparison is curve against curve. On the baseline, reaching the top ten costs the single-unit
install $0.0016$ nats against $0.0681$ for the closed-form rank-one update, a factor of forty, and
the target's rank elsewhere is $1.02$ times where it started against $0.81$. The published method,
with its gradient value, reaches rank one for $0.611$ nats and holds specificity at $0.80$.
Section~\ref{sec:mono-edit} runs the same comparison across the activation arms.

Two differences of kind survive the comparison, and they run in opposite directions. A rank-one
update needs a corpus sample to estimate $C$ and a gradient pass to obtain $v_*$, where both halves
of the install here are read from the parameters. Against that, the rank-one update writes through
the whole layer and therefore needs no spare component, where an install into a single unit consumes
one. The measurement supports treating them as complementary. The covariance term is worth adopting on
its own terms wherever a single unit cannot be silenced cheaply enough for a one-unit install to stay
local, which is the situation the shaped activations of Section~\ref{sec:mono} present.

A name, then, is an address that can be written to. A gain on one component moves the tokens its
parameters name and nothing else, and pushing harder buys no collateral. An association the model
does not hold installs into one spare unit, key and value both read from the parameters, for a
quarter of a percent of held-out loss. A unit the model trained for itself is driven from two layers
below through the row it already reads with, and carries eighty-six percent of the effect. A circuit
across two layers runs on a direction drawn at random, which the stream carries untouched because this model works in a quarter of its width, and what limits the circuit is the estimate of the
reader's frame rather than anything in the model. Against the rank-one editor the install reaches
the top ten for a fortieth of the cost, and the covariance term is the part of that method worth
keeping. A rank-one update replaces this with that. A circuit is a mechanism, and the components
here are read, built and driven in the model's own terms.

\secbarrier

%% file: T_edit_baseline.tex
\begin{tabular}{lrrrr}
\toprule
model & own scope & other scopes & random tokens & respects the name \\
\midrule
baseline (GELU) & 0.89 & 0.05 & -0.02 & 97\% \\
\bottomrule
\end{tabular}

%% file: T_put.tex
\begin{tabular}{lrrrrr}
\toprule
& & \multicolumn{2}{c}{rank of B where A appears} & \multicolumn{2}{c}{rank of B elsewhere} \\
\cmidrule(lr){3-4}\cmidrule(lr){5-6}
site & held-out loss & before & after & before & after \\
\midrule
feed-forward unit ($\beta{=}1$) & 3.181 & 578 & \textbf{1} & 778 & 782 \\
attention channel ($\beta{=}8$) & 3.501 & 108 & \textbf{1} & 826 & 1142 \\
no edit & 3.173 & --- & --- & --- & --- \\
\bottomrule
\end{tabular}

%% file: activation.tex
\section{What the activation function changes}
\label{sec:mono}

Sections~\ref{sec:circuit} to \ref{sec:edit} took a conventional transformer as it comes. This
section asks what changes if the activation function is chosen for legibility rather than inherited.
Four arms are trained beside the baseline, matched to it on everything but the unit: sigmoid,
softplus, ReLU, and a fourth built from bounded set operators with two operands per unit. The
diagnosis comes first, on the baseline's own pre-activations, since it says which functions to try.
Then the arms: their quality, the size of their circuits at one hop and at closure, how well their
parameters read, which property of the function was responsible, and what each will and will not
let an edit do.

\subsection{Why a conventional unit stops being readable}
\label{sec:naming-readout}

Section~\ref{sec:naming} left a diagnosis, and the intervention follows from it.
The reading procedure orders tokens by $w_u \cdot \tilde{E}[l,t]$, which estimates the ordering by
\emph{pre-activation}. The causal ground truth of Equation~\ref{eq:causal} orders by activation
\emph{averaged over contexts}. Between the two sit the activation function and an average, and either
can reorder. Take a trained baseline, capture its pre-activations on the same contexts and candidate
tokens, and hold the model, the weights, the tokens and the predictor fixed while changing only the
function applied before the average.

Every order-preserving choice reads at $96$ to $99$ percent and is flat in $K$
(Table~\ref{tab:readout}). GELU reads at $67$ and decays. The ordering the weights supply is very
nearly recoverable from the averaged behavior, unless the function in between is GELU.

\begin{figure}[htbp]
\centering
\includegraphics[width=0.52\textwidth]{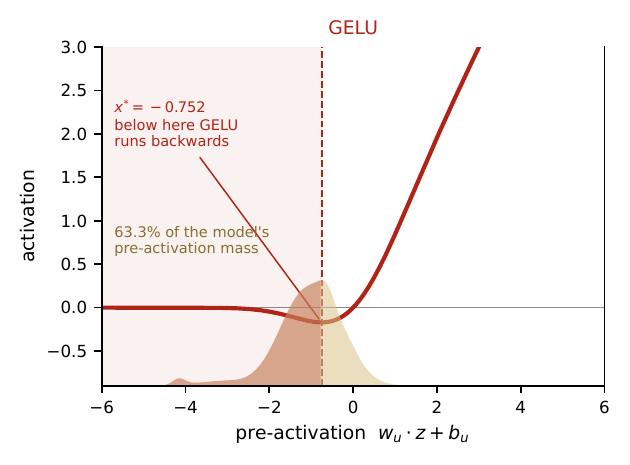}
\caption{Where the baseline actually operates on its activation function. The curve is GELU; the
shaded density under it is the \emph{measured} distribution of pre-activations $w_u \cdot z$
over a trained baseline, $1.2 \times 10^{9}$ samples at context length 2048. GELU decreases below
$x^{*} \approx -0.752$, and $63.3$ percent of the mass lies there on this denser estimate, against
the $59.8$ to $62.0$ percent the per-seed estimate in the text gives, so for most inputs a larger
pre-activation produces a smaller activation and the ordering the weights supply is reversed.
Section~\ref{sec:mono-model} puts a monotone function beside this one.}
\label{fig:dip}
\end{figure}

The cause is visible in the function itself. GELU decreases below $x^{*} \approx -0.752$: on that
interval a \emph{larger} $w_u \cdot z$ produces a \emph{smaller} activation, so the ordering the
weights supply is not blurred there but reversed. Across three converged seeds, $59.8$ to $62.0$
percent of all pre-activation mass sits below $x^{*}$ (Figure~\ref{fig:dip}). Nearly two thirds of
the time, a unit is operating on the part of its activation function that runs backward.

That explains the shape of Figure~\ref{fig:agreementk} as well as its level. Small $K$ selects the
strongest inputs, which are the positive excursions, where GELU is increasing and reads at $67$
percent. Large $K$ reaches the body of the set, where most of the mass lives, which is the reversed
interval, and that is exactly where agreement crosses below chance. A model that looked like it had a
few nameable inputs and an incoherent remainder is reproduced, on the same model, by the readout
function alone.

This is an ablation of the reading rather than a prediction about training: a network trained with a different $\phi$ arrives at a different distribution of
pre-activations, and nothing here guarantees the property survives. And the predictor is the
pre-activation mean rather than $w_u \cdot \tilde{E}$, which isolates $\phi$ cleanly at the cost of
putting these numbers on a different scale from the rest. Section~\ref{sec:mono-model} trains the
monotone model this predicts and reports what happens.
\begin{table}[htbp]
\centering
\small
\caption{One trained baseline model, one set of pre-activations, one predictor. Only the function
applied before averaging over contexts differs. Chance-corrected agreement, three converged seeds,
context length 2048.}
\label{tab:readout}
\begin{tabular}{@{}lllrrr@{}}
\toprule
applied before averaging & & & $K{=}8$ & $K{=}32$ & $K{=}128$ \\
\midrule
GELU     & not monotone & & 67.9, 67.7, 66.6 & 55.0, 54.9, 55.2 & 30.1, 29.7, 31.7 \\
softplus & monotone & unbounded & 96.4, 96.4, 96.8 & 96.9, 96.9, 97.3 & 97.2, 97.2, 97.6 \\
sigmoid  & monotone & bounded   & 98.7, 98.8, 98.9 & 98.8, 98.8, 99.0 & 98.6, 98.6, 98.8 \\
rank     & order only & & 89.7, 89.9, 90.8 & 94.1, 94.2, 94.7 & 97.3, 97.4, 97.6 \\
\bottomrule
\end{tabular}
\end{table}

Neither observation proves the activation is responsible. Changing it and holding everything else
fixed is how to find out, and Section~\ref{sec:stock} later supplies an independent check: among
twelve models trained by other people, the one built on ReLU has by far the least cancellation
between opposing contributions.

\subsection{Which activations make sense}
\label{sec:mono-which}

For the ordering the parameters supply to survive into the ordering the causal measurement sees,
$\phi$ must at least preserve order: if $\phi$ is strictly increasing, then within any single context
the two rankings are identical, and the only remaining discrepancy comes from averaging. GELU is not
strictly increasing, and most of a trained model's pre-activation mass sits where it decreases.

Two standard functions satisfy the requirement and sit on opposite sides of a second property that
might plausibly matter. The logistic function
\begin{equation}
  \phi_{\sigma}(z) = \frac{1}{1 + e^{-z}} \in (0,1)
  \label{eq:sigmoid}
\end{equation}
is strictly increasing and \emph{bounded}, so a unit's mean activation on a token is the fraction of
contexts in which it fired, and no single context can dominate. Softplus
\begin{equation}
  \phi_{+}(z) = \log\!\big(1 + e^{z}\big) \in (0,\infty)
  \label{eq:softplus}
\end{equation}
is strictly increasing and \emph{unbounded}, and behaves like the identity for large $z$. Both are
injective. If order-preservation is what matters they should behave alike; if boundedness is what
matters they should not.

A third, ReLU, is monotone but not injective: it maps the entire negative half-line
to zero, so all the tokens a unit is silent on become indistinguishable in its output. It is the control that separates \emph{order-preserving} from \emph{information-preserving}.

A fourth alternative arrives from a different direction and meets the requirement twice over. Three
earlier papers built feed-forward layers out of bounded fuzzy set operators --- $A \cap B$ and
$A \setminus B$ on operands squashed into $[0,1]$ --- on the argument that a unit computing a named
logical operation on two named sets is legible by
construction~\citep{oskin2026legible,oskin2026selectivity,oskin2026ncffn}. Every unit computes a pair
of operands from the residual stream,
\begin{equation}
  A_u(z) = \sigma\big(w^{a}_u \cdot z\big), \qquad
  B_u(z) = \sigma\big(w^{b}_u \cdot z\big),
  \label{eq:ab}
\end{equation}
and writes the two operations into separate columns,
\begin{equation}
  \mathrm{FFN}(z) = \sum_u \Big[ \underbrace{A_u B_u}_{A \cap B}\, c^{\cap}_u
                    \;+\; \underbrace{A_u (1 - B_u)}_{A \setminus B}\, c^{\setminus}_u \Big]
  \;=\; \sum_u A_u \Big[\, c^{\setminus}_u + B_u \big(c^{\cap}_u - c^{\setminus}_u\big) \Big].
  \label{eq:boolffn}
\end{equation}
Both operands are bounded and strictly increasing in their pre-activations. The rearrangement on the
right says what the two are for: $A$ is a gate deciding \emph{how much} the unit writes and $B$ is a
selector deciding \emph{which} of two fixed columns is written. The parameter budget is held to the
baseline's by narrowing the layer, since each unit now carries two read rows and two write columns.

Those papers could not run the construction everywhere. Below roughly half conversion, training
was indistinguishable from a conventional model; above it training diverged.
The arm reported here has no conventional units left in it,
which is what makes it a clean fourth point rather than a mixture of two designs. What removed the ceiling is an auxiliary objective borrowed from separate work on where a
transformer holds its intermediate content~\citep{oskin2026offaxis}: it prescribes, layer by layer,
the angle between a layer's residual decode and the final prediction --- orthogonal through the first
half of the stack, aligned through the second. Why it works is not explored here; that it does is
eight seeds of eight converging (Section~\ref{sec:mono-quality}). Appendix~\ref{sec:setmodel} gives
the construction, the objective and the recipe in full.

\subsection{The construction}
\label{sec:mono-model}

Everything in the baseline of Section~\ref{sec:circuit-model} is retained --- depth, width, head count, parameter
budget, data, schedule --- and the unit is
\begin{equation}
  a_u(z) = \phi\big(w_u \cdot z\big), \qquad
  \mathrm{FFN}(z) = \sum_{u} a_u(z)\, c_u ,
  \label{eq:monounit}
\end{equation}
with $\phi$ one of Equations~\ref{eq:sigmoid} or~\ref{eq:softplus}, or ReLU; the set-operator arm
replaces Equation~\ref{eq:monounit} by Equation~\ref{eq:boolffn} and is otherwise identical. Read
row, write column and the reading procedures of Sections~\ref{sec:naming-vocab} and~\ref{sec:naming-answer} are unchanged, which
is the point: the same instrument is pointed at a model differing only in what a unit does with its
pre-activation.

Two further departures from the baseline are part of the construction. The attention heads pass
their value vectors through a sigmoid, so what a head gathers from a position is a bounded quantity
in $[0,1]$ rather than an unbounded projection. The weighting over positions is the ordinary
softmax, unchanged from the baseline. That squashing is the same in the sigmoid, softplus and ReLU arms
whatever $\phi$ is, so it is held fixed across the comparison rather than varying with the activation
function; the set-operator arm bounds what its heads gather too, by applying
Equation~\ref{eq:boolffn}'s operations there as well. And a small auxiliary term rewards a bounded activation for sitting near either end of its range
rather than in the middle, at weight $0.003$ against the language-modeling loss
(Appendix~\ref{sec:setmodel} defines it). That term is defined only for a bounded activation, so the
softplus and ReLU arms carry it at zero and the sigmoid arm is trained both ways; the two sigmoid
variants are equivalent on every measurement here, so the cross-activation comparison is not
confounded by it. Section~\ref{sec:mono-what} separates the contribution of these two, neither of
which is an activation function, from $\phi$'s.

\subsection{Quality}
\label{sec:mono-quality}

Eight seeds of the sigmoid construction were trained under Section~\ref{sec:circuit-seeds}'s
protocol and screened by the same criterion, and they behave the same way the baseline did: five
converged and three did not. Whatever makes a run of this size fail at this budget is not something the activation function introduced or removed.

\begin{table}[htbp]
\centering
\small
\caption{The sigmoid construction, eight seeds at one epoch, screened by the criterion of
Section~\ref{sec:circuit-seeds} and reported the same way as Table~\ref{tab:baseline-seeds}. The
non-converged seeds are printed for the same reason they were there. $\mathrm{ov}@8$ here is taken under the ranked summary of Appendix~\ref{sec:pit-convention}, raw rather than chance-corrected.}
\label{tab:mono-seeds}
\begin{tabular}{@{}llrrrr@{}}
\toprule
seed & status & val.\ ppl & LAMBADA & $\mathrm{ov}@8$ & $\mathrm{PR}$ (L0--4) \\
\midrule
1 & converged & 19.30 & 0.258 & 66.6 & 112.7 \\
2 & converged & 19.33 & 0.274 & 67.0 & 112.7 \\
3 & converged & 19.35 & 0.271 & 68.7 & 113.9 \\
4 & converged & 19.36 & 0.260 & 68.2 & 109.5 \\
5 & converged & 19.40 & 0.243 & 71.1 & 112.8 \\
\midrule
6 & excluded  & 20.55 & 0.249 & 73.4 & \phantom{0}97.4 \\
7 & excluded  & 21.41 & 0.184 & 75.6 & \phantom{0}65.3 \\
8 & excluded  & 22.16 & 0.218 & 79.3 & \phantom{0}42.3 \\
\bottomrule
\end{tabular}
\end{table}

\begin{table}[htbp]
\centering
\small
\caption{Quality across activation functions at one epoch. The three arms whose operands are
strictly increasing sit inside the baseline's converged range on both measures. ReLU sits outside it on both.
Every row below the rule comes from an arm matched to the baseline on width, data and schedule.}
\label{tab:mono-quality}
\begin{tabular}{@{}llrr@{}}
\toprule
activation & property & val.\ ppl & LAMBADA \\
\midrule
GELU (baseline, converged) & not monotone & 19.00 -- 19.65 & 0.248 -- 0.274 \\
GELU (baseline, excluded)  & not monotone & 21.62 -- 39.51 & 0.152 -- 0.214 \\
\midrule
sigmoid   & monotone, injective, bounded   & 19.30 -- 19.40 & 0.243 -- 0.274 \\
softplus  & monotone, injective, unbounded & 19.28 -- 19.78 & 0.224 -- 0.269 \\
set operators & bounded, two operands per unit & 19.23 -- 19.43 & 0.234 -- 0.274 \\
ReLU      & monotone, \emph{not} injective & 21.01 -- 22.23 & 0.208 -- 0.219 \\
\bottomrule
\end{tabular}
\end{table}

Table~\ref{tab:mono-quality} settles quality. Both injective activations sit inside the
baseline's converged range on perplexity and on LAMBADA, and softplus was trained at eight seeds of
which all eight pass the screen.

The set-operator arm sits there too, at eight seeds of which all eight converged: $19.23$ to $19.43$
in perplexity, inside the baseline's range and tighter than it, and $0.234$ to $0.274$ on LAMBADA.
Eight of eight landing within $0.2$ nats of each other is not the signature of a model training at
the edge of stability, which is the state prior work left off at \citep{oskin2026legible}.

While the ranges overlap, the medians do not coincide.
The baseline leads sigmoid and softplus by about a point of accuracy, in a consistent direction, on
samples of five, five and eight seeds whose ranges span two to four points each. A real cost of that
size cannot be ruled out at this sample. What the experiment does establish is that replacing the
activation function costs nothing like what ReLU costs.

ReLU is not like other models. Its seeds sit about ten percent behind every other arm on perplexity and below every
converged seed of every other construction on accuracy. It trains stably to a ceiling and settles lower. The one function in this comparison that discards information rather than
merely reshaping it is the one that cannot reach the quality the others reach, which is the first
indication that injectivity is doing work beyond bookkeeping.

\begin{figure}[htbp]
\centering
\includegraphics[width=0.86\textwidth]{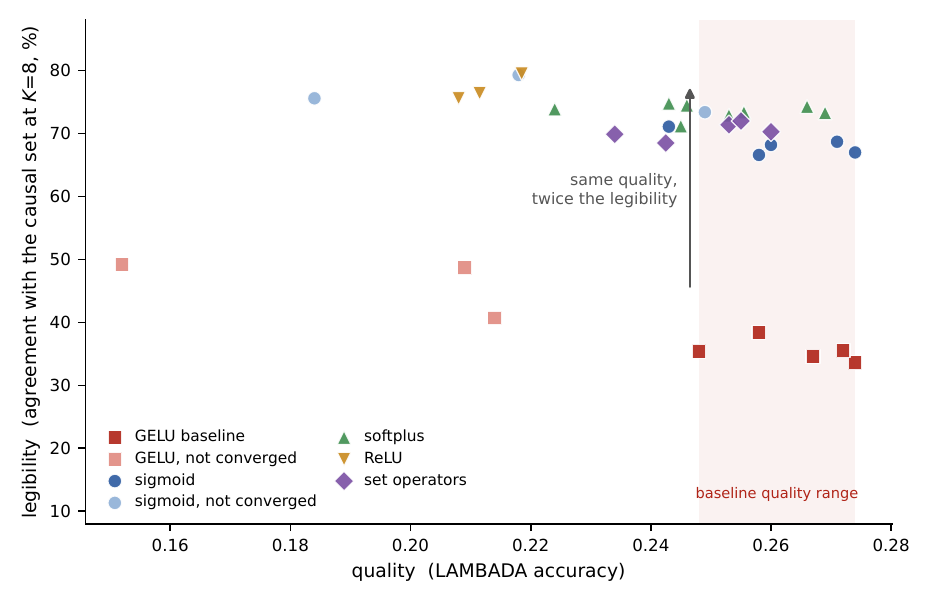}
\caption{Every trained model in this paper. Horizontal axis is quality; vertical axis is how well a
unit's parameters name the tokens that causally drive it \emph{in the vocabulary basis}, both under
one protocol. Read in a different basis, the same units order their drivers well above chance
(Section~\ref{sec:naming-provenance}); what this axis measures is whether the token table is the
right place to look, not whether anything is there. The shaded band is the baseline's converged
quality range. The baseline sits low and to the right --- good quality, poor legibility --- and every
order-preserving arm sits about twice as high at the same quality; ReLU sits higher still and to the
left, buying legibility with accuracy. In \emph{both} families the pale points, the seeds that
failed the convergence screen, sit \emph{above} their own converged siblings without exception,
which is why quality is on this axis at all.}
\label{fig:summary}
\end{figure}

Every model trained for this paper sits on one pair of axes in Figure~\ref{fig:summary}, quality
against legibility. Read Table~\ref{tab:mono-seeds} straight down and the legibility column rises monotonically as
quality falls, every excluded seed scoring above every converged one. That is why the sigmoid
figures quoted below come from the converged seeds alone.

\subsection{Circuit size across the arms}
\label{sec:mono-circuit}

The instrument of Section~\ref{sec:circuit} applies to these constructions unchanged, and this is the
cleanest comparison available: the arms share a width, a dataset, a schedule and a converged quality,
and differ only in $\phi$. Whatever separates them separates because of the activation function.

\begin{table}[htbp]
\centering\small
\caption{The accounting of Section~\ref{sec:circuit-accounting} across the arms. The inflation column
is what the absolute-value convention would report divided by what the signed one does. Every arm cancels heavily, the ReLU arm least, by a factor of five.}
\label{tab:accounting-mono}
\input{T_accounting_mono}
\end{table}

\begin{figure}[htbp]
\centering
\includegraphics[width=\textwidth]{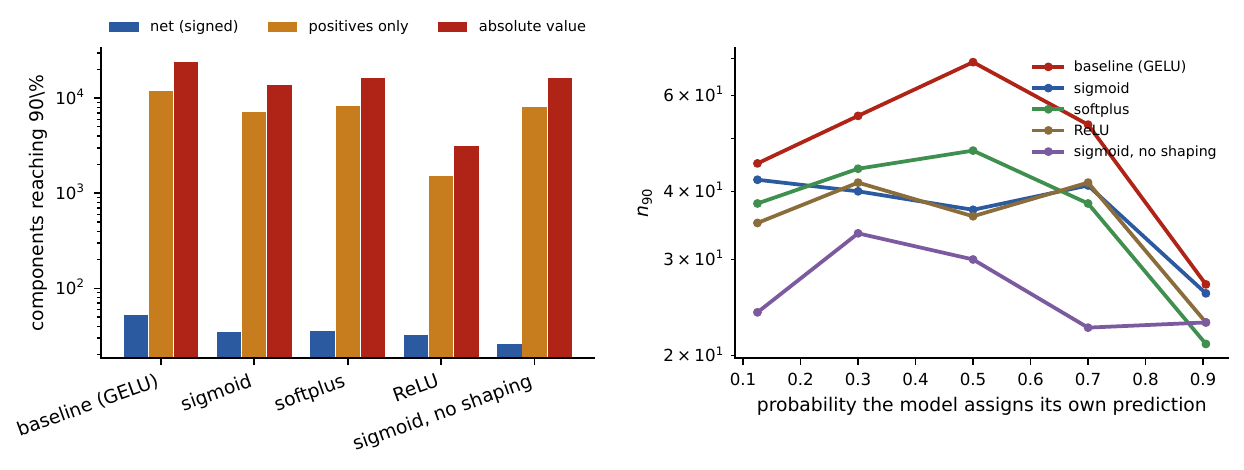}
\caption{Left: the three denominators across the arms, on a log axis. The gap between the signed
count and the other two is the same two-to-three orders of magnitude found in the baseline, and it
is present in every arm --- choosing the activation changes the size of the circuit, not whether the
convention matters. Right: circuit size against how confident the model is in its own prediction.
The arms keep their order at every confidence, so the comparison does not depend on where the
confidence floor of Table~\ref{tab:contract} is set.}
\label{fig:accounting-mono}
\end{figure}

\begin{figure}[htbp]
\centering
\includegraphics[width=0.86\textwidth]{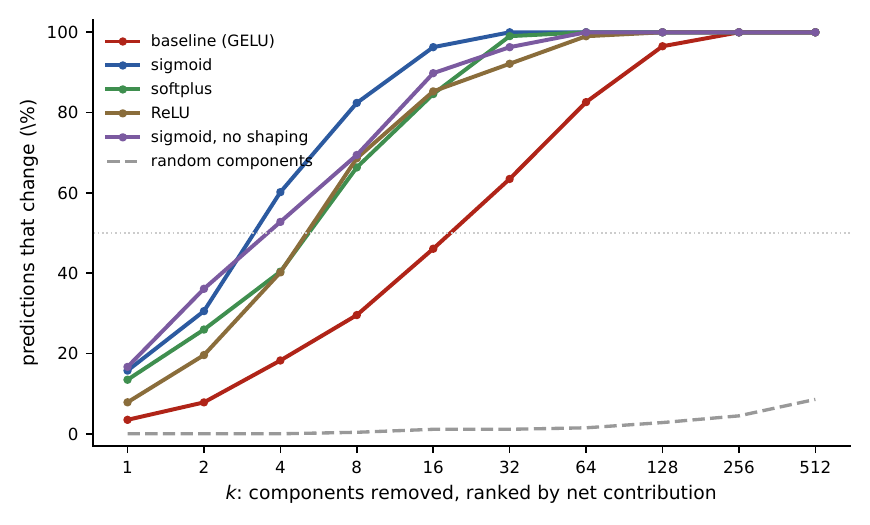}
\caption{Removing the $k$ highest-contributing components, across the arms. The random control is
pooled over all of them. Replacing the conventional activation moves the curve left: the median
prediction turns over at a quarter of the $k$.}
\label{fig:flipk-mono}
\end{figure}

\begin{table}[htbp]
\centering\small
\caption{Circuit size and the point at which the median prediction changes, with the range across
seeds. Four to eight seeds per arm where several were trained.}
\label{tab:headline}
\input{T_headline}
\end{table}

The accounting behaves the same way in every arm and differs only in what it counts (Table~\ref{tab:accounting-mono}). Two things separate (Table~\ref{tab:headline}). The circuit is smaller --- twenty-six to thirty-six components against the
baseline's fifty-three --- and it turns over sooner, the median prediction changing at a quarter of
the $k$ the baseline needs (Figure~\ref{fig:flipk-mono}). Both hold across every seed trained, and
Figure~\ref{fig:accounting-mono} shows they hold across the confidence range rather than only at the
floor these counts are measured at. The set-operator arm, measured under the same procedure and not
in these two floats, sits with the rest: $36$ components at ninety percent of the net and an
inflation of $277$ times against the baseline's $53$ and $452$.

\subsection{The whole dependency graph}
\label{sec:mono-branch}

Everything above counts components contributing directly to the readout --- one hop. The recursion
of Appendix~\ref{sec:circuit-branch} follows the graph the rest of the way, and reports two things:
the closure, which is what a prediction draws on, and the required graph, which is what ablation can
reduce it to. Every arm closes, at four to six levels, on ninety to a hundred predictions each.

\begin{table}[htbp]
\centering\small
\caption{The backward dependency graph of a prediction, closed. ``Attn heads'' is how many distinct
(layer, head) pairs the closure touches, out of the model's total: attention components are named
locations and are ablated with the rest, but are not followed. ``Ablated'' removes the closed graph,
against a size-matched random set.}
\label{tab:branch-mono}
\input{T_dep_mono}
\end{table}

\begin{table}[htbp]
\centering\small
\caption{The required graph: the smallest prefix of the one-hop set whose removal changes the
prediction, found by scanning prefix lengths upward on each prediction separately. Bisection is not
available here: the flip is not monotone in the prefix length, and on $11$ to $28$ percent of
predictions a larger prefix restores a prediction a smaller one broke. ``Defined on'' is the share
of predictions for which such a prefix exists at all; the columns to its left are medians over
those.}
\label{tab:req-mono}
\input{T_req_mono}
\end{table}

At one hop the arms sit within a factor of two of the baseline. At closure they separate from it by
four to seven times (Table~\ref{tab:branch-mono}). The recursion exposes an effect of the
activation function on how much machinery a prediction draws on that a one-hop measurement mostly
misses. The set-operator arm closes on $120$ components in six levels, between the sigmoid arm's
$99$ and the ReLU arm's $182$.

The required graph separates them too, at a small absolute size. Removing $13$ of the
baseline's components changes its prediction; for the arms it is $3$ to $5$. A prediction draws on
hundreds of components and cannot survive losing a handful of them, and the handful is smaller when
the activation is order-preserving.

The branching factor compounds far less than multiplication would suggest, because the sources of
different nodes overlap: multiplying $n_{90}$ by it overstates the measured two-level count by a
factor of four, so the counts here are the union taken at every level rather than a product.

Two caveats from Section~\ref{sec:circuit-bookkeeping} apply to the arms as they do to the baseline, one of them harder. Cancellation arriving at a component is far more severe than
cancellation leaving the model on the arms too, though by several hundred times rather than the
baseline's thousand. And a signed count of what reaches a node is undefined wherever that drive sums
to zero or below, which is a third of the baseline's nodes but more than half of the sigmoid model's,
so the branching medians above are taken over the nodes where the quantity exists and rest on a
smaller share of the sigmoid model's graph than of the baseline's.

\subsection{Legibility}
\label{sec:mono-legibility}

\begin{figure}[htbp]
\centering
\includegraphics[width=0.78\textwidth]{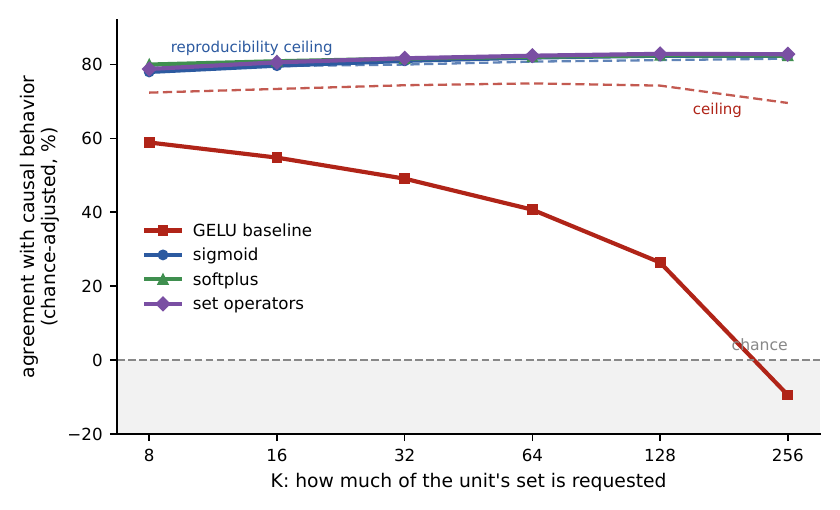}
\caption{Agreement between the parameter-derived set and the causal set, against how much of the set
is requested. The baseline curve is the one from Figure~\ref{fig:agreementk}. The three arms built on
order-preserving activations are flat in $K$ and sit near the dashed reproducibility ceiling --- the
rate at which the causal measurement agrees with \emph{itself} on independent contexts, which no
reading procedure can exceed --- and they land on each other. Set-operator agreement is taken on
operand $A$. Context length 2048, candidate pool $|V| = 512$, table and causal profile from 16
disjoint contexts each.}
\label{fig:mono-k}
\end{figure}

The measurement of Section~\ref{sec:naming-shape} was $\widehat{\mathrm{ov}}@K$ swept over $K$,
and it is repeated here unchanged. The baseline decays from $58.9$ percent at $K=8$ to
$-9.5$ at $K=256$ (Figure~\ref{fig:mono-k}). The other three do the opposite: they \emph{rise} slightly and then flatten,
sigmoid from $78.0$ to $82.4$, softplus from $80.0$ to $82.4$ and the set-operator arm from $78.8$ to
$82.8$ across the same sweep. At $K=256$,
where the baseline selects tokens overlapping the true set less often than a random draw would, they
recover it at better than four times chance.

\begin{figure}[htbp]
\centering
\includegraphics[width=0.42\textwidth]{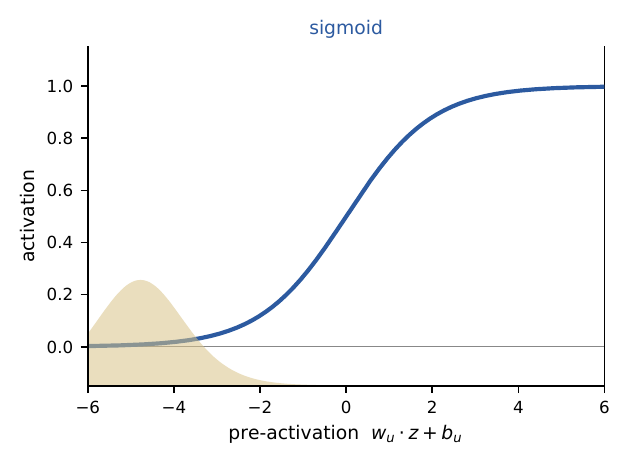}
\caption{The sigmoid arm's operating point, drawn on the same axes and to the same recipe as the
baseline's in Figure~\ref{fig:dip}: the curve is the function and the shaded density under it is the
measured distribution of pre-activations over a trained model of that arm. Its mass sits further left
than the baseline's, and it does not matter, because a monotone function preserves the ordering
wherever its argument happens to land. It is not the operating point that causes the baseline's
reversal --- it is that GELU has an interval where the ordering can be lost at all.}
\label{fig:dip-mono}
\end{figure}

The flatness and how little separates the three matter more than the level. A procedure
that reads the strongest few inputs and then fails produces a decaying curve, which is what the
baseline gives. A procedure that recovers the whole set produces a flat one. Section~\ref{sec:naming-readout} predicted exactly this from the
untrained readout ablation --- every monotone choice was flat in $K$ on the baseline's own
pre-activations --- and training with those activations reproduces it. Figure~\ref{fig:dip-mono}
rules out the operating point as the explanation: the sigmoid model's pre-activations sit further
into the tail than the baseline's, and its reading is unaffected.

The qualitative change behind those numbers is that the sigmoid model's token sets support a
description where the baseline's do not --- places, or the copula cluster around \emph{as} and
\emph{is}. Table~\ref{tab:mono-units} sets the two side by side in Appendix~\ref{sec:baseline-pitfalls}.

\paragraph{The numbers are at the instrument's ceiling.} A reading procedure cannot agree with the
causal set better than the causal set agrees with itself, and that self-agreement is finite because
it is estimated from a sample of contexts. Measured on independent halves and chance-corrected, the
baseline's causal top-8 reproduces itself at $72.4$ percent and its top-256 at $69.6$; the sigmoid
model reproduces at $79.2$ and $81.6$. Reading each model against its own ceiling:
\begin{center}
\small
\begin{tabular}{@{}lrrr@{}}
\toprule
& agreement & ceiling & fraction of ceiling \\
\midrule
GELU, $K=8$      & 58.9 & 72.4 & 0.81 \\
GELU, $K=256$    & $-9.5$ & 69.6 & --- \\
sigmoid, $K=8$   & 78.0 & 79.2 & 0.98 \\
sigmoid, $K=256$ & 82.4 & 81.6 & 1.01 \\
\bottomrule
\end{tabular}
\end{center}
The sigmoid model is at its ceiling, and so are the other two: softplus and the set-operator arm end
the sweep at $82.4$ and $82.8$ against sigmoid's $82.4$, differences smaller than the ceiling can
resolve, so nothing here separates the three from each other. The residual disagreement is the measurement's own noise, and collecting more contexts would raise both numbers together. The claim is that the parameters recover the unit's set as well as the set can be resolved, which is stronger than a ratio between two models.

\paragraph{What the gain costs.} The obvious objection to
these numbers is the one Appendix~\ref{sec:baseline-pitfalls} raised first: a
model reads as more legible when it has less to say. The dimensionality column of Table~\ref{tab:mono-seeds} answers it. The converged sigmoid seeds carry \emph{more} independent directions over the
first five layers than the baseline does, not fewer, while scoring roughly twice as high on
agreement. Whatever the change did, it did not do it by making the representation smaller.

The same column also disciplines the excluded seeds. Two of the three collapsed the way the baseline's
did, and their inflated scores are explained. The third, seed 6, is healthy on that column and still
scores above every converged seed. Collapse does not account for its elevated legibility, which is unexplained. It is excluded on
quality.

\paragraph{Smaller sets, read through the graph.} Agreement says the parameters name the right tokens without saying the tokens belong together. By Equation~\ref{eq:neff} the median effective set size falls by about a
third under sigmoid, a real but modest concentration. Coherence over the set as a whole stays out of reach in the token basis (Section~\ref{sec:naming-about}), so the graph supplies it instead. With the tokens sorted by which upstream source supplies each, a classifier
assigns a token to its source well above a shuffled control, and part-of-speech purity within a
source group exceeds a matched random partition. Both effects are larger than on the baseline
(Section~\ref{sec:naming-provenance}).

The same graph names the components the vocabulary leaves off its frame, separating the arms less cleanly than agreement does. Describing a component from what its incoming edges carry
recovers something it measurably fires on far more often than a control does, in every arm and in the
baseline alike (Table~\ref{tab:graphname}). Ranked by that controlled margin, only sigmoid and the set-operator arm clear the baseline. The set-operator arm's $10.0$ times
on components that name themselves is the largest margin in the table and the least well sampled
number in it: $42$ such components against the baseline's $119$, and a control of $5$ percent against
every other model's $13$ to $20$. Softplus reads highest of any arm
before the control is applied, and its control is twice every other model's because its edge labels
repeat --- the ten commonest hold a quarter of its label pool where the others hold an eighth --- so a
random description hits more often and the corrected margin collapses. Read through the graph rather
than the vocabulary, replacing the activation buys a large improvement in two arms and nothing
reliable in the other two.

\begin{table}[htbp]
\centering\small
\caption{Describing a component by what its inputs carry. A six-token description is built from the
incoming edges that pass the frame-agreement test of Appendix~\ref{sec:circuit-branch}, weighted by
each edge's share of the node's incoming drive, and scored on whether it contains a token the
component measurably fires on. The control builds the same description from edge labels drawn
elsewhere in the same model. ``Names itself'' splits the components by whether their own write
places the predicted token in its first hundred.}
\label{tab:graphname}
\input{T_graphname}
\end{table}

The two measurements ask different questions. Agreement
asks whether a unit's own parameters pick out the tokens it fires on, and the construction was
designed to fix exactly that; the graph asks whether a component's \emph{inputs} describe it, which
the activation function was never aimed at. The arms separate on the first alone, which places the construction's effect on the read side of a component rather than on its position in the graph.

A component whose own write is off the token frame stays readable in every arm. Its incoming edges carry tokens, and those tokens describe it: the
description contains something the component fires on $56$ percent of the time against a control at
$20$, and the rate is the same for components that can name themselves and components that cannot
($60$ against $56$ percent, Table~\ref{tab:graphname}). What a component's own parameters say about
it does not predict whether its inputs describe it. The description is weak in absolute terms --- it
covers $17$ percent of what the component responds to --- and it is a label rather than a name, but
it is available directly.

\paragraph{Two operands are two things to read.} Equation~\ref{eq:overlap} scores one read row
against one causal set, so a unit with two readable operands and a unit with one readable row are
indistinguishable to it however differently they are organized.  Both halves of every unit are readable, and
both at roughly twice the baseline: $68.5$ to $72.0$ for operand $A$ and $64.7$ to $68.9$ for operand
$B$, against the baseline's single row at $33.6$ to $38.4$, all under the ranked summary of
Appendix~\ref{sec:pit-convention}. They also carry different jobs rather than duplicating one. Take
each unit's top-40 tokens by $A$ and split them by whether $B$ is high or low: coherence over the
centered token embeddings rises by $0.0166$, a $31$ percent increase over the unsplit set, where a
random split of the same tokens into halves of the same sizes moves it by $0.0002$. The comparison is
between two ways of splitting the same tokens against a control that splits them at random, so a
similarity measure blind to some concept is blind to it in both arms equally. What it establishes is that $B$ separates $A$'s set along an axis the embedding can see rather than one a person would name.

\subsection{Which property was responsible}
\label{sec:mono-what}

Sigmoid and softplus differ in boundedness and agree on everything measured here: quality within
each other's range, agreement flat in $K$ at $82$, both at their ceiling. The operative property is that the function does not destroy the ordering the weights supply, which also disposes of the more intuitive explanation, that a bounded unit is legible because its activation reads as a probability.

The set-operator arm makes the same point from the far end of the design space. It carries two read
rows, two write columns and a named logical operation per unit,
where Equation~\ref{eq:monounit} carries one row, one column and a swapped function and is what a
feed-forward unit already was. On the $K$ sweep the two are within a point of each other everywhere;
on quality they overlap; at closure they sit twenty-one components apart. If the question is how much
better the parameter-derived set matches the causal set, the elaborate construction is not better.
Three constructions with nothing in common but an order-preserving activation land on one curve, and
a fourth that is monotone but not injective lands somewhere systematically different, which is the
strongest evidence here for which property is doing the work.

ReLU makes the point from the other side. It is order-preserving and scores \emph{highest} of all
on agreement, $75.6$ to $79.5$ percent at $K=8$ against sigmoid's $66.6$ to $71.1$, softplus's
$71.2$ to $74.8$ and the set-operator arm's $68.5$ to $72.0$, all under the ranked summary of Table~\ref{tab:mono-seeds}. It is also the only arm that loses quality, and its units respond to far fewer tokens:
arity $2.4$ to $2.7$ against sigmoid's $5.3$ to $6.7$. Its seeds also vary much more in how many
units answer to nothing --- $0.9$, $4.7$ and $10.2$ percent, against $0.2$ to $4.1$ for sigmoid ---
though those ranges overlap, so the dead-unit count is not a finding on its own. Discarding the negative half-line makes what remains easier to read and leaves
the model less to say. That is a legibility--quality exchange rate, visible in a single controlled
comparison, and it marks the edge of what this change buys for free.

\paragraph{What the construction separates.} A second reading of the same result does not concern the token ordering at all. Run against the carriage background of Section~\ref{sec:circuit-bookkeeping}, the search returns
$18$ components on the baseline and $4$ on the sigmoid model. The two are under one procedure, so
the comparison between them is the measurement, and neither is to be read against the
zeroed-background sizes of Table~\ref{tab:suf}. The same construction that makes a unit readable
also makes the prediction separable from the traffic it travels in. Whether the two effects are one mechanism or two is not settled here. The ordering
result is measured on what drives a unit and this one on what the model delivers, and a construction
could plausibly improve either alone.

Sigmoid, softplus and set-operator units are more legible than GELU units, then, and by the same
property. Whether that translates into editability is the next question.

\subsection{Editability}
\label{sec:mono-edit}

The standard promote edit comes first: add a multiple of a target token's embedding to one unit's
write column,
\begin{equation}
  c_u \;\leftarrow\; c_u + \beta\, E[T],
  \label{eq:edit}
\end{equation}
which credits the same mechanism as established weight-editing methods~\citep{meng2022rome} and
changes nothing about the model but one column. Two things are then measured on held-out text. The
\emph{efficacy} is how much the target token's logit rises at the positions where the unit fires:
\begin{equation}
  \mathrm{eff}(u) = \mathbb{E}_{\text{fires}}\big[\, \mathrm{logit}_T^{\text{edited}}
  - \mathrm{logit}_T^{\text{orig}} \,\big].
  \label{eq:eff}
\end{equation}
The \emph{collateral} is how much everything else moved, as the mean divergence between the original
and edited distributions at the positions where the unit does \emph{not} fire:
\begin{equation}
  \mathrm{coll}(u) = \mathbb{E}_{\text{not fires}}\big[\, \mathrm{KL}\big(P^{\text{orig}} \,\|\,
  P^{\text{edited}}\big) \,\big].
  \label{eq:coll}
\end{equation}
Restricting collateral to non-firing positions matters. Averaging over all positions rewards a unit
simply for being rare, since a unit that almost never fires disturbs almost nothing; asking what
happened where the unit was \emph{not} supposed to act removes that advantage.

Neither number is meaningful alone, as Section~\ref{sec:whatitmeans} noted: any edit is respected if
pushed hard enough and any edit is harmless if pushed gently. $\beta$ is swept over four
strengths per unit, scaled by each unit's own activation range so that the promote signal at a
typical firing position is comparable across architectures with different activation scales, and
collateral is read at \emph{matched efficacy} by interpolating each model's curve.

\begin{table}[htbp]
\centering
\small
\caption{Collateral damage at matched efficacy, median over 34--40 edited units per seed, target
token held fixed. Lower is better, and entries are per-seed values where there are two and ranges
where there are three. The last column is the range of efficacy each model's units reach across the
four strengths tested, which is where the families separate without overlap. On a set-operator unit
the edit adds to one of the two write columns, so it lands on one branch of the unit rather than on
the unit as a whole.}
\label{tab:edit-mono}
\begin{tabular}{@{}llrrrr@{}}
\toprule
& & \multicolumn{3}{c}{collateral at efficacy} & efficacy \\
\cmidrule(lr){3-5}
model & seeds & 1.5 & 2.0 & 3.0 & reached \\
\midrule
GELU baseline & 2 & 0.006, 0.023 & 0.011, 0.041 & 0.023, 0.085 & \phantom{0}3.3 -- \phantom{0}3.9 \\
\midrule
sigmoid       & 3 & 0.001--0.008 & 0.003--0.015 & 0.006--0.041 & \phantom{0}6.3 -- \phantom{0}8.4 \\
softplus      & 2 & 0.001, 0.002 & 0.001, 0.003 & 0.003, 0.006 & \phantom{0}8.7 -- \phantom{0}9.3 \\
set operators & 3 & 0.002        & 0.001--0.003 & 0.003--0.005 & 10.1 -- 14.3 \\
\bottomrule
\end{tabular}
\end{table}

The cleanest difference is reach rather than collateral (Table~\ref{tab:edit-mono}). Across the same four strengths, baseline units top out at an efficacy of $3.3$ and $3.9$, while the monotone units
reach $6.3$ to $9.3$ and the set-operator units $10.1$ to $14.3$. The ranges do not overlap. A baseline unit resists being told what to do past
a certain strength.  Pushing harder stops raising the target logit and only adds damage --- visible
in the per-unit curves, where efficacy peaks at the second of four strengths and then \emph{falls}
while collateral keeps climbing. The monotone units do not have that turning point in the range
tested.

Collateral points the same way. Four of the five monotone runs sit below both
baseline runs at every matched efficacy, by factors of four to fourteen. But the ranges overlap: one
sigmoid seed ($0.041$ at efficacy $3.0$) is worse than one baseline seed ($0.023$), so the statement is that the monotone models are typically more targeted rather than uniformly so. The
set-operator seeds are the one place the ranges do separate: $0.003$ to $0.005$ at efficacy $3.0$
against the baseline's $0.023$ and $0.085$, five to twenty times less disturbance for the same
effect, and below sigmoid's $0.006$ to $0.041$ as well. An edit to one of its write columns acts only
where $A$ opens \emph{and} $B$ selects that branch, which is the narrowest circumstance any
construction here puts a write in.

The two results are the same fact seen twice. An edit
to $c_u$ acts wherever $u$ fires, so its precision is bounded by how well-defined ``wherever $u$
fires'' is. If a unit's set is a grab-bag, an edit aimed at one part of it lands on all of it, and
the model resists the edit exactly to the extent that the unit was doing several unrelated jobs.
That is why the baseline's efficacy saturates and the monotone models' does not: the edit is
competing with the unit's other responsibilities. Legibility is what makes editing possible, which is why the edit checks the reading.

\paragraph{Changing facts.} Equation~\ref{eq:edit} adds to a write column a component already has.
Two further operations ask more of it. \emph{Installing} an association puts a key for one token and a value for another into a
component the model was not using, as Section~\ref{sec:edit-install} does on the baseline.
\emph{Rewiring} keeps the write column exactly as trained and replaces the read row, so the component
keys on a different token and delivers what it always delivered. Measured across the arms, the three order the models differently (Table~\ref{tab:editops}).

\begin{table}[htbp]
\centering\small
\caption{Three edit operations across the arms. \emph{Gain} is the share of edits moving a
component's own named scope more than another component's, at matched strength. \emph{Install} is the
share placing a new token at the top of the distribution where its key occurs \emph{and} not
elsewhere, at the strongest setting whose held-out loss stays within ten percent of the unedited
model. \emph{Rewire} is how far the component's own write moves
toward the token its read row was changed to. The operations are measured on one protocol each,
across both kinds of component.}
\label{tab:editops}
\input{T_editops}
\end{table}

\begin{table}[htbp]
\centering\small
\caption{The scope test of Section~\ref{sec:edit} on the sigmoid model: amplify a component and ask
where the logits move --- on the tokens its own write column names, on another component's, or on
random tokens --- at matched strength. ``Respects the name'' is the share of edits moving the
component's own scope more than another's. Same protocol as Table~\ref{tab:edit-baseline}, so the
rows are comparable across the two models.}
\label{tab:edit-scope-mono}
\input{T_edit_mono}
\end{table}

Amplification is indifferent to the activation. Every arm takes it, on both kinds of component, on
$68$ to $99$ percent of trials --- scaling a column commutes with whatever the activation does, so it
asks nothing of the function.
The scope test of Section~\ref{sec:edit},
however, asks whether the movement lands on the tokens the component's own column names rather than on
another component's.  The sigmoid model respects the name on every edit tried against $97$ percent for
the baseline, and moves its own scope by $1.62$ against the baseline's $0.89$
(Table~\ref{tab:edit-scope-mono}). The other two operations are not indifferent at all. No monotone arm
takes an installed association at any strength of either knob \emph{when the edited unit is the only
thing changed}, where the baseline manages $32$ percent; and rewiring, which the baseline's units
accept at $+2.94$, moves a sigmoid unit by $+0.002$.

\begin{table}[htbp]
\centering\small
\caption{Installing an association across the four activations, at the strongest setting whose
held-out loss stays within five percent of the unedited model, so the sites are compared at matched
damage rather than at matched strength. The last column is the same models' read side from
Table~\ref{tab:readname}. The unit beats the channel on the baseline; on the other three neither site takes the edit at any
strength, so the two are tied at zero. The ordering of the two columns is close to opposite.}
\label{tab:putarm}
\input{T_putarm}
\end{table}

The failure has one shape (Table~\ref{tab:putarm}): raise the write gain until $B$ arrives where $A$ is, and $B$ has arrived
everywhere else too, its rank away from $A$ falling from the high hundreds to $1$ with held-out loss
following.

Neither knob explains that away. The write gain sets how much of $B$ arrives and the read gain scales
the installed key; both were swept, over four orders of magnitude in the second. ReLU takes no
specific edit at any setting of either. Sigmoid and softplus reach a third of trials only at a read
gain of $256$ and a held-out loss of $11$ to $23$ against baselines near $3.2$ --- the association can
be forced in by saturating the unit, and what remains is not a model. The baseline needs neither knob
raised: at a read gain of $1$ it installs on $48$ percent of trials for five percent of held-out
loss, and on $24$ percent for one percent of it.

At least part of the reason is that an installed association stays local only where the component can
be silent on the tokens it was not installed for, and $\sigma(0) = \tfrac{1}{2}$ leaves a sigmoid unit
no such state. Measured on the candidate vocabulary, an installed key gives the baseline a
sixty-to-one contrast between the token it was installed for and the rest, and sigmoid two-to-one,
with no token silent at all. ReLU fails for a different reason: its contrast is fine, at forty to one, but the unit emits about a
ninth of what the baseline's does, so it must be driven nine times harder to be heard --- into a
residual that is smaller in the same proportion, which is why the same strength costs it four times
the loss. The baseline's advantage is sharper still: off its installed token it emits a slightly
\emph{negative} amount, so the edit pushes $B$ down everywhere it was not meant to act, which is
where the improving collateral in Table~\ref{tab:put} comes from.

The same ceiling
caps rewiring, since a unit whose output lies in $[0,1]$ can never deliver more than one column's
worth however its read row is changed. Softplus, unbounded above, sits between the two as that account predicts, at $+0.36$. The set-operator arm behaves like the bounded arms on all three, which is what its form implies: its outputs are products of sigmoids and so lie in $[0,1]$ too. Section~\ref{sec:mono-gate} works through the second write column it has and they do not.

\subsection{The edit these models will take}
\label{sec:mono-comp}

The previous section reads like all is lost for editing these models, but it also points to the repair.
What an installed unit emits away from its key is a
\emph{constant}: the resting value of the activation times the value column, at every position
alike. A constant added to the residual can be subtracted from it, and the subtraction need not come
from the edited unit. It has to come from somewhere, though, and that is where the architecture
binds. These models carry no bias --- neither the read projection nor the write projection has one
--- so there is no term in the layer whose whole job is to hold a constant. The only place the
correction can go is the write columns of the other units.  But it can be put there.

Install the key and the value as before, run the model on held-out text, and take
the positions where the key is \emph{absent}. Over those positions, ask the layer to reproduce the
constant the installed unit emits: solve for coefficients $\theta$ over the units whose activation
is most nearly constant off-key, and subtract $\theta_v \beta v$ from each of their write columns.
The fit is a ridge solve against activations the model supplies, with nothing trained and no gradient taken. The offset is removed and the excursion where the key occurs survives, because those positions were held out of the fit.

\begin{table}[htbp]
\centering\small
\caption{Installing an association, over $23$ token pairs, five depths and five strengths per model. The compensating fit and the threshold's quantile are both taken on a block of contexts disjoint from the one the edit is scored on.
An edit counts only if the target reaches the top ten where the key occurs, is \emph{not} promoted
anywhere else, and itself costs under ten percent of held-out loss --- the criterion of
Table~\ref{tab:editops}. \emph{Corrected} subtracts the leak from other units' write columns;
\emph{thresholded} instead gives the unit an off state by subtracting a threshold from its read row,
taken as a quantile of the key's own projection over positions where the key is absent. The two
repairs are independent and the table crosses them. ``Target elsewhere'' is the median ratio of the
target's rank away from the key after the edit to before it, where $1$ is untouched.}
\label{tab:comp}
\input{T_comp}
\end{table}

The leak is exactly what the diagnosis said it was, and it is confined to the constructions the
diagnosis is about (Table~\ref{tab:comp}). Uncompensated, a sigmoid unit promotes the target twelve
times over the rest of the corpus and softplus eight times, where the baseline moves it by three
percent: $\mathrm{GELU}(0)=0$, so a conventional unit is already silent off its key and has nothing
to subtract. Compensated, every arm's leak is gone --- the target's rank away from the key returns
to where it started or slightly worse.

On the baseline the correction is not the best repair.
What the unit emits
away from its key is a constant \emph{because its read row has no off point}: the row fires partway
wherever the key is even weakly present, and the activation turns that into a floor. A threshold in
the read row removes the cause rather than the symptom. These models have no bias, so it is built the
same way the correction is --- from the near-constant direction of the layer's input, with the offset
taken as a quantile of the key's own projection over positions where the key is \emph{absent}, so
nothing is fitted on the positions the edit is aimed at.

The threshold works
on every model: it takes the unit's mean activation off its key from $0.12$ to $-0.031$ on the
baseline and from $0.42$ to $0.004$ on the sigmoid arm, a hundredfold. But on the shaped arms the
install still fails, on $92$ to $99$ percent of attempts, and it fails on one clause --- the target is
promoted \emph{elsewhere}. What is left below the threshold is small on every model and its sign
differs: $\mathrm{GELU}$ is negative just below zero, so the residue demotes the target away from the
key, which the criterion permits; a bounded or rectified unit cannot go below zero, so however small
its residue it promotes the target everywhere, which the criterion forbids. That is what the
correction removes, and the numbers say so directly --- with the threshold alone the share of edits
that stay local is $1$ to $8$ percent on the shaped arms and $75$ on the baseline; adding the
correction takes the shaped arms to $27$ to $36$ and the baseline \emph{down} to $67$, because there
it subtracts something that was helping. So the baseline needs the threshold and not the correction,
reaching $75$ percent; the shaped arms need both, reaching $27$ to $36$ where the correction alone
reaches $5$ to $16$.

With the leak removed the association installs. Every order-preserving arm goes from
taking no edit at all to taking one, softplus on $9$ percent of attempts and ReLU on $16$, and
the edits are cheap: the least expensive passing edit costs a thousandth of a nat or less, which is
under a twentieth of a percent of held-out loss on any of them. The same correction helps the
baseline, from $25$ percent to $32$: a GELU unit is nearly silent off its key rather than exactly
silent, and the residue is worth removing.

One control decides how the correction has to be built. Given every unit in the layer, the fit
reconstructs the installed unit's activation at $R^2$ between $0.85$ and $1.00$ --- including at the
key positions it never saw, which are not linearly independent of the rest --- so what it subtracts
is the edit rather than the leak. On the sigmoid and set-operator models the target then falls past
rank eight thousand and rank thirty-eight thousand, where the capacity-limited fit had brought it
inside the top ten; on the others it is pushed back by an order of magnitude. That arm is worse than the capacity-limited fit on every model. The correction has to be limited to
units that cannot track the input: a unit with a large mean and little variance can supply a bias
and nothing else, which is precisely the term the architecture is missing.

So the construction changes what an edit has to consist of.
A conventional unit can be told something new on its own: a threshold in its own read row is enough,
and it is the best repair there is for that model. An order-preserving unit cannot, and the threshold
is what shows it --- the same repair, confined to the unit itself, leaves the shaped arms at $0$ to
$9$ percent, not because the unit goes on firing but because what it emits when it is off has the
wrong sign. The residue has to be taken back somewhere else in the same layer, which the layer can
express, at a cost too small to matter. These arms buy precision on the operations they support at the cost of installing a fact as a two-part edit rather than a one-part one. Section~\ref{sec:mono-gate} shows the set-operator arm taking the correction
out of its own second write column instead of out of the layer.

\paragraph{Their attention fails the other way.} The same question asked of these models'
\emph{attention} comes out differently, and it is a cost of the second departure of
Section~\ref{sec:mono-model} rather than of the activation function. Their heads bound the
\emph{value} rather than the weighting over positions, which is the ordinary softmax.
What a head gathers is therefore an average of bounded quantities, and a sigmoid rests at
$\sigma(0)=\tfrac{1}{2}$, so every position contributes that resting level whatever the head attends
to. Install a key into a channel and measure what it gathers where the key is
present and where it is absent. What an edit can use is the \emph{difference} between the two ---
what arrives at the key over and above what arrives everywhere. On the baseline the channel collects $-0.06$ where the key is absent
and $+0.38$ where it is present, a usable difference of $0.44$. On the sigmoid model the same
measurement returns $0.485$ against $0.524$: a difference of $0.04$, and on softplus $0.07$ and on
ReLU $0.02$. Both of those numbers sit on either side of $\tfrac{1}{2}$, which is where a sigmoid with
nothing driving it rests. So the usable signal is a sixth to a twentieth of the baseline's, and it
arrives on top of a resting level seven to twenty-six times its own size. That level is the same at
every position, so an installed association is delivered everywhere the head looks rather than only
where the key is: the target's rank away from the key improves fiftyfold on the sigmoid model and a
hundredfold on the other two, where on the baseline it does not move at all.

Subtracting that level removes the delivery everywhere and leaves too little behind. Compensated,
the target stops
arriving away from the key on all three, and no edit of this form passes at any strength. Nor is the
trouble that these models lack local heads: measured as Section~\ref{sec:edit-install} measures it,
their most local heads reach $0.87$ and $0.89$ against the baseline's $0.75$, and installing in them
sharpens what arrives --- the target reaches rank $2$ to $5$ at the key --- while the share of edits
that install \emph{and} stay local remains zero.

A head that puts the target at rank $2$ at
the key is a head that found the key --- what fails is not the choice of where to look.
What is flattened is what the head carries
there. A value squashed into $[0,1]$ and read by a key row that was not scaled for it sits in the
sigmoid's near-linear middle, so what arrives departs from the resting level by very little however
sharply the head attends. The two sublayers therefore fail for opposite reasons in these models. The
feed-forward unit discriminates and leaks, which a subtraction repairs. The attention channel
discriminates too, and delivers a value flattened toward its resting level; the subtraction removes
that level and leaves too little to work with. Section~\ref{sec:mono-gate} scales a key row to the
value's range and recovers a usable difference for a marker read by a feed-forward operand, though
not for an association read at the output.

\subsection{Placement, circuits and the rank-one editor across the arms}
\label{sec:mono-across}

\begin{figure}[htbp]
\centering
\includegraphics[width=\textwidth]{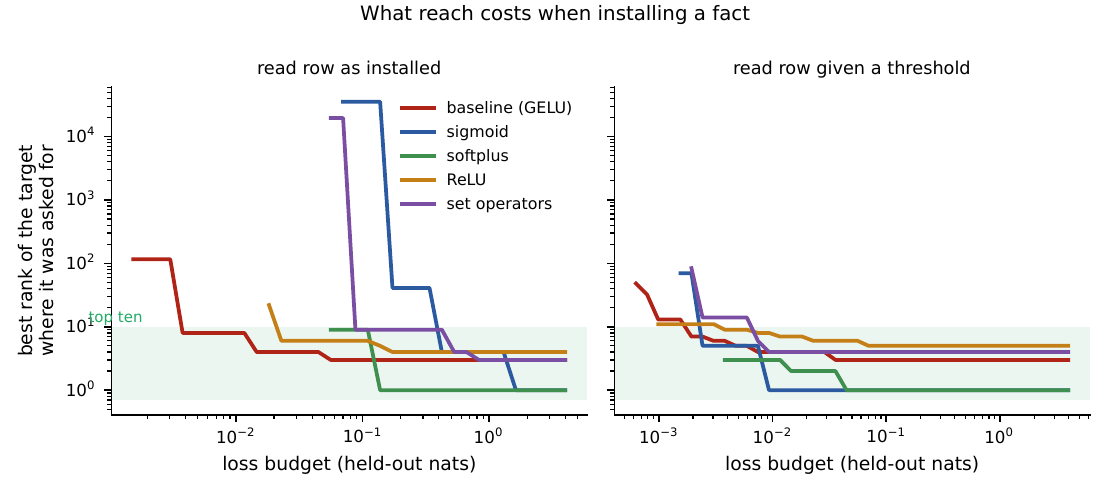}
\caption{The best rank the target reaches for a given held-out loss budget, at layer ten. Each
frontier is taken over four compensation arms and five write strengths. Left: the read row as
installed. Right: the same with a threshold subtracted from the read row, swept over three
quantiles.}
\label{fig:installcost}
\end{figure}

With the repair in hand, the rest of Section~\ref{sec:edit} runs across the arms. Reporting an install at one operating point hides the shape of what it buys, because there are two
independent prices rather than one. Figure~\ref{fig:installcost} reports the frontier: for each
budget of held-out loss, the best rank the target reaches over the compensation arm, the write
strength and the threshold quantile. Giving the read row a threshold moves every curve to the left,
by amounts that separate the activations. The baseline reaches the top ten for $0.0032$ nats without
one and $0.0016$ with one, a factor of two. Sigmoid needs $0.35$ without one and $0.0020$ with one, a
factor of a hundred and seventy, and its cheapest route to rank one falls from $1.31$ nats to
$0.0092$.

The placement results of Section~\ref{sec:place-depth} carry over to the arms without
qualification. On the sigmoid arm a write keeps $0.39$ of its direction at the readout when made
before the last layer against $0.10$ before layer two, spill falls from $0.40$ to $0.04$ over the
same range, and the readout frame is again the only family whose response is readout-visible above
chance, rising from $0.29$ to $0.52$. Injected content is found in the coordinates it was written in
there too, at $0.98$ with no gap. A unit the sigmoid model trained for itself is driven from two layers upstream with
$0.88$ of the effect passing through it, against $0.86$ on the baseline, and the read row's estimate
governs the result the same way, separation rising from $1.9$ at two positions to $24.8$ at the full
sample.

The same frontier drawn for the circuit of Section~\ref{sec:place-circuit} separates the arms less
than the install does. The baseline reaches the top ten for $0.121$ nats and the set-operator arm for
$0.240$, and imposing the specificity criterion takes those to $0.179$ and $0.551$, where ReLU pays
nothing for it at $0.139$ either way. The comparison of Section~\ref{sec:edit-rome} also holds
across the arms: the single-unit install reaches the top ten for less held-out loss than the
closed-form rank-one update on every one of the five models, by factors running from nine on sigmoid
to seventy-two on the set-operator arm.

\begin{figure}[htbp]
\centering
\includegraphics[width=\textwidth]{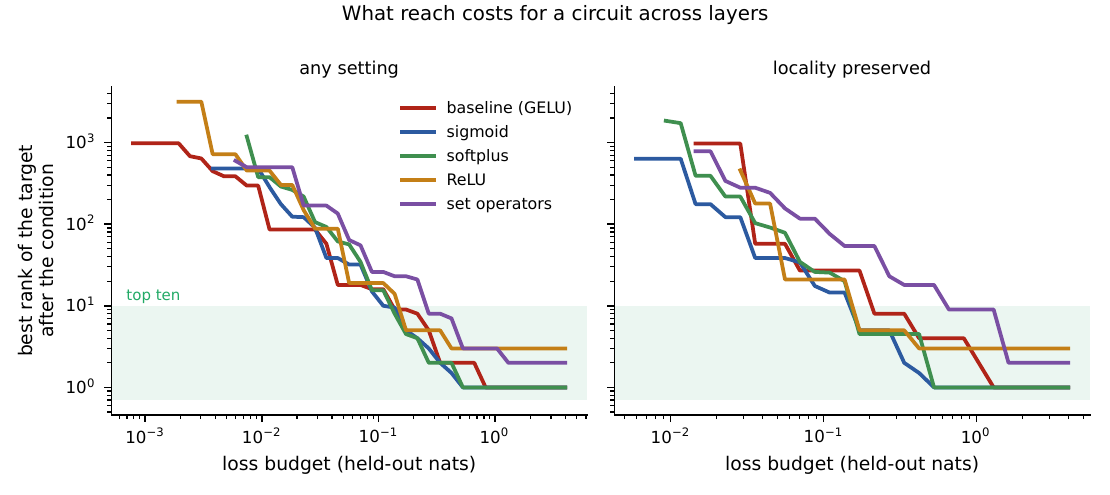}
\caption{The best rank the target reaches for a given held-out loss budget, over the full cross of
write strength, read gain, saturation, read-row calibration and marker tilt, with heads at layers
four, six, eight, nine and ten reading at layer ten. Left: any setting. Right: the same with the specificity criterion imposed, so the target
may not be promoted where it was not asked for. The gap between the panels is the price of
specificity.}
\label{fig:circuitcost}
\end{figure}

\subsection{Editing through the second operand}
\label{sec:mono-gate}

Everything above edits what a unit writes. The set-operator unit has a second place to act, on the
input side, and three edits use it. What they establish is what this construction expresses
directly; a conventional model can be made to express the same things, and the end of the section
says how. Leaving $W_a$ untouched, so the unit
keeps whatever it was already triggered by, and keying $W_b$ on a token $C$ makes the intersection
branch fire on \emph{its usual trigger and $C$} and the difference branch fire on \emph{its usual
trigger and not $C$}. One write installs a conjunction and a suppressor at the same time.

It works at no cost in held-out loss. Over twelve units per layer at every depth, the intersection branch fires
five times more often where $C$ is present than where it is not, and sharpens as the gate is driven
harder; the difference branch fires \emph{exactly} zero where $C$ is present, at every strength
tried; and held-out loss moves from $3.156$ to $3.157$. The exactness is the same property that costs
this family everywhere else. Section~\ref{sec:mono-edit} found that a bounded activation will not, on
its own, be told a new association, because $\sigma$ has no silent state. Here that is what makes the
gate clean: $B$ saturating at one makes $A \setminus B = A(1-B)$ exactly zero rather than
approximately zero, so the suppression is total. Boundedness is the mechanism the construction uses.

\paragraph{A unit that cancels its own leak.} Section~\ref{sec:mono-comp} installs an association into
an order-preserving unit by taking the constant it leaks back out of the other units in the layer.
This construction can do the same without leaving the unit. Set the intersection column to $\beta v$
and the difference column to $-\alpha\beta v$, and what the unit contributes is
\begin{equation}
  \big[A B - \alpha A (1 - B)\big]\,\beta v
  \;=\; A\big[(1+\alpha)B - \alpha\big]\,\beta v ,
  \label{eq:selfcancel}
\end{equation}
which is identically zero wherever the bracket vanishes, whatever $A$ is doing. Choosing
$\alpha = \bar{B}/(1-\bar{B})$ for the resting value $\bar{B}$ of the second operand puts the zero
where the unit is not wanted, and the only quantity measured is $\bar{B}$. $\alpha$ runs from $0.56$ to
$1.62$ with a median of $0.93$, which is what both operands resting near $\sigma(0)=\tfrac{1}{2}$
gives. Uncompensated, an installed unit here promotes the target six times over the rest of the
corpus; with the second column set this way it promotes it $0.82$ times. Taking $\alpha$ from a high
quantile of $B$ instead over-cancels --- the target is pushed back twelvefold away from the key ---
and passes: at $\beta = \tfrac{1}{2}$ the target moves from rank $46$ to rank $6$ where the key occurs
and from $346$ to $611$ elsewhere, for $0.018$ nats. It does not beat the fitted correction, which
passes on $3.7$ percent of attempts against this form's $2.3$, within each other's noise at this
sample. What it establishes is what the structure makes expressible: a single unit can be given a
fact and silenced off it using only its own parameters, where the other arms spend part of the layer
on the correction.

\paragraph{Installing a fact under a condition.} The third edit installs a fact \emph{under a
condition}: fire where token $A$ is current and token $C$ occurred in the
preceding forty-eight tokens. The feed-forward block cannot read that condition off its own position.
A row $\tilde{E}[l,C]$ of the layer-native table is the state the layer sees when $C$ \emph{is the
current token}, so keying an operand on it asks whether $C$ is here, not whether $C$ was there; at an
$A$-position $C$ is not here and the gate never opens. Carrying $C$ forward is attention's office, so
the condition is delivered by an attention channel installed in the same edit: the channel reads $C$
into its value row and writes what it gathered along a private direction, and the unit's second
operand reads that direction. Five weight writes, no gradient, the two sublayers joined by hand.

Two settings are needed, and both follow from the construction rather than from tuning. The channel's
value is bounded in $[0,1]$, so a key row not scaled for it sits in the sigmoid's near-linear middle
and what the channel gathers barely leaves its resting level: the contrast between gathering at $C$
and away from it is $1.02$ to $1.45$ times. Choosing a threshold and gain from the measured
distribution of $\mathrm{key} \cdot z$ instead --- one quantile and one scale, nothing fitted against
the target --- takes that contrast to $901$. And the operand that reads the marker has no off state,
because $\sigma(0) = \tfrac{1}{2}$ and these models carry no bias parameter; supplying one from the
layer's near-constant direction takes the operand's firing after $C$ against elsewhere from $1.2$
times to $43$. The first of the two is the question Section~\ref{sec:mono-comp} leaves open, where an
association installed into these models' attention channels arrived everywhere at the resting
level and would not separate: scaled here, a channel does deliver a usable difference --- for a marker read by a
feed-forward operand, which is a weaker thing to ask of a channel than an association read at the
output.

With both set, the edit lands. Over six triples, three depths and five strengths, the unit reaches the
top ten where $A$ is current and $C$ fell in the preceding forty-eight tokens, stays out of it in both
other groups, and holds held-out loss inside ten percent on $31$ of $90$ attempts. Its operands read
$A = 0.924$ at those positions against $0.020$ and $0.023$ at the other two, and $B = 0.997$ against
$0.002$, so the intersection separates by about $130$ and $90{,}000$ times. The sharpest circuit puts
the target at rank $1$ where $A$ follows $C$, against $10{,}170$ where $A$ occurs without it and
$14{,}215$ elsewhere, for $1.6$ percent of loss; the cheapest moves it from rank $3902$ to $9$ where
the condition holds, leaves it at $4869$ where $A$ occurs without $C$, and costs $0.2$ percent. An
independent replication over a different range of strengths passes $17$ of $100$.

The head is what carries the condition. Repeat every weight write and leave the
attention head alone, so the operand reads a direction nothing writes, and the conditional ratio falls
from $17.65$ times to $1.24$, with one attempt of ninety passing rather than thirty-one.

\paragraph{Two can play at this game.}
Two operands are worth less within this model than the construction suggests. Pin the
second operand open, which is the same circuit with one operand and no threshold, and $8$ of $90$
pass --- at the \emph{same} conditional ratio, $18.49$ against $17.65$. The second operand does not
sharpen the condition at all. What it does is narrow where the write happens: with one operand the
target is promoted wherever $C$ occurred regardless of $A$, and its rank away from both groups sits at
$764$ against $1616$ with both operands. Put both detectors into a single read row of the same model
and raise the threshold, so that a thresholded sum performs the conjunction, and $18$ of $90$ pass at
a conditional ratio of $9.74$, against the two-operand form's $31$ of $90$ at $17.65$
($p = 0.044$ on the counts). That margin is understated rather than inflated, because the thresholded
sum was given the best of five thresholds and the two-operand form was not tuned that way.

None of that makes the edit a capability of this construction. Run the same circuit on the
conventional baseline --- one read row carrying both detectors over a raised threshold, the head
installed the same way --- and $28$ of $90$ attempts pass, against the set-operator arm's $31$, which
is not a difference at this sample; its own no-head control passes none, at a ratio of $1.34$. On the
sharpness of the condition the conventional unit is an order of magnitude ahead: a conditional ratio
of $194.5$ against $17.65$, and a median rank of $1$ where $A$ follows $C$ against $110$. What the second operand buys, on the evidence here, is locality rather than the condition itself.

One part of the construction does \emph{not} earn its place here. The self-cancellation of Equation~\ref{eq:selfcancel}, which suggested this edit, contributes nothing to it: $31$ attempts pass with the difference column set and $31$ without. The algebra says why it cannot.
$A \setminus B = A(1-B)$ is nonzero only where $B$ is off, which is at \emph{non}-$A$ positions, while
the leak that breaks the conjunction is at $A$-positions, where the difference column is already zero.
The cancellation works for the case Equation~\ref{eq:selfcancel} was derived for, both operands keyed
on one token; it does not extend to a condition carried in from somewhere else.

So what the second operand is worth is narrow. It buys
no better reading --- the $K$ sweep of Section~\ref{sec:mono-legibility} put every order-preserving
arm on one curve --- and on the write side it buys locality rather than capability: the second operand
narrows where a write lands without changing how sharply the condition binds. Two smaller things come
with it, a suppressor that is exactly off rather than approximately off and a leak the unit takes back
out of its own parameters instead of out of the layer.  The cleverness is not in the set-operator model so much as in the fact that a conventional model
can be compelled to perform the same tricks.

What the activation function changes, then, is this. An order-preserving unit reads at the
instrument's ceiling, flat in $K$, at a quality inside the baseline's converged range; a conventional
unit reads its strongest few inputs and then reverses, because nearly two thirds of its
pre-activations sit where GELU runs backward. The order-preserving arms' circuits are half to two thirds the size at one
hop and four to seven times smaller at closure, and the handful a prediction cannot lose shrinks from
thirteen to a few. The promote edit reaches twice as far before it saturates. What the change costs
is the install: a bounded unit has no silent state, so a new fact leaks until the layer takes the
constant back, and installing becomes a two-part edit. Three constructions with nothing in common
but an order-preserving activation land on one curve, and the one that discards the negative
half-line buys its extra legibility with quality --- so the property doing the work is that the
function does not destroy the ordering the weights supply, and nothing more elaborate is needed to
get it.

\secbarrier

%% file: T_accounting_mono.tex
\begin{tabular}{lrrrr}
\toprule
model & net (signed) & positives only & absolute value & inflation \\
\midrule
baseline (GELU) & 53 & 11,974 & 23,971 & 452$\times$ \\
sigmoid & 34 & 7,207 & 13,851 & 401$\times$ \\
softplus & 36 & 8,311 & 16,403 & 456$\times$ \\
ReLU & 32 & 1,527 & 3,133 & 98$\times$ \\
sigmoid, no shaping & 26 & 8,097 & 16,429 & 632$\times$ \\
\bottomrule
\end{tabular}

%% file: T_headline.tex
\begin{tabular}{lrrrrr}
\toprule
model & predictions & $n_{90}$ & across seeds & flips at $k$ & attention \\
\midrule
baseline (GELU) & 115 & 53 \small{[42, 63]} & 45--107 & 32 & 27\% \\
sigmoid & 108 & 34 \small{[30, 40]} & 9--44 & 4 & 25\% \\
softplus & 104 & 36 \small{[29, 40]} & 10--28 & 8 & 22\% \\
ReLU & 102 & 32 \small{[24, 38]} & 22--41 & 8 & 44\% \\
sigmoid, no shaping & 108 & 26 \small{[23, 30]} & 12--30 & 4 & 46\% \\
\bottomrule
\end{tabular}

%% file: T_dep_mono.tex
\setlength{\tabcolsep}{4pt}
\begin{tabular}{lrrrrrrrr}
\toprule
& & \multicolumn{2}{c}{graph} & & & & \multicolumn{2}{c}{ablated} \\
\cmidrule(lr){3-4}\cmidrule(lr){8-9}
model & preds & one hop & closed & levels & \% of model & attn heads & closure & random \\
\midrule
baseline (GELU) & 100 & 53 & 736 & 6 & 1.60\% & 62/144 & 98\% & 9\% \\
sigmoid & 95 & 35 & 99 & 5 & 0.21\% & 27/144 & 97\% & 9\% \\
softplus & 90 & 36 & 123 & 4 & 0.27\% & 30/144 & 99\% & 2\% \\
ReLU & 91 & 30 & 182 & 5 & 0.39\% & 51/144 & 100\% & 5\% \\
sigmoid, no shaping & 94 & 26 & 122 & 6 & 0.26\% & 30/144 & 100\% & 0\% \\
set operators & 98 & 36 & 120 & 6 & 0.43\% & 24/144 & 93\% & 0\% \\
\bottomrule
\end{tabular}

%% file: T_req_mono.tex
\setlength{\tabcolsep}{4.5pt}
\begin{tabular}{lrrrrrr}
\toprule
& \multicolumn{2}{c}{graph} & \multicolumn{4}{c}{required} \\
\cmidrule(lr){2-3}\cmidrule(lr){4-7}
model & one hop & closed & components & \% of one hop & \% of model & defined on \\
\midrule
baseline (GELU) & 53 & 736 & 13 & 25\% & 0.028\% & 73\% \\
sigmoid & 35 & 99 & 4 & 11\% & 0.009\% & 99\% \\
softplus & 36 & 123 & 5 & 14\% & 0.011\% & 87\% \\
ReLU & 30 & 182 & 5 & 17\% & 0.011\% & 86\% \\
sigmoid, no shaping & 26 & 122 & 3 & 12\% & 0.007\% & 95\% \\
set operators & 36 & 120 & 5 & 14\% & 0.018\% & 84\% \\
\bottomrule
\end{tabular}

%% file: T_graphname.tex
\setlength{\tabcolsep}{4.5pt}
\begin{tabular}{lrrrrrrrr}
\toprule
& \multicolumn{4}{c}{names itself} & \multicolumn{4}{c}{does not} \\
\cmidrule(lr){2-5}\cmidrule(lr){6-9}
model & nodes & hit & control & lift & nodes & hit & control & lift \\
\midrule
baseline (GELU) & 119 & 58\% & 13\% & 4.6$\times$ & 489 & 50\% & 17\% & 2.9$\times$ \\
sigmoid & 118 & 62\% & 13\% & 4.9$\times$ & 391 & 61\% & 18\% & 3.4$\times$ \\
softplus & 110 & 54\% & 15\% & 3.7$\times$ & 377 & 63\% & 37\% & 1.7$\times$ \\
ReLU & 116 & 59\% & 20\% & 3.0$\times$ & 281 & 53\% & 20\% & 2.7$\times$ \\
set operators & 42 & 48\% & 5\% & 10.0$\times$ & 394 & 51\% & 16\% & 3.2$\times$ \\
unshaped sigmoid & 144 & 69\% & 17\% & 4.0$\times$ & 446 & 57\% & 18\% & 3.1$\times$ \\
\bottomrule
\end{tabular}

%% file: T_editops.tex
\setlength{\tabcolsep}{5pt}
\begin{tabular}{lrrrrrr}
\toprule
& \multicolumn{2}{c}{gain} & \multicolumn{2}{c}{install} & \multicolumn{2}{c}{rewire} \\
\cmidrule(lr){2-3}\cmidrule(lr){4-5}\cmidrule(lr){6-7}
model & unit & chan. & unit & chan. & unit & chan. \\
\midrule
baseline (GELU) & 94\% & 91\% & 32\% & 10\% & +2.939 & +0.016 \\
sigmoid & 90\% & 85\% & 0\% & 0\% & +0.002 & +0.027 \\
softplus & 99\% & 95\% & 0\% & 0\% & +0.363 & +0.102 \\
ReLU & 82\% & 75\% & 0\% & 0\% & +0.050 & +0.027 \\
set operators & 77\% & 68\% & 0\% & 0\% & +0.044 & +0.036 \\
\bottomrule
\end{tabular}

%% file: T_edit_mono.tex
\begin{tabular}{lrrrr}
\toprule
model & own scope & other scopes & random tokens & respects the name \\
\midrule
sigmoid & 1.62 & -0.05 & -0.10 & 100\% \\
\bottomrule
\end{tabular}

%% file: T_putarm.tex
\begin{tabular}{lrrl}
\toprule
& installs \emph{and} stays local & cheapest loss at which & read side, \\
model & within five percent of loss & \emph{any} edit takes & $K{=}8 \to K{=}128$ \\
\midrule
baseline (GELU) & \textbf{48\%} & 3.21 (from 3.19) & $0.598 \to -0.116$ \\
softplus & \textbf{0\%} & 9.23 (from 3.22) & --- \\
ReLU & \textbf{0\%} & never & $0.513 \to 0.139$ \\
sigmoid & \textbf{0\%} & 5.66 (from 3.30) & $0.743 \to 0.680$ \\
\bottomrule
\end{tabular}

%% file: T_comp.tex
\setlength{\tabcolsep}{5pt}
\begin{tabular}{lrrrrrr}
\toprule
& \multicolumn{4}{c}{installs \emph{and} stays local} & target & cheapest \\
\cmidrule(lr){2-5}
model & plain & corrected & thresholded & both & elsewhere & passing edit \\
\midrule
baseline (GELU) & 25\% & 32\% & \textbf{75\%} & 67\% & 1.03 & $-0.000$ \\
sigmoid & 1\% & 6\% & 4\% & \textbf{33\%} & 1.00 & $-0.000$ \\
softplus & 0\% & 9\% & 1\% & \textbf{27\%} & 1.00 & $-0.000$ \\
ReLU & 1\% & 16\% & 8\% & \textbf{36\%} & 1.00 & $-0.000$ \\
set operators & 0\% & 5\% & 6\% & \textbf{33\%} & 1.00 & $+0.000$ \\
\bottomrule
\end{tabular}

%% file: othermodels.tex
\section{Models trained by other people}
\label{sec:stock}

Everything so far has been measured on models trained for this paper. The instrument does not
depend on how a model was built, so the more useful question is what it says about models built by
other people for other reasons. This section applies it unchanged to twelve of them: seven
architecture families, three activation functions, eight organizations, and a range of $124$ million
to $7$ billion parameters. None was trained, fine-tuned or modified for this. The procedure is the
one in Section~\ref{sec:circuit}, at the same confidence floor, with the same matched random
control.\footnote{Two of the twelve, OLMo-2 and Gemma-3, normalize a sublayer's output before adding it to the residual rather than normalizing its input; an RMS norm divides by a scalar, so with that scalar held at its measured value the per-layer factor folds into the readout direction, which is the same approximation already made for the final norm.} The accounting, the removal curves, the closure, the required
graph and the sufficiency search run on all twelve; the composition, edit, margin and horizon
measurements were run on subsets, and each says which.

\begin{table}[htbp]
\centering\small
\caption{Twelve off-the-shelf models under the accounting of
Section~\ref{sec:circuit-accounting}. $n_{90}$ is the number of components reaching ninety percent
of the net; ``absolute'' is the same quantity under the absolute-value convention; ``inflation'' is
their ratio. The last column is where the median prediction changes when the top-ranked components
are removed. Components are feed-forward units and attention (head, channel) pairs pooled together,
so the totals range from $46{,}080$ for the smallest model to $630{,}784$ for the largest.}
\label{tab:stock}
\input{T_stock}
\end{table}

\begin{figure}[htbp]
\centering
\includegraphics[width=0.66\textwidth]{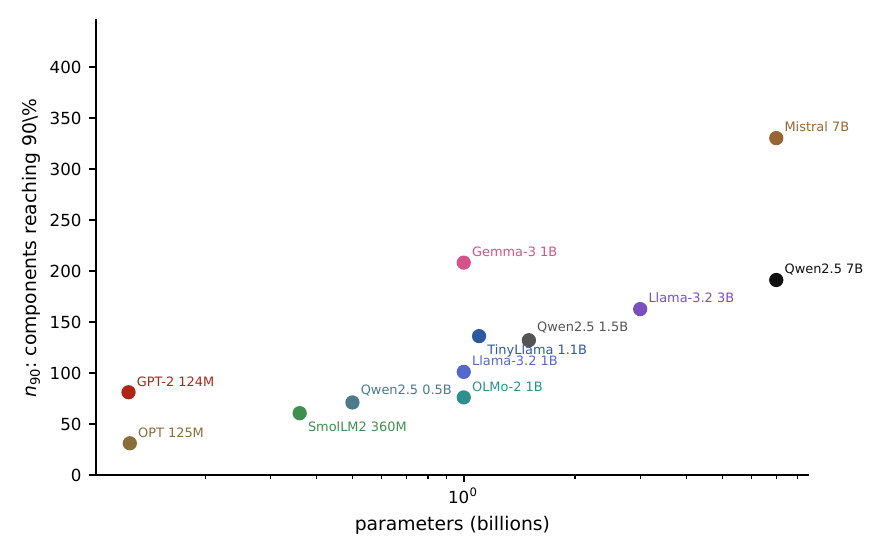}
\caption{Circuit size against parameter count, one point per model. Across a $56\times$ span in
parameters $n_{90}$ spans $10.6\times$, and the spread within a size band is wide: the three $1$B
models range from $76$ to $208$. Circuit size grows far more slowly than the model, so the
\emph{fraction} of a model carrying one prediction falls with scale. The removal curves these
points summarize are Figure~\ref{fig:flipk-stock}.}
\label{fig:stock}
\end{figure}

\begin{figure}[htbp]
\centering
\includegraphics[width=\textwidth]{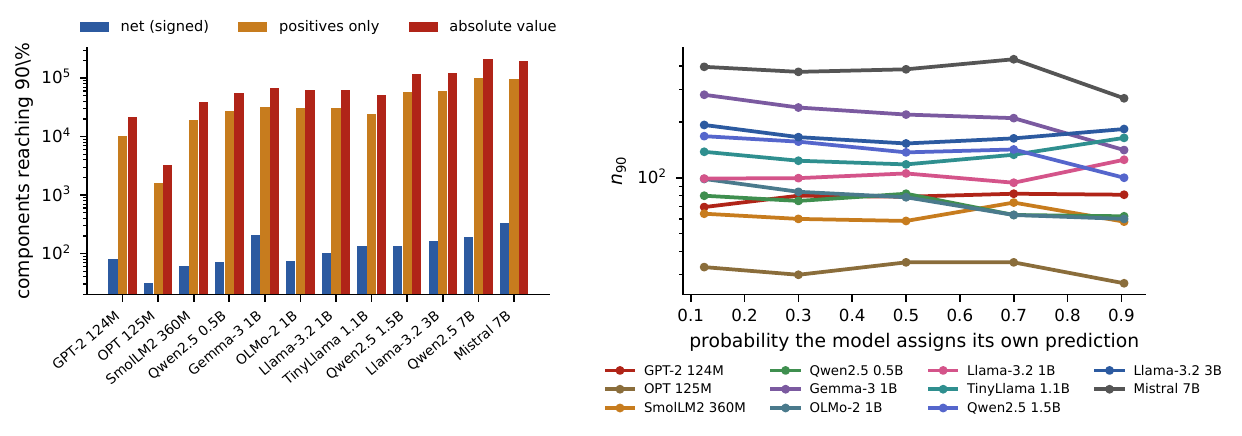}
\caption{Left: the three denominators across the twelve models, on a log axis. The gap between the
signed convention and the others is two to three orders of magnitude in every one. Right: circuit size
against the confidence of the prediction, which is why all figures here are read at one floor.}
\label{fig:accounting-stock}
\end{figure}

The accounting makes small counts visible everywhere. Every model reaches ninety percent of its prediction on a few dozen to a
few hundred components, and every model reports two to three orders of magnitude more under the
convention in common use (Figure~\ref{fig:accounting-stock}). That the same gap appears in a $124$
million parameter model from 2019 and a $7$ billion parameter model from 2024, across seven
independent implementations of the transformer, is the strongest evidence available here that it is a
property of transformers more generally.

Llama-3.2 3B reaches ninety percent on fewer components than Gemma-3 1B, a model three times
smaller (Figure~\ref{fig:stock}). Across the set the share of a model carrying a single prediction falls with
scale, from about two tenths of a percent in GPT-2 to a few hundredths in the largest models here.
On this evidence larger models are more concentrated, though the sample does not support a curve.

The activation function shows up here too, on models nobody chose for the purpose. OPT is the only stock model in the set built on ReLU, and it
has by far the least cancellation of the twelve --- close to the figure measured on the ReLU arm
trained for Section~\ref{sec:mono}, which was built and trained independently. Whatever the bounded and half-wave activations do to how much a
model's contributions cancel, they do it in models trained by other people as readily as in ours.

\begin{figure}[htbp]
\centering
\includegraphics[width=\textwidth]{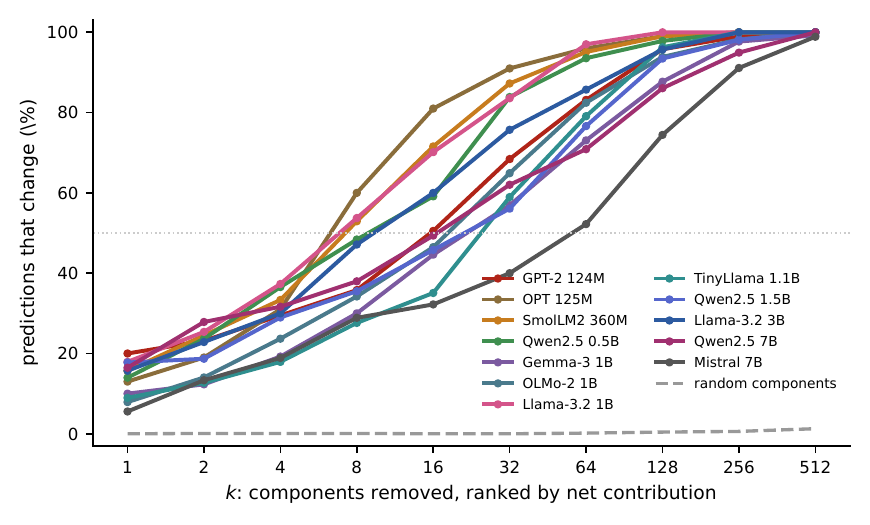}
\caption{Removing the top-ranked components changes the prediction in all twelve models; removing
the same number at random does not. The random control is pooled across the set and stays below five
percent until $k$ exceeds a hundred.}
\label{fig:flipk-stock}
\end{figure}

\subsection{Composition}
\label{sec:stock-composition}

\begin{table}[htbp]
\centering\small
\caption{What the \emph{one-hop} sets are made of, for the models where that measurement was run.
``Last quarter'' is the share of members in the final quarter of layers, against $25$ percent if
membership were spread evenly; these are shares of direct contribution, which is where a prediction
is credited rather than necessarily where it is decided. ``Seen once'' is the share of distinct
components appearing in exactly one prediction. Figure~\ref{fig:sufdepth} asks the same two
questions of the sufficient sets, which are the smaller object and cover fewer models.}
\label{tab:stockdepth}
\input{T_stockdepth}
\end{table}

\begin{figure}[htbp]
\centering
\includegraphics[width=\textwidth]{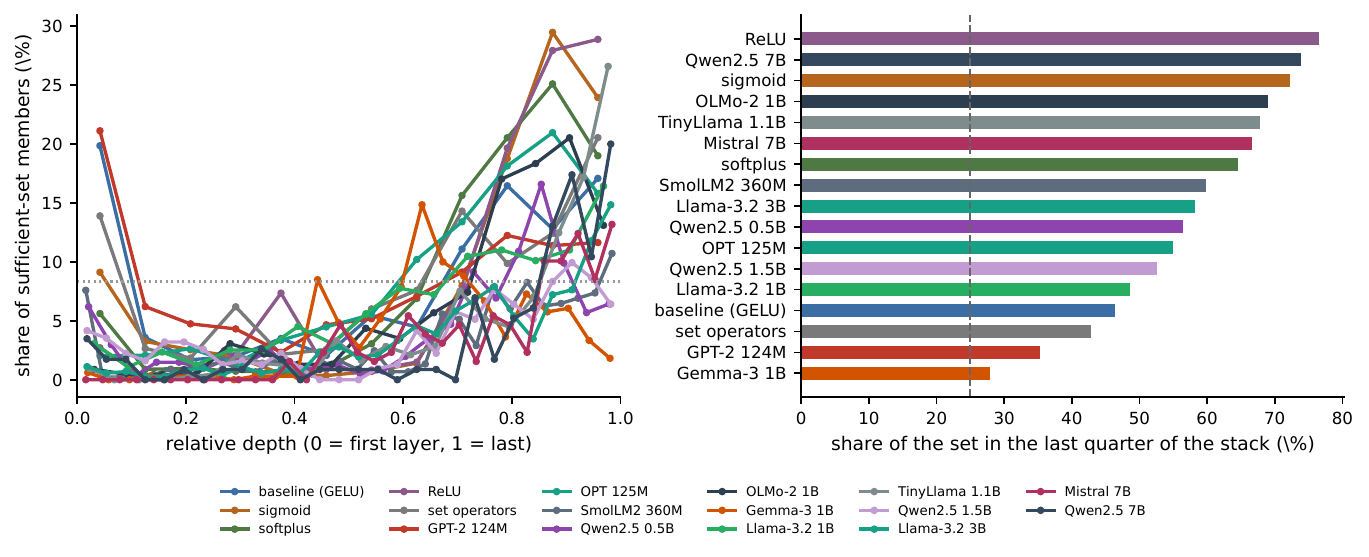}
\caption{Where the \emph{sufficient} circuits sit, across every model the search was run on. Left:
share of the set's members by relative depth, one line per model; the dotted line is what an even
spread would give. Right: the share falling in the last quarter of the stack, against the $25$
percent an even spread would put there. Members are those of the pruned sufficient sets of
Section~\ref{sec:circuit-suf}, not of the one-hop accounting.}
\label{fig:sufdepth}
\end{figure}

The circuits sit late, in every model. The last quarter of the stack holds $28$ to $76$ percent of
a sufficient set where an even spread would put $25$, and no model here falls below that line
(Figure~\ref{fig:sufdepth}). They are also mostly specific to the prediction they were found for:
between $79$ and $100$ percent of the distinct components in a model's sets appear in exactly one of
them, so a small shared remainder does the recurring work and the rest does not recur at all.

Attention supplies between $22$ and $40$ percent of the components in the one-hop sets
(Table~\ref{tab:stockdepth}), so a trace restricted to the feed-forward stack would be missing
between a fifth and two fifths of the machinery. Neither this share nor the two above sort by architecture family, by
size, or by activation: the two Llama-3.2 models sit at $34$ and $37$ percent and the two Qwen
models at $24$ and $34$, ranges that overlap. Whatever sets any of them is not visible at this sample size.

\subsection{The whole dependency graph}
\label{sec:stock-branch}

The recursion of Appendix~\ref{sec:circuit-branch} extends to these models unchanged. Every one of
them closes, at seven to ten levels, on fifty to a hundred predictions each.

\begin{table}[htbp]
\centering\small
\caption{The backward dependency graph, closed, for the twelve models trained by other people. Columns as in
Table~\ref{tab:branch-mono}.}
\label{tab:branch-stock}
\input{T_dep_stock}
\end{table}

\begin{table}[htbp]
\centering\small
\caption{The required graph for the twelve models trained by other people, defined as in Table~\ref{tab:req-mono}.}
\label{tab:req-stock}
\input{T_req_stock}
\end{table}

Tables~\ref{tab:branch-stock} and~\ref{tab:req-stock} say three things. The first is that the graph closes at all, in every model. Following ``what does this need'' from a prediction all the way back terminates on a few percent of the model's components. The share does not grow with size: Mistral-7B, sixty times the
size of GPT-2, closes on a slightly smaller fraction of itself than GPT-2 does.

The second is that the required graph is two orders of magnitude smaller than the closure, and does
not grow with the model either. A prediction from a seven-billion-parameter model can be changed by
removing thirty-four components. Removing the closure changes the prediction almost always, where
removing a size-matched random set almost never does.

\begin{figure}[htbp]
\centering
\includegraphics[width=0.80\textwidth]{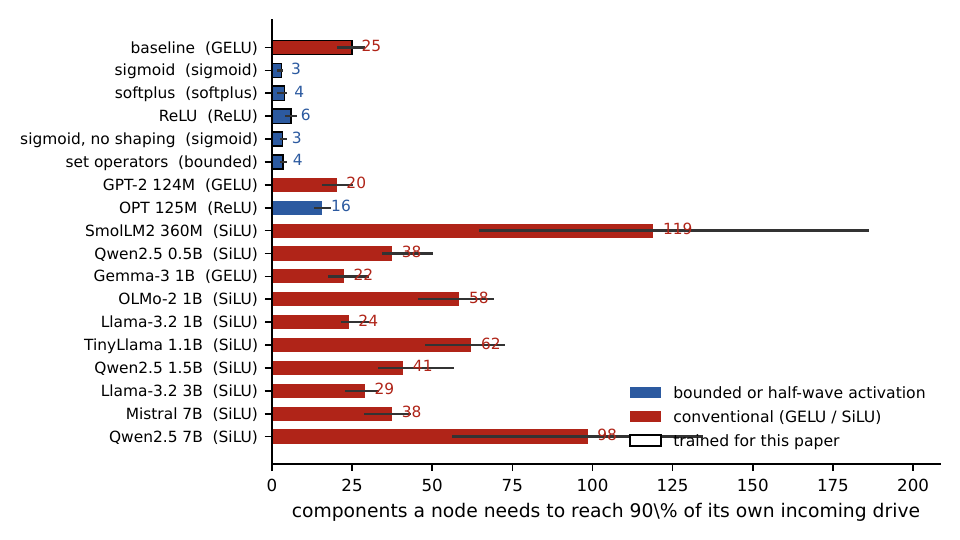}
\caption{Branching factor with bootstrap intervals, every model in this paper. Outlined bars are the
models trained here, which differ from one another only in the feed-forward unit; the rest hold
nothing fixed.}
\label{fig:branch}
\end{figure}

The third is that the branching factor in the wild does not sort by activation function. It runs from
$16$ to $119$, almost an order of magnitude among conventional activations alone, and the intervals
interleave (Figure~\ref{fig:branch}): OPT's ReLU at $16$ $[14, 18]$ overlaps GPT-2's GELU at $20$
$[16, 25]$, and Llama-3.2 1B's SiLU at $24$ $[22, 30]$ overlaps Gemma-3's GELU at $22$ $[18, 30]$. The controlled comparison of
Section~\ref{sec:mono-branch} holds everything but the feed-forward unit fixed and finds a clean separation; these
models hold nothing fixed and show none. Of the two statements about the same quantity, only the first isolates the activation function. Nothing here supports reading a branching factor off a
model's activation function without the rest of the training held fixed.

\subsection{The smallest sufficient set across models}
\label{sec:compare-suf}

Section~\ref{sec:circuit-suf} measures, on the baseline, the smallest set of components that
reproduces a prediction when everything else at that position is zeroed. The measurement depends on neither the construction nor who trained the model.

\begin{table}[htbp]
\centering\small
\caption{Sufficiency on GPT-2, one of the twelve. ``Keep-more'' is the rise in preservation from
keeping nothing to keeping $2{,}048$ components, the check that keeping more helps.}
\label{tab:closure-stock}
\input{T_closure_stock}
\end{table}

Sufficiency reads the same way here as on the models trained for this paper: the closure does not
reproduce a prediction reliably and a same-sized set taken further down the one-hop ranking does
better (Table~\ref{tab:closure-stock}, in full in Appendix~\ref{sec:appgraph}). What does reproduce a
prediction is the small set of Section~\ref{sec:circuit-suf}, which behaves on these models
as it does on the ones trained here.

\begin{table}[htbp]
\centering\small
\caption{The smallest set that reproduces a prediction, across models. Columns as in
Table~\ref{tab:suf}. ``Floor'' is the share of predictions the token embedding alone already
returns; those are excluded from the columns to its right, because no set kept on top of the floor
can be credited with the answer. ``Converged'' is how many of the sampled predictions Algorithm~\ref{alg:suf} verified, and the size
and the controls are over those.}
\label{tab:suf-all}
\input{T_recsuf}
\end{table}

Every model the search was run on has such sets, of two to twenty-one components out of tens of thousands (Table~\ref{tab:suf-all}). The range holds from $124$ million parameters to seven billion, and the largest models sit at
its small end: Mistral 7B and Qwen2.5 7B rest on two components each, three parts in a million of
the model.

The search does not finish on every prediction. The repair step is given a fixed number of rounds,
and where it runs out the kept set never becomes sufficient and there is nothing to report. Those predictions are counted in the ``converged'' column and excluded from everything to its right. The measurement covers all six models trained here and all twelve taken off the shelf. The controls behave
the same way everywhere too. A set of that size drawn at random reproduces the prediction on
\emph{no} prediction of any model --- not one --- so whatever these sets are doing is not a property
of their size, and not an artifact of a test that deletes the competition along with everything else.
The plain contribution ranking at the same size manages $0$ to $22$ percent, which is the closest any
control comes.

\subsection{Naming a component and acting on the name}
\label{sec:stock-edit}

Saying what a component is \emph{for} is a stronger claim than identifying which components carry a prediction, and the way to test it is to act on the name and see where the consequences land.
A component's parameters name a scope with nothing fitted: the tokens its write column promotes,
read through the unembedding. Amplifying the component and asking which logits move then has a right
answer fixed in advance.

Dropping the sign that scope carries loses the result on the gated models. A component's effect on the logits is $a_u c_u$ rather than $c_u$. In a gated feed-forward the write-site
value is $\phi(W_g x) \odot (W_v x)$ and the value branch is linear and unbounded, so $a_u$ is freely
signed: on Qwen2.5-1.5B and Llama-3.2-1B, $47$ and $48$ percent of the components carrying a
prediction fire negative. For those, reading the scope off $c_u$ alone inverts the targets, so amplifying the component moves the named tokens the wrong way. Table~\ref{tab:sign} shows the
consequence, which is not subtle: read without the sign, edits on negative-firing components land
correctly $4$ to $11$ percent of the time, and the models as a whole score barely above chance.
Ungated models are untouched, because almost none of their components fire negative --- zero percent
in GPT-2. Reading the scope with the sign the component actually has fixes it everywhere.

\begin{table}[htbp]
\centering\small
\caption{What the sign of the activation is worth. ``Fire negative'' is the share of
prediction-carrying components whose write-site value is negative. The middle columns score edits
against a scope read from the write column alone; the last reads it with the activation's sign. The
failure is confined to gated feed-forwards and to exactly the components that fire negative.}
\label{tab:sign}
\input{T_sign}
\end{table}

\begin{table}[htbp]
\centering\small
\caption{An edit installed in one component, scored against the scope its own parameters name and
against the scopes of the other components edited in the same predictions, at matched strength. The
last column is the share of edits moving their own scope more than another component's, across all
strengths and all edits.}
\label{tab:edit}
\input{T_edit}
\end{table}

With the scope read correctly, Table~\ref{tab:edit} reports the test on the three models taken off
the shelf it was run on. In every one, an edit moves what the parameters said it would and leaves the
other components' targets alone, in $94$ to $97$ percent of cases --- $97$ on the baseline
(Table~\ref{tab:edit-baseline}) and $100$ on the sigmoid model (Table~\ref{tab:edit-scope-mono}). This asks the parameters to predict the consequence of an intervention rather than merely to rank components.

\subsection{Where the instruments are stressed}
\label{sec:stock-instrument}

Two of the choices made in Section~\ref{sec:circuit} looked like bookkeeping on the model they were
demonstrated on. Across twelve models they are not, and this is where that shows.

\begin{table}[htbp]
\centering\small
\caption{The raw logit drop against the margin, on models where they diverge. Sign disagreement is how
often the two disagree about the direction of an effect; non-monotone steps count the times the median
drop \emph{falls} as more components are removed. Target logits span two orders of magnitude across
these models, so raw drops are not comparable between rows either.}
\label{tab:margin-stock}
\input{T_margin_stock}
\end{table}

The raw drop disagrees with the margin about the direction of the effect in up to forty-five percent of
cases, and in two models it is not monotone in $k$, reporting the impossibility that removing more components did less damage. Both failures are invisible on the baseline. Numbers elsewhere in
this paper would be impossible without the correction: measured in raw logits, the ratio of realized
to predicted effect is negative for GPT-2.

\subsection{What the constructions are worth}
\label{sec:compare-standing}

\begin{figure}[htbp]
\centering
\includegraphics[width=\textwidth]{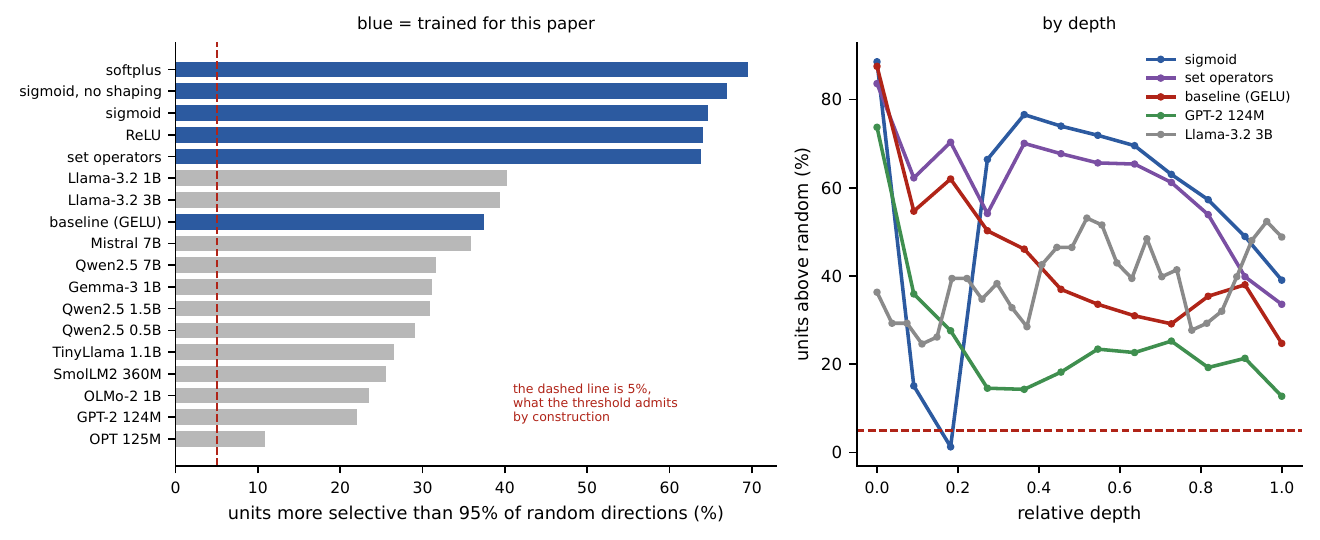}
\caption{Left: the share of feed-forward units whose response to candidate tokens is more
concentrated than $95$ percent of random directions in the same layer. The dashed line is the $5$
percent such a threshold admits by construction, so it is the level a model with no token-selective
units would show. Right: the same quantity against relative depth. The constructions trained here hold their level through the body of the network, and the
conventional baseline declines with depth.}
\label{fig:sel}
\end{figure}

The twelve models above show the lens works on models other groups have trained; what the
constructions of Section~\ref{sec:mono} add is a population that differs sharply by construction, in
the direction this paper argues. Figure~\ref{fig:sel} and Table~\ref{tab:sel} show it across every
model in the paper. The five constructions trained here for legibility all sit well above their own
GELU baseline, which matches them in width, data and schedule, and the models trained by other
people sit lower still --- most of them below that baseline. At the median, roughly two thirds of
units are token-shaped in the models built to be read, against about a third in the rest.

\begin{table}[htbp]
\centering\small
\caption{Token selectivity by model. ``Top-8 share'' is the fraction of a unit's positive response
carried by its eight strongest tokens, and ``random'' is the same statistic for random directions in
the same layer --- the comparison that makes the first column mean anything. ``Rank recovery'' is
how well a unit's top eight reproduce its full ranking over tokens, with the random value in
parentheses. Rows are medians over layers.}
\label{tab:sel}
\input{T_sel}
\end{table}

\begin{table}[htbp]
\centering\small
\caption{Whether a circuit is worth more than an equally sized random set for the tokens that
\emph{follow} the prediction, at six horizons nested on one set of positions. Columns as in
Table~\ref{tab:horizon}; $1.00$ means it is worth no more. Measured at each model's training
context, except GPT-2 and OPT, which cap at $1024$ positions.}
\label{tab:horizon-all}
\input{T_horizon}
\end{table}

The reading of Section~\ref{sec:circuit-bookkeeping} does not depend on the construction either.
Across nine models the circuit is worth no more than an equally sized random set for the tokens that
follow: the per-prediction ratio stays between $0.73$ and $1.07$ at every horizon from sixteen to
five hundred and twelve (Table~\ref{tab:horizon-all}).

The lens transfers unchanged. The gap between the signed count and the conventions in common use
is two to three orders of magnitude in every implementation from 2019 to 2024; the set that builds a
prediction runs from two components to sixteen, with the largest models at the small end and the
fraction of a model behind one prediction falling with scale; on gated feed-forwards the sign of the
activation has to travel with the name, and with it an edit lands where the parameters said it
would. The constructions' extra legibility is real, and none of it is needed to read a stock model.
Conventional transformers can be read at the level of which components carry a prediction. They
are not opaque blobs of computation. They are large collections of interwoven circuits whose
predictions mostly cancel, carrying the architecture's own management work alongside the few
components that decide the token.

\secbarrier

%% file: T_stock.tex
\begin{tabular}{llllrrrrr}
\toprule
model & size & family & activation & preds & $n_{90}$ & absolute & inflation & flips at $k$ \\
\midrule
GPT-2 & 124M & GPT-2 & GELU & 95 & 81 \small{[62, 85]} & 21,390 & 264$\times$ & 16 \\
OPT & 125M & OPT & ReLU & 100 & 31 \small{[26, 38]} & 3,291 & 106$\times$ & 8 \\
SmolLM2 & 360M & Llama & SiLU & 102 & 60 \small{[54, 72]} & 39,802 & 658$\times$ & 8 \\
Qwen2.5 & 0.5B & Qwen2 & SiLU & 93 & 71 \small{[65, 82]} & 55,140 & 777$\times$ & 16 \\
Gemma-3 & 1B & Gemma & GELU & 130 & 208 \small{[180, 232]} & 66,695 & 321$\times$ & 32 \\
OLMo-2 & 1B & OLMo-2 & SiLU & 114 & 76 \small{[65, 85]} & 62,299 & 820$\times$ & 32 \\
Llama-3.2 & 1B & Llama & SiLU & 67 & 101 \small{[92, 121]} & 63,285 & 627$\times$ & 8 \\
TinyLlama & 1.1B & Llama & SiLU & 134 & 136 \small{[124, 153]} & 50,564 & 372$\times$ & 32 \\
Qwen2.5 & 1.5B & Qwen2 & SiLU & 107 & 132 \small{[115, 147]} & 115,729 & 877$\times$ & 32 \\
Llama-3.2 & 3B & Llama & SiLU & 70 & 162 \small{[142, 183]} & 122,212 & 752$\times$ & 16 \\
Qwen2.5 & 7B & Qwen2 & SiLU & 79 & 191 \small{[168, 216]} & 210,394 & 1102$\times$ & 32 \\
Mistral & 7B & Mistral & SiLU & 90 & 330 \small{[282, 360]} & 194,783 & 590$\times$ & 64 \\
\bottomrule
\end{tabular}

%% file: T_stockdepth.tex
\begin{tabular}{lrrrrr}
\toprule
model & layers & $n_{90}$ & last quarter & attention & seen once \\
\midrule
GPT-2 124M & 12 & 77 & 67\% & 22\% & 72\% \\
OPT 125M & 12 & 30 & 76\% & 33\% & 72\% \\
SmolLM2 360M & 32 & 59 & 93\% & 26\% & 79\% \\
Qwen2.5 0.5B & 24 & 75 & 87\% & 24\% & 77\% \\
Gemma-3 1B & 26 & 193 & 55\% & 40\% & 74\% \\
OLMo-2 1B & 16 & 64 & 92\% & 38\% & 80\% \\
Llama-3.2 1B & 16 & 95 & 85\% & 34\% & 78\% \\
TinyLlama 1.1B & 22 & 126 & 92\% & 37\% & 71\% \\
Qwen2.5 1.5B & 28 & 130 & 87\% & 34\% & 82\% \\
Llama-3.2 3B & 28 & 153 & 80\% & 37\% & 80\% \\
Mistral 7B & 32 & 327 & 81\% & 30\% & 79\% \\
\bottomrule
\end{tabular}

%% file: T_dep_stock.tex
\setlength{\tabcolsep}{4pt}
\begin{tabular}{lrrrrrrrr}
\toprule
& & \multicolumn{2}{c}{graph} & & & & \multicolumn{2}{c}{ablated} \\
\cmidrule(lr){3-4}\cmidrule(lr){8-9}
model & preds & one hop & closed & levels & \% of model & attn heads & closure & random \\
\midrule
GPT-2 124M & 95 & 81 & 984 & 7 & 2.14\% & 100/144 & 97\% & 2\% \\
OPT 125M & 100 & 31 & 700 & 8 & 1.52\% & 82/144 & 100\% & 1\% \\
SmolLM2 360M & 86 & 62 & 3822 & 8 & 3.39\% & 276/480 & 100\% & 7\% \\
Qwen2.5 0.5B & 74 & 74 & 2406 & 8 & 1.74\% & 210/336 & 100\% & 3\% \\
Gemma-3 1B & 84 & 215 & 3224 & 8 & 1.56\% & 103/104 & 99\% & 6\% \\
OLMo-2 1B & 77 & 75 & 3483 & 9 & 2.13\% & 189/256 & 99\% & 6\% \\
Llama-3.2 1B & 67 & 101 & 1831 & 8 & 1.12\% & 258/512 & 100\% & 3\% \\
TinyLlama 1.1B & 89 & 141 & 4540 & 9 & 2.69\% & 364/704 & 100\% & 6\% \\
Qwen2.5 1.5B & 73 & 127 & 4076 & 8 & 1.39\% & 250/336 & 99\% & 3\% \\
Llama-3.2 3B & 52 & 158 & 3726 & 9 & 1.18\% & 348/672 & 100\% & 2\% \\
Qwen2.5 7B & 50 & 188 & 8602 & 9 & 1.36\% & 486/784 & 100\% & 6\% \\
Mistral 7B & 56 & 356 & 11114 & 10 & 1.88\% & 620/1024 & 98\% & 7\% \\
\bottomrule
\end{tabular}

%% file: T_req_stock.tex
\setlength{\tabcolsep}{4.5pt}
\begin{tabular}{lrrrrrr}
\toprule
& \multicolumn{2}{c}{graph} & \multicolumn{4}{c}{required} \\
\cmidrule(lr){2-3}\cmidrule(lr){4-7}
model & one hop & closed & components & \% of one hop & \% of model & defined on \\
\midrule
GPT-2 124M & 81 & 984 & 12 & 15\% & 0.026\% & 92\% \\
OPT 125M & 31 & 700 & 6 & 19\% & 0.013\% & 93\% \\
SmolLM2 360M & 62 & 3822 & 8 & 13\% & 0.007\% & 97\% \\
Qwen2.5 0.5B & 74 & 2406 & 8 & 11\% & 0.006\% & 93\% \\
Gemma-3 1B & 215 & 3224 & 22 & 10\% & 0.011\% & 93\% \\
OLMo-2 1B & 75 & 3483 & 14 & 19\% & 0.009\% & 91\% \\
Llama-3.2 1B & 101 & 1831 & 7 & 7\% & 0.004\% & 100\% \\
TinyLlama 1.1B & 141 & 4540 & 26 & 18\% & 0.015\% & 99\% \\
Qwen2.5 1.5B & 127 & 4076 & 12 & 9\% & 0.004\% & 89\% \\
Llama-3.2 3B & 158 & 3726 & 14 & 9\% & 0.004\% & 100\% \\
Qwen2.5 7B & 188 & 8602 & 9 & 5\% & 0.001\% & 94\% \\
Mistral 7B & 356 & 11114 & 34 & 10\% & 0.006\% & 86\% \\
\bottomrule
\end{tabular}

%% file: T_closure_stock.tex
\begin{tabular}{lrrrrrrr}
\toprule
& \multicolumn{2}{c}{set size} & \multicolumn{4}{c}{prediction preserved keeping only} & \\
\cmidrule(lr){2-3}\cmidrule(lr){4-7}
model & one hop & closure & one hop & closure & top-$|C|$ & random & keep-more \\
\midrule
GPT-2 124M & 69 & 130 & 50\% & \textbf{71\%} & 64\% & 4\% & $+93$ \\
\bottomrule
\end{tabular}

%% file: T_recsuf.tex
\setlength{\tabcolsep}{4pt}
\begin{tabular}{lrrrrrrr}
\toprule
& & & & & \multicolumn{2}{c}{same size, other sets} & \\
\cmidrule(lr){6-7}
model & preds & floor & converged & size & top-$k$ & random & \% of model \\
\midrule
baseline (GELU) & 150 & 12\% & 115 & \textbf{8} & 7\% & 0\% & 0.017\% \\
sigmoid & 150 & 12\% & 86 & \textbf{12} & 6\% & 0\% & 0.026\% \\
softplus & 150 & 12\% & 102 & \textbf{10} & 6\% & 0\% & 0.023\% \\
ReLU & 150 & 12\% & 125 & \textbf{13} & 2\% & 0\% & 0.028\% \\
sigmoid, no shaping & 150 & 12\% & 102 & \textbf{15} & 2\% & 0\% & 0.033\% \\
set operators & 150 & 12\% & 101 & \textbf{21} & 1\% & 0\% & 0.076\% \\
GPT-2 124M & 150 & 5\% & 131 & \textbf{6} & 1\% & 0\% & 0.013\% \\
OPT 125M & 150 & 2\% & 141 & \textbf{16} & 7\% & 0\% & 0.035\% \\
SmolLM2 360M & 150 & 5\% & 97 & \textbf{3} & 8\% & 0\% & 0.003\% \\
Qwen2.5 0.5B & 150 & 1\% & 71 & \textbf{4} & 15\% & 0\% & 0.003\% \\
OLMo-2 1B & 150 & 0\% & 100 & \textbf{2} & 19\% & 0\% & 0.001\% \\
Gemma-3 1B & 150 & 0\% & 134 & \textbf{2} & 22\% & 0\% & 0.001\% \\
Llama-3.2 1B & 150 & 0\% & 60 & \textbf{6} & 2\% & 0\% & 0.004\% \\
TinyLlama 1.1B & 150 & 4\% & 121 & \textbf{2} & 12\% & 0\% & 0.001\% \\
Qwen2.5 1.5B & 150 & 1\% & 65 & \textbf{3} & 12\% & 0\% & 0.001\% \\
Llama-3.2 3B & 150 & 0\% & 102 & \textbf{4} & 8\% & 0\% & 0.001\% \\
Mistral 7B & 60 & 0\% & 47 & \textbf{2} & 21\% & 0\% & 0.000\% \\
Qwen2.5 7B & 60 & 0\% & 32 & \textbf{2} & 22\% & 0\% & 0.000\% \\
\bottomrule
\end{tabular}

%% file: T_sign.tex
\begin{tabular}{llrrrr}
\toprule
& & & \multicolumn{2}{c}{scope read without the sign} & with \\
\cmidrule(lr){4-5}
model & feed-forward & fire negative & all & negative-firing & the sign \\
\midrule
GPT-2 124M & ungated & 0\% & 97\% & --- & 97\% \\
baseline (GELU) & ungated & 4\% & 92\% & 0\% & 96\% \\
Llama-3.2-1B & gated & 48\% & 52\% & 4\% & 96\% \\
Qwen2.5-1.5B & gated & 47\% & 57\% & 11\% & 97\% \\
\bottomrule
\end{tabular}

%% file: T_edit.tex
\begin{tabular}{lrrrr}
\toprule
model & own scope & other scopes & random tokens & respects the name \\
\midrule
GPT-2 124M & 1.94 & 0.63 & 0.14 & 97\% \\
Llama-3.2-1B & 1.05 & 0.02 & -0.01 & 96\% \\
Qwen2.5-1.5B & 0.69 & -0.06 & -0.06 & 94\% \\
\bottomrule
\end{tabular}

%% file: T_margin_stock.tex
\begin{tabular}{lrrrrr}
\toprule
& & \multicolumn{2}{c}{sign disagreement} & \multicolumn{2}{c}{non-monotone steps} \\
\cmidrule(lr){3-4}\cmidrule(lr){5-6}
model & target logit & median & worst & raw & margin \\
\midrule
GPT-2 124M & -77.8 & 39\% & 45\% & 4 & 0 \\
OLMo-2 1B & 5.3 & 30\% & 40\% & 2 & 0 \\
Qwen2.5 1.5B & 23.6 & 5\% & 17\% & 0 & 0 \\
\bottomrule
\end{tabular}

%% file: T_sel.tex
\setlength{\tabcolsep}{5pt}
\begin{tabular}{lrrrrr}
\toprule
model & layers & top-8 share & random & units above random & rank recovery \\
\midrule
\multicolumn{6}{l}{\emph{trained here}} \\
\quad softplus & 12 & 0.120 & 0.098 & 70\% & 0.61 (0.48) \\
\quad sigmoid, no shaping & 12 & 0.118 & 0.097 & 67\% & 0.60 (0.49) \\
\quad sigmoid & 12 & 0.117 & 0.098 & 65\% & 0.63 (0.49) \\
\quad ReLU & 12 & 0.115 & 0.098 & 64\% & 0.57 (0.50) \\
\quad set operators & 12 & 0.117 & 0.097 & 64\% & 0.63 (0.48) \\
\quad baseline (GELU) & 12 & 0.108 & 0.097 & 38\% & 0.58 (0.48) \\
\addlinespace
\multicolumn{6}{l}{\emph{off the shelf}} \\
\quad Llama-3.2 1B & 16 & 0.108 & 0.098 & 40\% & 0.53 (0.44) \\
\quad Llama-3.2 3B & 28 & 0.139 & 0.125 & 39\% & 0.54 (0.45) \\
\quad Mistral 7B & 32 & 0.138 & 0.126 & 36\% & 0.49 (0.44) \\
\quad Qwen2.5 7B & 28 & 0.138 & 0.126 & 32\% & 0.50 (0.47) \\
\quad Gemma-3 1B & 26 & 0.108 & 0.098 & 31\% & 0.48 (0.47) \\
\quad Qwen2.5 1.5B & 28 & 0.135 & 0.125 & 31\% & 0.51 (0.49) \\
\quad Qwen2.5 0.5B & 24 & 0.104 & 0.098 & 29\% & 0.53 (0.49) \\
\quad TinyLlama 1.1B & 22 & 0.104 & 0.098 & 27\% & 0.47 (0.42) \\
\quad SmolLM2 360M & 32 & 0.105 & 0.098 & 26\% & 0.51 (0.47) \\
\quad OLMo-2 1B & 16 & 0.106 & 0.101 & 23\% & 0.48 (0.45) \\
\quad GPT-2 124M & 12 & 0.098 & 0.098 & 22\% & 0.50 (0.48) \\
\quad OPT 125M & 12 & 0.083 & 0.081 & 11\% & 0.44 (0.41) \\
\addlinespace
\bottomrule
\end{tabular}

%% file: T_horizon.tex
\setlength{\tabcolsep}{4.5pt}
\begin{tabular}{lrrrrrrrr}
\toprule
& & & \multicolumn{6}{c}{tokens after the prediction} \\
\cmidrule(lr){4-9}
model & preds & at it & 16 & 32 & 64 & 128 & 256 & 512 \\
\midrule
baseline (GELU) & 206 & 0.25 & 0.98 & 0.98 & 0.99 & 0.99 & 0.99 & 0.98 \\
sigmoid & 206 & -0.04 & 0.97 & 1.00 & 1.00 & 0.99 & 0.99 & 1.00 \\
softplus & 202 & 0.00 & 1.00 & 0.99 & 0.99 & 0.99 & 0.99 & 0.99 \\
ReLU & 194 & 0.05 & 1.00 & 1.00 & 1.00 & 0.99 & 0.99 & 0.98 \\
set operators & 200 & 0.08 & 0.97 & 0.96 & 0.94 & 0.91 & 0.94 & 0.83 \\
GPT-2 124M & 201 & 0.25 & 1.00 & 1.02 & 1.04 & 1.07 & 1.07 & 1.07 \\
OPT 125M & 213 & 0.08 & 0.94 & 0.95 & 0.94 & 0.94 & 0.95 & 0.95 \\
SmolLM2 360M & 240 & 0.70 & 0.73 & 0.83 & 0.86 & 0.87 & 0.83 & 0.80 \\
OLMo-2 1B & 245 & 0.88 & 1.00 & 0.99 & 0.99 & 0.99 & 0.99 & 0.90 \\
\bottomrule
\end{tabular}

%% file: related.tex
\section{Related work}
\label{sec:related}

Every ingredient this paper uses has been built before. What differs from prior work is one accounting choice and the object it is applied to. This section says which parts are inherited and from where.

\paragraph{Reading a component from its parameters.} Projecting a feed-forward component's write
into vocabulary space is established practice. \citet{geva2021kv} characterize feed-forward layers as
key--value memories, and \citet{geva2022promoting} read their value vectors as promoting concepts in
the vocabulary; \citet{dar2023embeddingspace} treat the parameters themselves as objects in embedding
space. The logit lens \citep{nostalgebraist2020logitlens} and the tuned lens
\citep{belrose2023tunedlens} apply the same projection to intermediate residual states, and
\citet{bommasani2020static} distill static token representations by averaging contextual ones. Of the layer-native table of Section~\ref{sec:circuit}, a member of this family, only the substitution protocol is specific to this work.

\paragraph{Decomposing a prediction.} The residual stream framework of
\citet{elhage2021framework} makes a logit an exact sum of per-component terms, which is what allows
a prediction to be attributed at all. Two lines qualify what that sum means. \citet{janiak2023dlaadversarial}
show that components write in opposing directions, measuring in a four-layer trained transformer one
head whose output is about ninety percent removed downstream and a correlation of $-0.70$ between
writing and erasing components. They propose no corrected attribution metric, recommending instead
that direct attribution be paired with activation patching, which is the procedure this paper
adopts. Separately, \citet{mcgrath2023hydra} and \citet{rushing2024selfrepair}
show that ablating a component causes others to compensate. Together these predict that a linear
attribution will overstate what removing a component actually does, which
Table~\ref{tab:linear} measures directly. \citet{mcdougall2023copysuppression} describe attention
heads whose function is to suppress a token the model is about to copy. Appendix~\ref{sec:pit-opposing}
asks whether suppression of that kind is visible above its floor in the models here, and finds the
negative side of a token's accounting indistinguishable from a control token's.

\paragraph{Prediction-specific graphs built backward.} The recursion of
Appendix~\ref{sec:circuit-branch} --- start at the prediction, keep what feeds it above a threshold,
repeat --- is a shape that already exists. \citet{ferrando2024flowroutes} build exactly such a graph
top-down for a single prediction, with nodes as token representations, edges as whole operations
(an attention head, a feed-forward layer), and importance given by an ALTI proximity score that is
non-negative by construction. \citet{abnar2020attentionflow} aggregate attention across layers as a
maximum flow, choosing that formulation because paths through a transformer share edges and summing
path weights would double-count. Neither the backward construction nor the observation that paths
overlap originates here.

Three things differ in what this section measures rather than in what it constructs. The
unit is a single component --- one feed-forward unit, one attention channel --- rather than an
operation or a token representation, so an edge relates two components rather than two layers. The edge
weight keeps its sign, which is the distinction Section~\ref{sec:circuit-accounting} is about; a
score defined as non-negative, or a capacity in a flow network, is the convention that section
argues changes the answer by orders of magnitude. And the object reported is the \emph{size} of the
graph and what survives removing it: that the recursion reaches a fixed point, on how many
components, what fraction of the model that is, and which part of it a prediction cannot survive
losing. \citet{ferrando2024flowroutes} report routes and visualizations rather than sizes, and
validate against previously published patched circuits rather than by ablating their own graphs.

\paragraph{Circuit discovery.} Automated methods locate the components responsible for a behavior:
ACDC prunes a computational graph \citep{conmy2023acdc}, edge attribution patching approximates that
search with gradients \citep{syed2023eap}, and AtP* makes the localization efficient at scale
\citep{kramar2024atp}. The reported sizes live in different bases and are not directly comparable to
one another. \citet{wang2023ioi} identify twenty-six attention heads for indirect object
identification; \citet{marks2024sparsefeature} report on the order of a hundred sparse-autoencoder
features, and separately about fifteen hundred raw neurons, under dataset-level faithfulness and
completeness measured on a mean-ablation baseline; the
attribution graphs of \citet{ameisen2025circuittracing} operate on cross-layer transcoder features
and prune a representative graph from 236 nodes to 55. Those graphs rank paths by the absolute value
of edge weights, and the pipelines built on them accumulate absolute influence against its total.
This paper keeps the sign and divides by the net. That normalization is not new:
\citet{chen2026unpack} call the signed-sum share the standard one, and \citet{salek2025leastresistance}
take the net change in the output as the reference quantity an attribution should sum to, noting that
when attributions cancel their absolute total exceeds it. What
Table~\ref{tab:accounting} adds is the consequence for circuit size, which does not appear to have
been measured: on the same predictions and the same attribution, the conventions differ by two to
three orders of magnitude in how many components reach ninety percent.

\citet{chen2026unpack} also supply the sharpest caution. Their credit is propagated recursively
through a multi-layer path expansion, where a small signed denominator compounds with depth, and they
report that a pure signed denominator degrades their ranking to chance; their remedy floors its
magnitude at a fraction of the absolute mass. The quantity here is a single sum over components at one position, with no recursion. A signed denominator is demonstrably unsafe once it is iterated.

\paragraph{The neuron basis.} Whether raw neurons are an adequate basis is unsettled.
\citet{ameisen2025circuittracing} report that thresholded neurons underperform their transcoder
dictionaries on interpretability metrics, which is part of the motivation for working in a learned
basis at all. \citet{arora2026neuronbasis} reach the opposite conclusion, reporting that
approximately one hundred MLP neurons suffice to control behavior on a subject--verb agreement
benchmark and that neuron-basis circuits are as sparse as feature-basis ones. They also trace a
neuron circuit for a single prompt on an off-the-shelf model, narrowing 257 attributed neurons to 23
by manual curation and steering, and identify one neuron that flips the top output. That is the
closest published work to Section~\ref{sec:circuit}. It differs in what is asked of the set: their
nodes are selected by a per-component attribution threshold and validated by steering, where the
sets here are required to change the prediction when removed and are read against a control that
removes the same number of components at random. Their attribution also targets a sum over several
candidate tokens rather than one logit, and does not place attention on the same footing as
feed-forward units.

\paragraph{What ablation replaces.} What a removed component is replaced with is a consequential choice. \citet{li2024optimalablation} show that zero, mean, and resample
baselines can differ by a factor of three on the same components, attribute the gap to the
replacement value carrying information of its own, and propose an optimal constant instead. Circuit
evaluation toolkits accordingly report several baselines rather than committing to one. This paper
takes the same care and finds an asymmetry: the necessity measurement is nearly unchanged across the
three counterfactuals while the sufficiency measurement moves severalfold
(Table~\ref{tab:asymmetry}), which is why the claims here are necessity claims. A related choice is
what a component's \emph{constant} write counts as. \citet{sun2024massiveactivations} show that a few
activations of very large and nearly fixed magnitude act as bias terms in the residual stream, and
\citet{xiao2024attentionsinks} that heads park their attention on early tokens whose values then add
a constant at every position; either makes a component's mean write large. Appendix~\ref{sec:pit-opposing} reruns its tests
with that constant part removed, and has a seed where it dominates the signed sum.

\paragraph{What the feed-forward stack is for.} A line of work asks whether the feed-forward
layers are load-bearing at all. \citet{elhage2021framework} study attention-only transformers as a
tractable simplification; \citet{sukhbaatar2019augmenting} fold the feed-forward computation into
attention as persistent memory and lose little; and against that,
\citet{dong2021attention} show that attention alone degenerates --- without the skip connections and
feed-forward layers, a stack of pure attention loses rank doubly exponentially in depth.
\citet{geva2021kv} give the feed-forward layers a positive role as key--value memories.

Section~\ref{sec:circuit-suf} speaks to this from an unusual direction. The question here is how much of an existing stack one prediction needs. The answer depends on depth. The components that reproduce a
prediction sit mostly in the last third of the stack, $43$ to $81$ percent of members, and the
middle is the thinnest region on eight of the ten models measured, holding $12$ to $55$ percent.
Attention's share of what is kept is nevertheless larger through the middle than at the top on most
of them, $22$ to $54$ percent there against $19$ to $35$ percent in the last third. For one token,
most of the middle of the stack is not among the components that rebuild the answer. That is consistent with the memories reading of \citet{geva2021kv}, since a memory is
addressed on some inputs and not others, and it does not contradict \citet{dong2021attention},
whose result is about what a stack can represent rather than what one prediction consumes.

\paragraph{Editing a component.} Model editing installs an association a model does not hold, and
its methods split the job the same way this paper does: choose a site, then compute what to write.
\citet{meng2022rome} locate the site by causal mediation --- corrupting the subject tokens and
restoring hidden states one at a time --- and obtain the update by optimizing a value vector against
a target probability, normalized by an activation covariance estimated over a large text sample.
MEMIT and its successors inherit those traced layers rather than choosing their own, and the
knowledge-neuron line \citep{dai2022kn} scores neurons by integrated gradients. The construction
they share descends from \citet{bau2020rewriting}, who treat a layer as a linear associative memory
and derive the covariance-weighted rank-one update that \citet{meng2022rome} inherit.
\citet{huang2023tpatcher} take the closest structural approach to the one here, adding a single
neuron per correction, with that neuron trained by gradient rather than written. Every one of these
reads the site off what the model \emph{does} on some input.

A separate line does read a component's function from weights alone. \citet{avrahamy2026rotate}
rotate a unit's weight matrices into vocabulary space with no forward passes, and
\citet{dunefsky2024transcoders} perform circuit analysis through feed-forward sublayers with an
explicitly input-invariant component. Section~\ref{sec:edit-install} contributes the conjunction of
the two lines: a site read from the weights, an association installed at it, and the result scored
on a criterion an edit can fail.

The write half is where the overlap is closest. \citet{dai2022kn} already update a unit's output
column in closed form using token embeddings, and \citet{hakimi2026ove} do so through unembedding
columns, explicitly a per-unit version of the rank-one mechanism. Section~\ref{sec:edit-install}
takes that construction as given and contributes the other half: a site chosen from the read row of
Section~\ref{sec:naming-answer}, decoded in the layer's own frame, with nothing traced and nothing
fitted.

Two results from the critique line bear on whether that is worth having.
\citet{hase2023localization} find that where causal tracing localizes a fact predicts almost nothing
about where editing it succeeds, which makes the expensive step of the standard pipeline hard to
justify on its own terms. And \citet{yang2025mirage} show that headline success rates fall sharply
under evaluation conditions closer to deployment, which is a caution this paper's own edit numbers
inherit: they are measured on single edits, at one position, on a model of $124$ million parameters.
One more thing is worth noting about scope. This literature edits feed-forward layers
predominantly, with \citet{liu2025intattn} an exception that writes to attention parameters as well,
motivated by evidence that attention carries a substantial share of factual storage
\citep{wei2024localization}. Section~\ref{sec:edit-install} finds a reason for that beyond convention: what
attention writes at a position is a mixture over other positions, so an association installed in a
channel cannot be made local. The restriction does not extend to a gain, which attention takes about
as reliably as a feed-forward unit does, and Section~\ref{sec:place-circuit} shows that it does not
extend to an installed head used as a marker for a unit downstream.

Writing a transformer's weights by hand rather than training them has its own line.
\citet{lindner2023tracr} compile programs into transformer weights, producing a model whose
mechanism is known by construction, and \citet{vergarabrowne2025tracrinj} place an algorithm inside
a pretrained model's residual stream by distillation. Section~\ref{sec:place-circuit} writes two components into an otherwise untouched model trained by gradient descent.

\paragraph{Legibility by construction.} A separate line changes the model rather than the analysis.
\citet{gao2025weightsparse} train transformers with most weights constrained to zero and recover
circuits small enough to reverse-engineer, verified as both necessary and sufficient, at a cost in
capability; \citet{tamkin2023codebook} replace activations with entries from a learned discrete
codebook. The constructions in Section~\ref{sec:mono} pursue the same
goal along a different axis, changing what a feed-forward unit does with its pre-activation while
leaving connectivity dense, and are evaluated here by the same instrument as the unmodified
baseline.

\secbarrier

%% file: discussion.tex
\section{Discussion}
\label{sec:discussion}

On every model measured here the same three questions have answers --- which components carry a
prediction, which it cannot survive losing, and which set builds it --- and the numbers that come
back are small. What follows is what that makes possible and what it asks.

\paragraph{Reading.} The first consequence is that ``what is this model doing here'' becomes a
measurement with an error bar. A model that fails on one input can be audited for the components
that produced the failure; a capability can be located before anyone argues about whether it exists;
and the circuit for one token, traced through two architectures, can be compared.

\paragraph{Writing.} An edit installed on a named component moves what the name says it should and
leaves other components' targets alone, and that is a different operation from fine-tuning.
Fine-tuning changes a model by showing it data and accepting whatever internal change follows; an
edit of this kind changes a stated thing and can be checked against the statement.

Where an edit can be placed is a property of the architecture rather than of the component chosen
--- a write is lost to depth, not to distortion --- and it leaves an editor a choice it has not
usually been offered: write in the model's own frame so that components already present respond, or
write in a frame nothing uses and supply both ends of the circuit. The second is cheap for a reason
worth generalizing, since the space a trained model operates in occupies a modest fraction of its
width, leaving room for channels that collide with nothing. How much room, and whether it shrinks
with scale, is left for future work.

The tap of Section~\ref{sec:place-tap} is the direction with the longest reach. Driving a component
the model trained for itself, from two layers upstream, using a read row taken from that component's
own weights, turns a legibility claim into a control surface. The limit measured here is the
accuracy with which that read row can be estimated rather than anything the model imposes, which
suggests the constructions get better with better instruments rather than requiring a different
architecture.

An attention channel refuses an installed association for a structural reason rather than a tuning
one: what a channel writes at a position is gathered from other positions, so the edit cannot be
made local. That explains as a constraint rather than a habit the pattern in the editing literature
of writing predominantly to feed-forward layers. The restriction is specific to installing content
into a channel; a gain on a channel's existing scope works about as well as a gain on a unit's, and
an installed head serving as a marker for a unit downstream carries a condition that neither
component holds alone.

The forward-looking version of all of this is a model that can be programmed and not merely
trained. New skills, factual updates, changes to behavior that would ordinarily require retraining
can be installed; if the component carrying the behavior can be named and the edit's reach measured,
the collateral becomes a quantity rather than a surprise. The measurements here are a beginning:
they cover single components and single predictions.

\subsection{What the carriage result asks of the architecture}

To read a component the lens has to separate predictive work from carriage, and what it separates
out is most of the model. At any position the great majority of contribution mass cancels, leaving a few dozen components to carry what survives. The canceling mass is neither noise nor waste.  It is moving information into the few places that use it, and keeping a superposed residual stream
organized so that many things can share it.

That reframes the efficiency question. The usual form is ``which computations can be skipped'', and
the honest answer from these measurements is that the surplus is not idle, so skipping it is not
free. The better form is the one this instrument makes askable: how much of a
model's capacity goes into carriage rather than into deciding, is that ratio a constant, and is it a
property of the architecture or of the training? The evidence here says it is not a constant. The
arms trained for this paper share width, data, schedule and converged quality and differ only in the
activation function, and their prediction graphs differ by a factor of four to seven. Whatever determines how much machinery a prediction needs is a design variable.

That raises the question of whether the carried part is compressible. A fixed operator reproducing $74$ to $79$
percent of the update needs neither attention nor a feed-forward block to supply it, and a layer
whose sublayers only had to produce the remainder could be considerably thinner. What the
measurement says against this is specific: carriage and content are not separable
subspaces. The prediction lives inside the span the carriage map acts on, and a node's write column
lands there too, so factoring the map out of the architecture is not the same operation as factoring
it out of the algebra. The compressible object is the shared \emph{map} rather than a \emph{subspace}. That distinction is what a serious attempt would have to respect, and
it is why the obvious version --- project the residual into a carriage subspace and a content
subspace, and give each its own machinery --- is the version that already fails.

Redundancy is the other half of that question. Started from the required graph and started from a random draw, the same search reaches sufficient
sets with \emph{no member in common} on a fifth of predictions (Section~\ref{sec:circuit-suf}), so
training leaves more than one set of components able to carry the same answer. What that costs, and whether a model could be built to carry less of it, is what
the carriage asks in a different currency. The caution from Appendix~\ref{sec:pit-ablate} applies
to any attempt: keeping only a pruned graph is not monotone in how much is kept, so an instrument
that trims redundancy has to be checked against that failure before its numbers mean anything.

This instrument can pose a stronger version of the same thought. A component
plausibly does double duty: a little prediction and a great deal of carriage. If that is right, and
if carriage admits many equivalent implementations, then where a design puts its parameters should
matter less than how many it has, because the carriage absorbs the difference while the deciding part
stays small and much the same. Something like that is what the field observes --- models of a given
size land in a narrow band on downstream tasks across quite different designs --- and it has not
had a measurement attached. Two measurements bear on it. The closure sizes of
Section~\ref{sec:stock} are $1.1$ to $3.4$ percent of a model's components across seven architecture
families, with no trend in size. And about three quarters of the residual update is predictable from
the state before it (Section~\ref{sec:circuit-bookkeeping}), which bounds how much of a model's work
between depths could be carriage. If the carrying is that large a share and admits many
implementations, the deciding part is what little is left, and it would take a great deal of design
difference to move it.

If that holds beyond the comparison made here, several things follow.
A measured per-prediction dependency graph could replace the proxies for importance that currently guide conditional computation and early exit. Architecture search optimizes quality against
parameters and latency; the size of a prediction's circuit is a third axis, it is now measurable,
and nothing says the three are aligned. And an architecture whose carriage is cheaper --- one that
needs less machinery to get information to the components that use it --- would show up on this instrument as a smaller closure at equal quality, which is a target a
designer can aim at directly.

One reading of all this is uncomfortable. A model that spends most of each layer moving information rather
than computing with it may simply be inefficient, and the architecture may be the reason. The opposing reading is at least as plausible and follows from the same data. Carriage is everywhere
because putting it everywhere is what makes the residual readable from anywhere, so whichever few
components a prediction happens to need can find their inputs without the model having known in
advance which those would be. On that account the carriage is not overhead but the price of a
general-purpose bus, and a design that spent less on it would need to know its circuits ahead of
time. Distinguishing the two means building something that carries less and seeing what it loses,
which is an architecture question this instrument can only set up.

\subsection{What a one-component circuit says about the vocabulary}

Some predictions rest on a single component. On Gemma-3 1B, $49$ of the $134$ predictions
Algorithm~\ref{alg:suf} verifies are carried by one component out of $206{,}336$ --- $43$ of them a
feed-forward unit and six an attention channel. On $31$ of the $49$ the contribution ranking's own
top choice does not reproduce the token, so on those the component is found rather than nominated.
The sample examined below is the prefix-and-prune search's draw on the same model, $37$ predictions of $114$,
and what those units predict is mostly punctuation and function words: eleven of the thirty-seven
write a mark rather than a word.

That a comma can be produced by one unit invites a question about what the unit is for, and the sets
answer it differently in different cases. When each unit is transplanted to another position with the same
target token, the general ones carry it and the specific ones do not: a period transfers on eight of
twelve attempts and \emph{in}, \emph{and} and \emph{a} transfer on nine of ten, against a control in
which a component drawn at random carries no prediction on any model. Commas do not transfer at all.
The two comma units in this sample are distinct components and neither produces the other's comma, the behavior of a unit meaning ``\emph{this} word takes a comma'' rather than one meaning ``a comma belongs here''.

If that reading survives a larger sample --- and two units are an anecdote, not a measurement --- it
bears on where a vocabulary should put its boundaries. Byte-pair encoding forbids merges across
whitespace and punctuation, and the work that lifts that restriction \citep{superbpe, boundlessbpe}
argues for it from compression and downstream accuracy. None of the models measured here uses such a
vocabulary: across the six tokenizers checked, spanning $32{,}000$ to $262{,}145$ entries, not one
token joins a word to a following mark. A word-specific punctuation unit is what a model learning that association in parameters looks like. That is evidence of a
different kind from compression: the model has spent a component on knowing that a particular word
takes a particular mark, which is exactly what a merged token would have supplied for free. The
measurement here is small and one-sided, but it is the sort of evidence that has been missing, and
this instrument produces it directly --- for any candidate merge, ask whether the model devotes
components to the association, and whether those components are specific to the pair.

\subsection{Where the components have no name}

Which components carry a prediction can be found in every model measured. What a component is
\emph{for} is harder, and the limit is the vocabulary: a component operating on a direction no token
induces has no name there, and no lens can supply one, because the thing to be named is not lexical.
What such a component has is an address --- the components upstream that write it
(Section~\ref{sec:naming-provenance}) --- which is what the gallery prints in place of a word.
Naming those quantities is the open problem; what follows is what is known about them.

On the baseline, nearly three quarters of a unit's response
falls on tokens no concept explains, and Appendix~\ref{sec:naming-remainder} rules out six accounts of it: sampling noise,
several smaller concepts, a relational class, carriage seen from the write side, carriage seen from
the read side, and weights the gradient never reached. What survives is reproducible, single, and no
more context-bound than the part that can be named.

Superposition, the obvious remaining guess, does not survive contact \citep{elhage2022superposition}. If a token elicits a response through a direction the unit
shares with many others, the response is leakage, and leakage is shared by construction: a small
basis fitted across all units at once should account for the remainder while leaving the nameable
part alone. It does the reverse. At every rank tried, on both models, the shared basis explains the
\emph{coherent} response better than the remainder --- $68$ against $55$ percent at thirty-two
directions on the baseline. The nameable tokens are the shared structure, which is what makes them
recognizable as concepts. What is left over is idiosyncratic to the unit.

Section~\ref{sec:naming-provenance} says what that idiosyncratic part is. It is written by identifiable
components upstream, it arrives having been carried further than the nameable part, and it separates
into categories once the terms are sorted by where they came from. A component whose response is
unit-specific, reproducible, single, context-independent, and assembled from several upstream sources
is behaving like a condition with a list: fire if the input is any of these, then write the one thing
this component writes. The consequent cannot depend on which member fired, because a unit has one
write column. Under that reading the remainder is not a failure of naming. The disjuncts of a
disjunction need not resemble each other, and Equation~\ref{eq:coh} tests precisely for resemblance,
so a well-formed condition with an unlovely list scores as incoherent while being perfectly well
defined. The quarter of a unit's response that can be named is the part of its list that
happens to arrive with few enough rotations to still lie near the vocabulary.

What the categories are made of is the part this lens still does not settle. They are partly
grammatical --- part-of-speech purity within a source group exceeds a random partition of the same
sizes --- and grammatical classes are exactly what Equation~\ref{eq:coh} cannot see, since members of
one are substitutable in a position rather than close in distribution. Whether the rest is
morphological, orthographic, or positional needs a method that names \emph{features} rather than
tokens, which is what a sparse autoencoder
provides \citep{cunningham2023sparse,bricken2023monosemanticity}.

\subsection{Is this as good as it gets?}

Finding a sufficient set efficiently is an open problem. Algorithm~\ref{alg:suf} was built to find
small sufficient sets in reasonable time, not provably smallest ones, and it converges on $96$
percent of predictions on some models and $44$ on others. Every size reported here is an upper bound
from that search.  We presume someone more clever will close this gap to $100$ percent, possibly with minimally
sized circuits.

The dependency graph follows an attention channel back to what it reads at the position being
explained and no further (Section~\ref{sec:naming-attn}). What a channel contributes is that value
aggregated over the sequence, and following the pattern that does the aggregating leaves the
position being explained and opens a graph over the whole context. Every circuit in this paper is
for one position; the graph over the context is the next instrument.

The component pool is attention channels and
feed-forward units reading a sublayer's input, so architectures shaped differently need extensions
not built here: models computing their two sublayers in parallel from the same state, models
gating a third class of writer into the residual at every layer, and sparse mixtures of experts,
where a layer holds many feed-forward experts and routes a few of them per token. The first two
want a wider pool. The third asks something harder, because the router writes nothing into the
residual and is credited nothing by an accounting that scores a component by what its own write
places on the readout, while deciding which components exist at a position at all; and a circuit
quoted as a share of the model no longer has one denominator to be a share of. This paper had to
draw the line somewhere.

\secbarrier

%% file: conclusion.tex
\section{Conclusion}
\label{sec:conclusion}

This paper presents a lens: a way to look inside a transformer and say \emph{which} components made a
particular prediction. It is derived from the model's own parameters and activations, it fits nothing,
and it ran unmodified over eighteen models. What it returns is an accounting of which components
carry a prediction, which of them it cannot survive losing, and which set is \emph{sufficient} to
produce it: on the baseline fifty-three, thirteen and eight of $46{,}080$. The sufficient set is
small everywhere --- two to sixteen across twelve models trained by other people --- and far below
what the conventions in common use report for the same predictions. Every size here is an upper
bound returned by a search rather than a minimum, which leaves the smallest circuit an open question
rather than a settled one.

Identifying components is one thing and \emph{naming} them is another. A name requires knowing the
geometry the computation runs in, and much of this paper is spent there. Most of what a layer writes
is not the prediction but the management of that geometry: three quarters of a layer's update is a
fixed map of the state it received, applied whatever the model is about to say, and at the readout
the mass pushing away from the prediction is seven times the mass carrying it. Once that is separated
out, close to half of a model's components can be named by the predictions they drive, and the
strongest few inputs to a unit can be read from its own weights in the frame of its own layer. Most
of what drives a unit has no name in the vocabulary at all, and there the dependency graph supplies
provenance in place of a word: sorted by which component upstream writes it, the remainder separates
into categories a classifier recovers.

A named component can be acted on.  A gain
stays inside the scope the parameters name. An association the model never held can be installed into
one spare component with both halves read off the weights rather than traced or optimized. A
component the model trained for itself can be driven from upstream layers. And a circuit can be
built from components that did not exist before, carrying its intermediate state on a channel nothing
in the model is oriented to read and committing its answer on the token axis, which is the one frame
the readout answers. And replacing the activation function with an order-preserving one --- the one
controlled comparison here --- puts a unit's inputs at the instrument's ceiling at parity quality, at
the price of turning an install into a two-part edit.

Finding, naming, and editing, taken together, are a toolkit --- and a roadmap for
different ways to build a transformer, to train it, and to use it.

\secbarrier

%% file: gallery.tex
\section{Gallery}
\label{sec:gallery}

Two predictions for each of the eighteen models, drawn mechanically. Each prediction is shown in
full: every member of its set is a row of the table, and every edge among them is in the graph.
Nothing is trimmed to fit. Where a set is large, the table takes a page of its own and the graph a
page turned sideways; smaller ones share a page. What a table shows is the set that \emph{builds}
the prediction: keep these components, zero
every other feed-forward unit and attention channel, let the survivors recompute, and the model
emits the same token (Section~\ref{sec:circuit-suf}). Nothing is spliced in --- a component drawn
here fires from the residual that remains, which is the token embedding plus whatever the other
survivors have written. These sets come from a prefix-and-prune search rather than from
Algorithm~\ref{alg:suf}, so a panel is a verified sufficient set but is typically larger than the
medians in Table~\ref{tab:suf}; a sufficient set is an upper bound, and either search returns a
valid one.

\subsection{How to read a page}

\paragraph{The header.} The context, the token the model predicts, and four numbers: the
probability it assigns, how many components reach ninety percent of the net contribution, how many
of them its prediction cannot survive losing, and what share of the total contribution cancels.

\paragraph{The rows.} Each row is one member of the set. The unindented rows run from the components
nearest the readout down to the ones furthest from it, and beneath each, indented, are its three
strongest suppliers among the members, labeled with their share of \emph{its} own incoming drive;
the graph beneath the table carries every edge, not only those three. A chain therefore reads
down the page: a component that promotes the predicted token, then what fed it, then what fed that.
A node reached twice is written once and back-referenced, because duplicating shared ancestors makes
a prediction look like it rests on more machinery than it does. Printed in \textbf{bold} are the
members whose \emph{removal} from the intact model changes the prediction, so one page carries both
the components that build the answer and the smaller set it cannot lose. Where the two do not
coincide, a prediction's necessary components are not the ones that reproduce it --- which is worth
seeing, and is most likely a consequence of there being more than one sufficient circuit.

\paragraph{The columns.} \emph{Strength} is how hard the component fires, as a magnitude; which way
it pushes is already carried by the write list. \emph{Fires on} lists the tokens that drive it,
scored from its read row against a layer-native token table. A driver is printed in ink where it
belongs to a group the token embedding can see and in gray where it does not, which is the split of
Section~\ref{sec:naming-about} appearing on the page, token by token; a dash means the read row
names nothing. \emph{Writes toward} lists the tokens
its write column promotes, signed by how the component is firing.\footnote{The sign matters. In a
gated feed-forward the write value is $\phi(\text{gate}) \cdot \text{up}$ and the up branch is
linear, so a negatively firing unit whose column points away from the target is the one promoting
it, and an unsigned decode would report the opposite.} \emph{Answer} is where the predicted token
sits in that list, read at the readout --- the same quantity on every row, because every write list
here is quoted at the readout. A row reading \#1 promotes the answer; a row reading \#$>$999 carries
the prediction without promoting it, and is still read through its incoming edges, which say what
its inputs carry.

A row marked $\leftarrowtail$ is an attention channel, whose read side is what it picks up from the
position it attends to rather than what sits at its own. Its read side is measured from the value
projection and its incoming edges are followed the same way a feed-forward unit's are
(Section~\ref{sec:naming-attn}). What is \emph{not} followed is the attention pattern: the value a
channel reads is aggregated over the sequence by weights this lens does not read.

On an edge row the label beside the share names the tokens the writer and the reader share, meaning
tokens both rank in their first eight, up to three of them; where they share none that highly, the
edge prints its share alone. That bar is met on between a quarter and three fifths of edges,
depending on the arm. Where the reader's fires-on list is printed, it is led by the first shared
token, drawn in the gloss color. The restraint is deliberate. The agreement result of
Section~\ref{sec:naming-provenance} is a statement about the population of edges, and it does not
license reading one edge's best token as a fact about that edge. The writes-toward list is the
component's own strongest tokens at the readout and is not led by anything.

In the graph beneath each table, every component in the net at the readout, whether or not it also
feeds another component in the set, is joined to an answer node at the right, and that edge is labeled with
where the answer sits in what the writer promotes, the table's last column.

\paragraph{Glosses.} Where a component's tokens fall into a nameable category, a short gloss is
printed beside them in italics. The gloss is ours and the tokens are the model's; it sits next to
them rather than in place of them, so a reader can see what it was derived from and disagree.

\paragraph{Concepts.} An edge is drawn dashed and led by $\langle A \rangle$, $\langle B \rangle$
when writer and reader share no token at all: a quantity with no word but with an address, named
by the component that wrote it, which carries the same tag in its writes-toward column.
Tagging by source rather than with a single placeholder makes reuse visible, since one internal
quantity feeding three readers is a fact about the circuit that a uniform label hides.
Section~\ref{sec:naming-provenance} is what licenses the address --- these quantities are written by
identifiable components even where no vocabulary describes them.

\paragraph{The graph.} Beneath each table, the same components in the same relation, with depth
running left to right and the answer at the right. Every member is drawn, since every member has
at least its edge to the answer.

\subsection{How these traces were chosen}

Up to twenty predictions were drawn at random from held-out text under filters fixed in advance, and
the draw stops early where the filters exhaust the sampled contexts first --- eleven predictions on
the baseline, twenty on most of the models taken off the shelf. The model's top choice must carry
probability at least $0.30$, so there is a decision to explain rather than a coin flip; the context
must not be a run of one repeated token; and the predicted token must contain a letter, so the
sample is not dominated by punctuation. Of those twenty, the two whose sufficient set is nearest the
\emph{median} of the draw were kept. Selecting for typicality avoids flattering the claim this paper
makes about circuit size, as selecting the smallest would have. Nothing about a trace's content, or
how readable it turned out to be, entered the choice.

That holds for the first group, two predictions per model. Most of them use no concepts at all: on the
great majority of the predictions drawn this way every edge has some token that both its writer and
its reader rank, whether or not the agreement clears the bar for printing it. A prediction that needs
nothing the vocabulary cannot say is the common case, and a gallery that showed only the interesting
minority would misrepresent how often it arises. So the mechanically drawn pages come first and are
left as they fell. The predictions after them are selected \emph{because} they contain edges no
token describes, and are labeled as chosen for it.

The six models trained for this paper come first, in the order of Section~\ref{sec:mono}, followed
by the twelve we did not train, ordered by size.

\subsection{What the conventions rest on}

An edge is drawn as a concept when its writer and reader share no token at all among a few hundred
candidates, so that there is no token-level content to report and the row does not show a bad list
in the same style as a good one. That happens on a median of $2$ percent of the edges on these
pages per model, and on one model far more: OPT-125M runs at $65$ percent, and the pooled figure
across all eighteen, $11$ percent, is almost entirely that one model.

The edge tokens carry a caveat the pages cannot show. Scored against twenty random controls rather than
one, $13$ percent of the baseline's edges and $11$ percent of the sigmoid model's have a writer that
beats its controls at $p \le 0.05$; the median edge sits at $p \approx 0.35$. The aggregate claim is
solid --- across twenty-three thousand edges the real writer wins on $63$ percent against $32$,
which is not a close call --- but it rests on many weakly informative edges rather than on a subset
of strong ones, and the token on any \emph{single} edge is correspondingly weak evidence. These pages
illustrate what a trace looks like rather than claiming anything about particular edges.

Two things a reader should expect to see. On the conventional baseline the required components are
often attention channels, whose read side is what they collect from elsewhere in the sequence rather
than what sits at the position --- the boundary of Appendix~\ref{sec:circuit-branch} appearing in a
concrete trace rather than as a caveat. And a
source's share of a node's drive can exceed one hundred percent, because the denominator is the
\emph{net} drive and other sources cancel against it; that is the same accounting
Section~\ref{sec:circuit-accounting} is about, visible one level down.

\input{galpages}

\clearpage
\subsection{Traces where the vocabulary runs out}
\label{sec:gallery-concept}

The pages above were drawn without regard to their content, and most of them use no concepts. These
were selected for the opposite reason: each carries edges that no token describes, so that the case
the rest of this paper argues about can be seen rather than described. They are not a sample of
anything. OPT-125M leads them because it is genuinely unlike the others here --- about two thirds of
the edges in its traces carry no word, against a few percent everywhere else --- and it is the one model
in the set built on ReLU, whose units are silent on the whole negative half-line.

\input{galpages_concept}

%% file: galpages.tex
\clearpage
\begin{figure}[H]
\centering
\includegraphics[width=\textwidth,height=0.92\textheight,keepaspectratio]{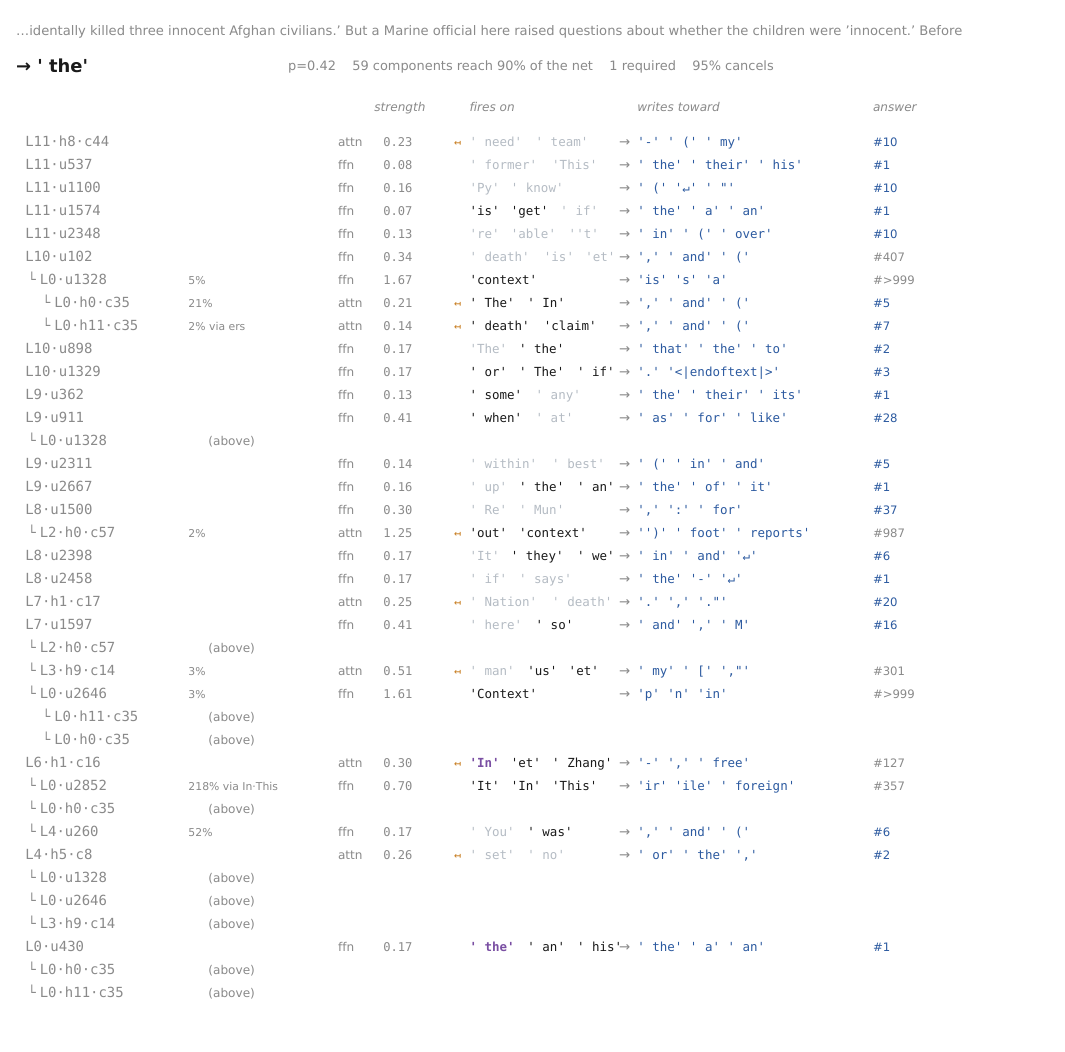}
\caption{baseline (GELU), prediction 1: the set, 28 members. Of 11 predictions drawn at random, this is one of the four whose sufficient set is nearest the median, 24.}
\end{figure}
\clearpage
\begin{landscape}
\begin{figure}[H]
\centering
\includegraphics[width=\linewidth,height=0.88\textheight,keepaspectratio]{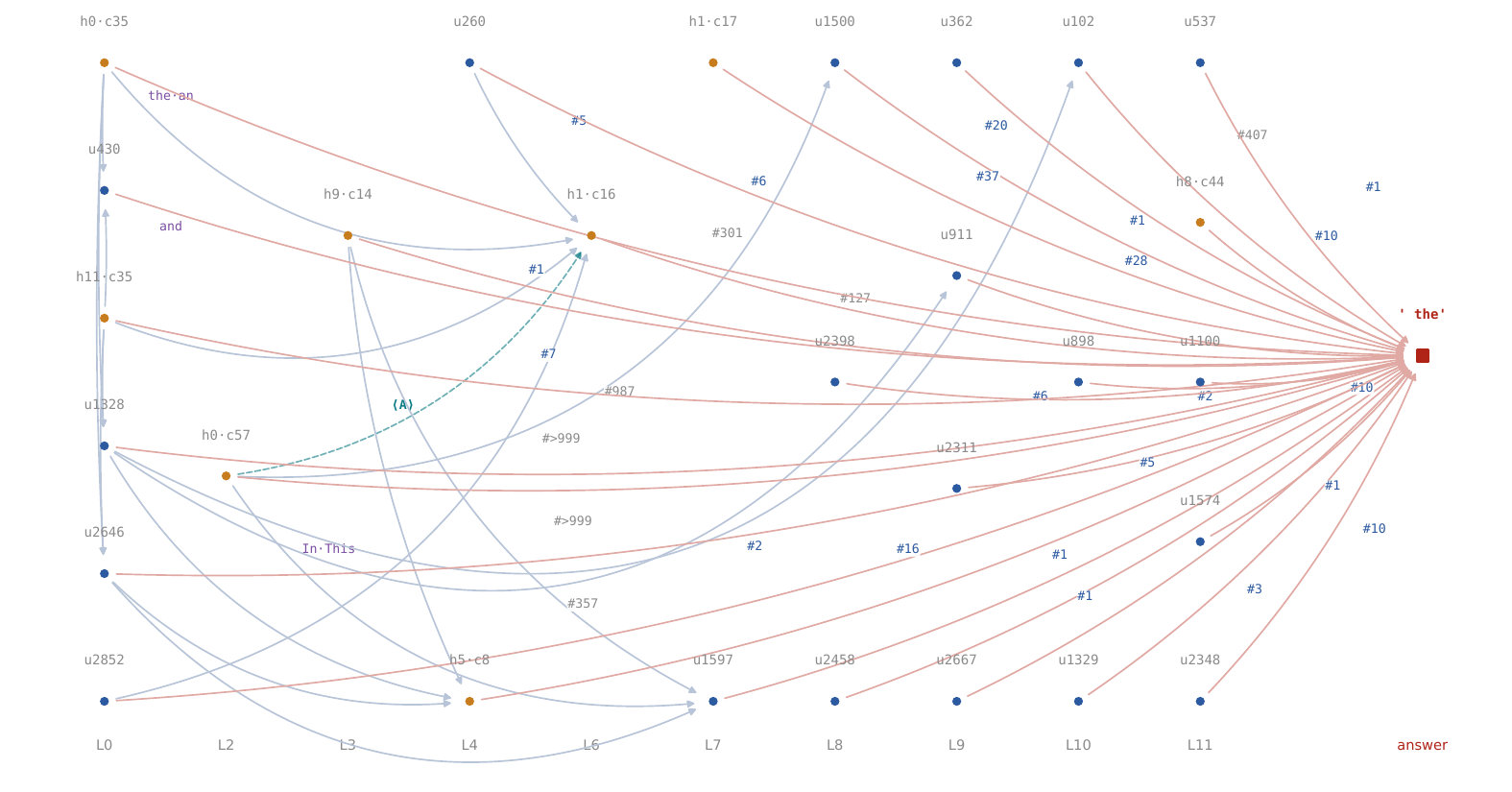}
\caption{baseline (GELU), prediction 1: the graph, 28 members and 20 edges among them.}
\end{figure}
\end{landscape}
\clearpage
\begin{figure}[H]
\centering
\includegraphics[width=\textwidth,height=0.92\textheight,keepaspectratio]{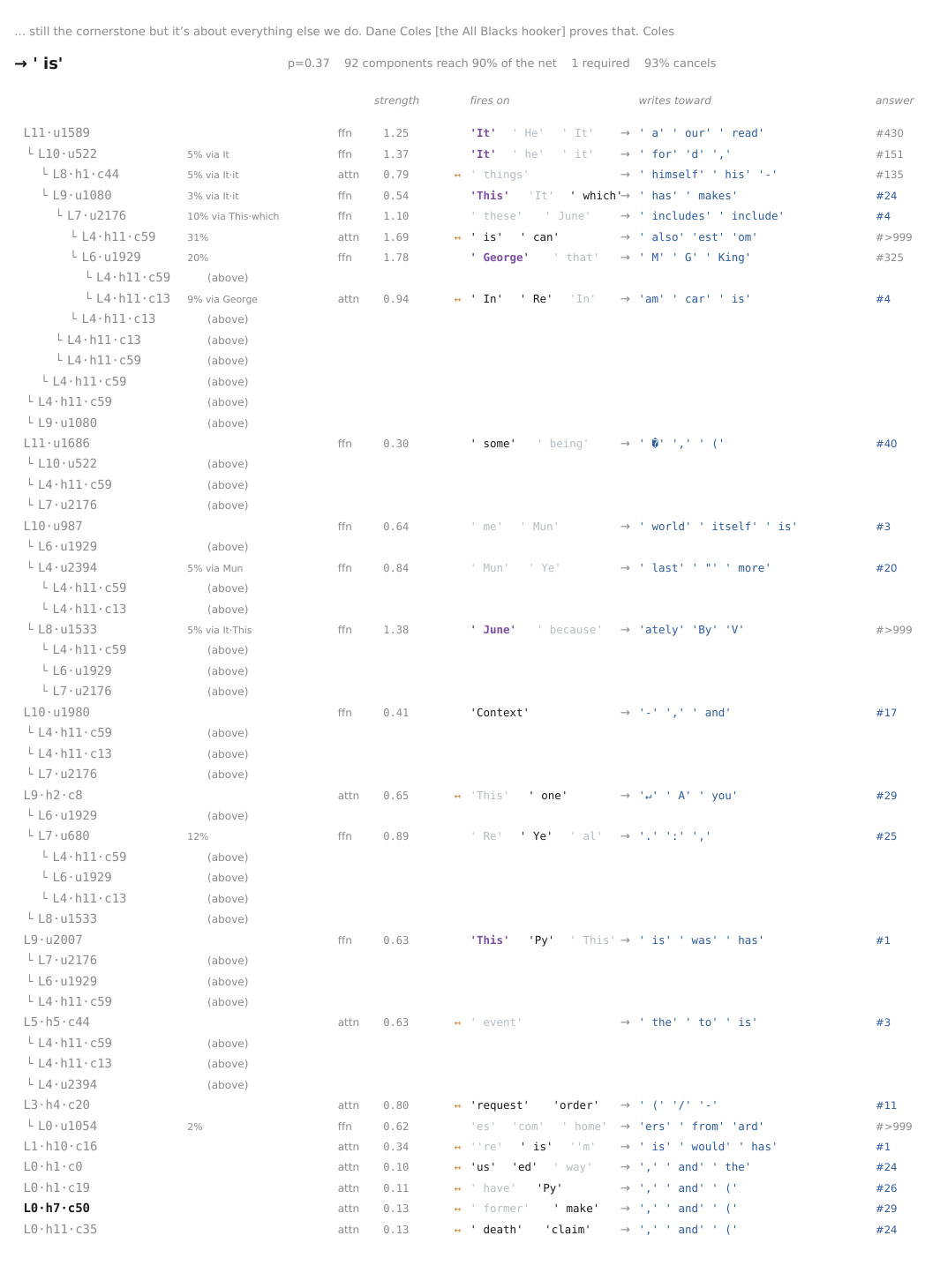}
\caption{baseline (GELU), prediction 2: the set, 24 members. Of 11 predictions drawn at random, this is one of the four whose sufficient set is nearest the median, 24.}
\end{figure}
\clearpage
\begin{landscape}
\begin{figure}[H]
\centering
\includegraphics[width=\linewidth,height=0.88\textheight,keepaspectratio]{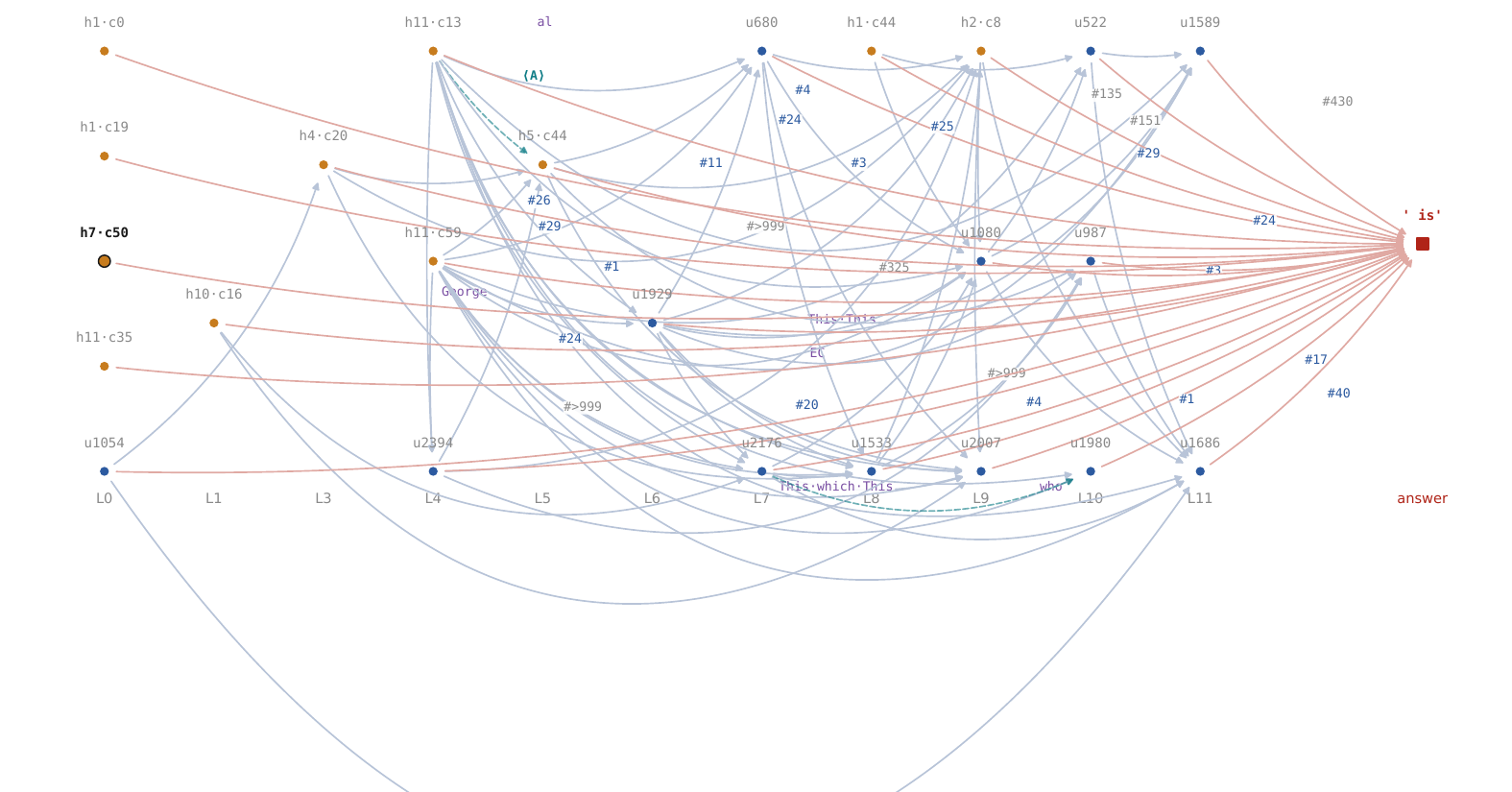}
\caption{baseline (GELU), prediction 2: the graph, 24 members and 68 edges among them.}
\end{figure}
\end{landscape}
\clearpage
\begin{figure}[H]
\centering
\includegraphics[width=\textwidth,height=0.92\textheight,keepaspectratio]{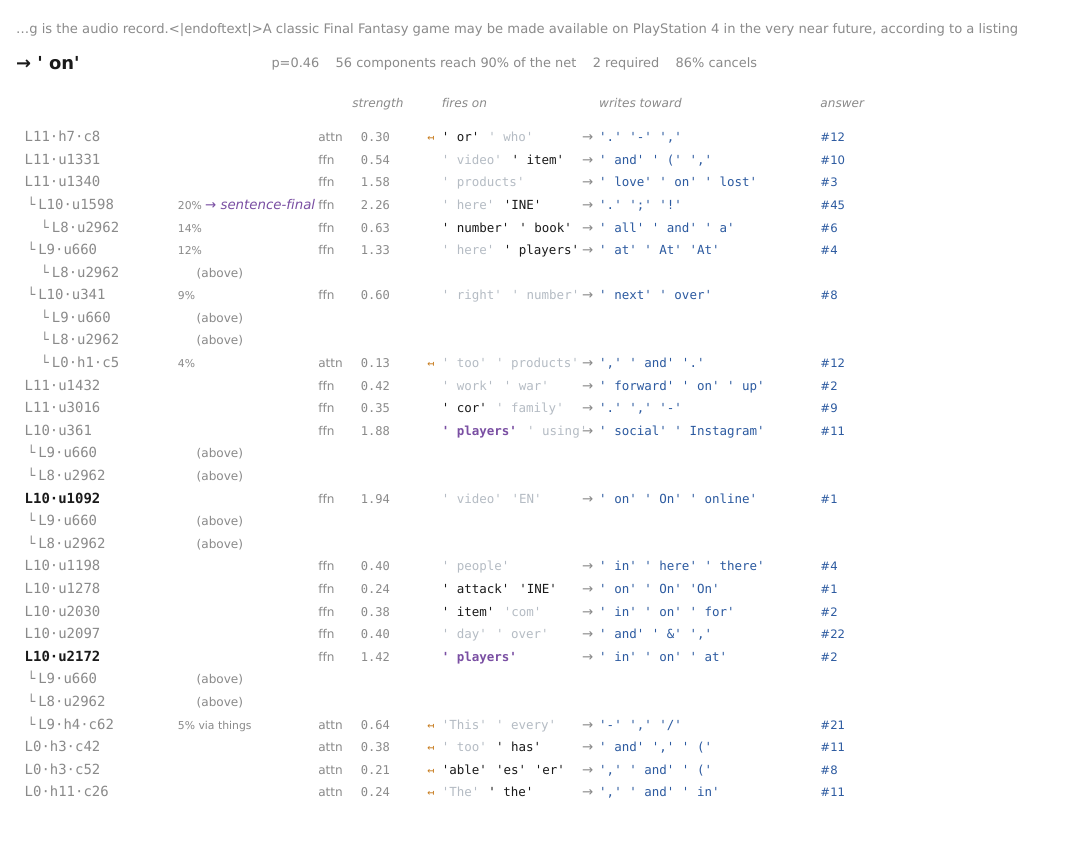}
\caption{sigmoid, prediction 1: the set, 21 members. Of 18 predictions drawn at random, this is one of the four whose sufficient set is nearest the median, 19.}
\end{figure}
\clearpage
\begin{landscape}
\begin{figure}[H]
\centering
\includegraphics[width=\linewidth,height=0.88\textheight,keepaspectratio]{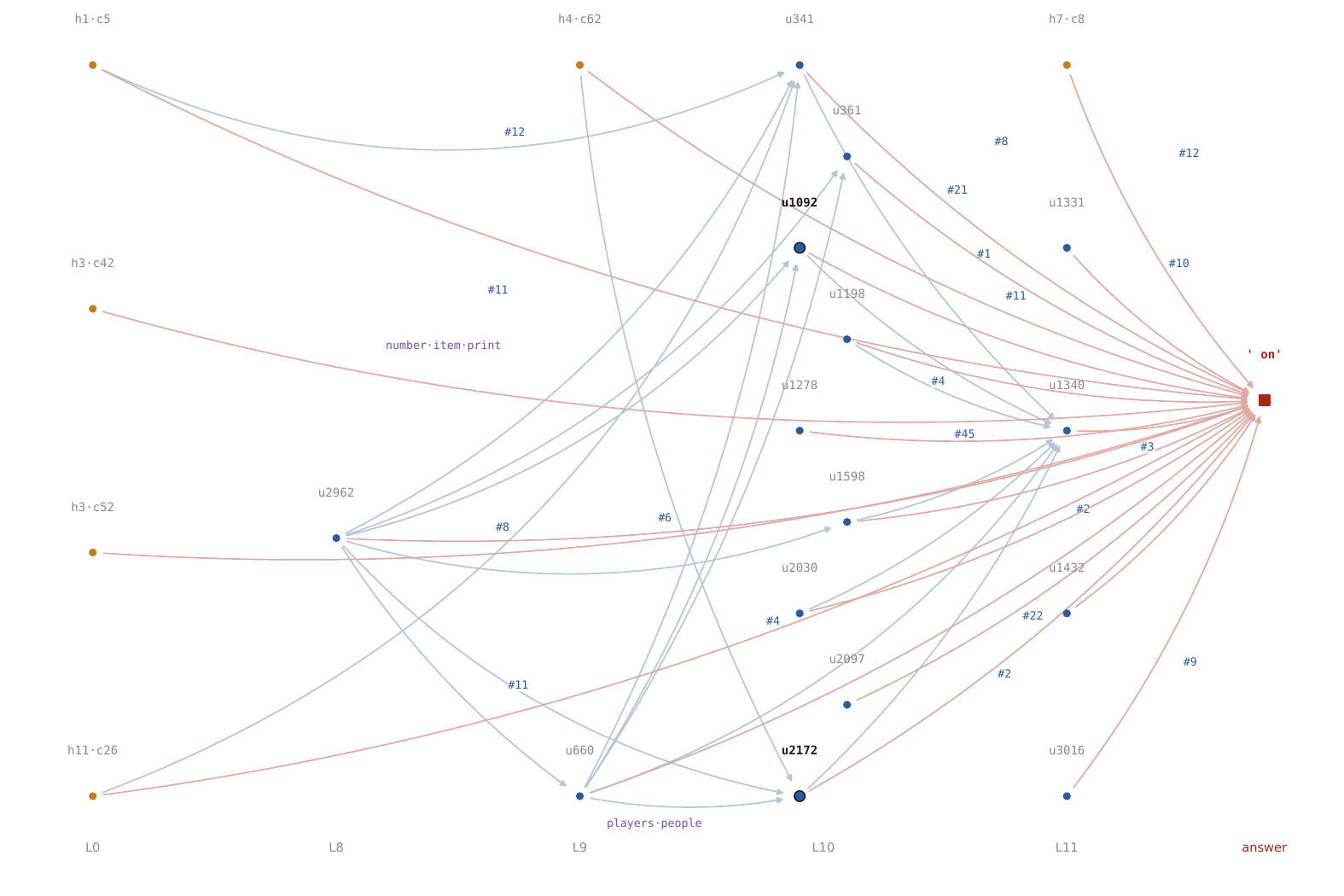}
\caption{sigmoid, prediction 1: the graph, 21 members and 20 edges among them.}
\end{figure}
\end{landscape}
\clearpage
\begin{figure}[H]
\centering
\includegraphics[width=\textwidth,height=0.92\textheight,keepaspectratio]{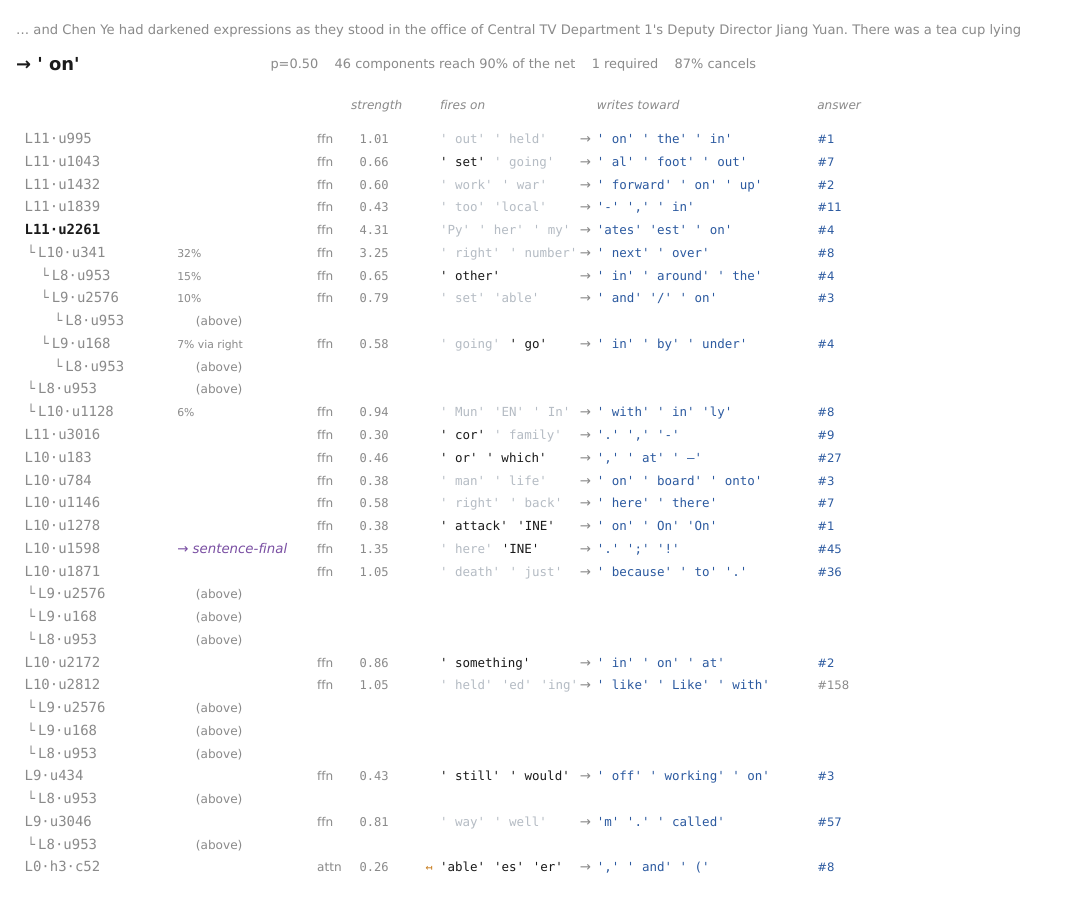}
\caption{sigmoid, prediction 2: the set, 22 members. Of 18 predictions drawn at random, this is one of the four whose sufficient set is nearest the median, 19.}
\end{figure}
\clearpage
\begin{landscape}
\begin{figure}[H]
\centering
\includegraphics[width=\linewidth,height=0.88\textheight,keepaspectratio]{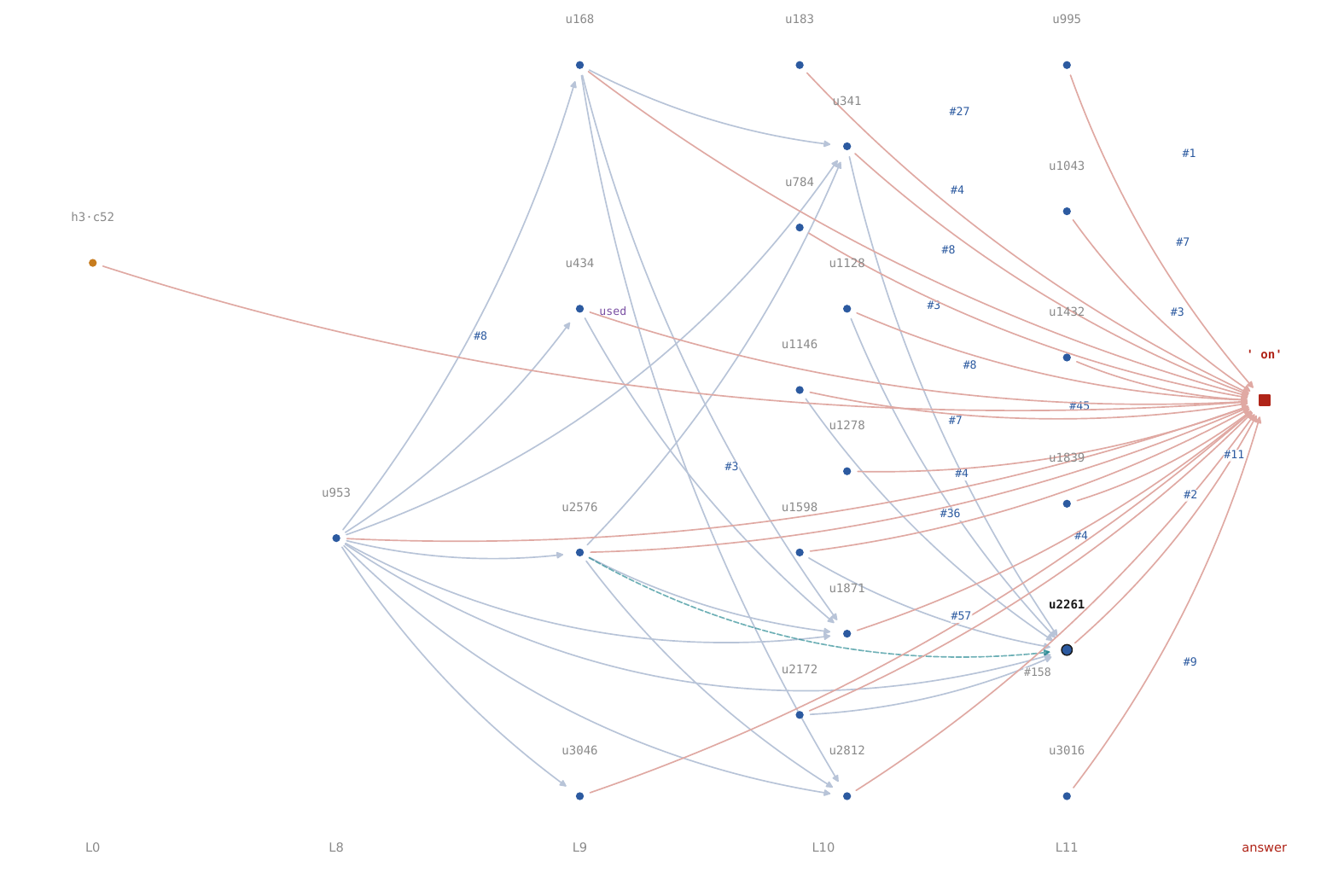}
\caption{sigmoid, prediction 2: the graph, 22 members and 21 edges among them.}
\end{figure}
\end{landscape}
\clearpage
\begin{figure}[H]
\centering
\includegraphics[width=\textwidth,height=0.92\textheight,keepaspectratio]{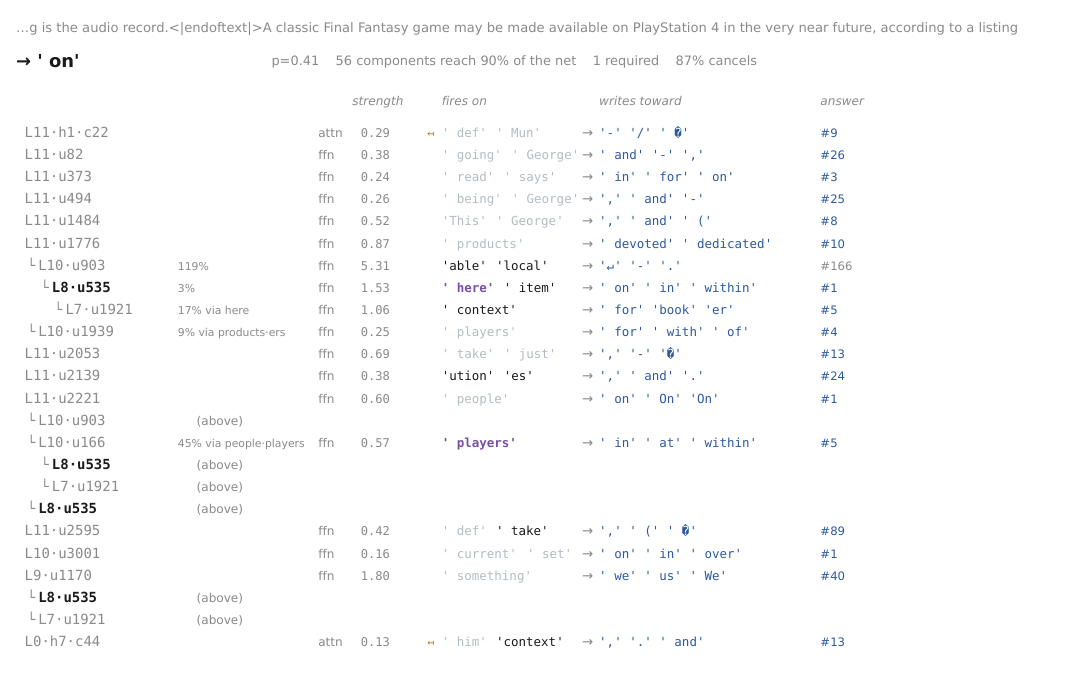}
\caption{softplus, prediction 1: the set, 18 members. Of 15 predictions drawn at random, this is one of the four whose sufficient set is nearest the median, 18.}
\end{figure}
\clearpage
\begin{landscape}
\begin{figure}[H]
\centering
\includegraphics[width=\linewidth,height=0.88\textheight,keepaspectratio]{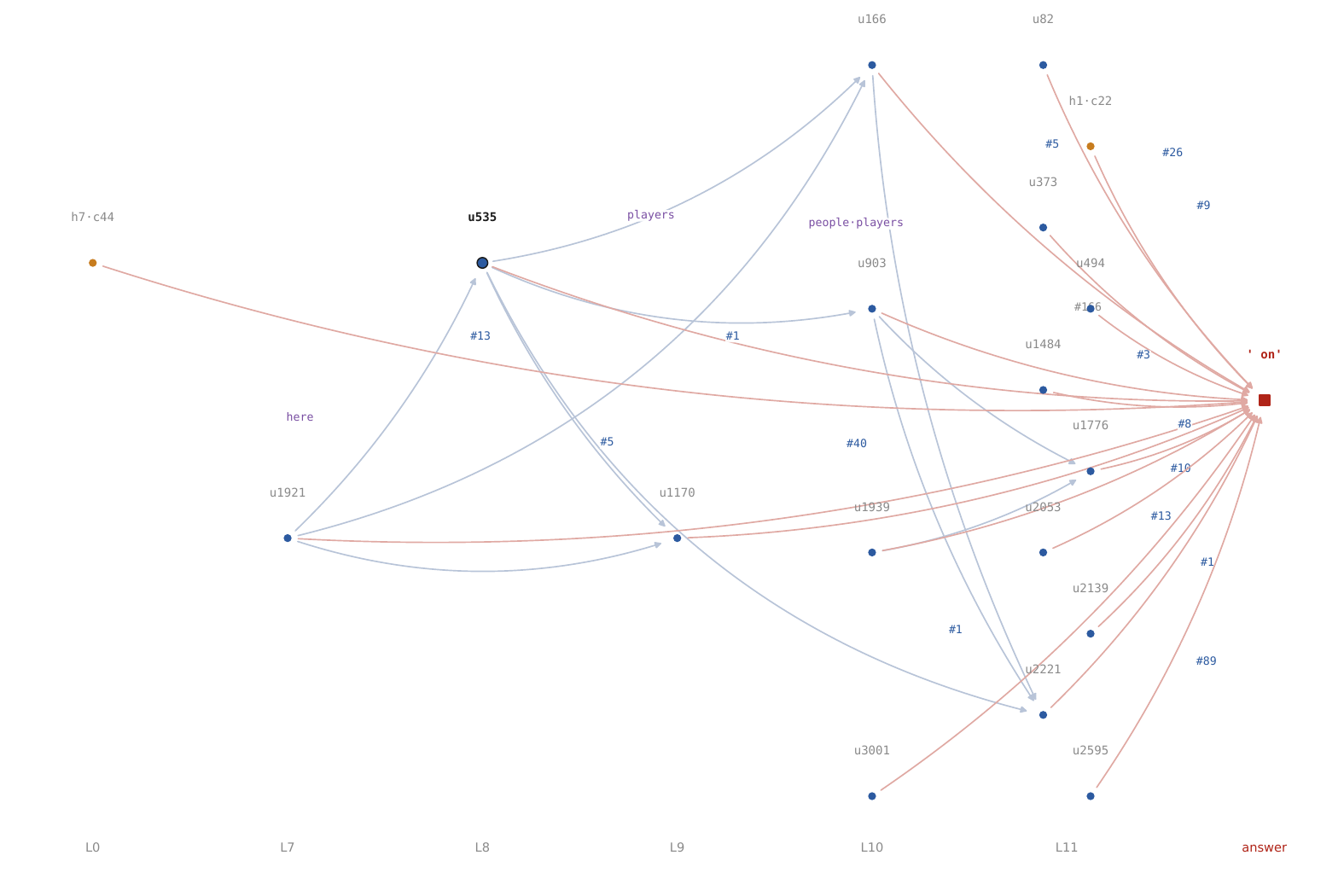}
\caption{softplus, prediction 1: the graph, 18 members and 11 edges among them.}
\end{figure}
\end{landscape}
\clearpage
\begin{figure}[H]
\centering
\includegraphics[width=\textwidth,height=0.92\textheight,keepaspectratio]{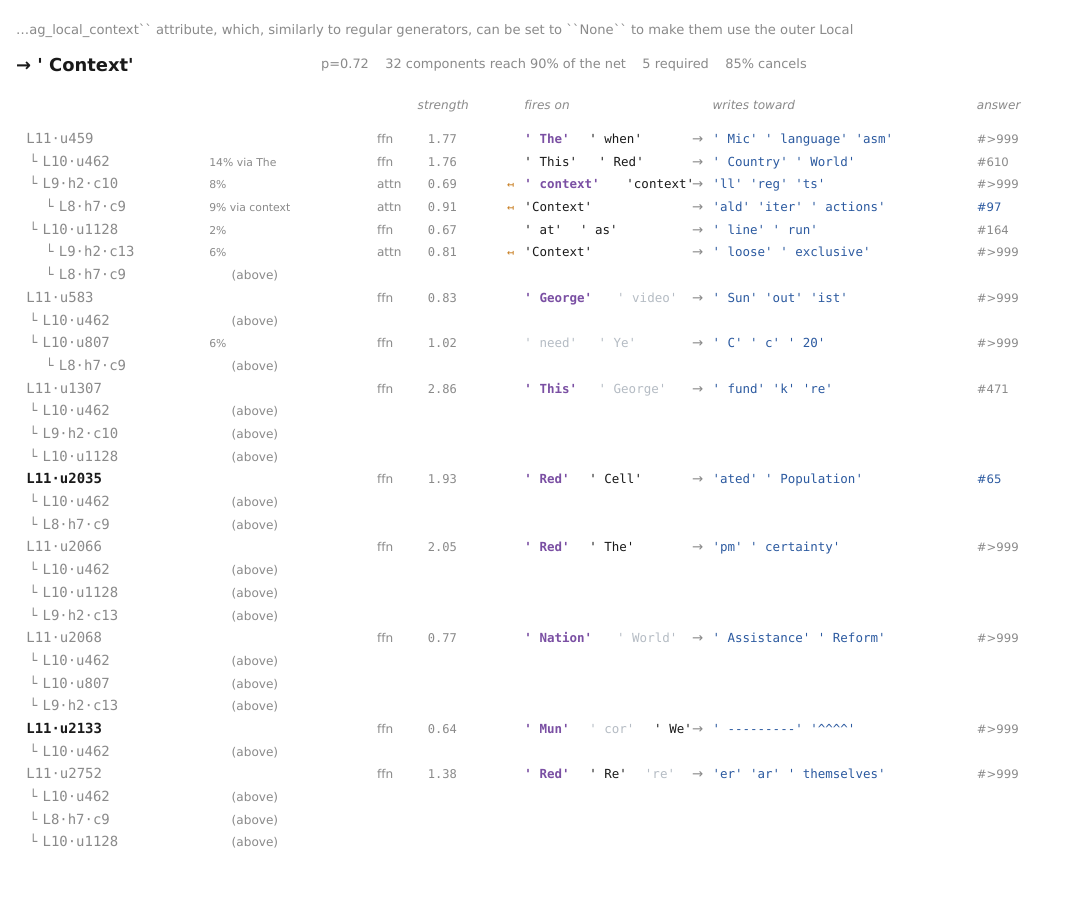}
\caption{softplus, prediction 2: the set, 14 members. Of 15 predictions drawn at random, this is one of the four whose sufficient set is nearest the median, 18.}
\end{figure}
\clearpage
\begin{landscape}
\begin{figure}[H]
\centering
\includegraphics[width=\linewidth,height=0.88\textheight,keepaspectratio]{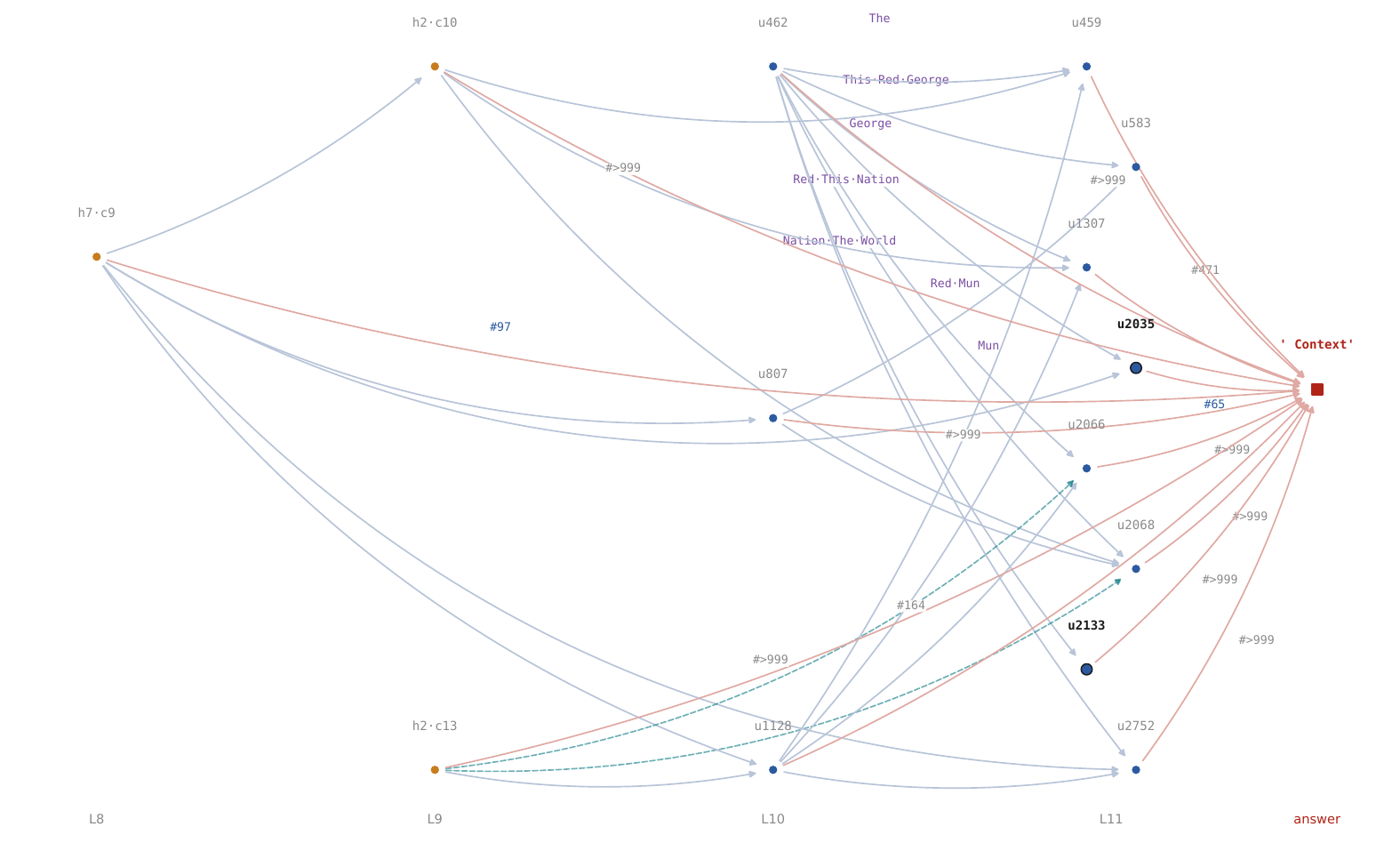}
\caption{softplus, prediction 2: the graph, 14 members and 25 edges among them.}
\end{figure}
\end{landscape}
\clearpage
\begin{figure}[H]
\centering
\includegraphics[width=\textwidth,height=0.92\textheight,keepaspectratio]{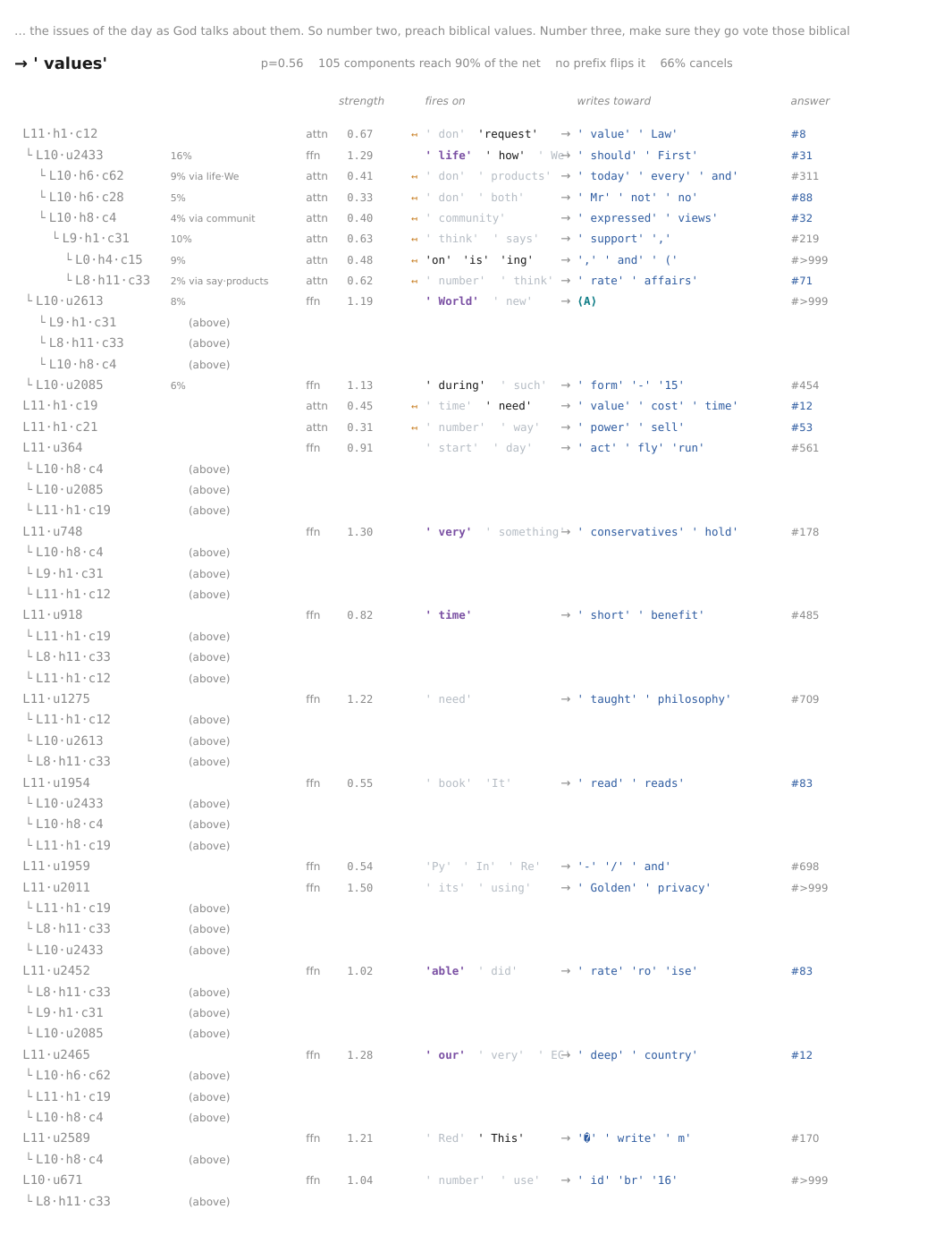}
\caption{ReLU, prediction 1: the set, 23 members. Of 14 predictions drawn at random, this is one of the four whose sufficient set is nearest the median, 22.}
\end{figure}
\clearpage
\begin{landscape}
\begin{figure}[H]
\centering
\includegraphics[width=\linewidth,height=0.88\textheight,keepaspectratio]{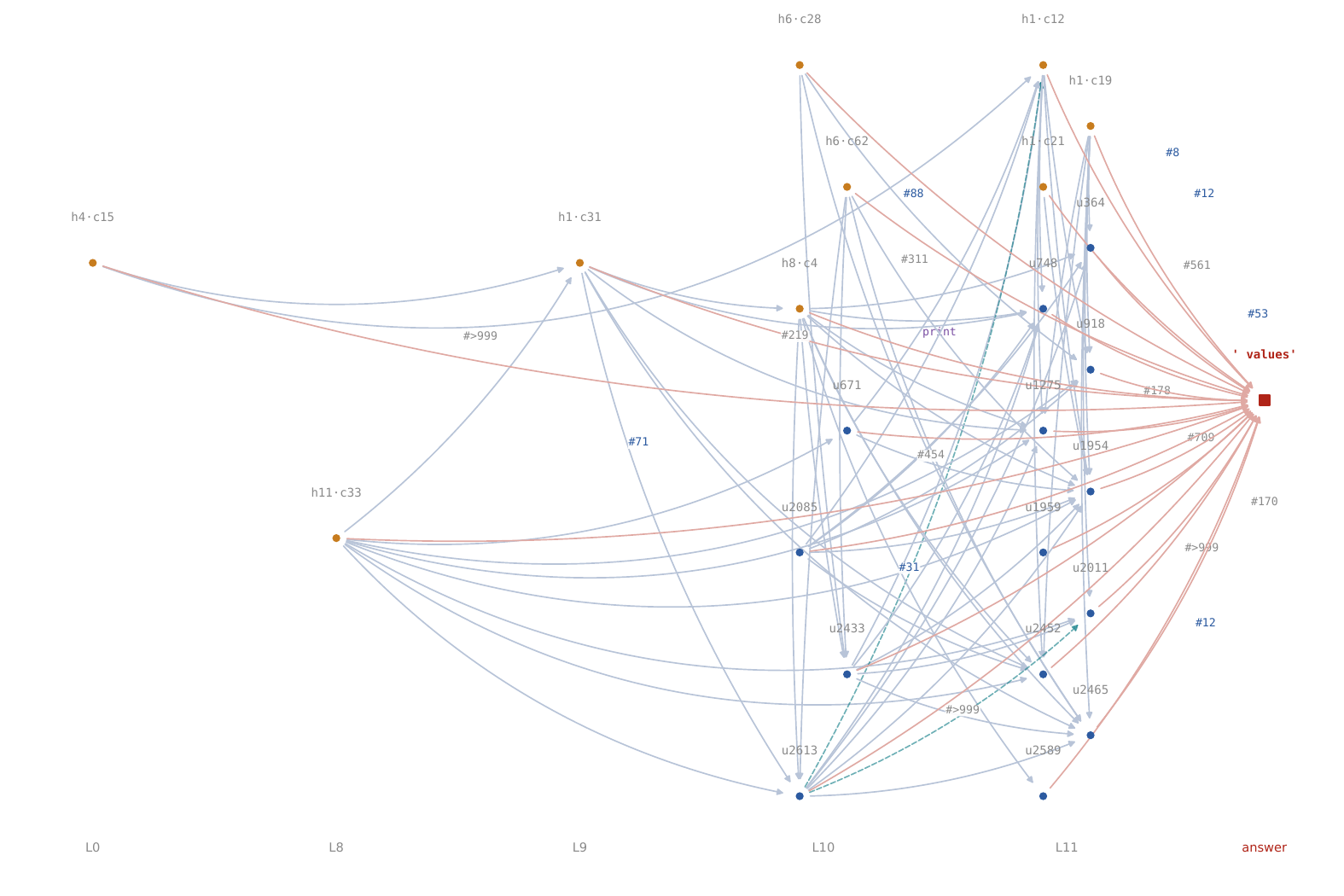}
\caption{ReLU, prediction 1: the graph, 23 members and 65 edges among them.}
\end{figure}
\end{landscape}
\clearpage
\begin{figure}[H]
\centering
\includegraphics[width=\textwidth,height=0.92\textheight,keepaspectratio]{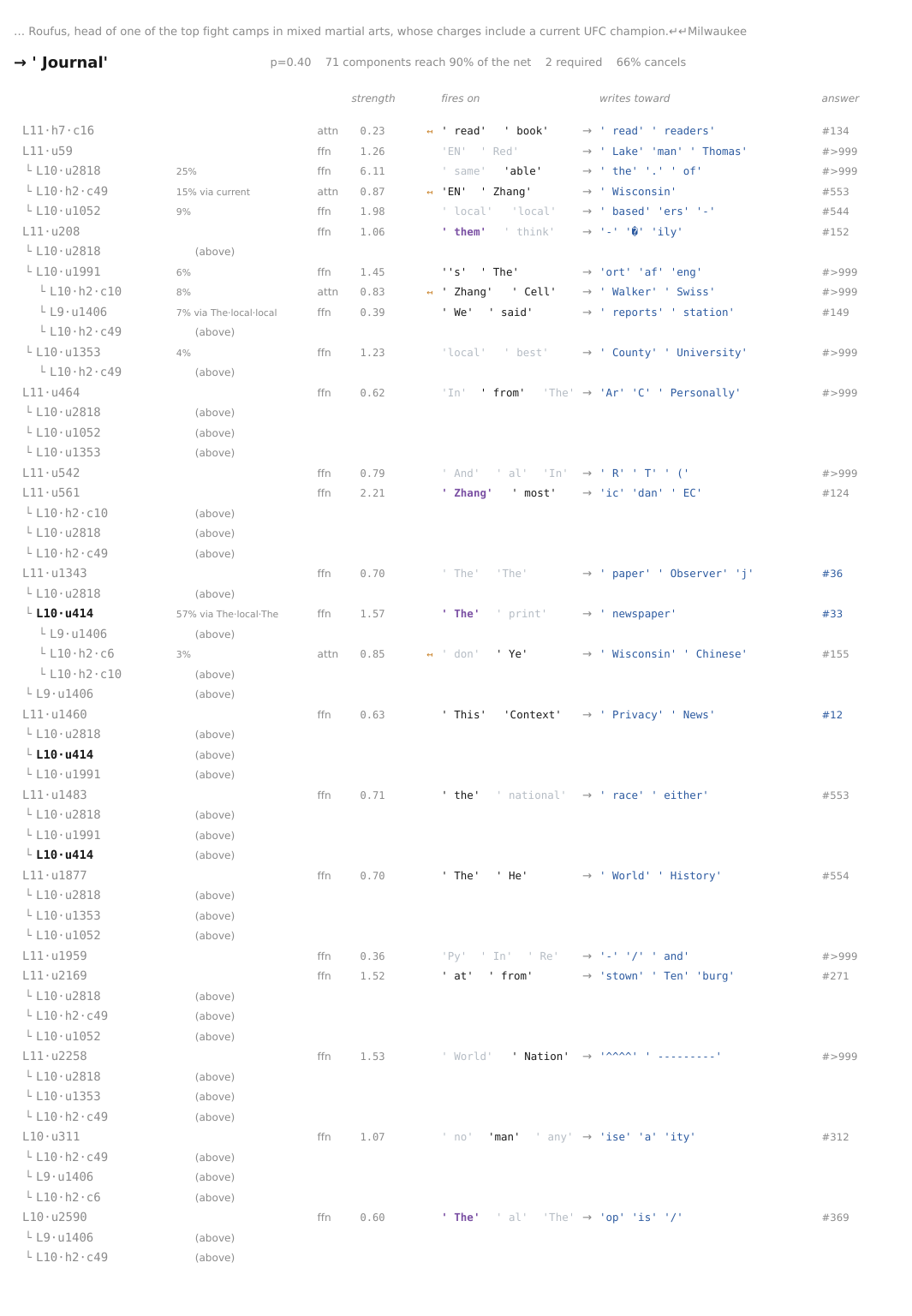}
\caption{ReLU, prediction 2: the set, 24 members. Of 14 predictions drawn at random, this is one of the four whose sufficient set is nearest the median, 22.}
\end{figure}
\clearpage
\begin{landscape}
\begin{figure}[H]
\centering
\includegraphics[width=\linewidth,height=0.88\textheight,keepaspectratio]{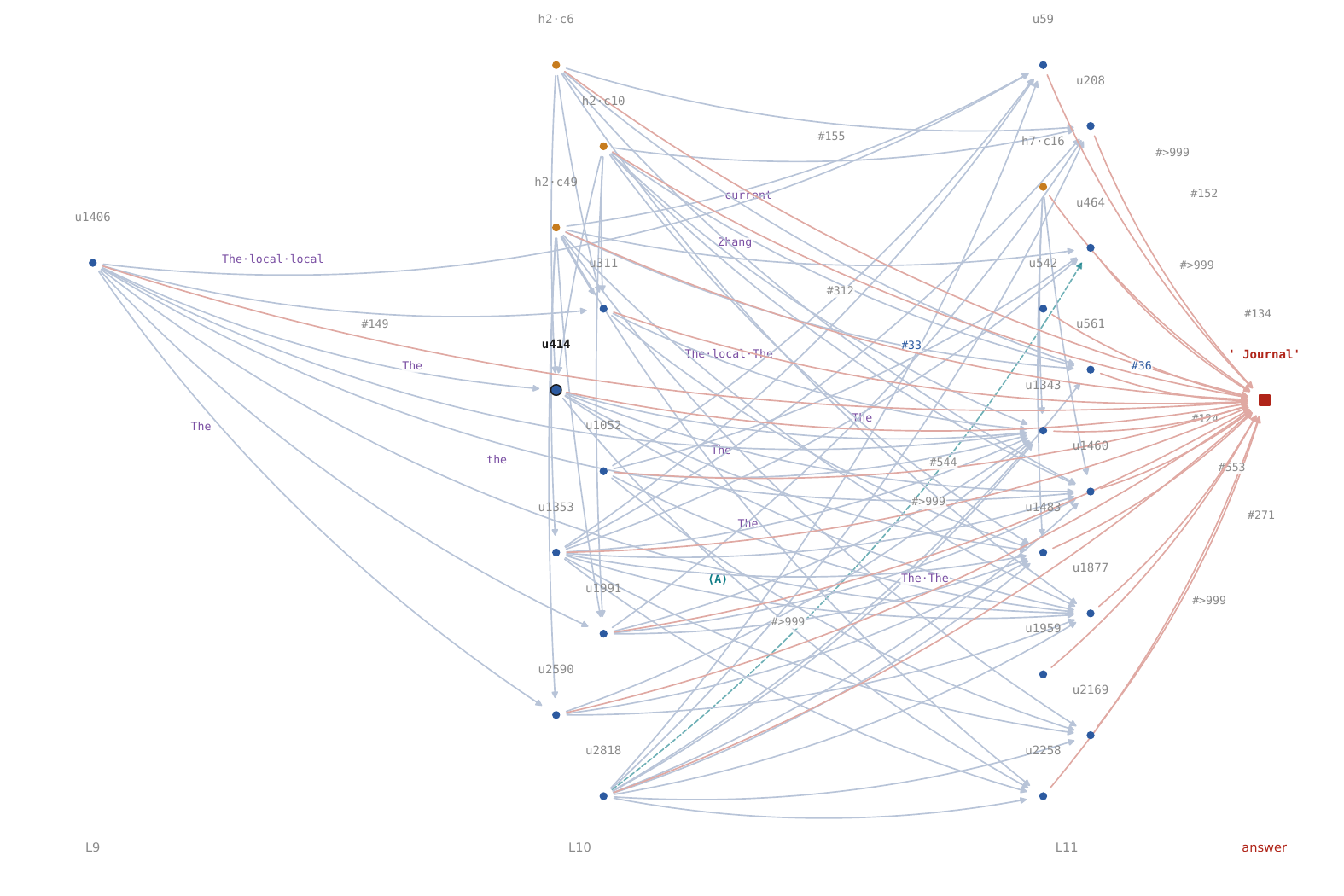}
\caption{ReLU, prediction 2: the graph, 24 members and 73 edges among them.}
\end{figure}
\end{landscape}
\clearpage
\begin{figure}[H]
\centering
\includegraphics[width=\textwidth,height=0.92\textheight,keepaspectratio]{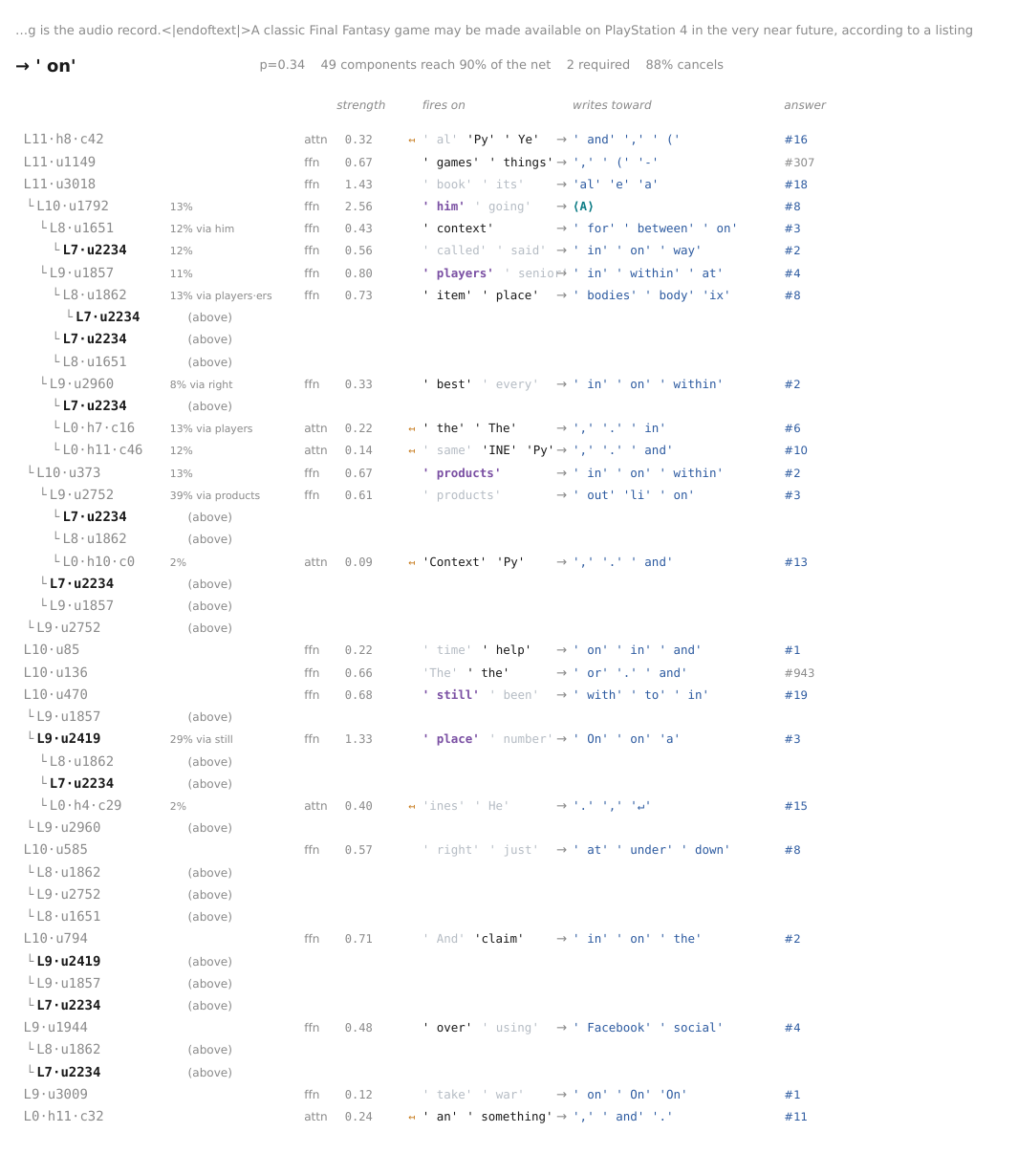}
\caption{sigmoid, no shaping, prediction 1: the set, 24 members. Of 16 predictions drawn at random, this is one of the four whose sufficient set is nearest the median, 25.}
\end{figure}
\clearpage
\begin{landscape}
\begin{figure}[H]
\centering
\includegraphics[width=\linewidth,height=0.88\textheight,keepaspectratio]{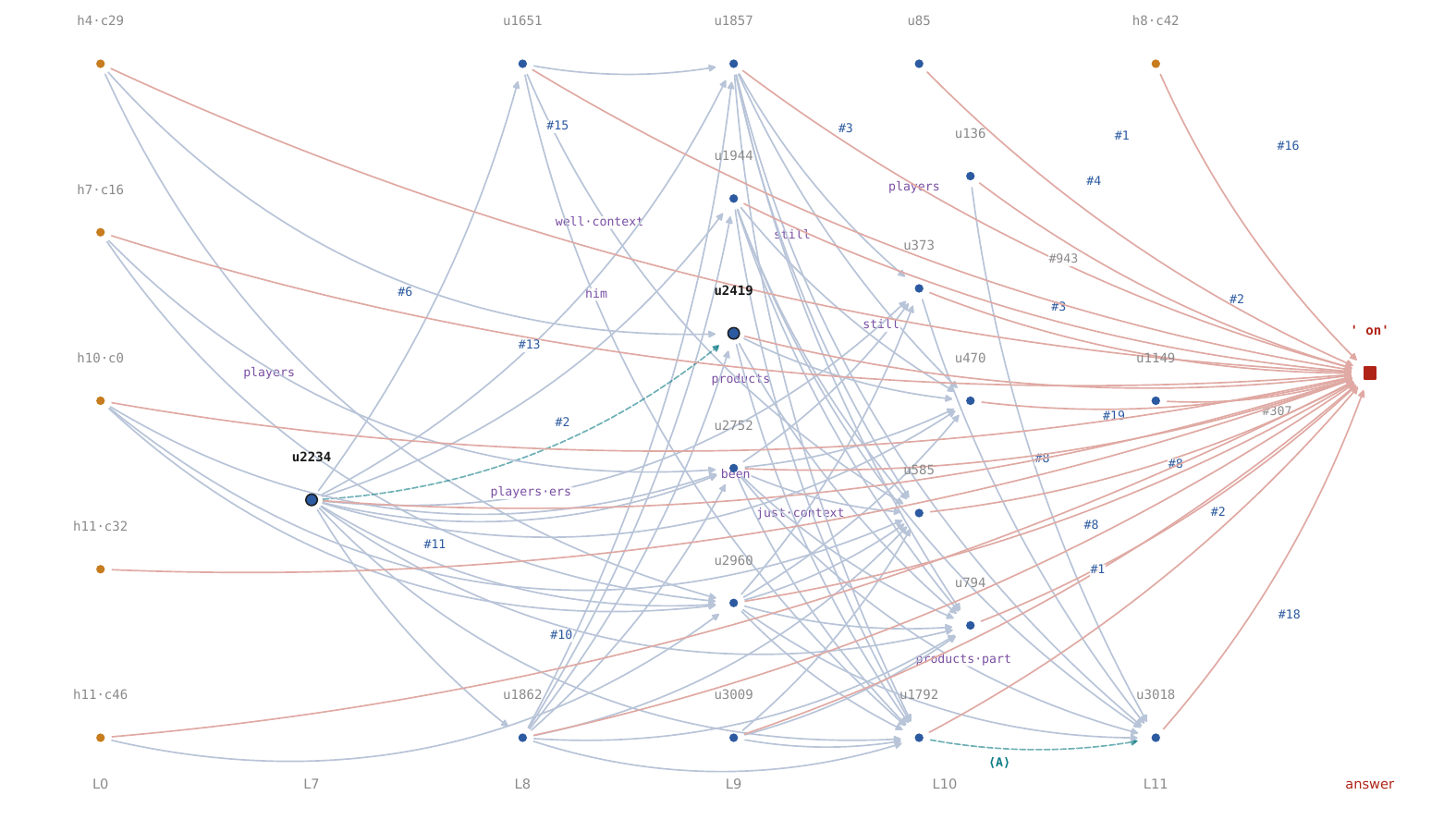}
\caption{sigmoid, no shaping, prediction 1: the graph, 24 members and 61 edges among them.}
\end{figure}
\end{landscape}
\clearpage
\begin{figure}[H]
\centering
\includegraphics[width=\textwidth,height=0.92\textheight,keepaspectratio]{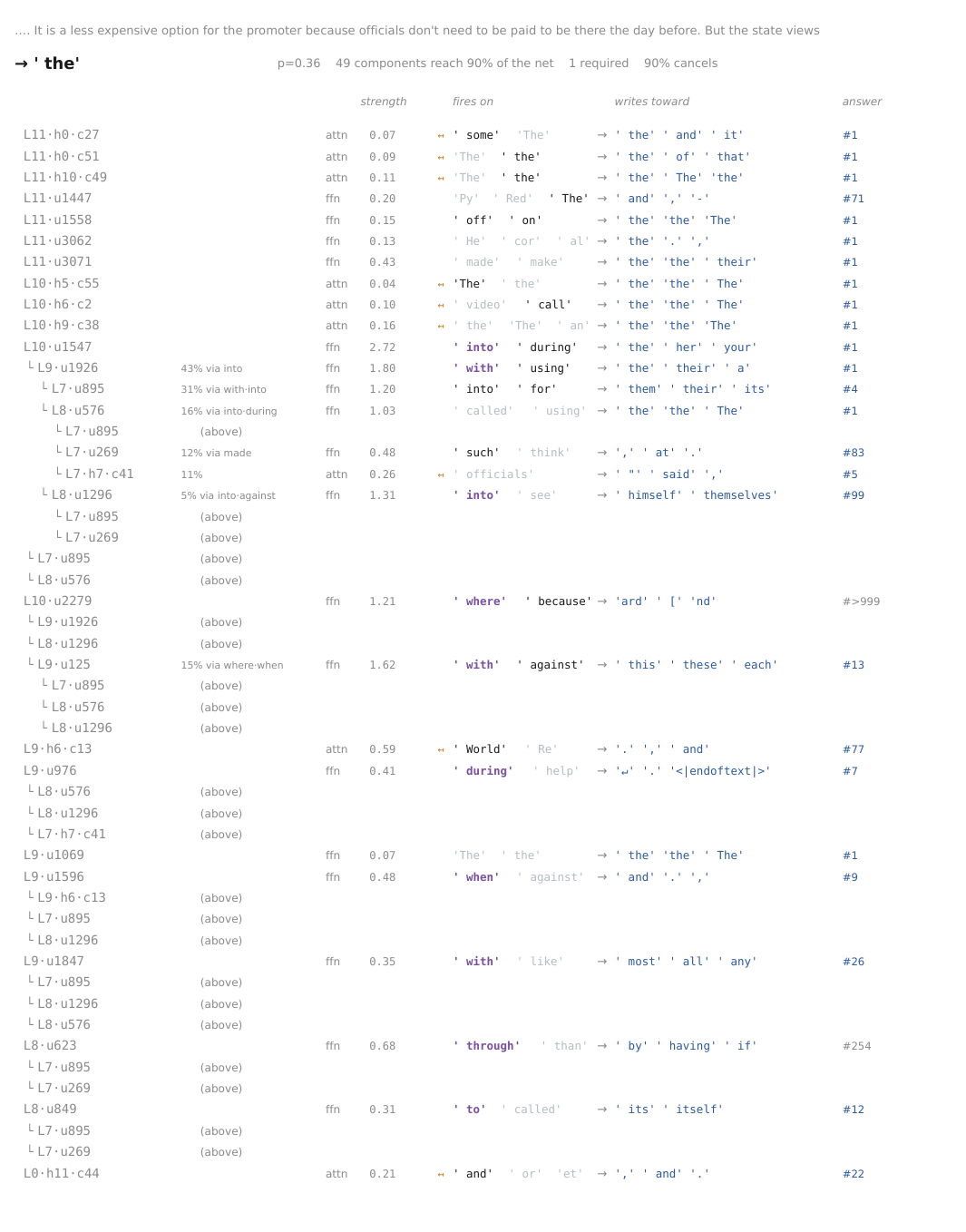}
\caption{sigmoid, no shaping, prediction 2: the set, 27 members. Of 16 predictions drawn at random, this is one of the four whose sufficient set is nearest the median, 25.}
\end{figure}
\clearpage
\begin{landscape}
\begin{figure}[H]
\centering
\includegraphics[width=\linewidth,height=0.88\textheight,keepaspectratio]{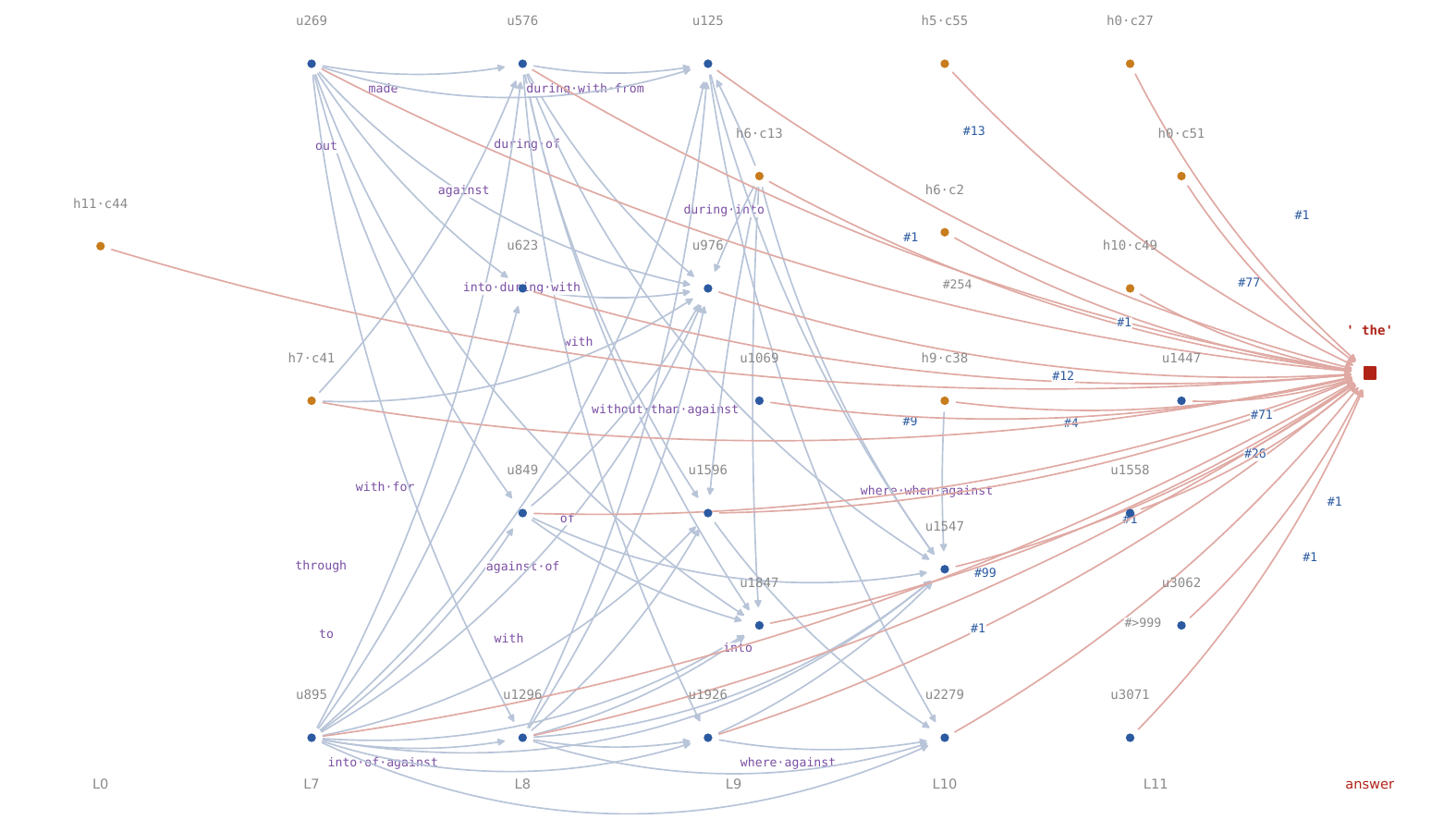}
\caption{sigmoid, no shaping, prediction 2: the graph, 27 members and 48 edges among them.}
\end{figure}
\end{landscape}
\clearpage
\begin{figure}[H]
\centering
\includegraphics[width=\textwidth,height=0.92\textheight,keepaspectratio]{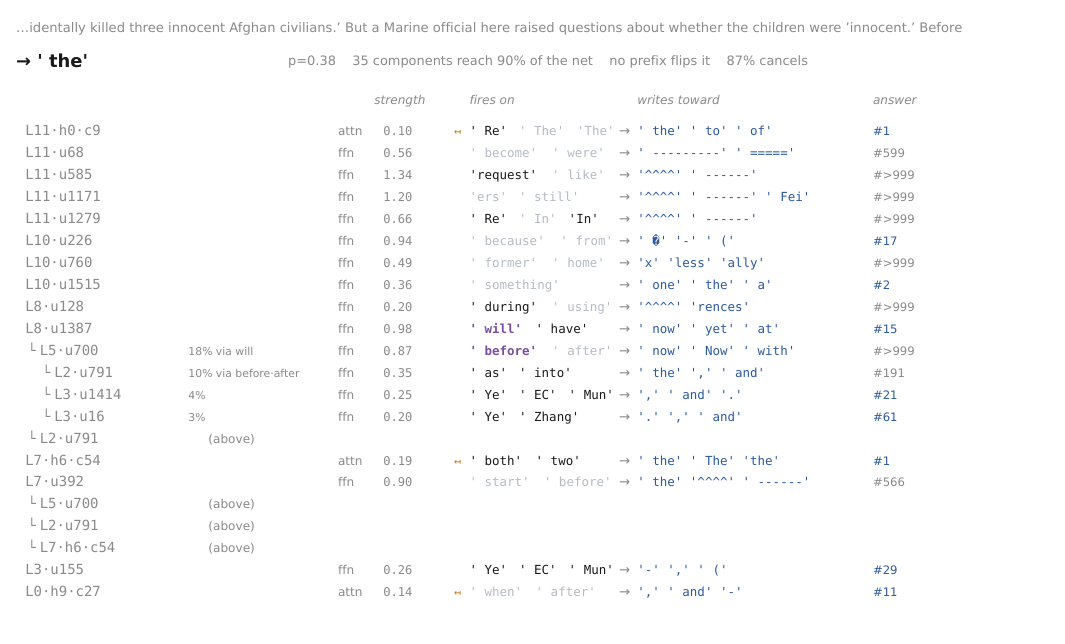}
\caption{set operators, prediction 1: the set, 18 members. Of 13 predictions drawn at random, this is one of the four whose sufficient set is nearest the median, 22.}
\end{figure}
\clearpage
\begin{landscape}
\begin{figure}[H]
\centering
\includegraphics[width=\linewidth,height=0.88\textheight,keepaspectratio]{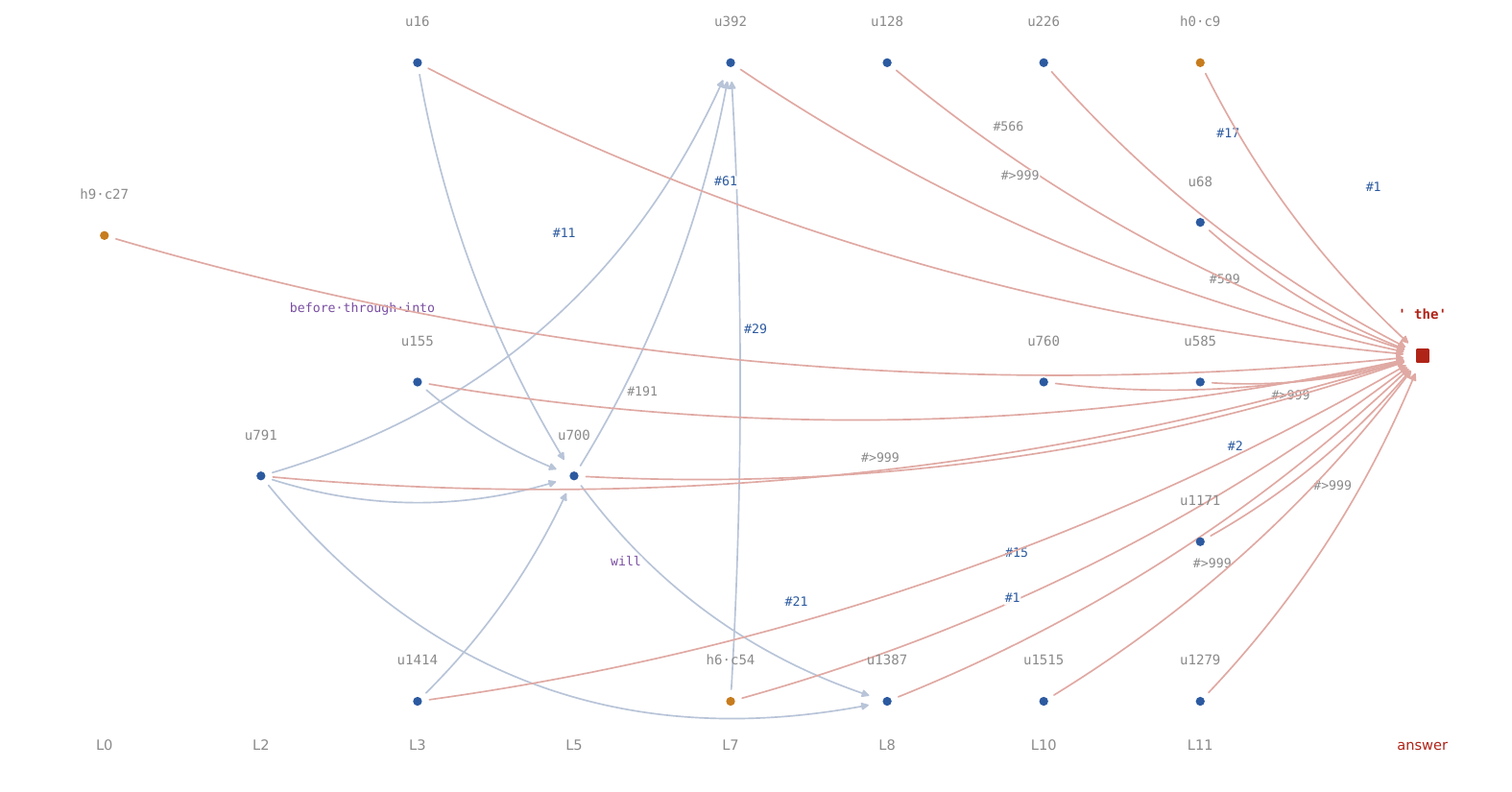}
\caption{set operators, prediction 1: the graph, 18 members and 9 edges among them.}
\end{figure}
\end{landscape}
\clearpage
\begin{figure}[H]
\centering
\includegraphics[width=\textwidth,height=0.92\textheight,keepaspectratio]{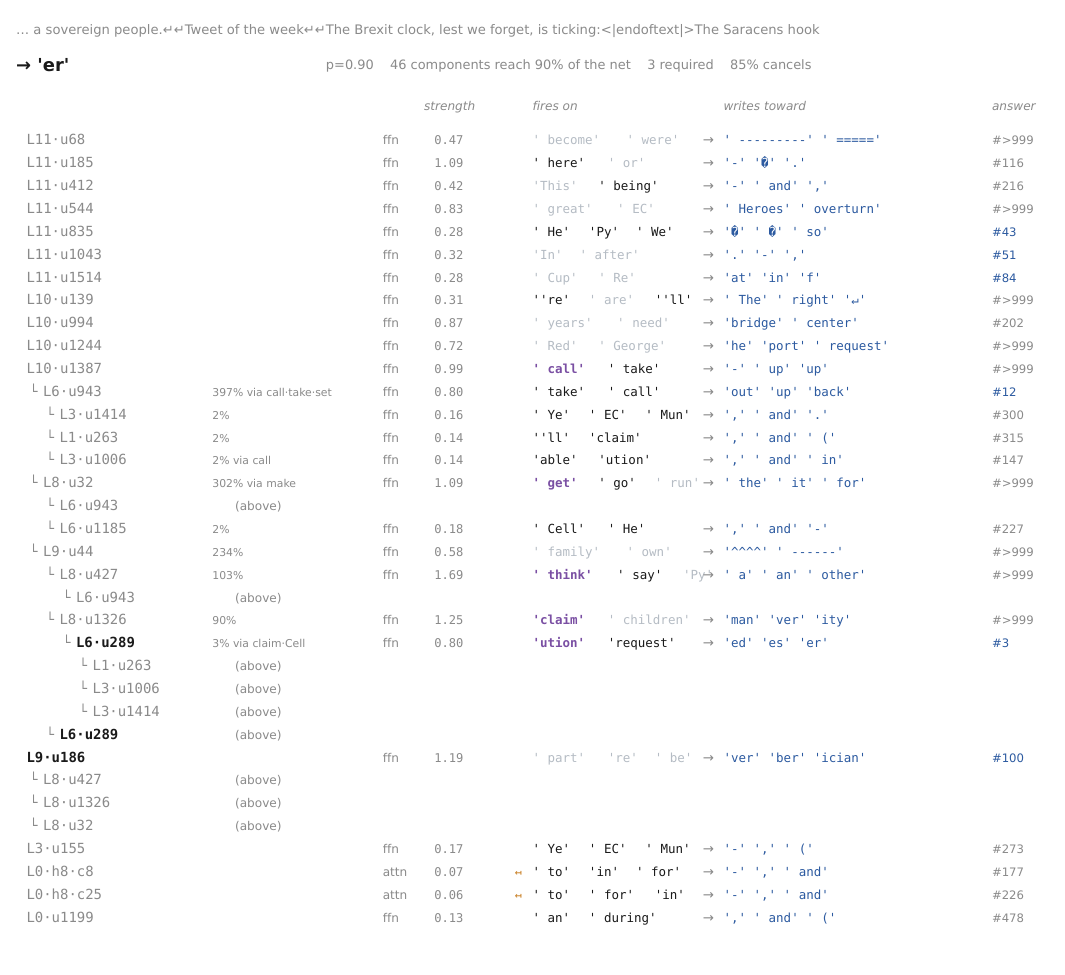}
\caption{set operators, prediction 2: the set, 26 members. Of 13 predictions drawn at random, this is one of the four whose sufficient set is nearest the median, 22.}
\end{figure}
\clearpage
\begin{landscape}
\begin{figure}[H]
\centering
\includegraphics[width=\linewidth,height=0.88\textheight,keepaspectratio]{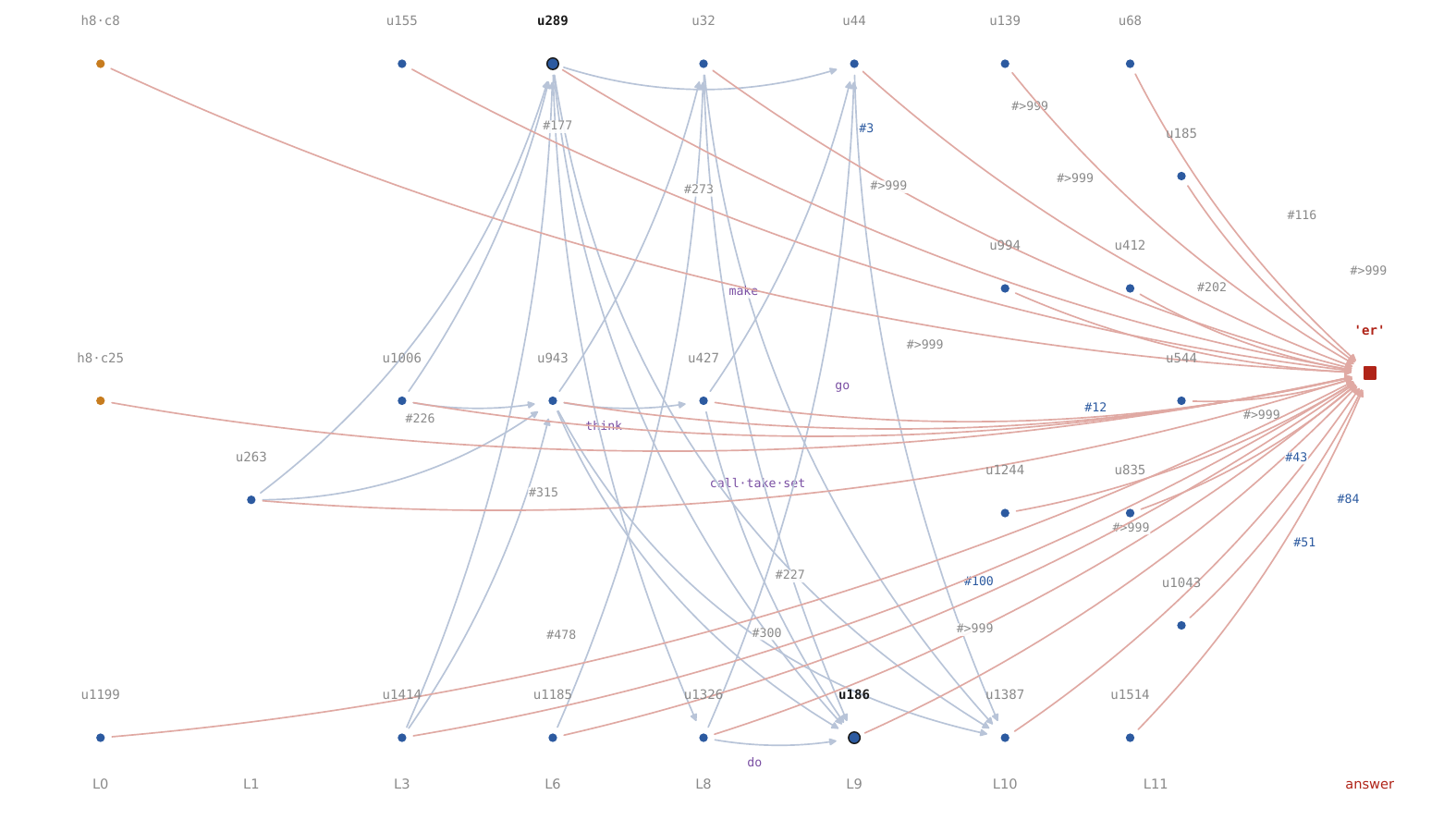}
\caption{set operators, prediction 2: the graph, 26 members and 22 edges among them.}
\end{figure}
\end{landscape}
\clearpage
\begin{figure}[H]
\centering
\includegraphics[width=\textwidth,height=0.45\textheight,keepaspectratio]{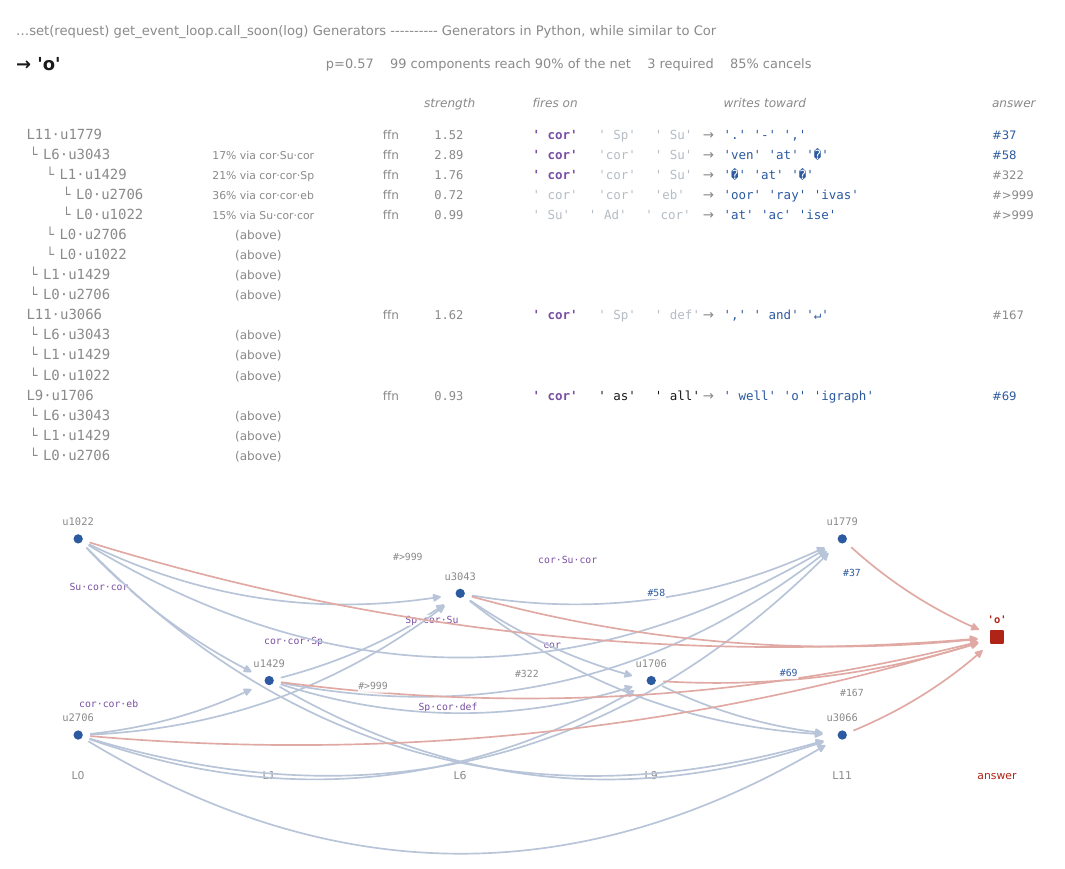}\\[1.2em]
\includegraphics[width=\textwidth,height=0.45\textheight,keepaspectratio]{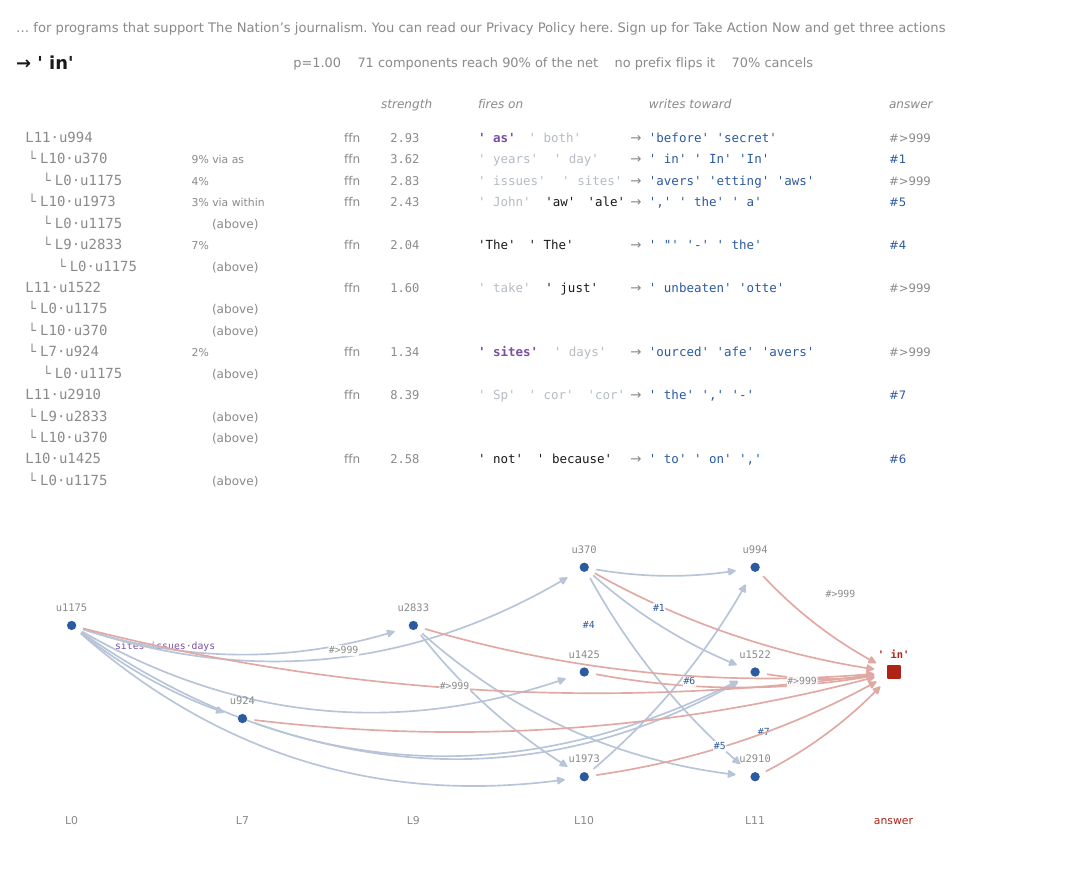}
\caption{GPT-2 124M, predictions 1 and 2. Of 20 predictions drawn at random, these are two of the four whose sufficient set is nearest the median, 8.}
\end{figure}
\clearpage
\begin{figure}[H]
\centering
\includegraphics[width=\textwidth,height=0.92\textheight,keepaspectratio]{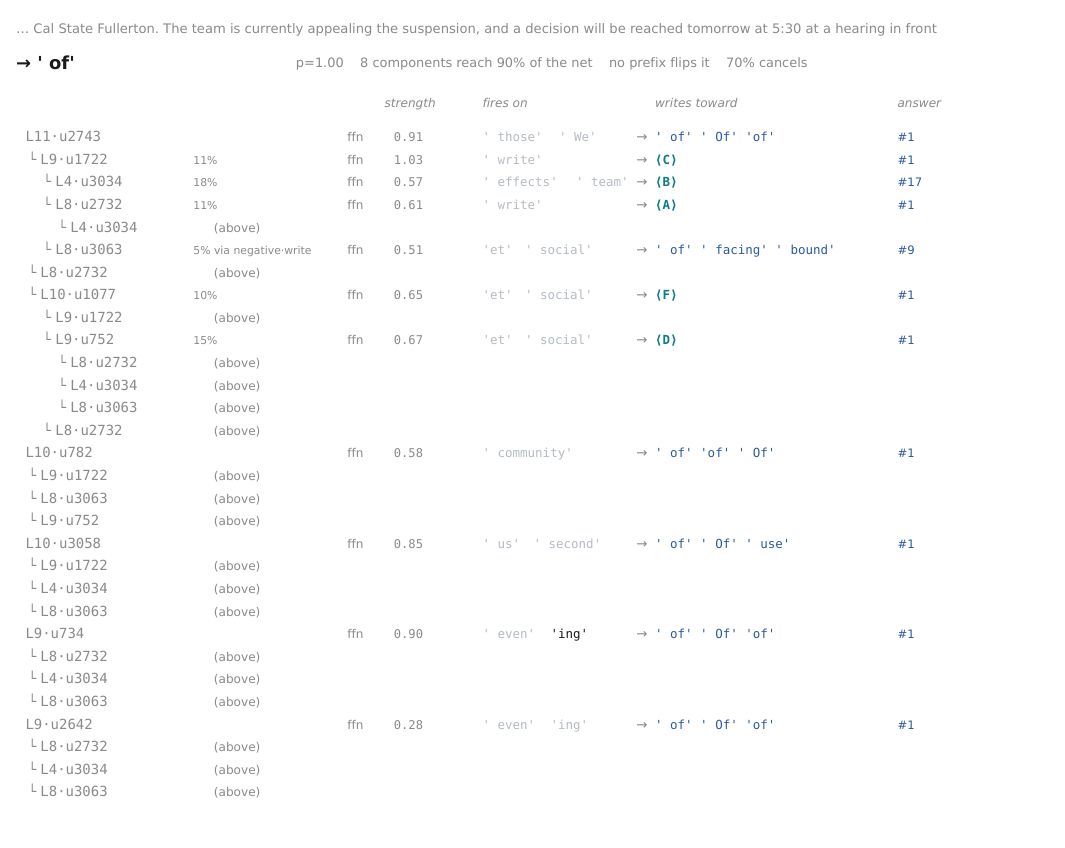}
\caption{OPT 125M, prediction 1: the set, 11 members. Of 20 predictions drawn at random, this is one of the four whose sufficient set is nearest the median, 13.}
\end{figure}
\clearpage
\begin{landscape}
\begin{figure}[H]
\centering
\includegraphics[width=\linewidth,height=0.88\textheight,keepaspectratio]{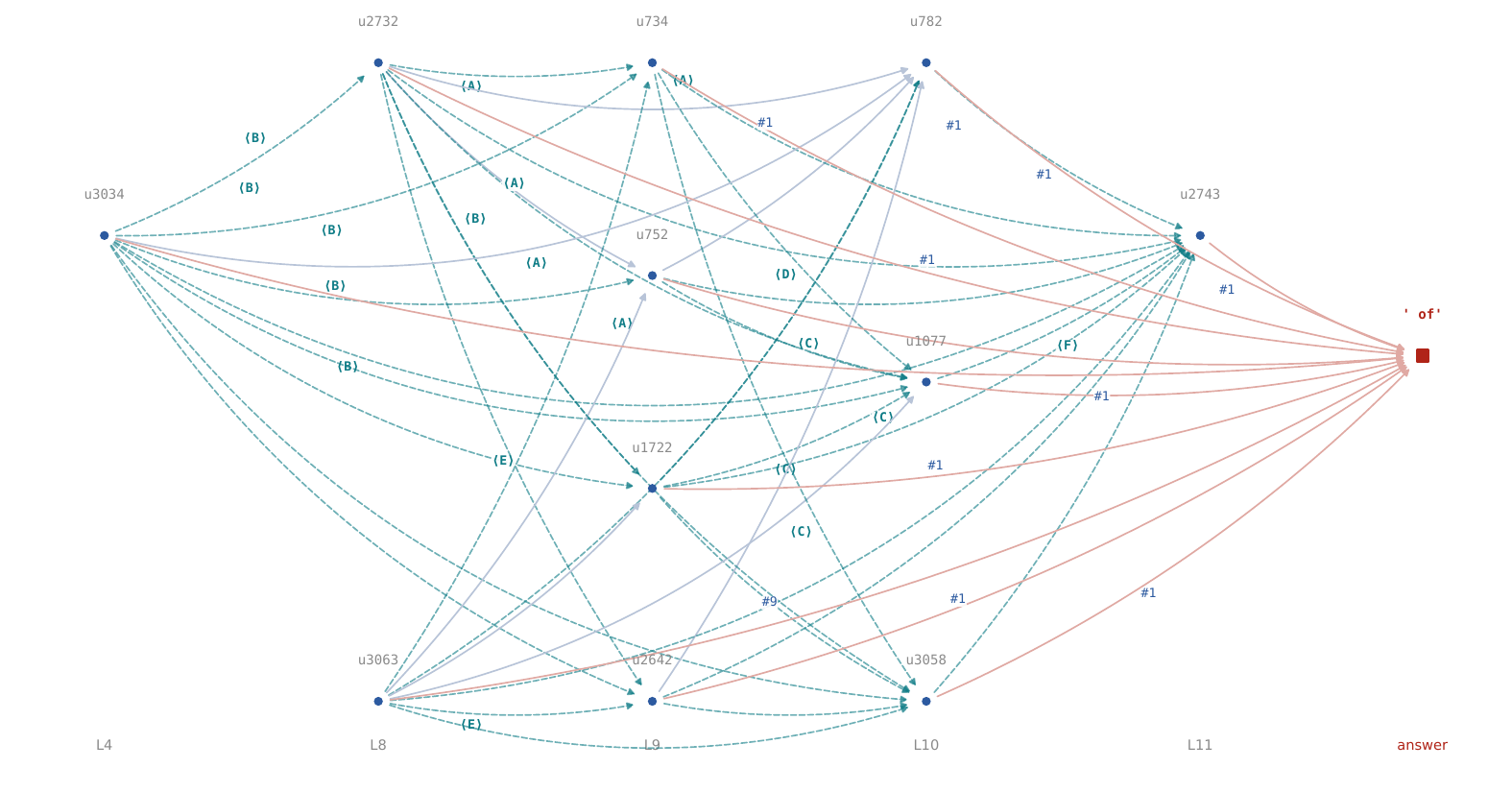}
\caption{OPT 125M, prediction 1: the graph, 11 members and 41 edges among them.}
\end{figure}
\end{landscape}
\clearpage
\begin{figure}[H]
\centering
\includegraphics[width=\textwidth,height=0.92\textheight,keepaspectratio]{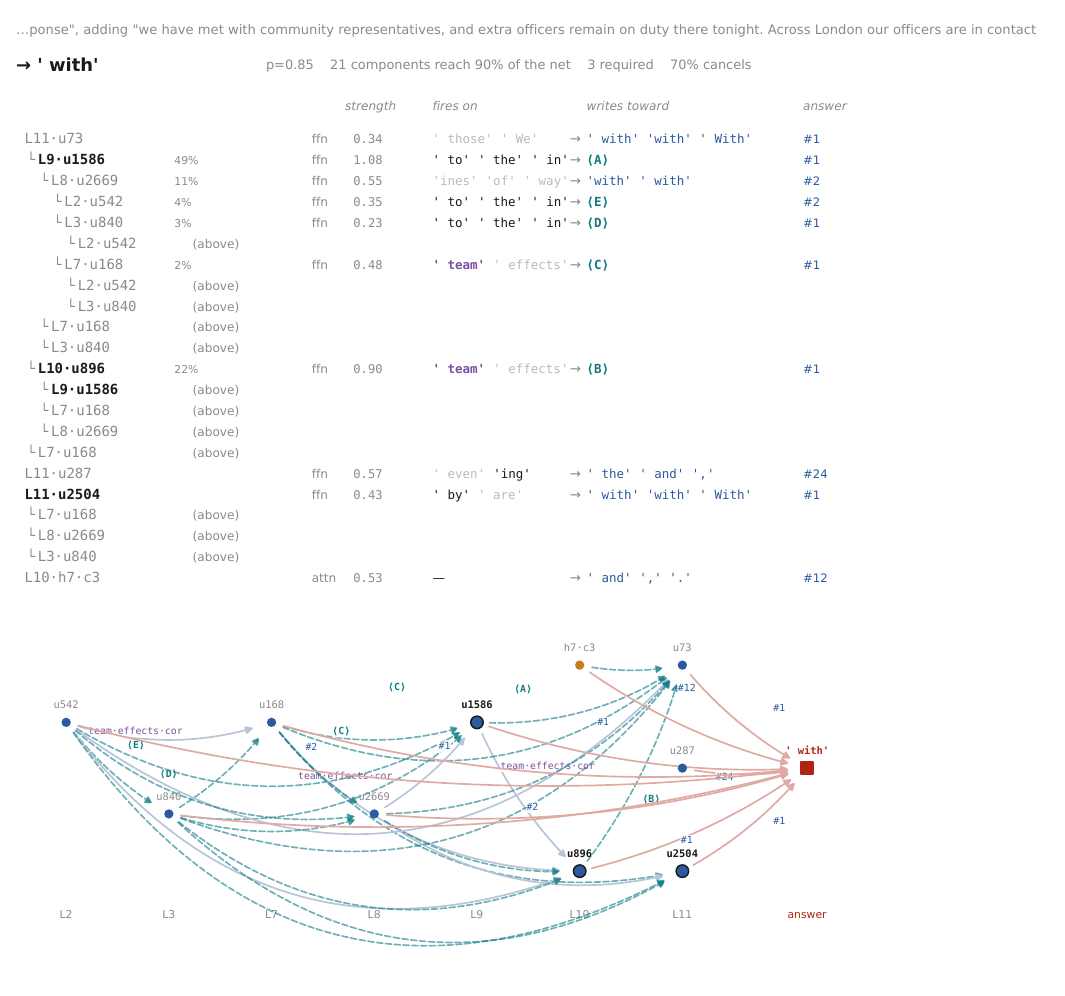}
\caption{OPT 125M, prediction 2. Of 20 predictions drawn at random, this is one of the four whose sufficient set is nearest the median, 13.}
\end{figure}
\clearpage
\begin{figure}[H]
\centering
\includegraphics[width=\textwidth,height=0.45\textheight,keepaspectratio]{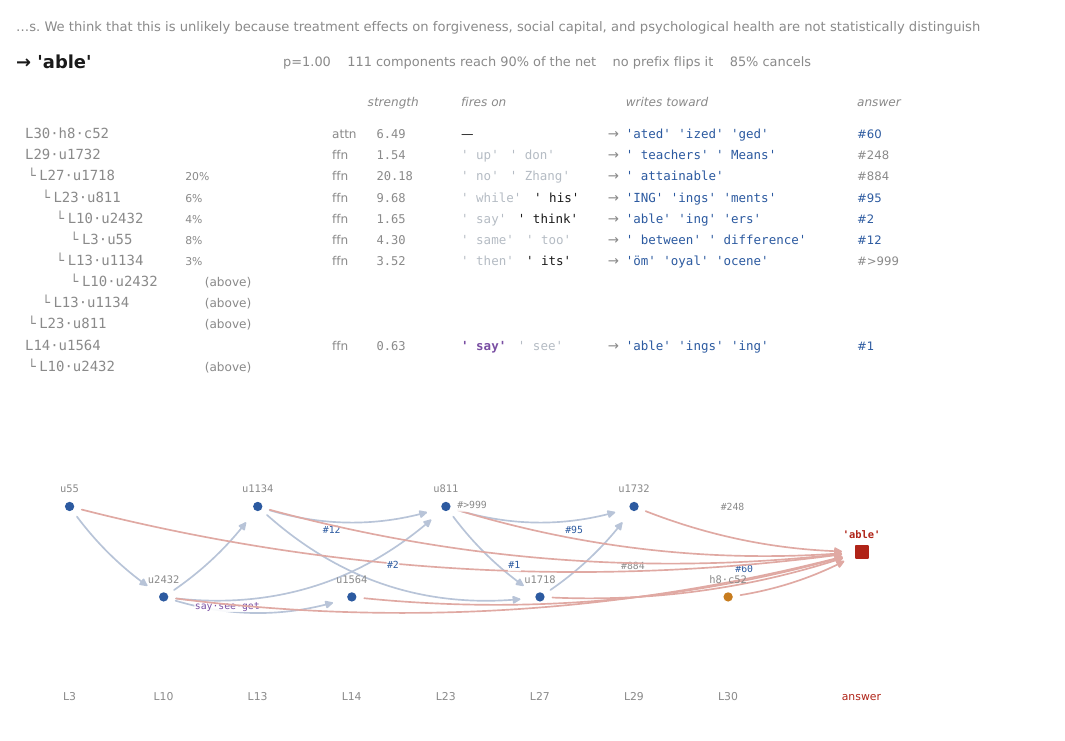}\\[1.2em]
\includegraphics[width=\textwidth,height=0.45\textheight,keepaspectratio]{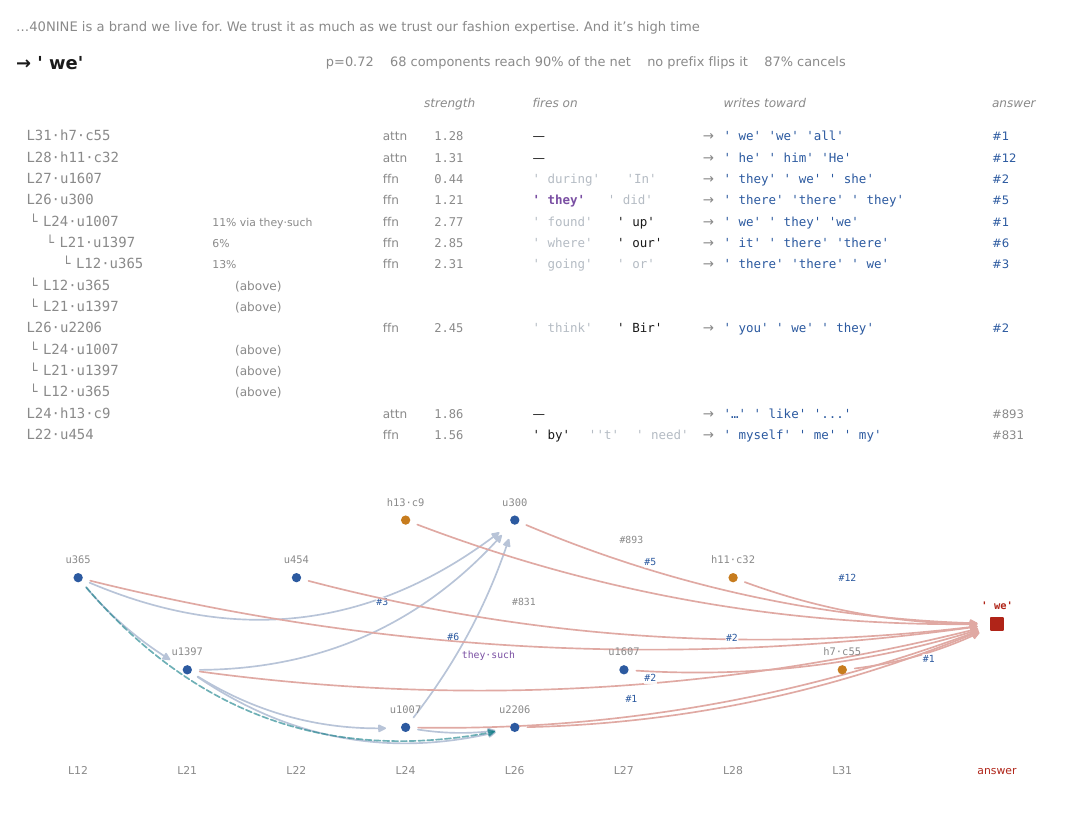}
\caption{SmolLM2 360M, predictions 1 and 2. Of 20 predictions drawn at random, these are two of the four whose sufficient set is nearest the median, 10.}
\end{figure}
\clearpage
\begin{figure}[H]
\centering
\includegraphics[width=\textwidth,height=0.45\textheight,keepaspectratio]{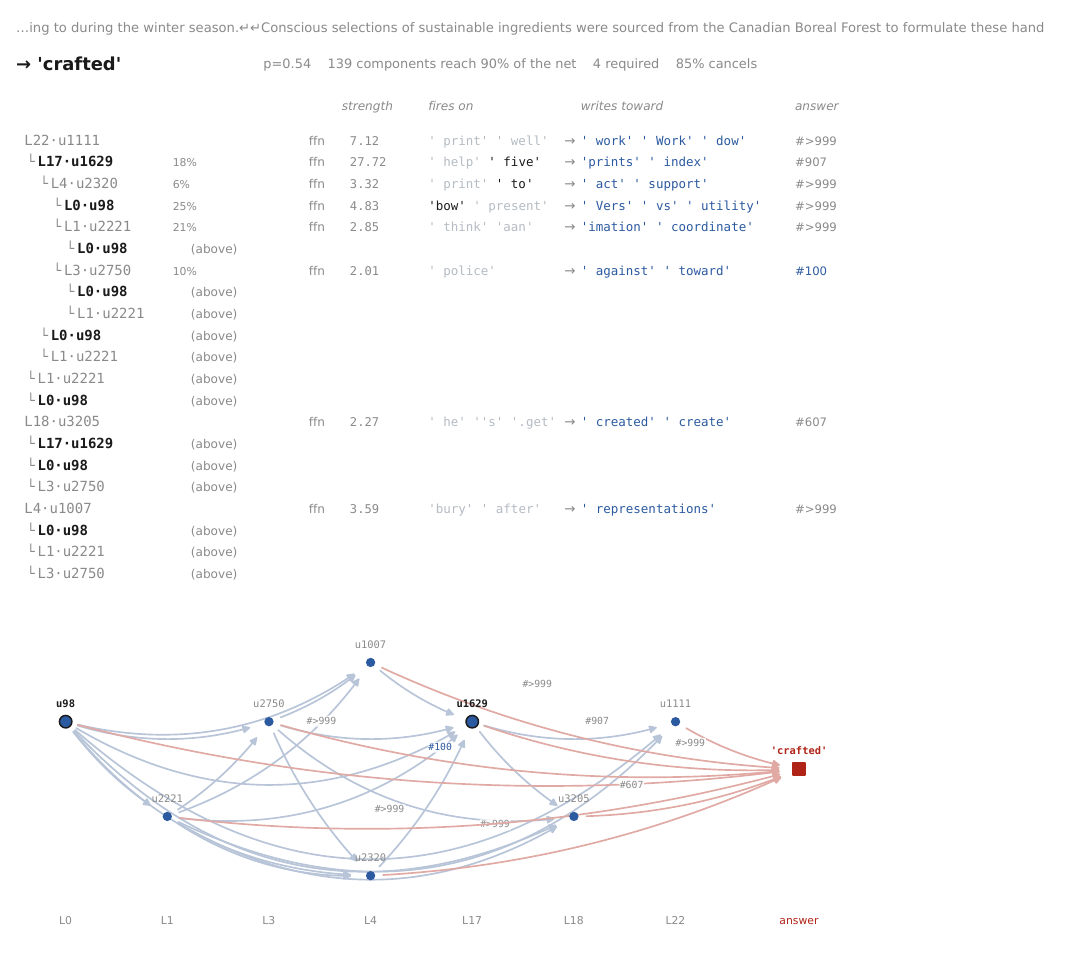}\\[1.2em]
\includegraphics[width=\textwidth,height=0.45\textheight,keepaspectratio]{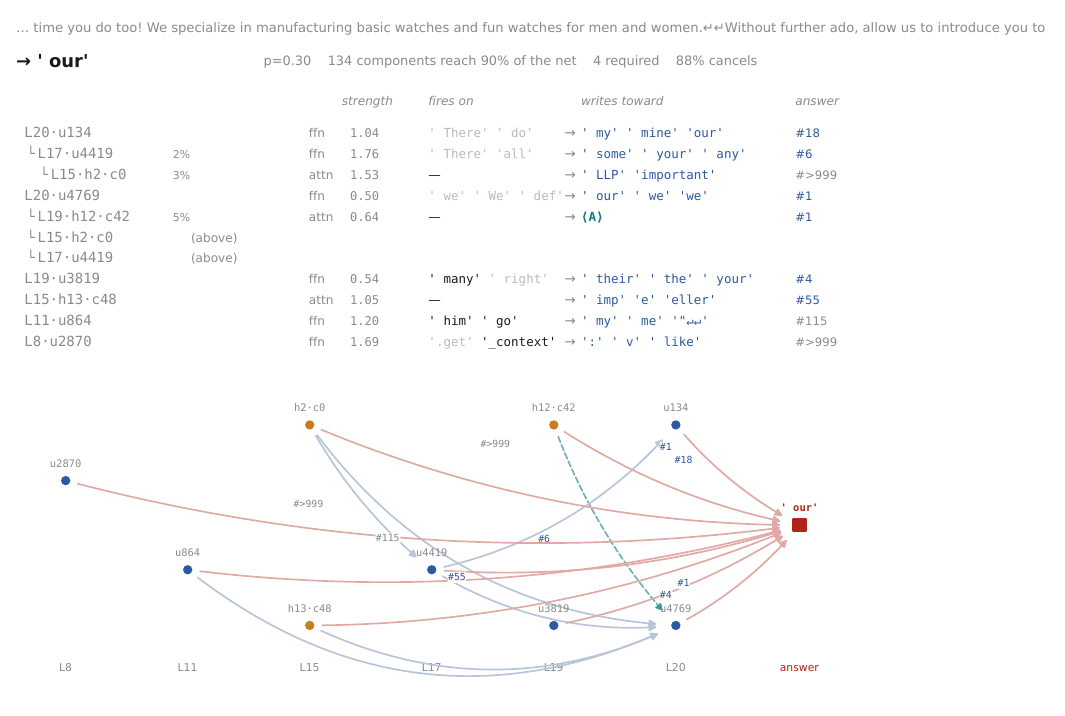}
\caption{Qwen2.5 0.5B, predictions 1 and 2. Of 20 predictions drawn at random, these are two of the four whose sufficient set is nearest the median, 9.}
\end{figure}
\clearpage
\begin{figure}[H]
\centering
\includegraphics[width=\textwidth,height=0.45\textheight,keepaspectratio]{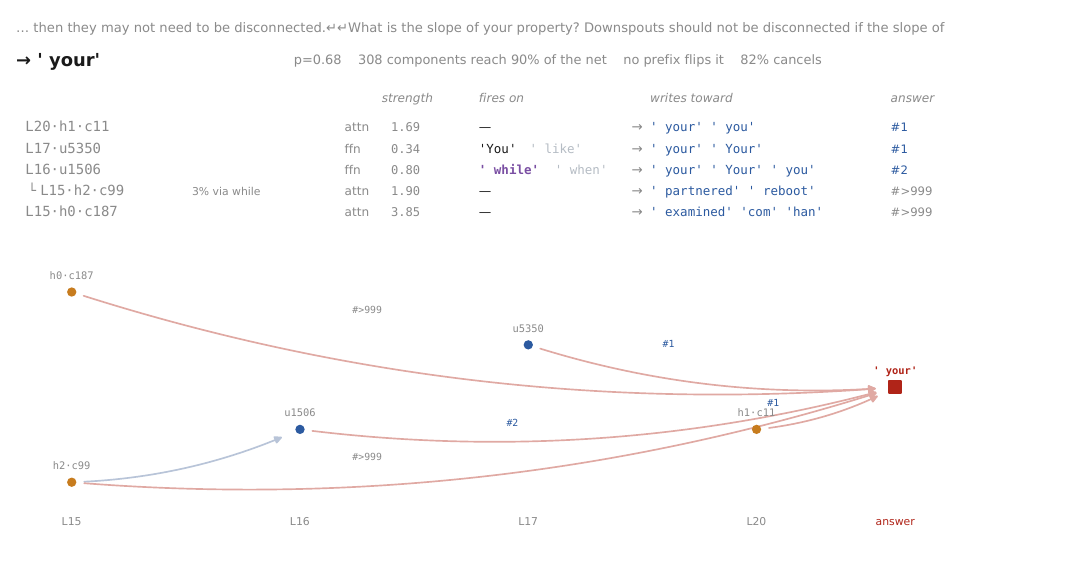}\\[1.2em]
\includegraphics[width=\textwidth,height=0.45\textheight,keepaspectratio]{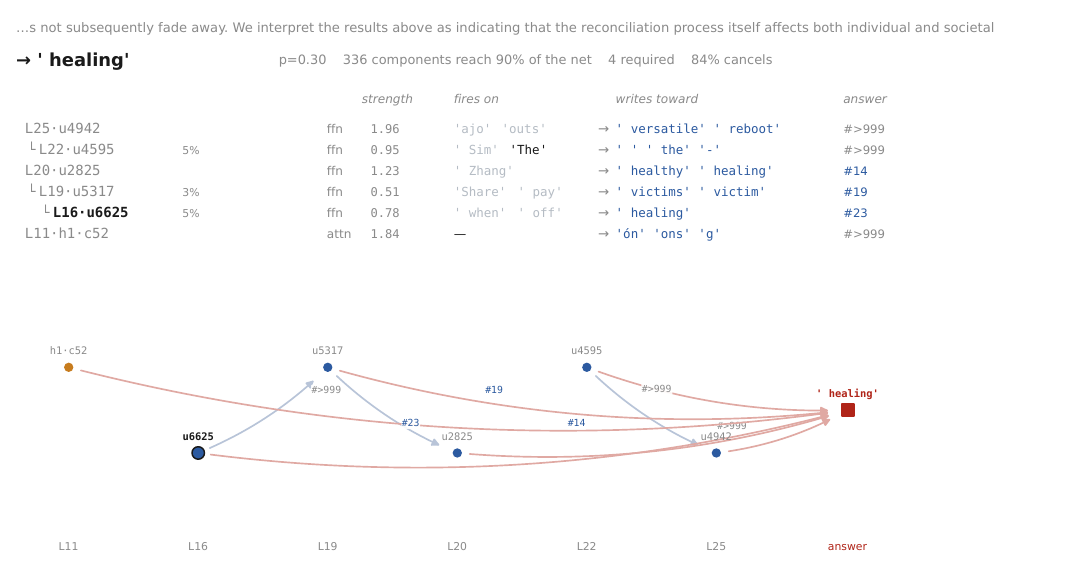}
\caption{Gemma-3 1B, predictions 1 and 2. Of 20 predictions drawn at random, these are two of the four whose sufficient set is nearest the median, 6.}
\end{figure}
\clearpage
\begin{figure}[H]
\centering
\includegraphics[width=\textwidth,height=0.45\textheight,keepaspectratio]{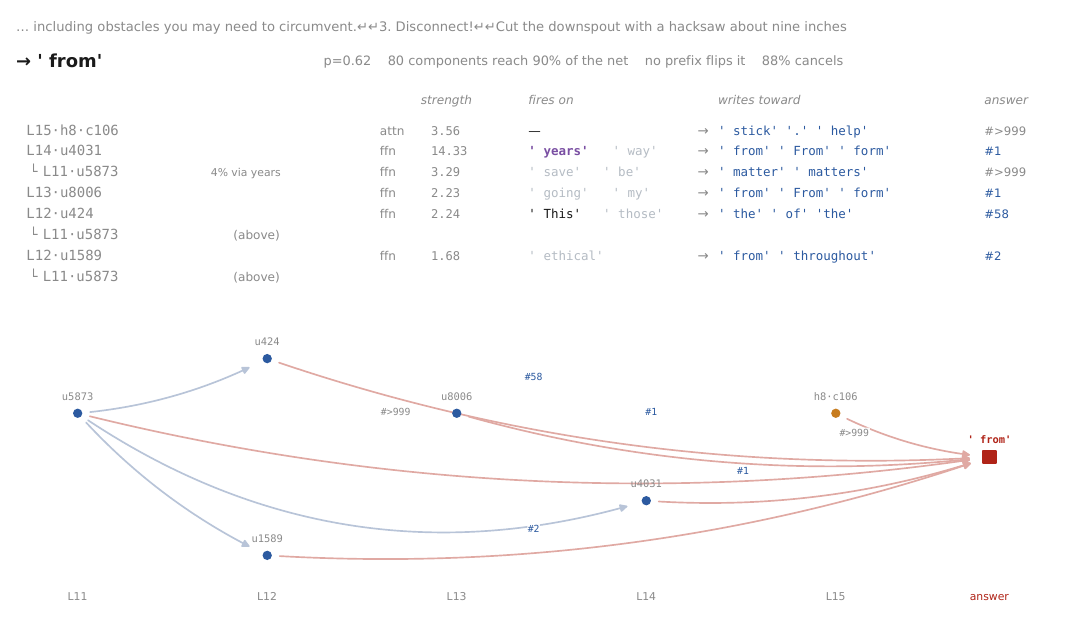}\\[1.2em]
\includegraphics[width=\textwidth,height=0.45\textheight,keepaspectratio]{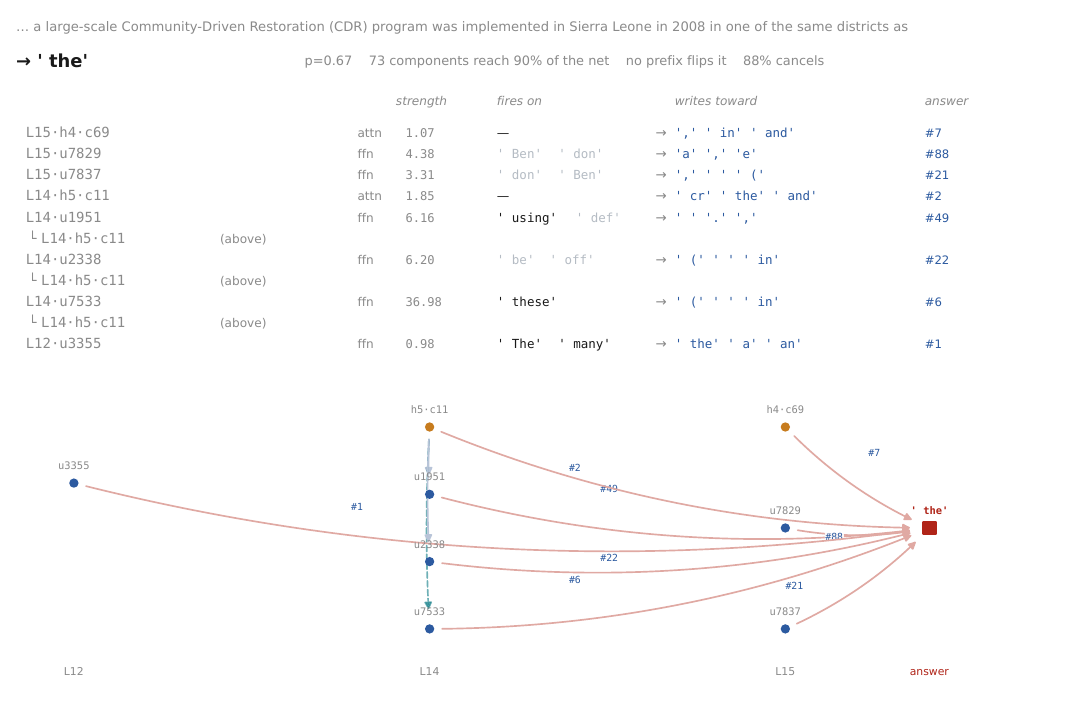}
\caption{OLMo-2 1B, predictions 1 and 2. Of 20 predictions drawn at random, these are two of the four whose sufficient set is nearest the median, 6.}
\end{figure}
\clearpage
\begin{figure}[H]
\centering
\includegraphics[width=\textwidth,height=0.92\textheight,keepaspectratio]{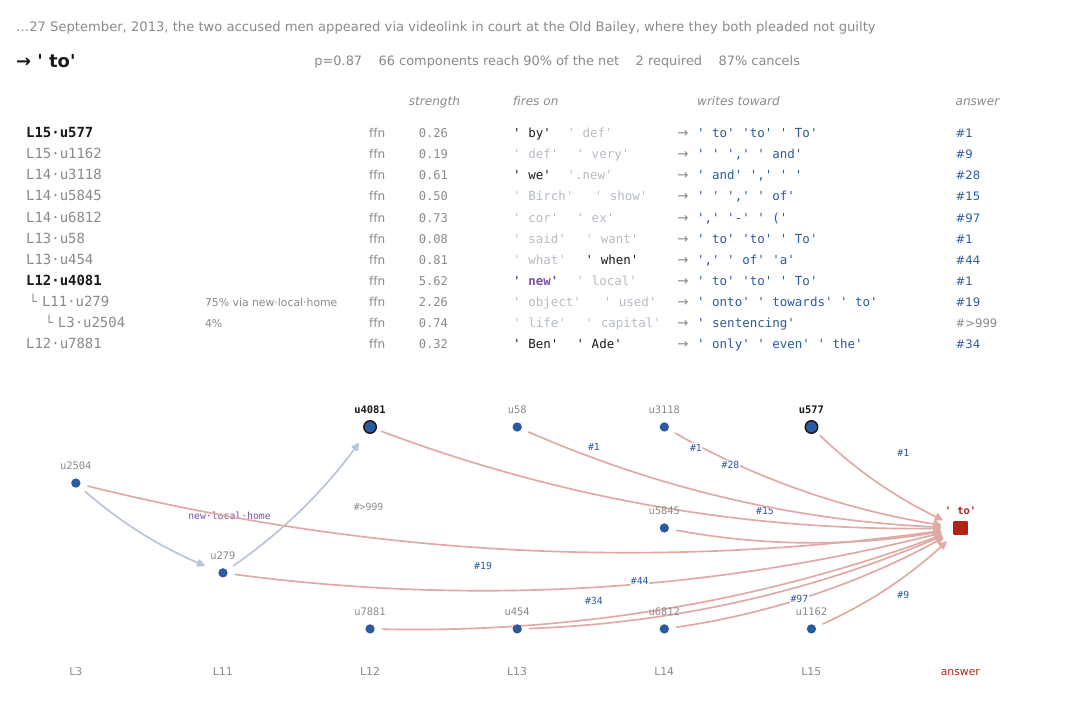}
\caption{Llama-3.2 1B, prediction 1. Of 20 predictions drawn at random, this is one of the four whose sufficient set is nearest the median, 13.}
\end{figure}
\clearpage
\begin{figure}[H]
\centering
\includegraphics[width=\textwidth,height=0.92\textheight,keepaspectratio]{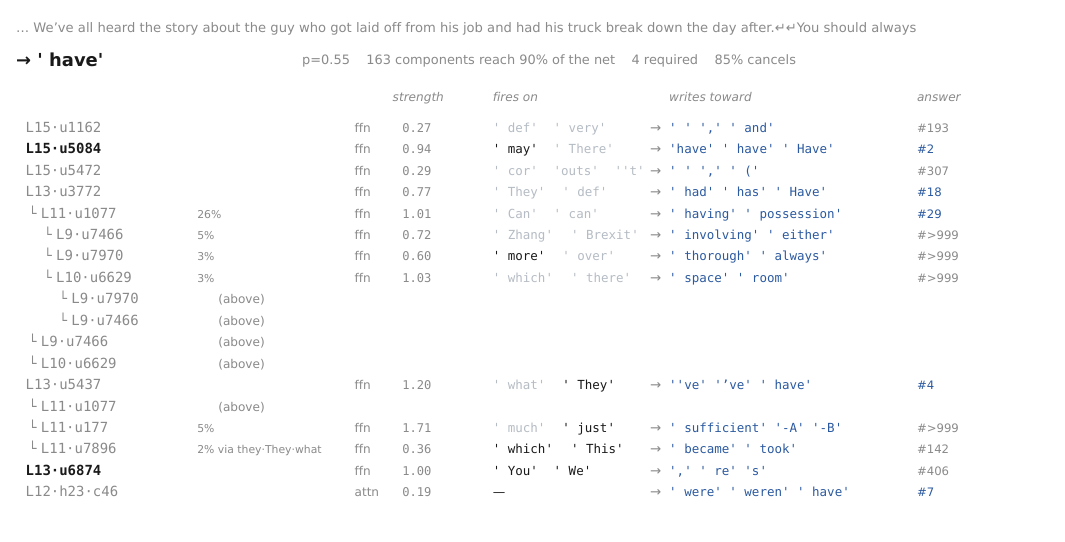}
\caption{Llama-3.2 1B, prediction 2: the set, 13 members. Of 20 predictions drawn at random, this is one of the four whose sufficient set is nearest the median, 13.}
\end{figure}
\clearpage
\begin{landscape}
\begin{figure}[H]
\centering
\includegraphics[width=\linewidth,height=0.88\textheight,keepaspectratio]{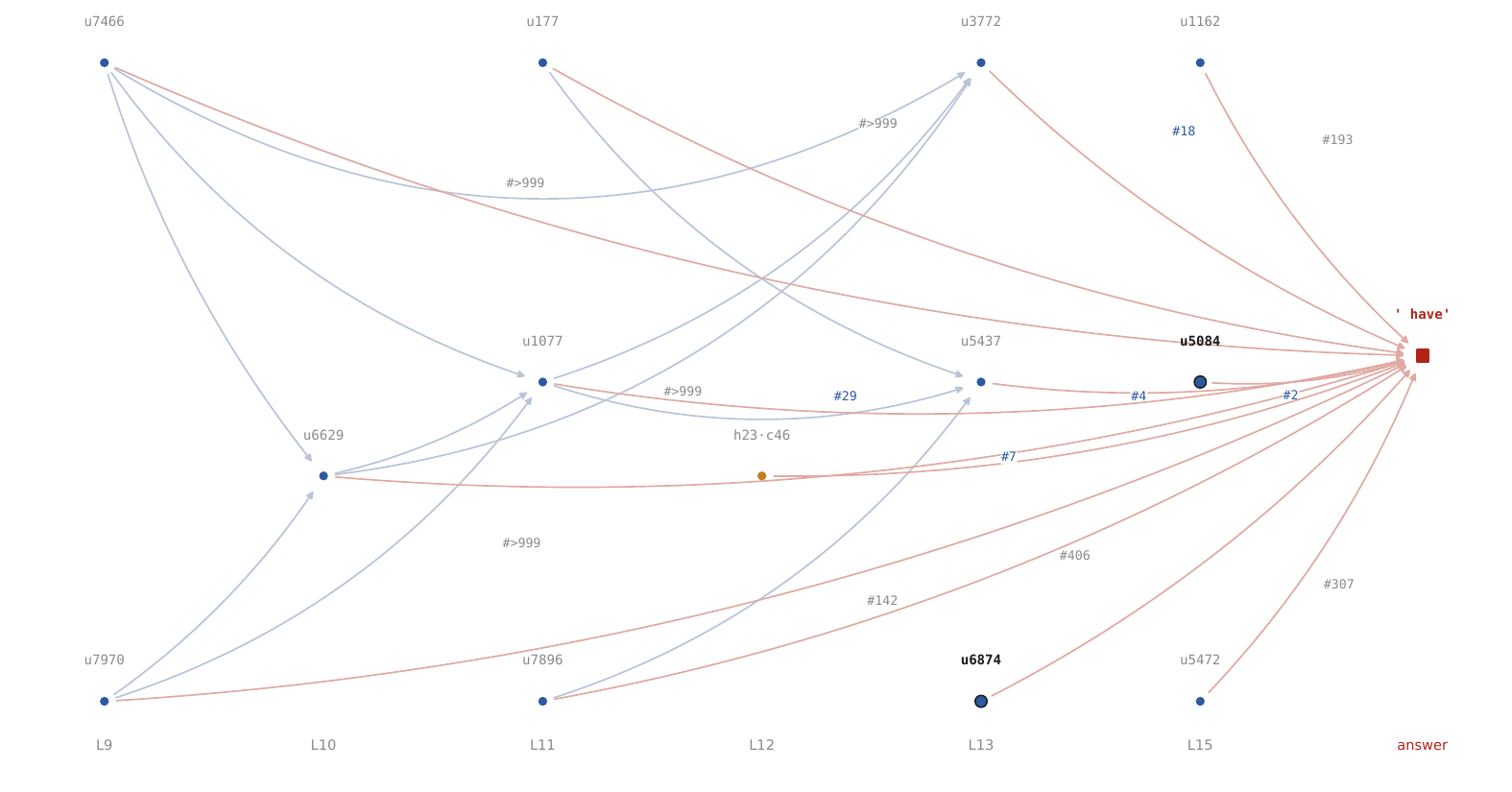}
\caption{Llama-3.2 1B, prediction 2: the graph, 13 members and 11 edges among them.}
\end{figure}
\end{landscape}
\clearpage
\begin{figure}[H]
\centering
\includegraphics[width=\textwidth,height=0.45\textheight,keepaspectratio]{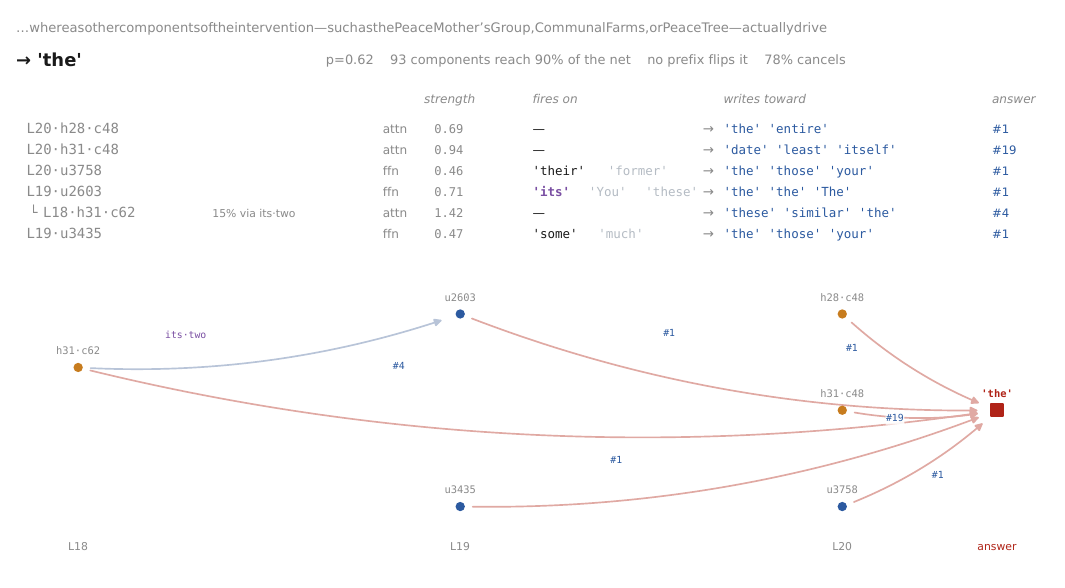}\\[1.2em]
\includegraphics[width=\textwidth,height=0.45\textheight,keepaspectratio]{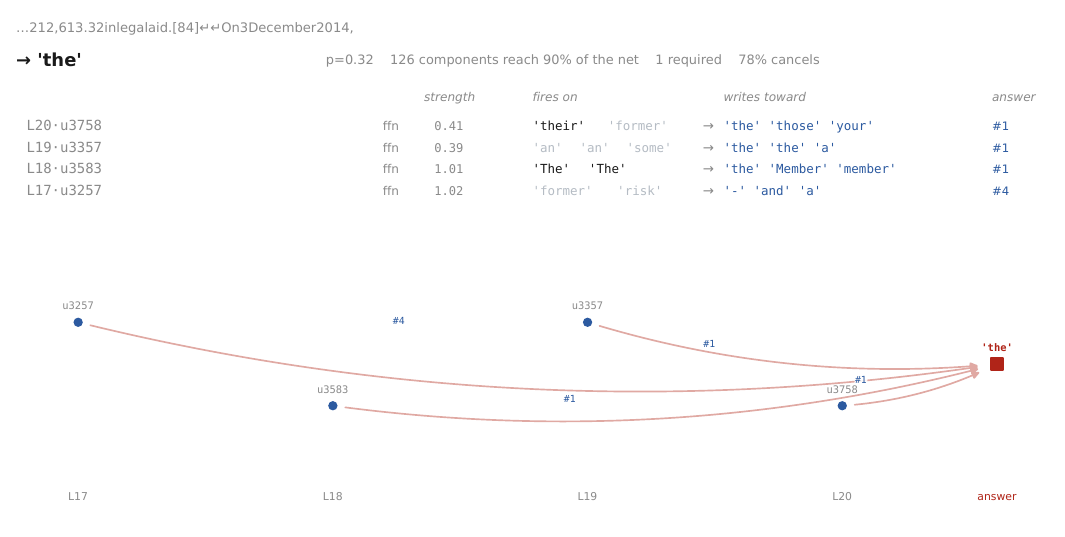}
\caption{TinyLlama 1.1B, predictions 1 and 2. Of 20 predictions drawn at random, these are two of the four whose sufficient set is nearest the median, 5.}
\end{figure}
\clearpage
\begin{figure}[H]
\centering
\includegraphics[width=\textwidth,height=0.45\textheight,keepaspectratio]{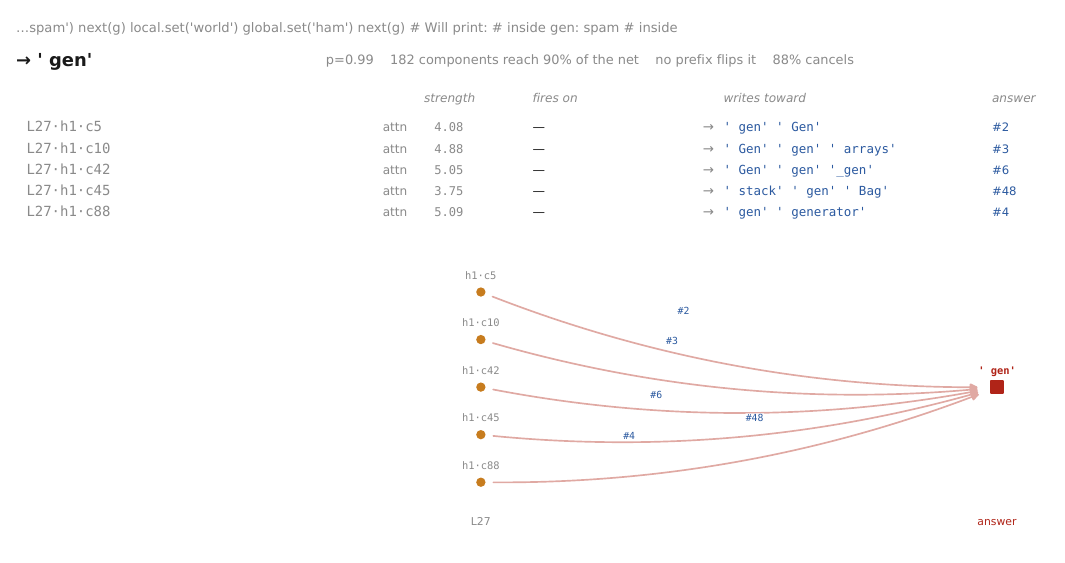}\\[1.2em]
\includegraphics[width=\textwidth,height=0.45\textheight,keepaspectratio]{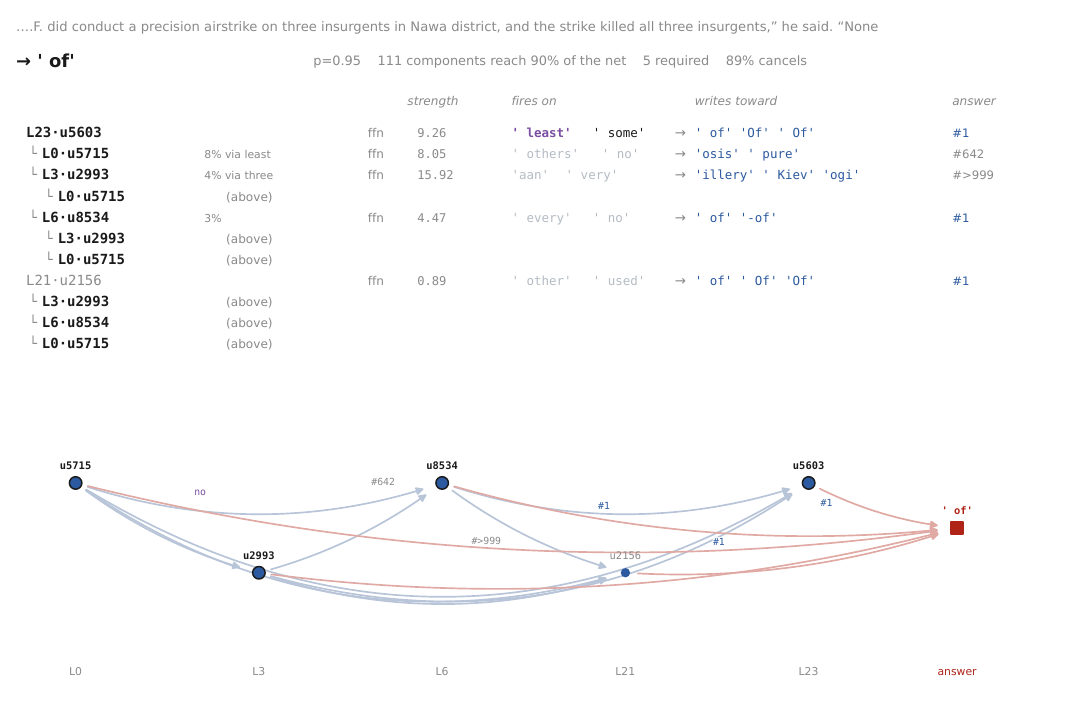}
\caption{Qwen2.5 1.5B, predictions 1 and 2. Of 19 predictions drawn at random, these are two of the four whose sufficient set is nearest the median, 6.}
\end{figure}
\clearpage
\begin{figure}[H]
\centering
\includegraphics[width=\textwidth,height=0.45\textheight,keepaspectratio]{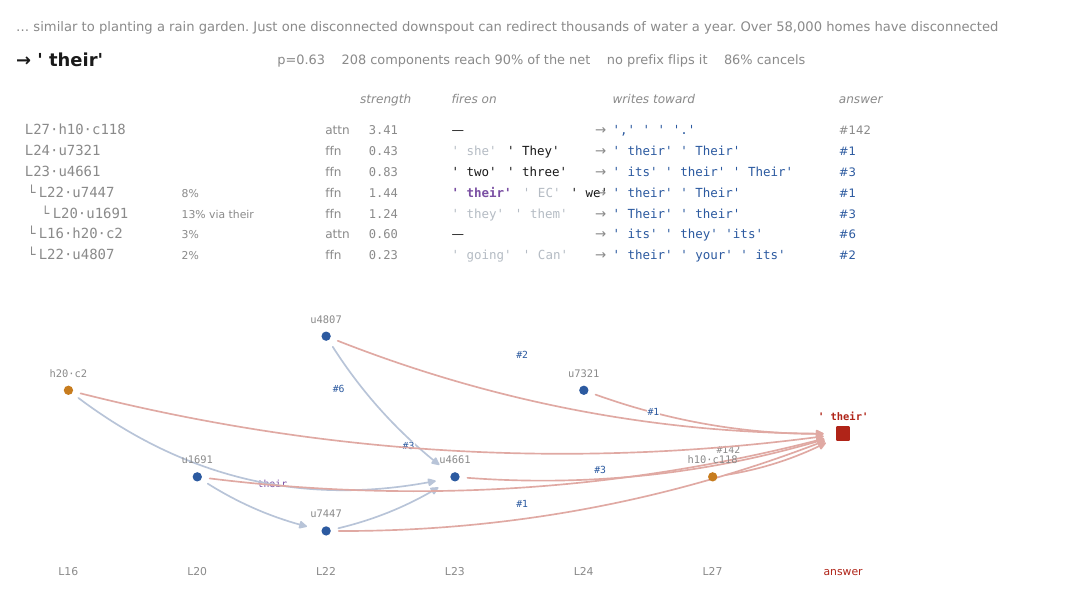}\\[1.2em]
\includegraphics[width=\textwidth,height=0.45\textheight,keepaspectratio]{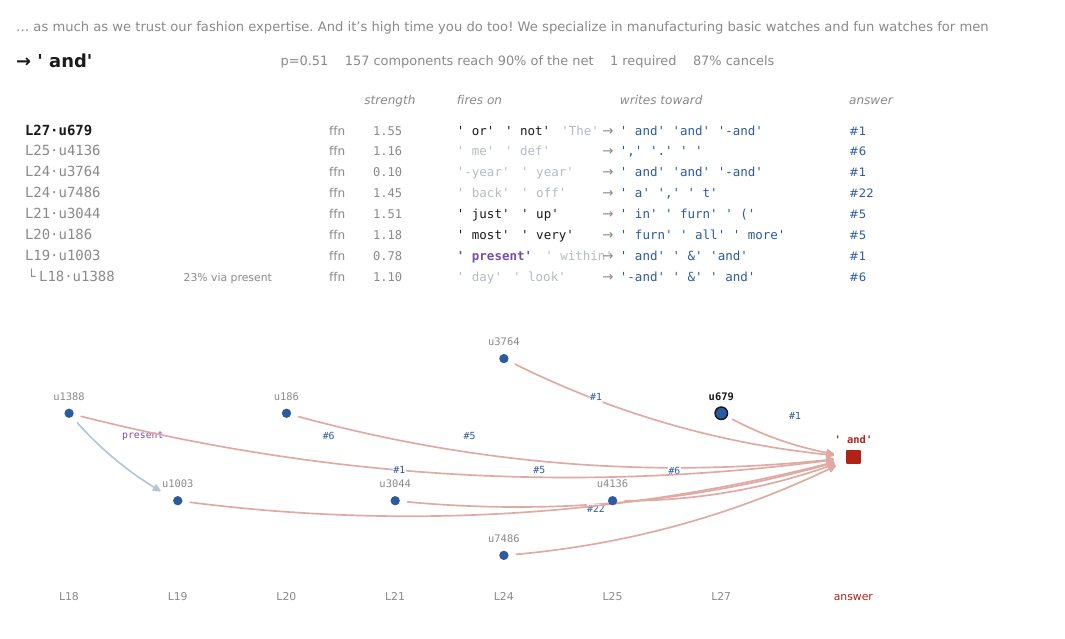}
\caption{Llama-3.2 3B, predictions 1 and 2. Of 20 predictions drawn at random, these are two of the four whose sufficient set is nearest the median, 8.}
\end{figure}
\clearpage
\begin{figure}[H]
\centering
\includegraphics[width=\textwidth,height=0.45\textheight,keepaspectratio]{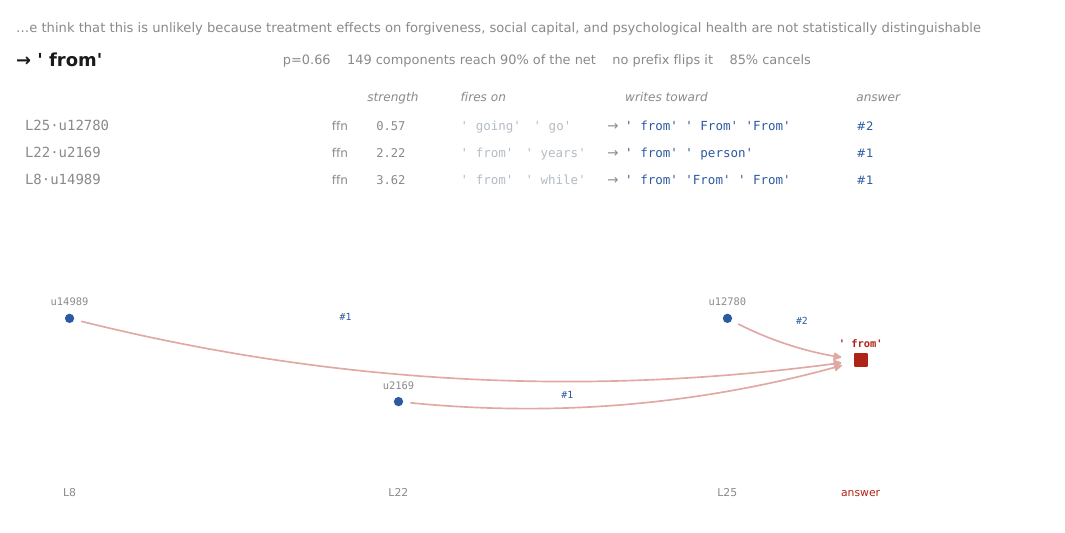}\\[1.2em]
\includegraphics[width=\textwidth,height=0.45\textheight,keepaspectratio]{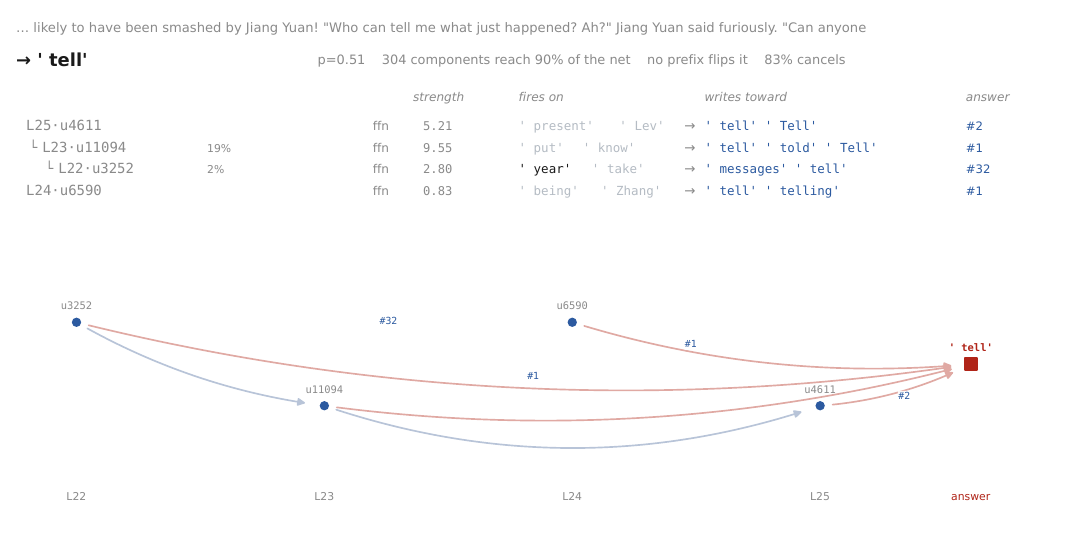}
\caption{Qwen2.5 7B, predictions 1 and 2. Of 17 predictions drawn at random, these are two of the four whose sufficient set is nearest the median, 4.}
\end{figure}
\clearpage
\begin{figure}[H]
\centering
\includegraphics[width=\textwidth,height=0.45\textheight,keepaspectratio]{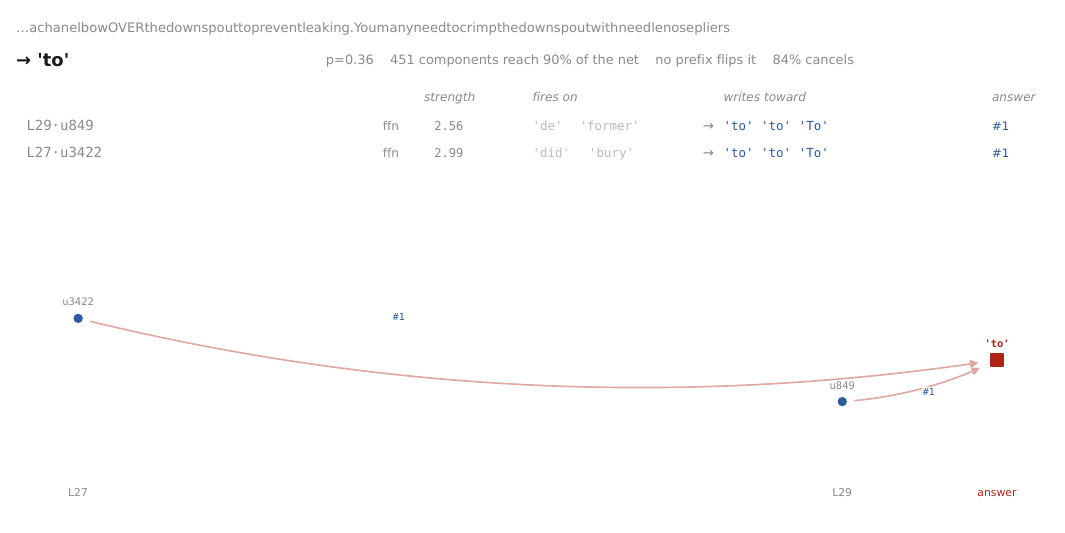}\\[1.2em]
\includegraphics[width=\textwidth,height=0.45\textheight,keepaspectratio]{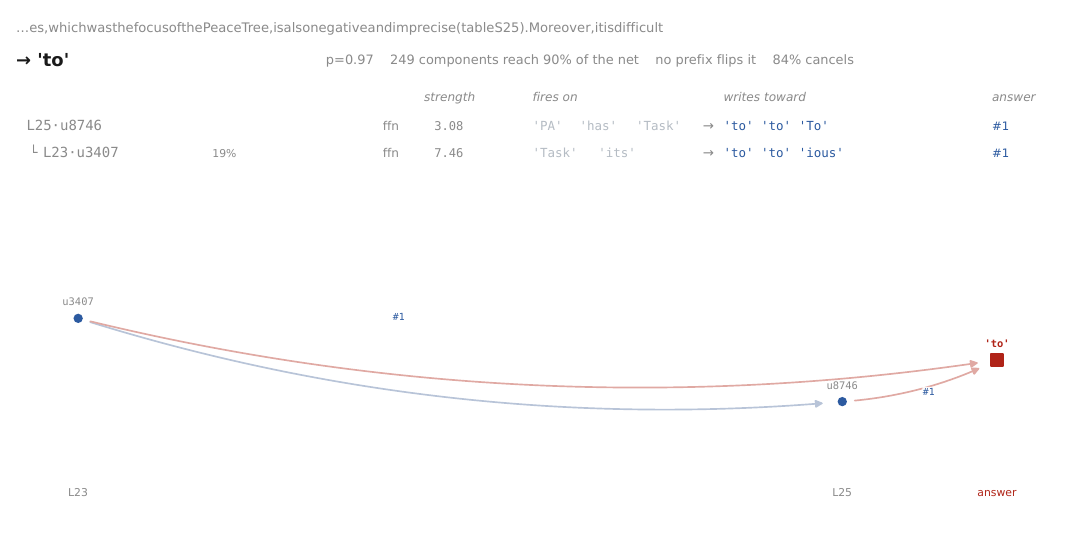}
\caption{Mistral 7B, predictions 1 and 2. Of 20 predictions drawn at random, these are two of the four whose sufficient set is nearest the median, 2.}
\end{figure}

%% file: galpages_concept.tex
\clearpage
\begin{figure}[H]
\centering
\includegraphics[width=\textwidth,height=0.92\textheight,keepaspectratio]{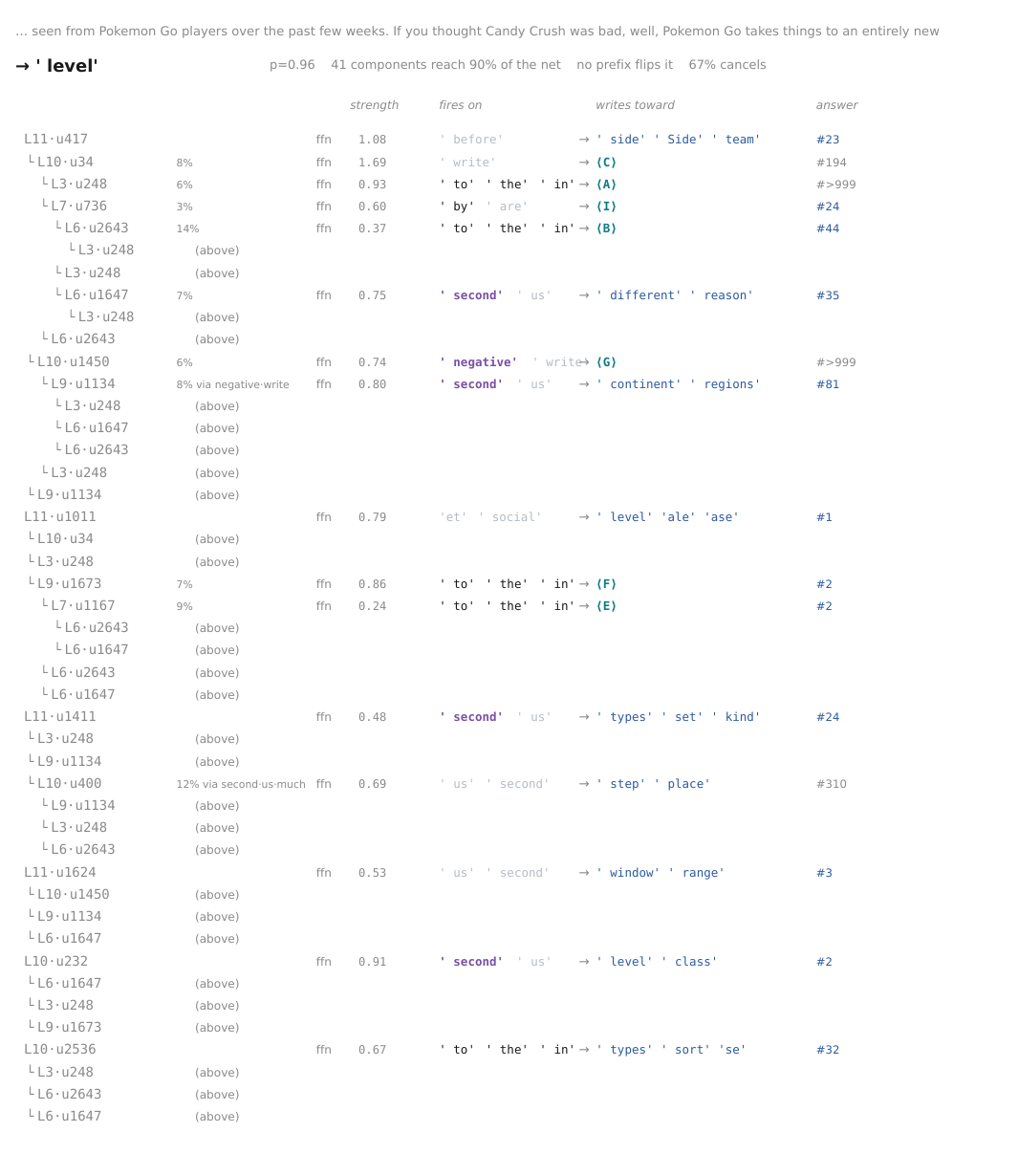}
\caption{OPT 125M, prediction 3: the set, 16 members. Selected for containing edges no token describes. Of 20 predictions drawn at random, this is one of the four whose sufficient set is nearest the median, 13.}
\end{figure}
\clearpage
\begin{landscape}
\begin{figure}[H]
\centering
\includegraphics[width=\linewidth,height=0.88\textheight,keepaspectratio]{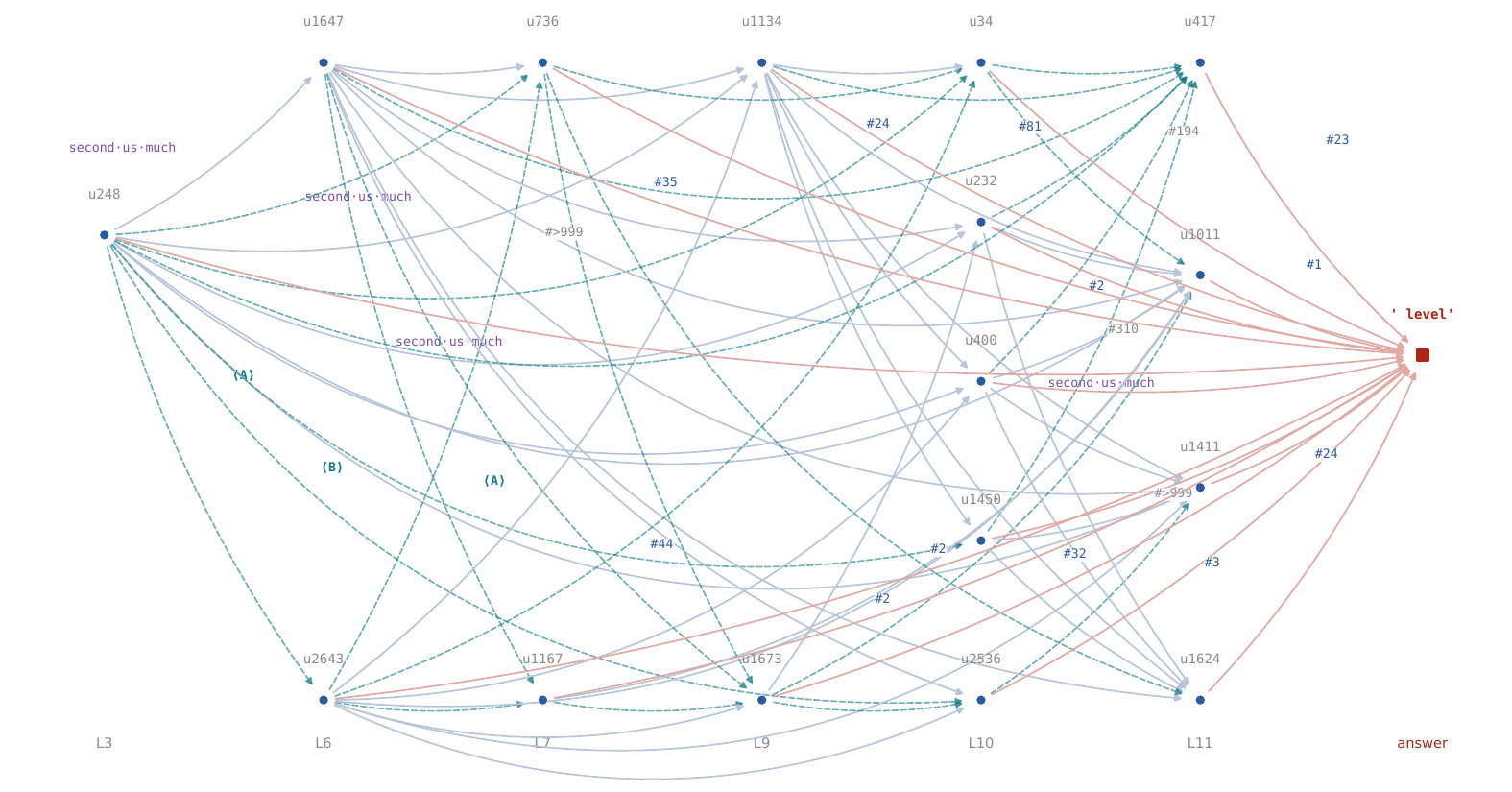}
\caption{OPT 125M, prediction 3: the graph, 16 members and 59 edges among them.}
\end{figure}
\end{landscape}
\clearpage
\begin{figure}[H]
\centering
\includegraphics[width=\textwidth,height=0.92\textheight,keepaspectratio]{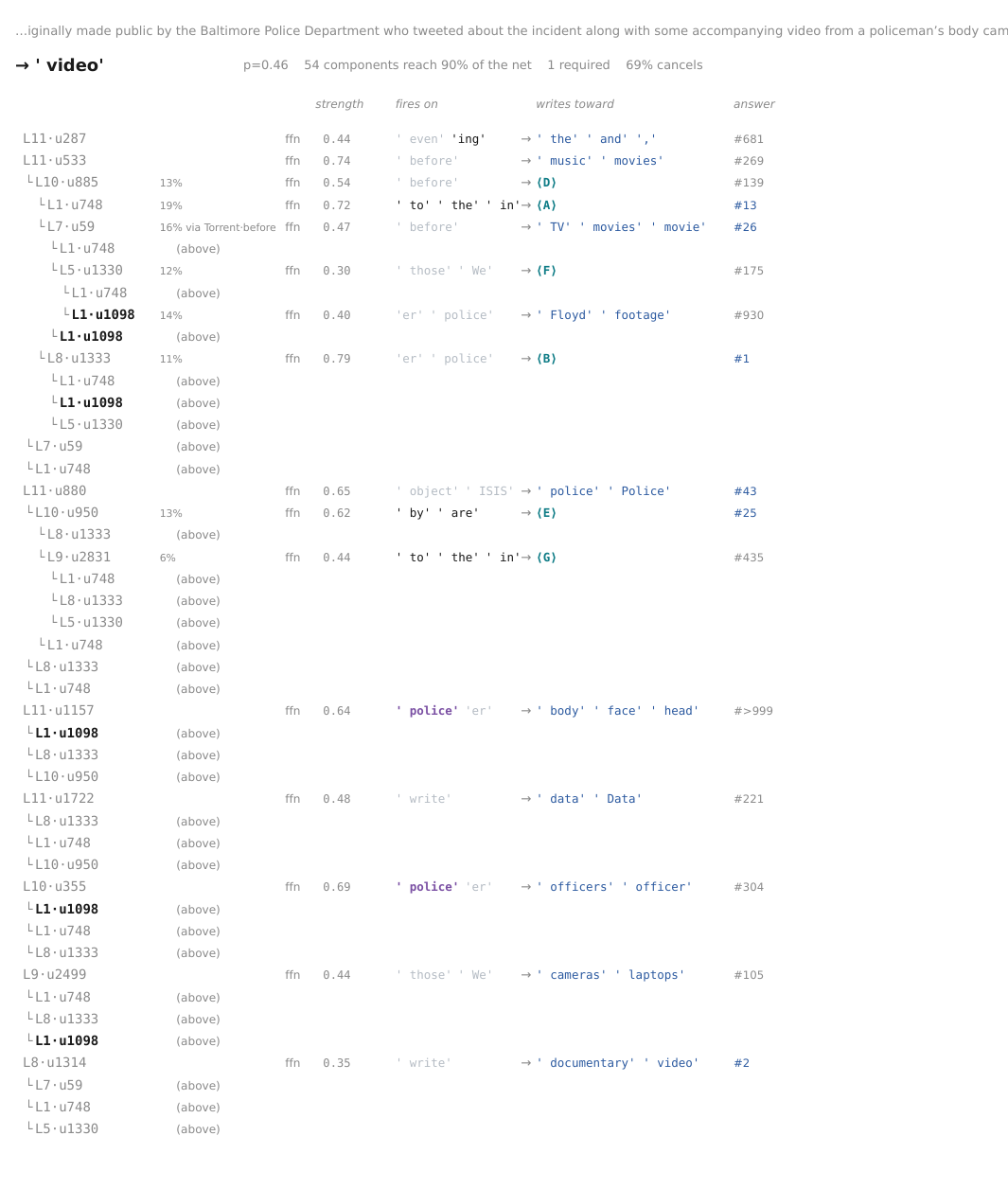}
\caption{OPT 125M, prediction 4: the set, 16 members. Selected for containing edges no token describes. Of 20 predictions drawn at random, this is one of the four whose sufficient set is nearest the median, 13.}
\end{figure}
\clearpage
\begin{landscape}
\begin{figure}[H]
\centering
\includegraphics[width=\linewidth,height=0.88\textheight,keepaspectratio]{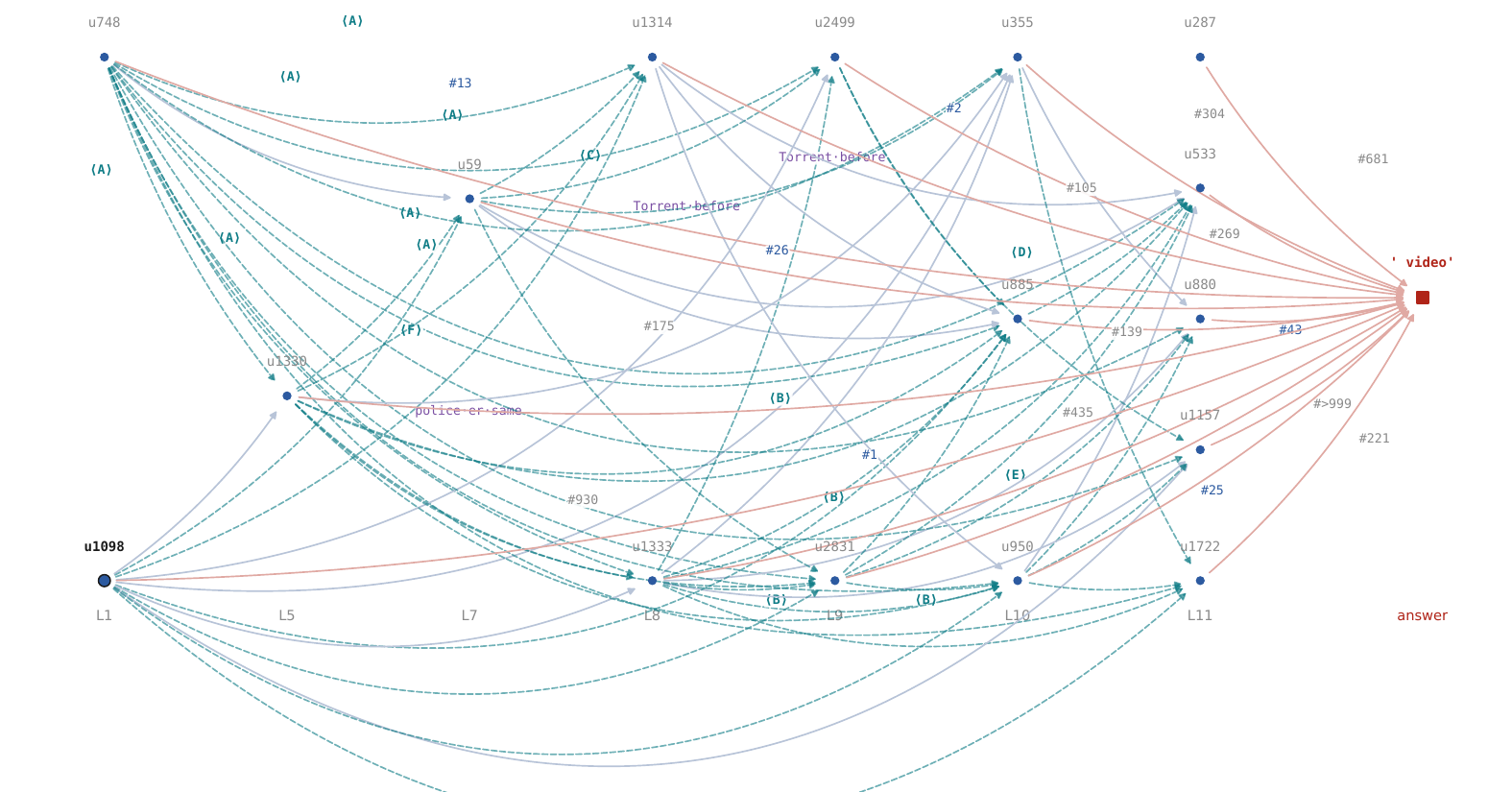}
\caption{OPT 125M, prediction 4: the graph, 16 members and 63 edges among them.}
\end{figure}
\end{landscape}
\clearpage
\begin{figure}[H]
\centering
\includegraphics[width=\textwidth,height=0.92\textheight,keepaspectratio]{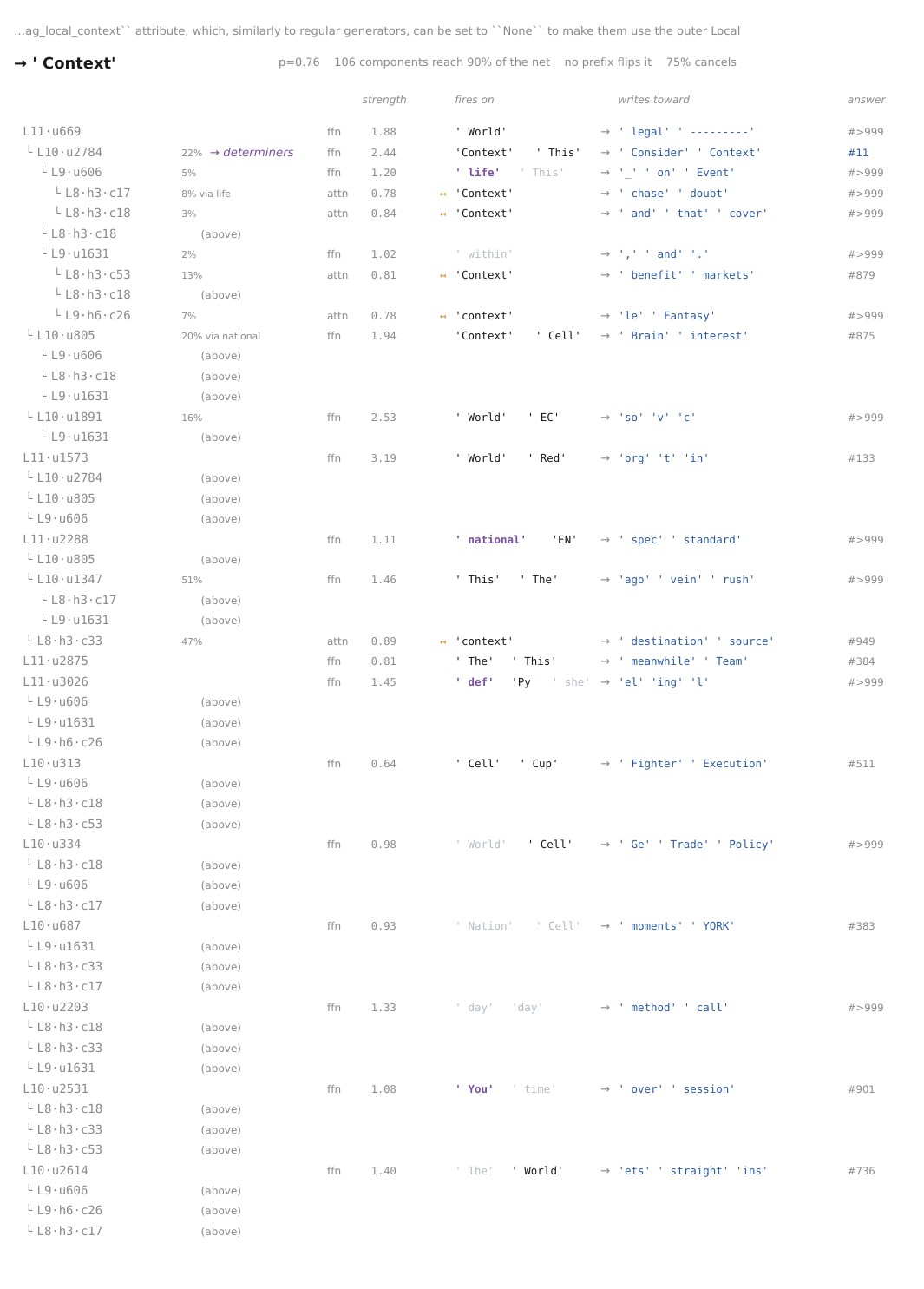}
\caption{sigmoid, no shaping, prediction 3: the set, 22 members. Selected for containing edges no token describes. Of 16 predictions drawn at random, this is one of the four whose sufficient set is nearest the median, 25.}
\end{figure}
\clearpage
\begin{landscape}
\begin{figure}[H]
\centering
\includegraphics[width=\linewidth,height=0.88\textheight,keepaspectratio]{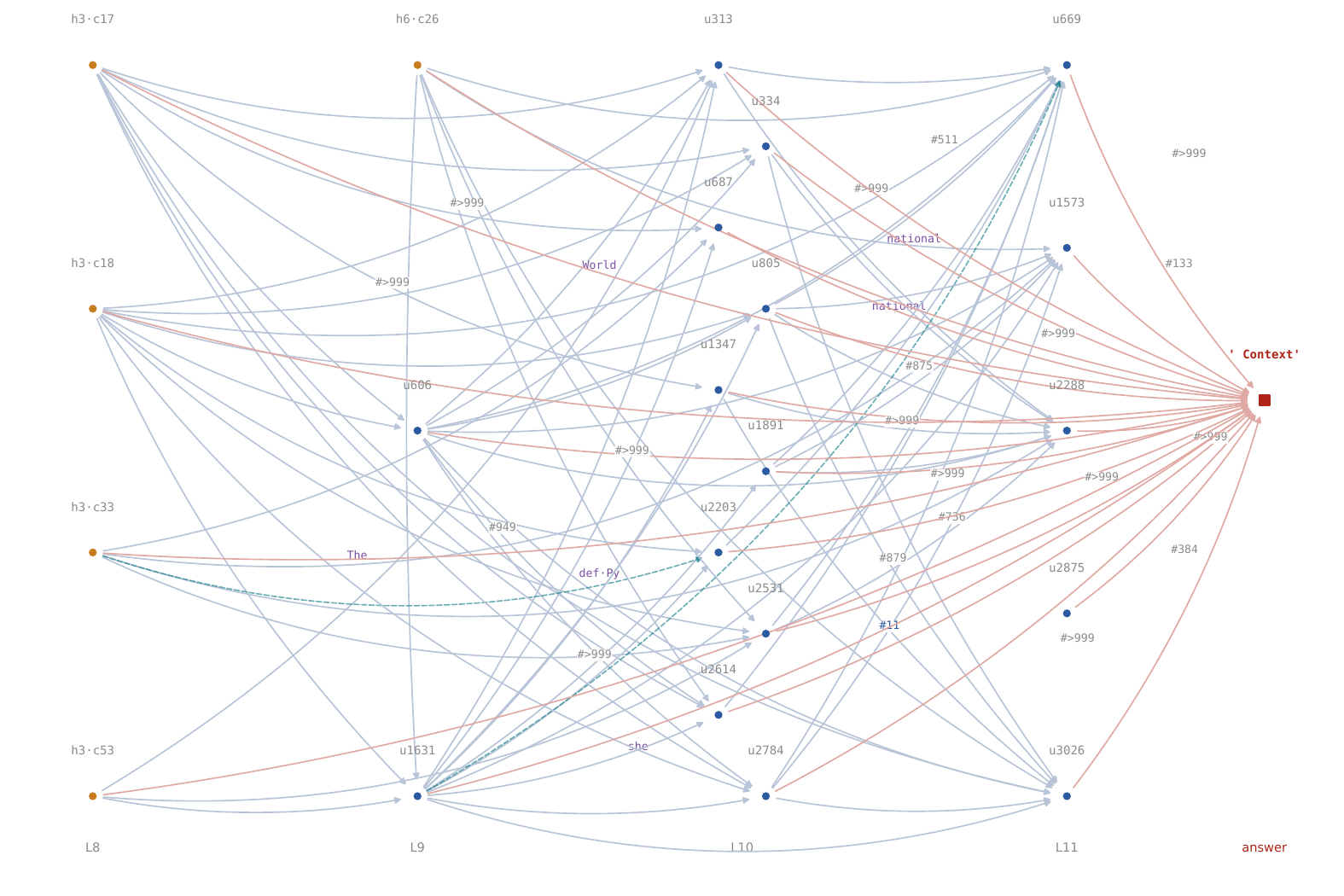}
\caption{sigmoid, no shaping, prediction 3: the graph, 22 members and 71 edges among them.}
\end{figure}
\end{landscape}
\clearpage
\begin{figure}[H]
\centering
\includegraphics[width=\textwidth,height=0.92\textheight,keepaspectratio]{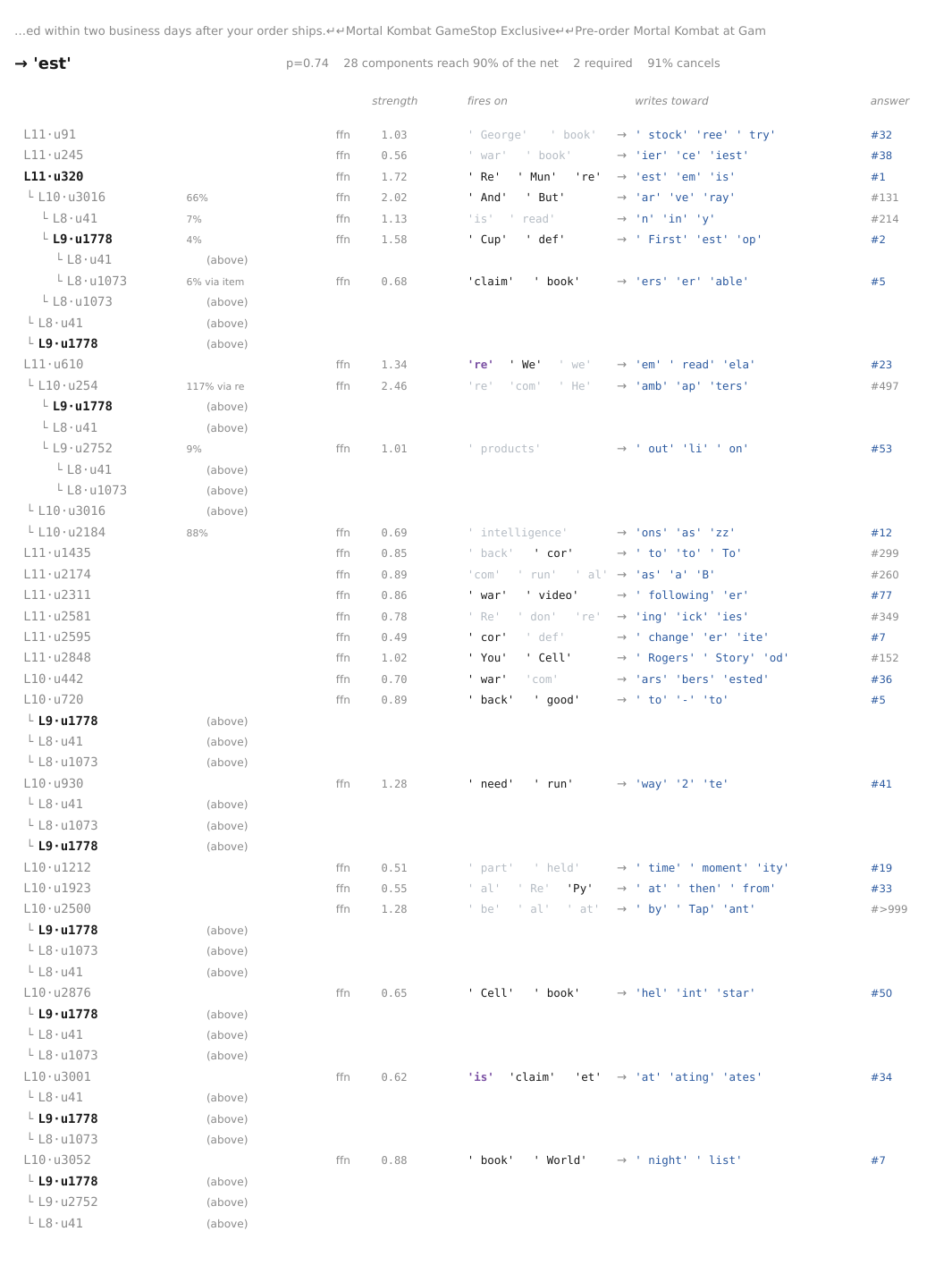}
\caption{sigmoid, no shaping, prediction 4: the set, 26 members. Selected for containing edges no token describes. Of 16 predictions drawn at random, this is one of the four whose sufficient set is nearest the median, 25.}
\end{figure}
\clearpage
\begin{landscape}
\begin{figure}[H]
\centering
\includegraphics[width=\linewidth,height=0.88\textheight,keepaspectratio]{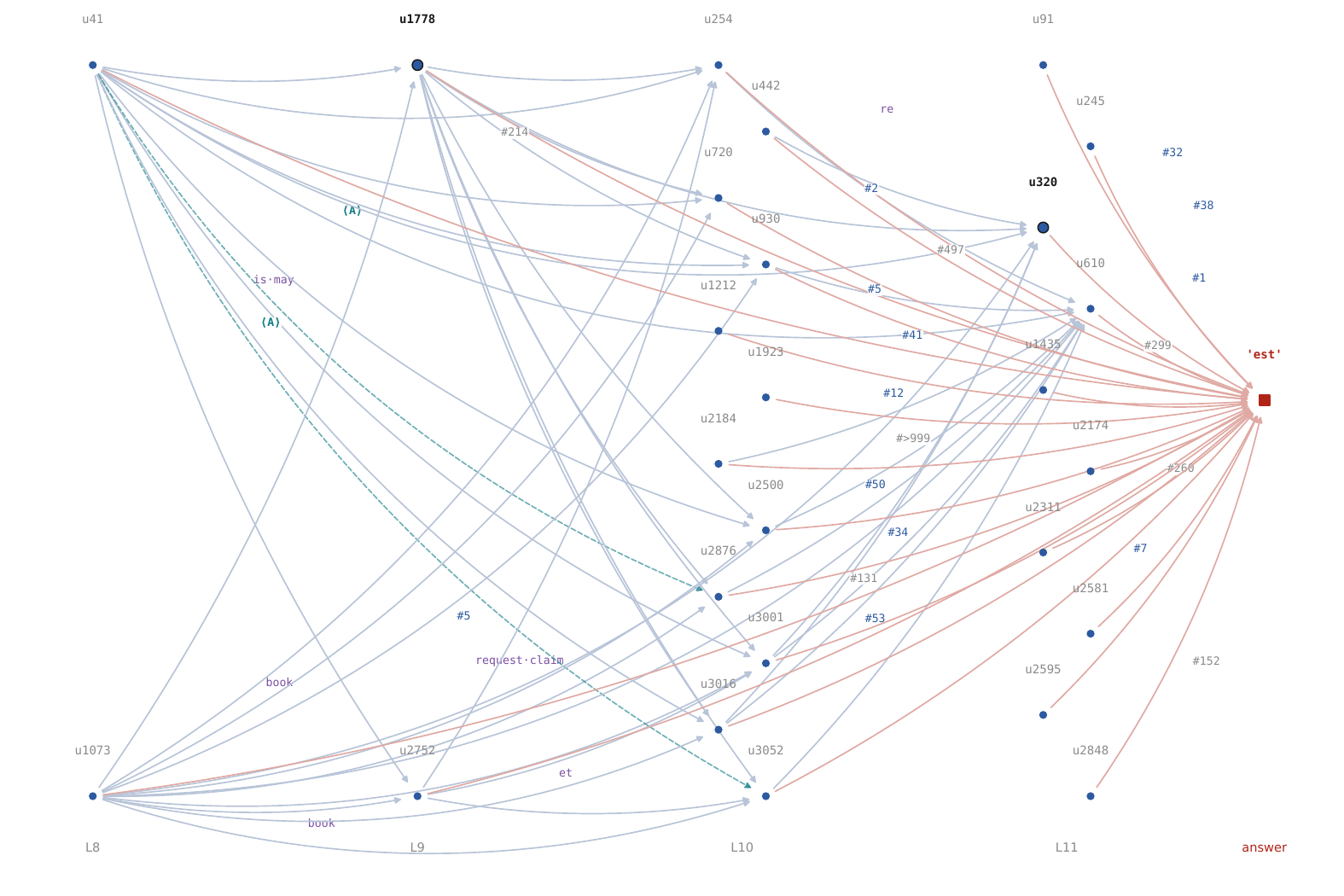}
\caption{sigmoid, no shaping, prediction 4: the graph, 26 members and 47 edges among them.}
\end{figure}
\end{landscape}
\clearpage
\begin{figure}[H]
\centering
\includegraphics[width=\textwidth,height=0.92\textheight,keepaspectratio]{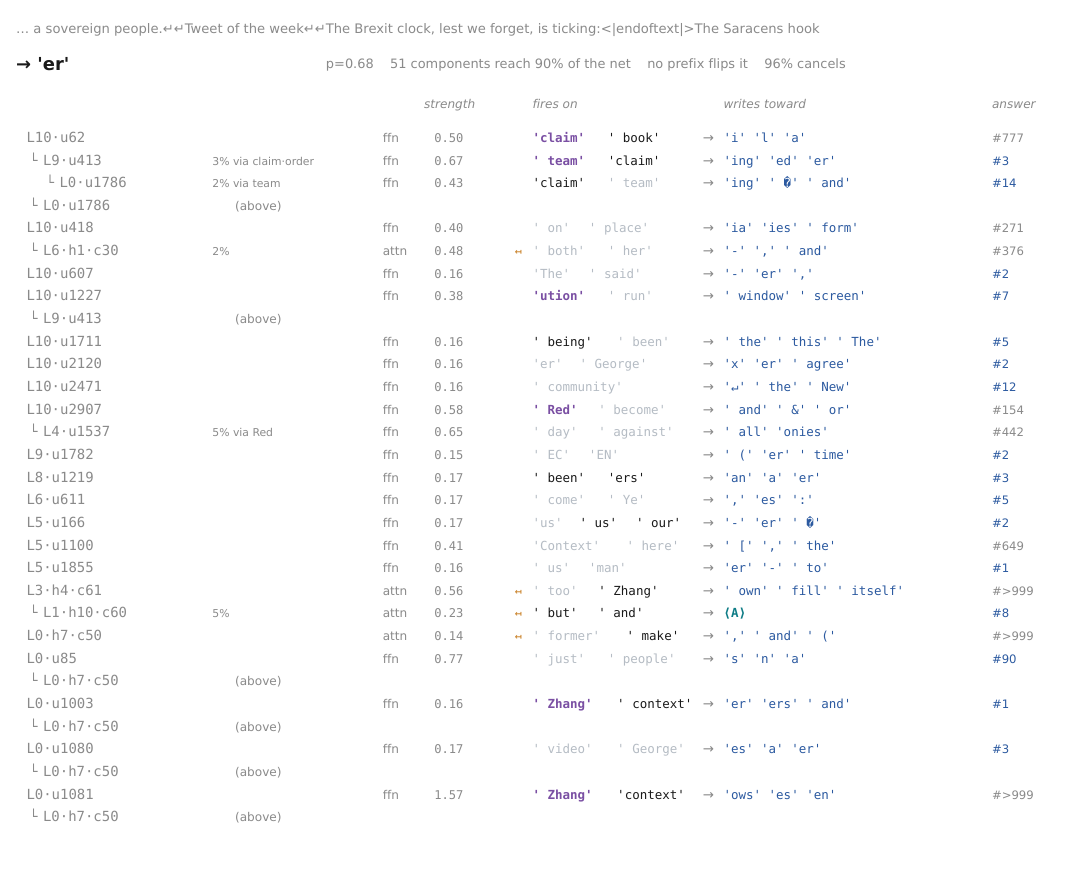}
\caption{baseline (GELU), prediction 3: the set, 25 members. Selected for containing edges no token describes. Of 11 predictions drawn at random, this is one of the four whose sufficient set is nearest the median, 24.}
\end{figure}
\clearpage
\begin{landscape}
\begin{figure}[H]
\centering
\includegraphics[width=\linewidth,height=0.88\textheight,keepaspectratio]{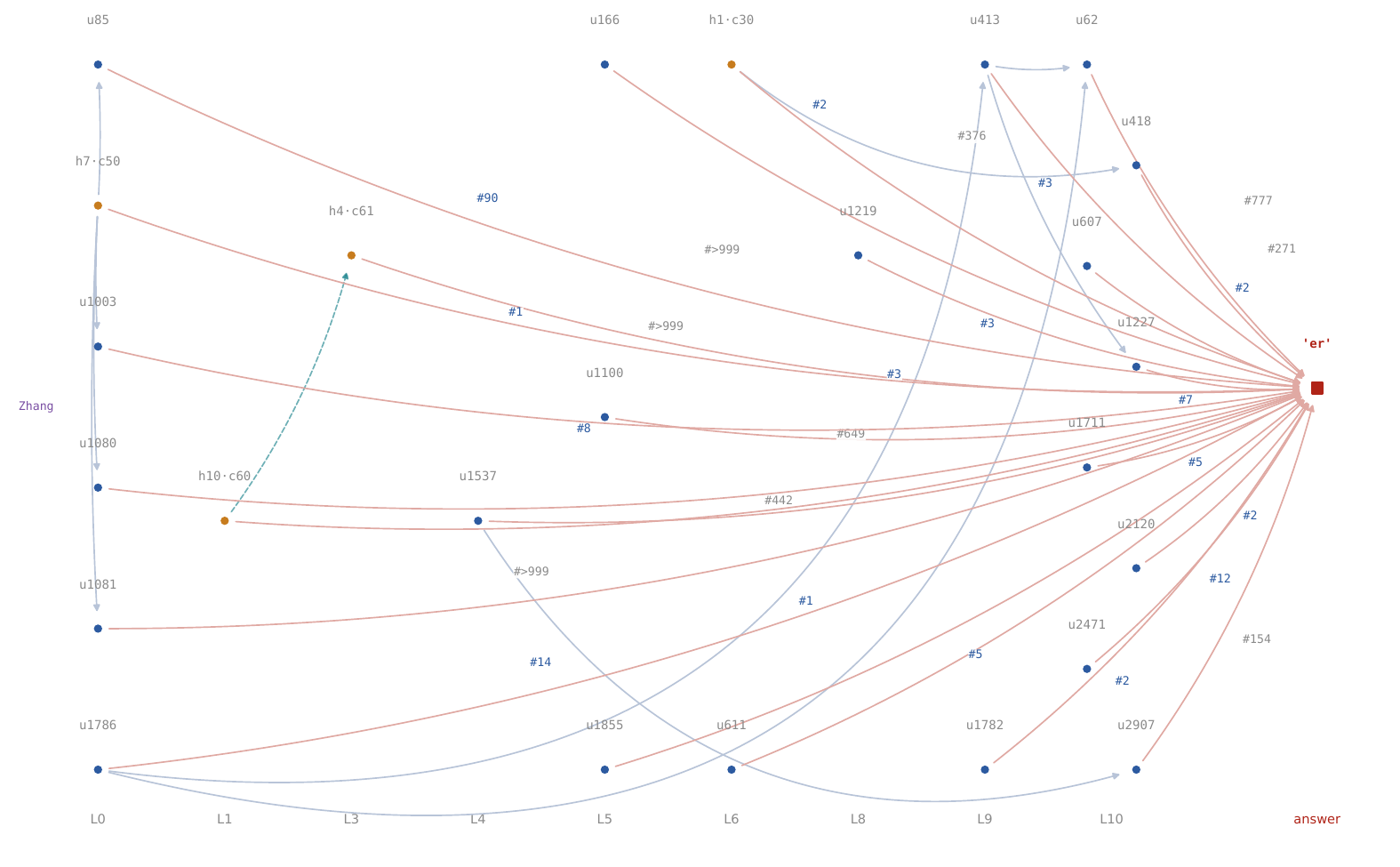}
\caption{baseline (GELU), prediction 3: the graph, 25 members and 11 edges among them.}
\end{figure}
\end{landscape}
\clearpage
\begin{figure}[H]
\centering
\includegraphics[width=\textwidth,height=0.92\textheight,keepaspectratio]{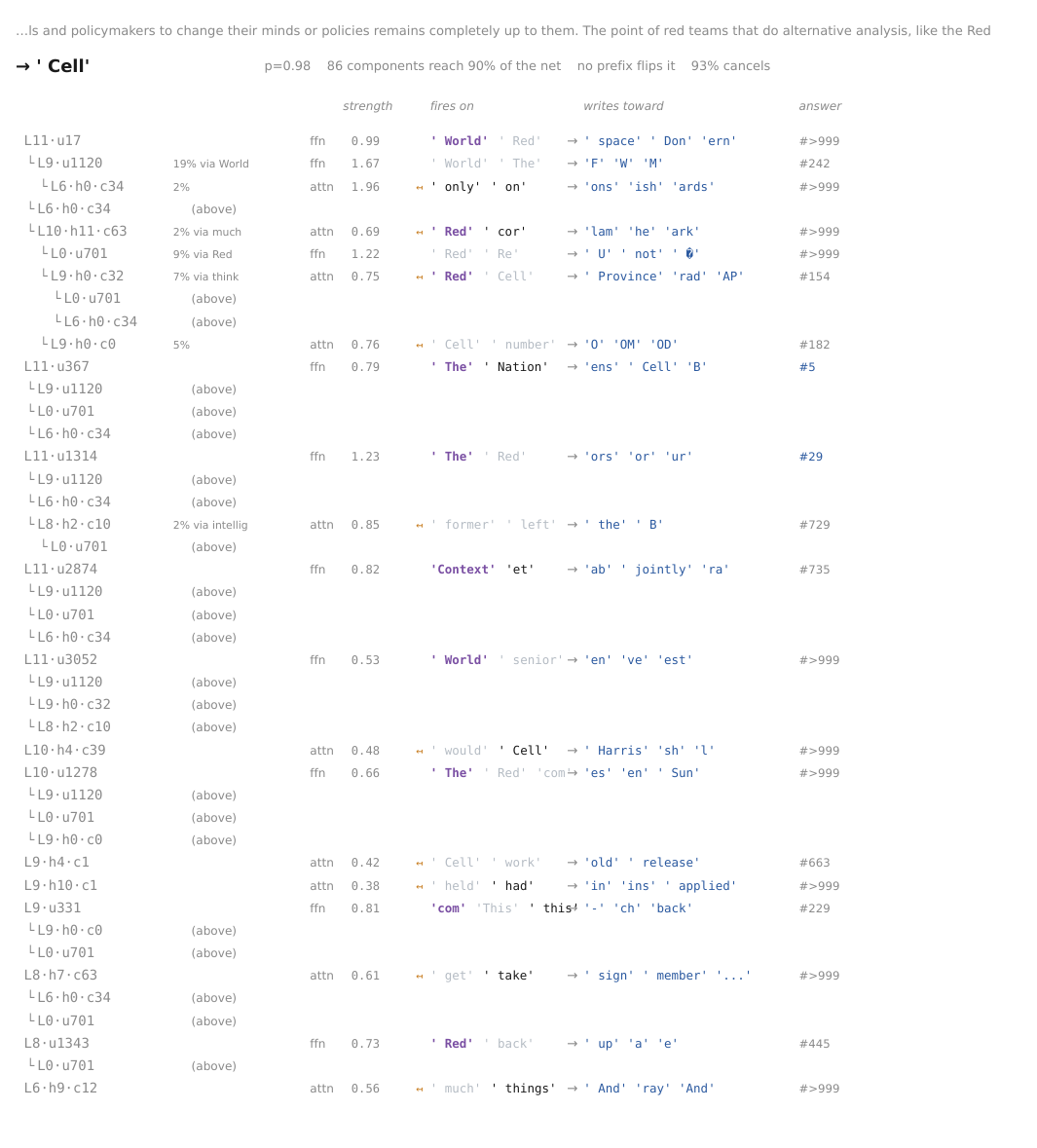}
\caption{baseline (GELU), prediction 4: the set, 20 members. Selected for containing edges no token describes. Of 11 predictions drawn at random, this is one of the four whose sufficient set is nearest the median, 24.}
\end{figure}
\clearpage
\begin{landscape}
\begin{figure}[H]
\centering
\includegraphics[width=\linewidth,height=0.88\textheight,keepaspectratio]{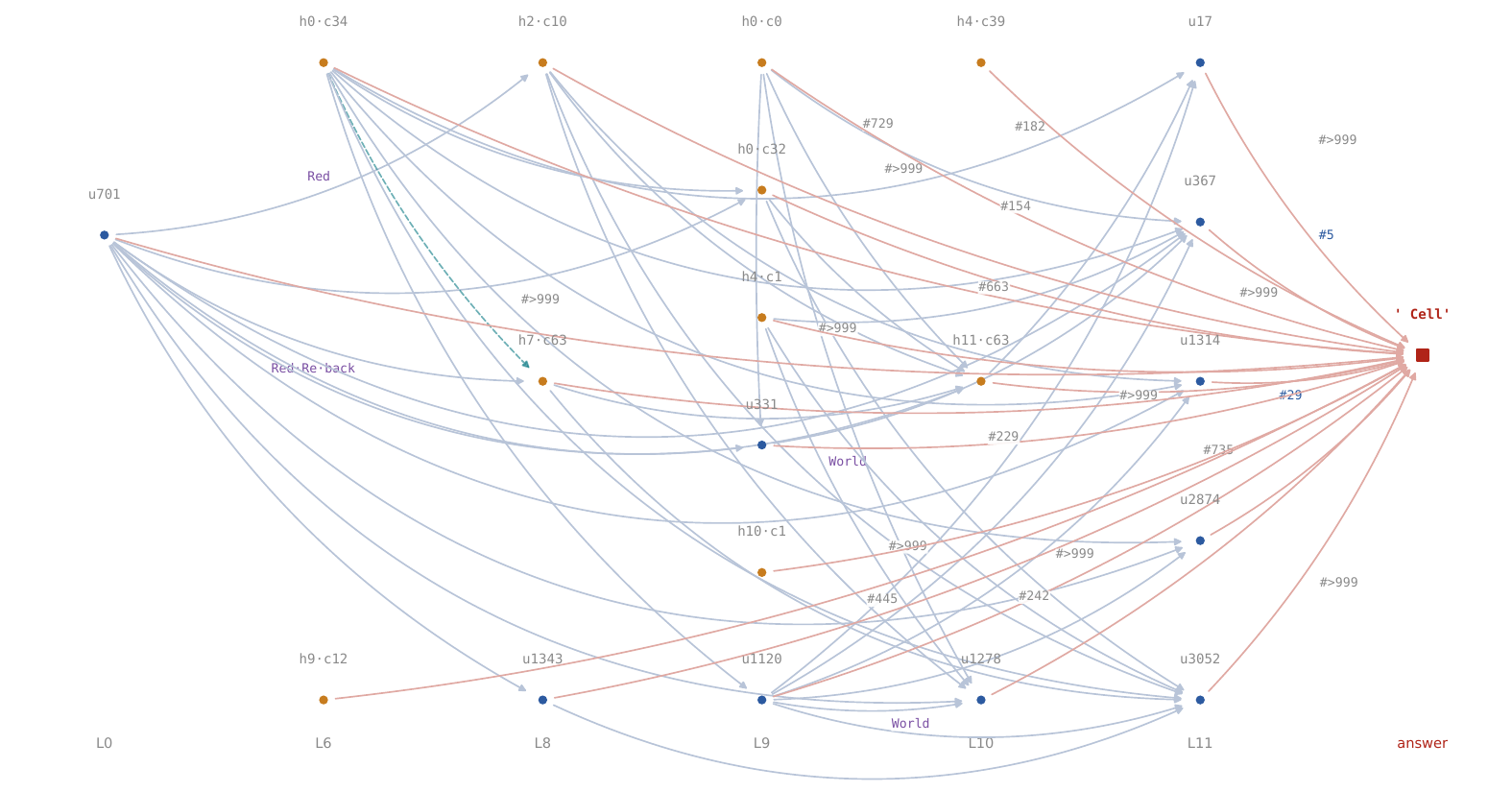}
\caption{baseline (GELU), prediction 4: the graph, 20 members and 42 edges among them.}
\end{figure}
\end{landscape}
\clearpage
\begin{figure}[H]
\centering
\includegraphics[width=\textwidth,height=0.92\textheight,keepaspectratio]{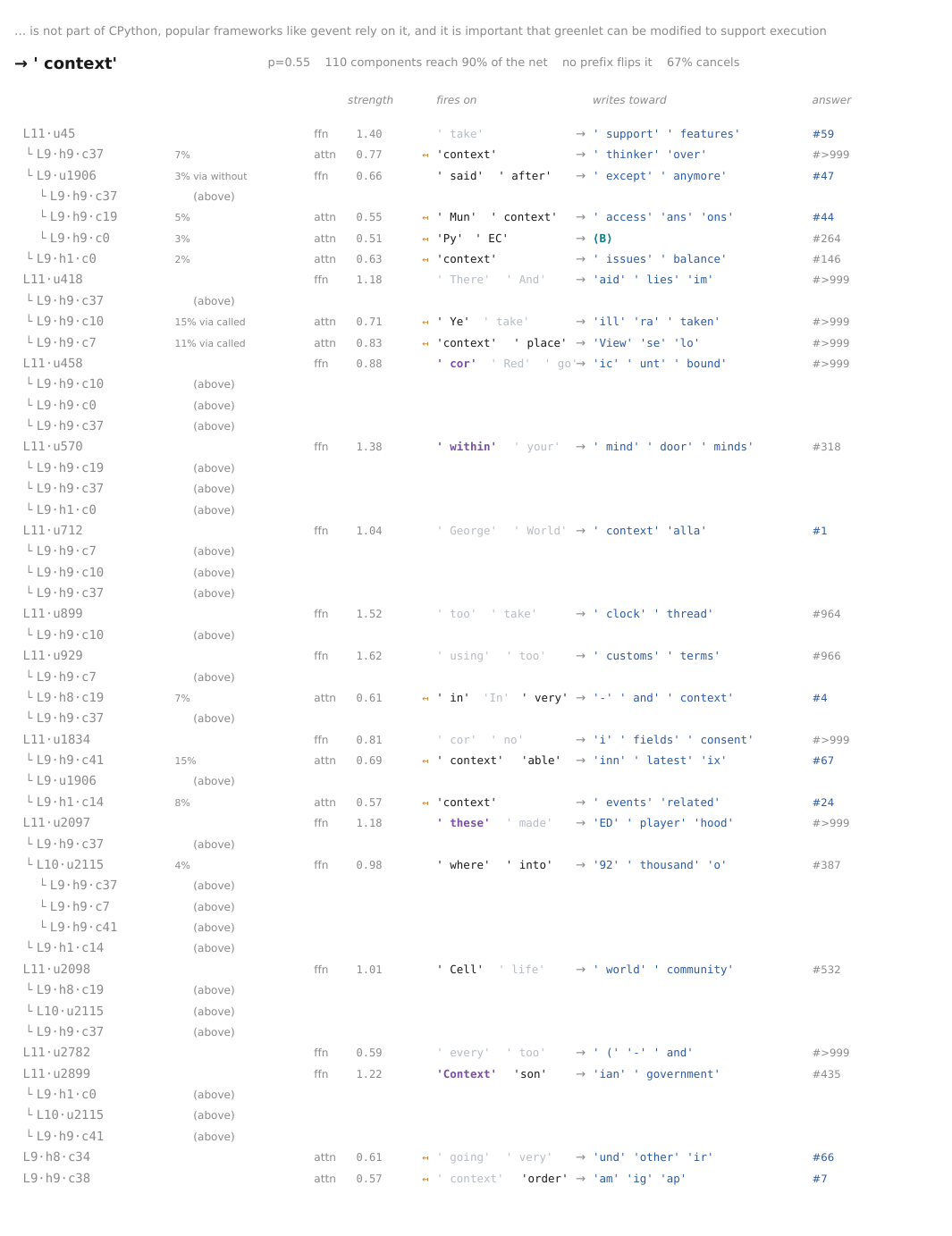}
\caption{ReLU, prediction 4: the set, 25 members. Selected for containing edges no token describes. Of 14 predictions drawn at random, this is one of the four whose sufficient set is nearest the median, 22.}
\end{figure}
\clearpage
\begin{landscape}
\begin{figure}[H]
\centering
\includegraphics[width=\linewidth,height=0.88\textheight,keepaspectratio]{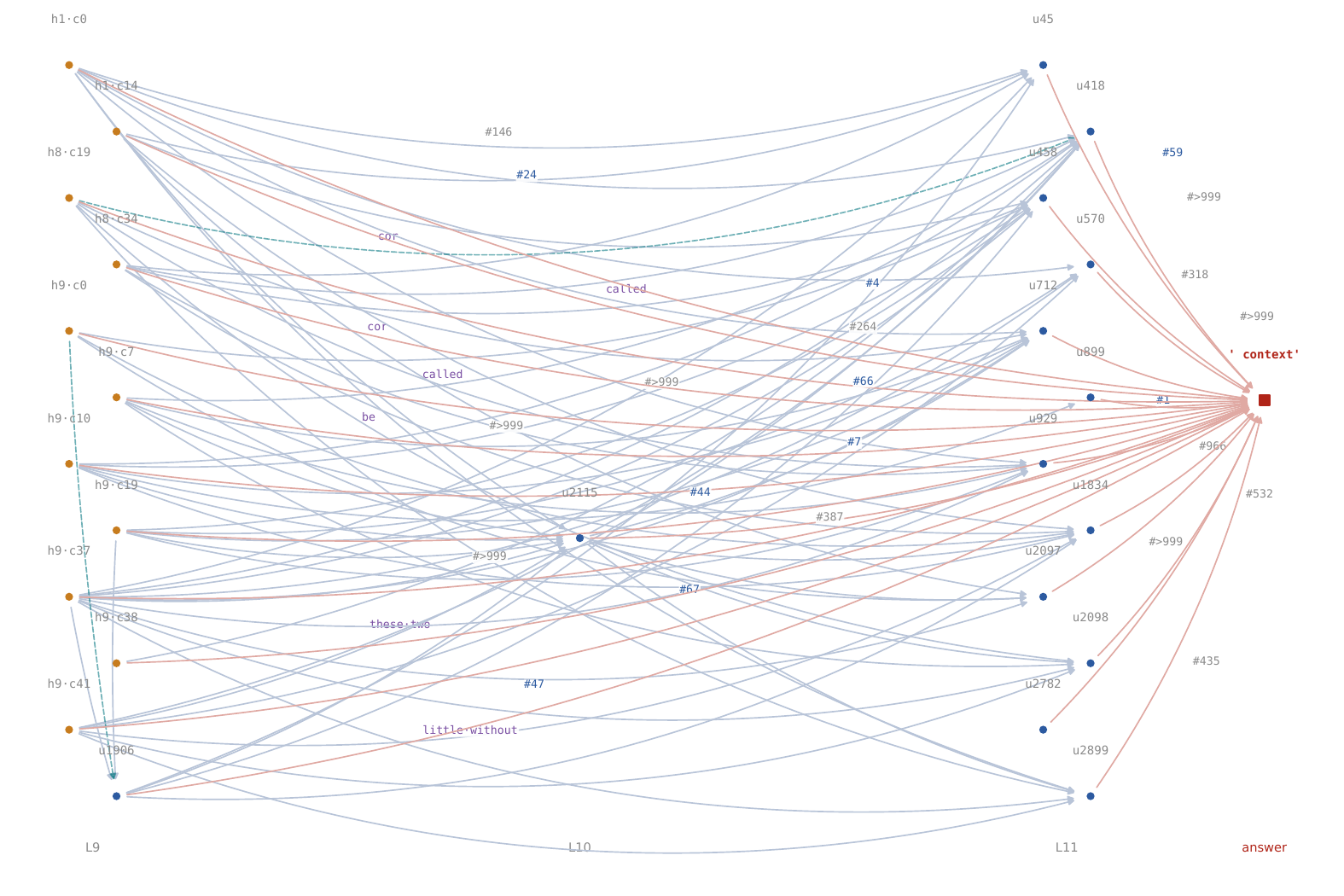}
\caption{ReLU, prediction 4: the graph, 25 members and 70 edges among them.}
\end{figure}
\end{landscape}
\clearpage
\begin{figure}[H]
\centering
\includegraphics[width=\textwidth,height=0.92\textheight,keepaspectratio]{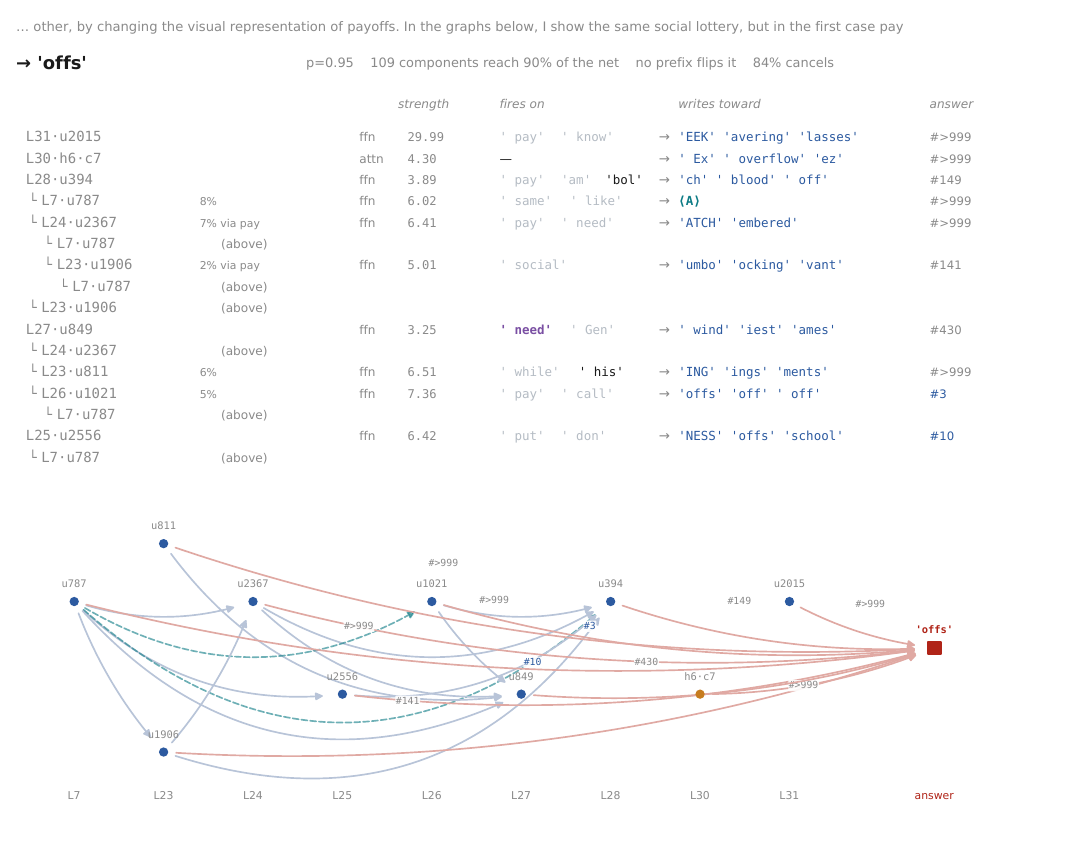}
\caption{SmolLM2 360M, prediction 4. Selected for containing edges no token describes. Of 20 predictions drawn at random, this is one of the four whose sufficient set is nearest the median, 10.}
\end{figure}
\clearpage
\begin{figure}[H]
\centering
\includegraphics[width=\textwidth,height=0.92\textheight,keepaspectratio]{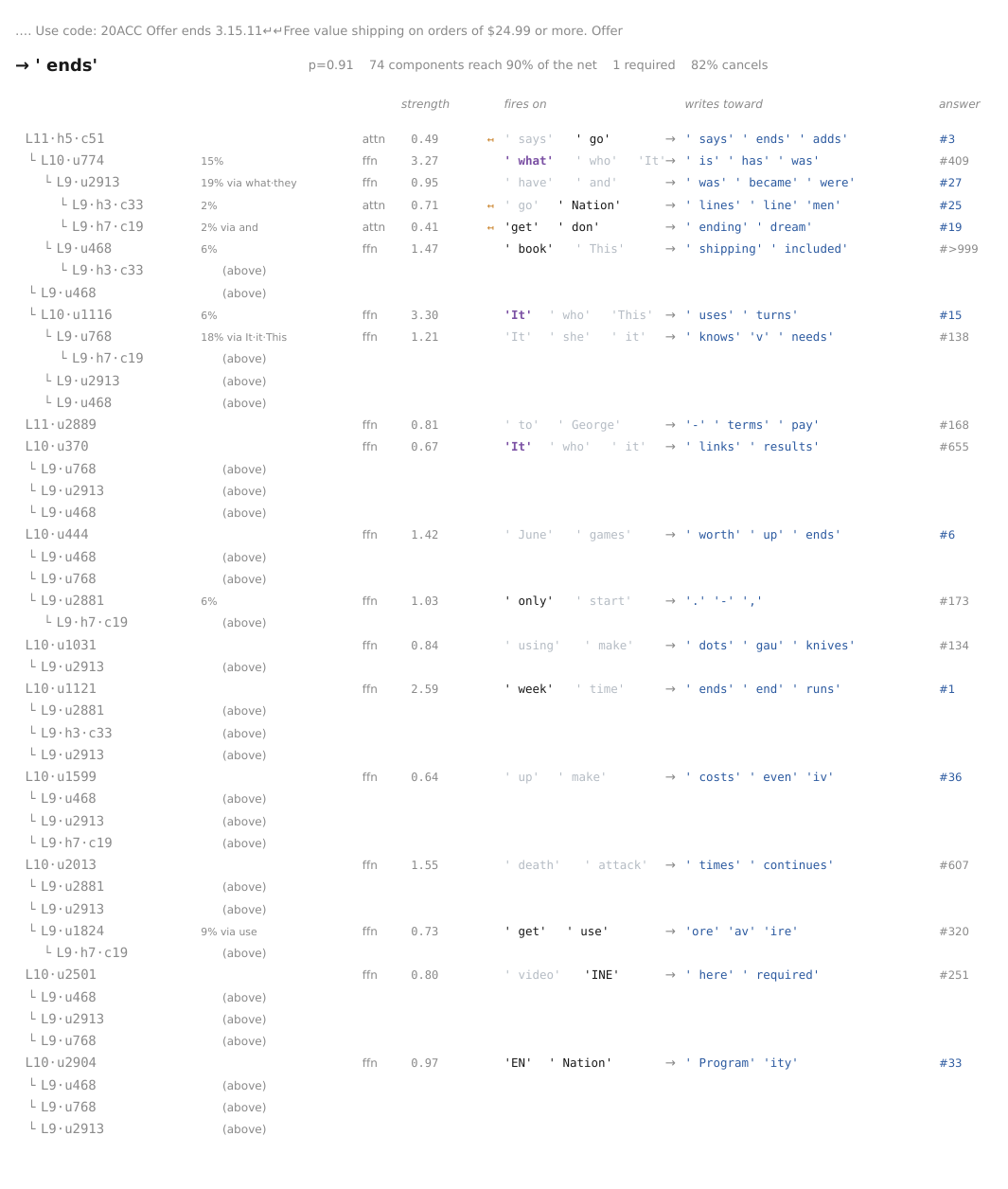}
\caption{sigmoid, prediction 4: the set, 19 members. Selected for containing edges no token describes. Of 18 predictions drawn at random, this is one of the four whose sufficient set is nearest the median, 19.}
\end{figure}
\clearpage
\begin{landscape}
\begin{figure}[H]
\centering
\includegraphics[width=\linewidth,height=0.88\textheight,keepaspectratio]{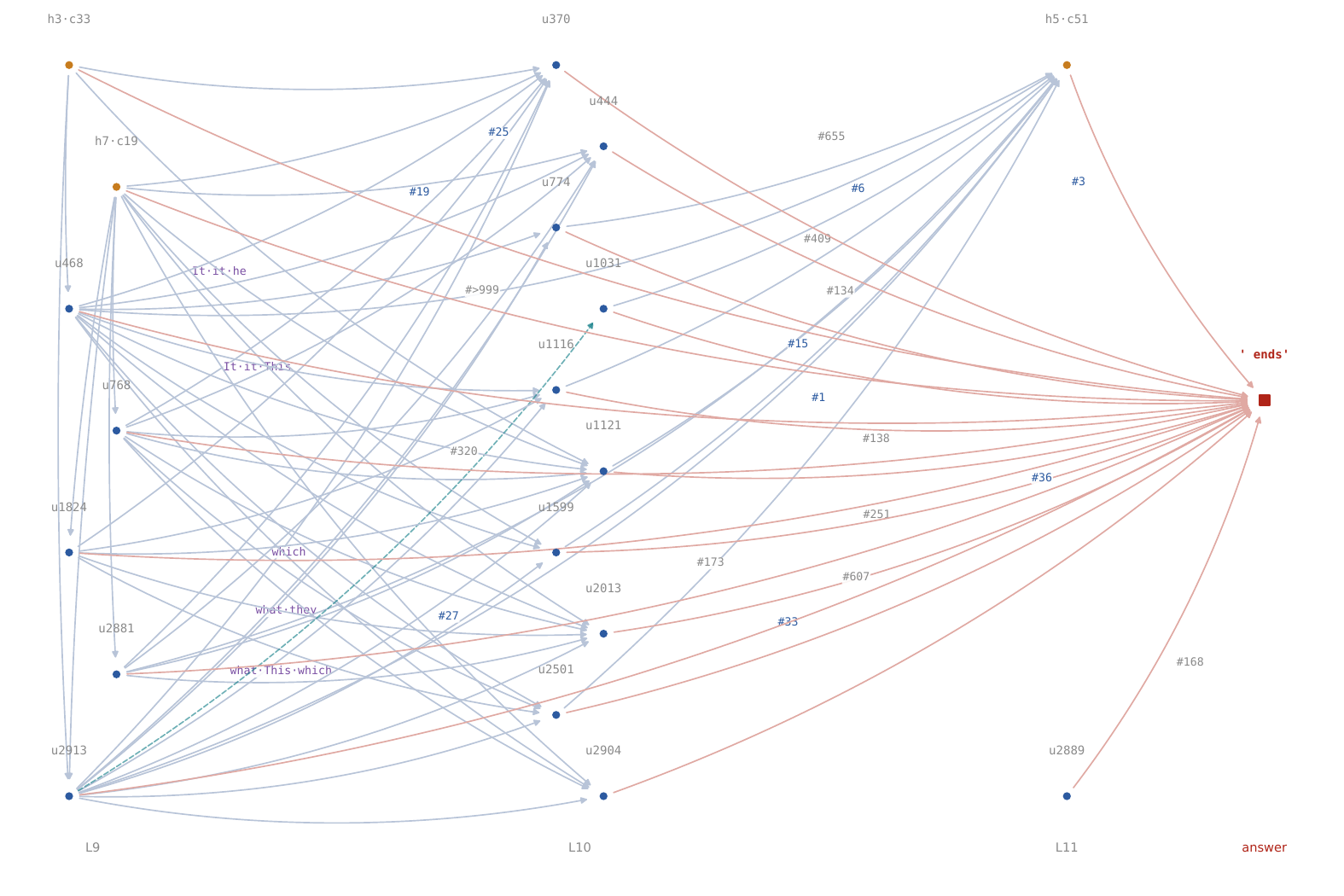}
\caption{sigmoid, prediction 4: the graph, 19 members and 58 edges among them.}
\end{figure}
\end{landscape}
\clearpage
\begin{figure}[H]
\centering
\includegraphics[width=\textwidth,height=0.92\textheight,keepaspectratio]{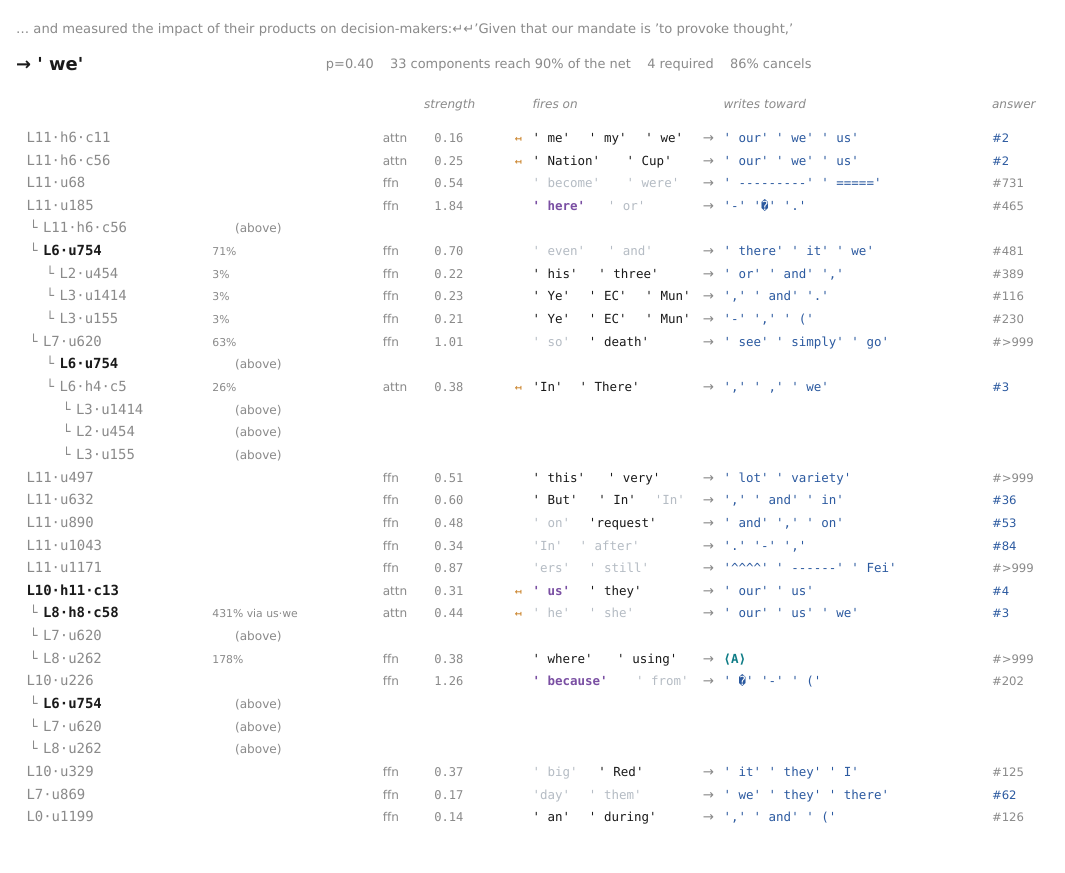}
\caption{set operators, prediction 3: the set, 22 members. Selected for containing edges no token describes. Of 13 predictions drawn at random, this is one of the four whose sufficient set is nearest the median, 22.}
\end{figure}
\clearpage
\begin{landscape}
\begin{figure}[H]
\centering
\includegraphics[width=\linewidth,height=0.88\textheight,keepaspectratio]{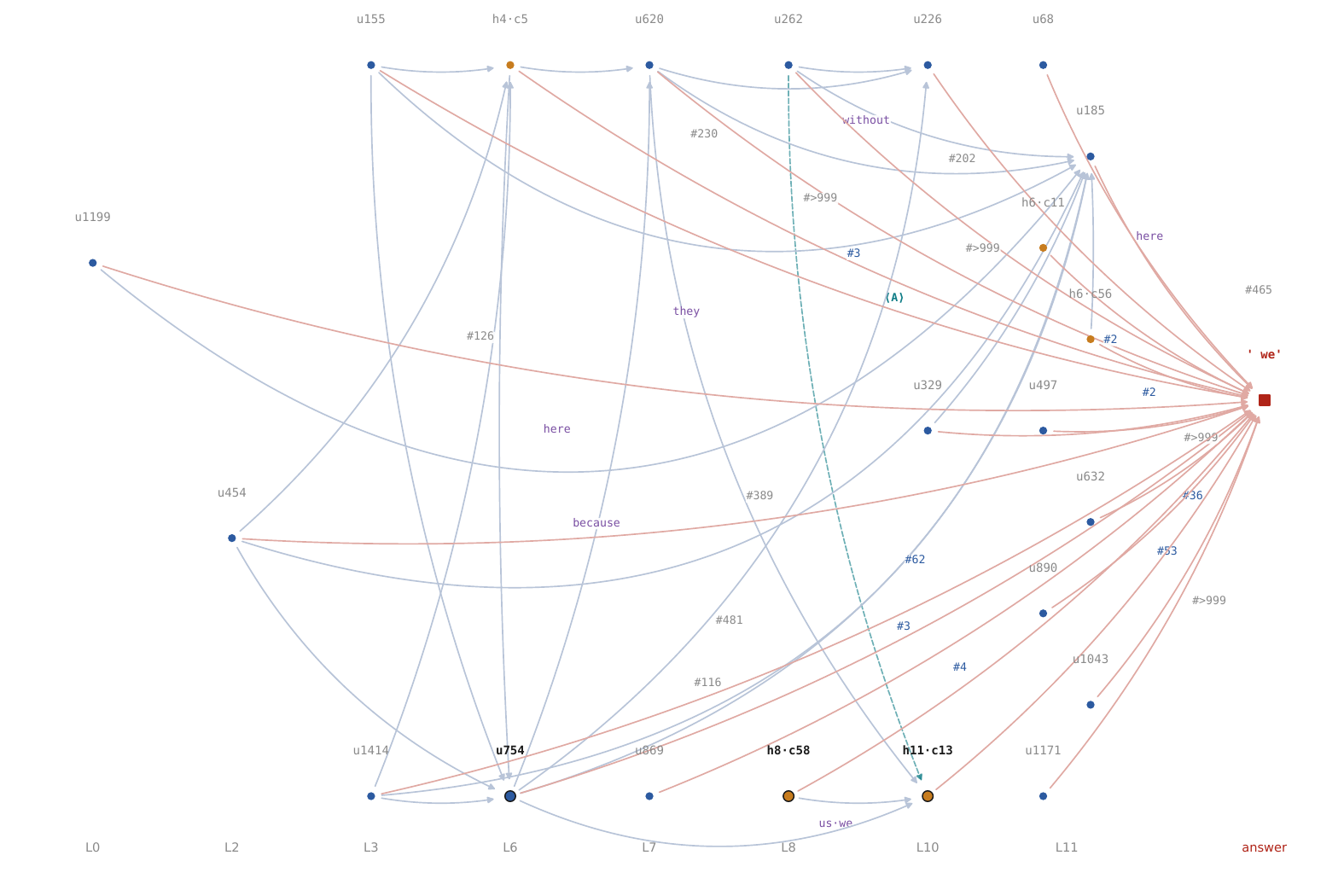}
\caption{set operators, prediction 3: the graph, 22 members and 25 edges among them.}
\end{figure}
\end{landscape}
\clearpage
\begin{figure}[H]
\centering
\includegraphics[width=\textwidth,height=0.92\textheight,keepaspectratio]{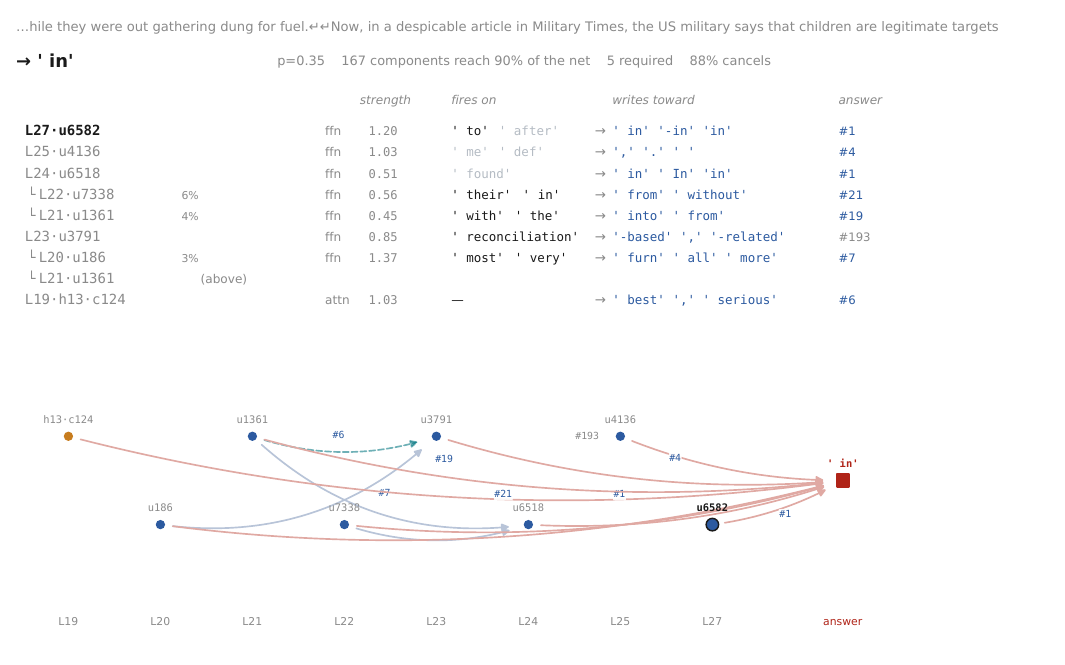}
\caption{Llama-3.2 3B, prediction 4. Selected for containing edges no token describes. Of 20 predictions drawn at random, this is one of the four whose sufficient set is nearest the median, 8.}
\end{figure}
\clearpage
\begin{figure}[H]
\centering
\includegraphics[width=\textwidth,height=0.92\textheight,keepaspectratio]{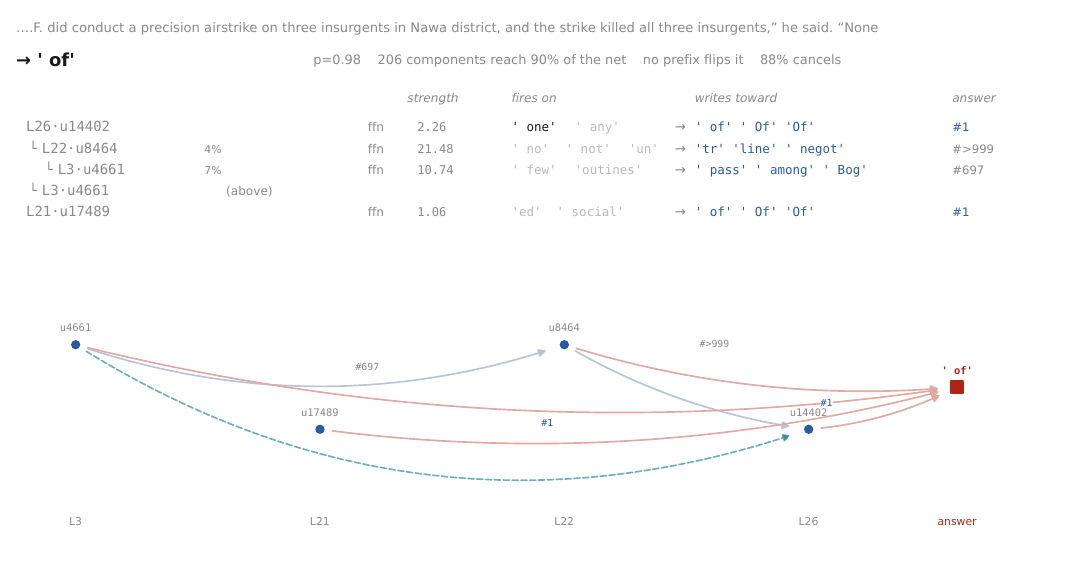}
\caption{Qwen2.5 7B, prediction 4. Selected for containing edges no token describes. Of 17 predictions drawn at random, this is one of the four whose sufficient set is nearest the median, 4.}
\end{figure}

%% file: setmodel.tex
\section{The set-operator model and how it is trained}
\label{sec:setmodel}

Section~\ref{sec:mono} treats the set-operator construction as one arm among four, and says only as
much about it as the comparison needs. It was built across three earlier papers
\citep{oskin2026legible,oskin2026selectivity,oskin2026ncffn}, and the ingredient that lets it run at
every unit of every layer comes from a fourth \citep{oskin2026offaxis}. This appendix collects the
construction and its training in one place: the unit, the attention head, the pressure applied
during training, the two shaping terms, and the recipe. Nothing here is new to those papers except
the assembly.

\subsection{The feed-forward unit}

A conventional unit reads one row and writes one column through a scalar nonlinearity. A
set-operator unit reads two rows and writes two columns, and what it computes between them is a pair
of set operations on two fuzzy memberships. With $z$ the normalized residual at a position, unit $u$
forms its operands as in Equation~\ref{eq:ab},
\begin{equation}
  A_u = \sigma\big(w^{a}_u \cdot z\big), \qquad B_u = \sigma\big(w^{b}_u \cdot z\big),
  \qquad A_u, B_u \in (0,1),
\end{equation}
and writes the intersection and the difference of the two, each through its own column:
\begin{equation}
  \mathrm{FFN}(z) \;=\; \frac{g_\ell}{r(z)}\sum_{u}\Big[\underbrace{A_u B_u}_{A \cap B}\; c^{\cap}_u
  \;+\; \underbrace{A_u\,(1 - B_u)}_{A \setminus B}\; c^{\setminus}_u\Big],
  \qquad r(z) = \operatorname{rms}\big(A \cap B \,\|\, A \setminus B\big),
  \label{eq:set-ffn}
\end{equation}
where $r(z)$ is the root-mean-square of the $3072$ operation values at that position and $g_\ell$ is
one learned scalar per layer. That normalization is the block's only departure from a plain sum: the
operation values are bounded in $[0,1]$ where a GELU activation is not, and the normalization lets
the block set its own scale rather than inherit the interval's. It is a scalar the model computes,
which is how Table~\ref{tab:write} treats it.

The choices in that line each do a job. The memberships are squashed into $[0,1]$ because that
interval is closed under complement, which is what makes negation a primitive: $1 - B$ is ``not
$B$'' in the same units as $B$, where a raw direction has no absence. The operations are the product
t-norm rather than $\min$ and $\max$, because the product trains and the lattice operations saturate.
And the two operations chosen are intersection and difference, which between them span conjunction
and negation-under-a-condition, the two things a single-row unit cannot express at all.

Rearranged, and setting the scale aside, the sum in Equation~\ref{eq:set-ffn} reads
\begin{equation}
  \sum_{u} A_u \Big[\, c^{\setminus}_u + B_u\,\big(c^{\cap}_u - c^{\setminus}_u\big) \Big],
  \label{eq:set-gate}
\end{equation}
which is how the two operands divide the work. $A_u$ is a gate: it scales the unit's whole
contribution and is zero when the unit is off. $B_u$ is a selector: it moves the write continuously
between the two fixed columns, delivering exactly $c^{\setminus}_u$ at $B_u = 0$ and exactly
$c^{\cap}_u$ at $B_u = 1$. Both endpoints are exact rather than approximate, and
Section~\ref{sec:mono-gate} is built on that: keying $B_u$ on a token makes the difference branch
vanish identically where the token is present.

The layer is held to the baseline's parameter count by narrowing it. A conventional layer of width
$d_{\mathrm{ff}} = 3072$ has $2\,d\,d_{\mathrm{ff}}$ parameters in its two projections. A set-operator
layer with $p$ operand pairs has $2\,d\,p$ in the read rows and $2\,p\,d$ in the write columns, so
$p = d_{\mathrm{ff}}/2 = 1536$ pairs make the two equal. The write projection is therefore the same
$3072 \times d$ matrix as the baseline's, with its first $1536$ columns carrying $c^{\cap}$ and its
last $1536$ carrying $c^{\setminus}$; only the activation feeding it has changed. Every unit in every
layer is of this kind. The earlier papers ran the construction on a fraction of each layer with
conventional units alongside as a gradient path, because the pure form did not train; the arm here
has none.

\subsection{The attention head}

The same construction is applied on the value side of every attention head, so that what a head
gathers from a position is bounded in the same way a unit's operands are. A head's value vector
$v \in \mathbb{R}^{64}$ is split in half and each half becomes a membership,
\begin{equation}
  A = \sigma\big(v_{1:32}\big), \qquad B = \sigma\big(v_{33:64}\big), \qquad
  \tilde{v} = \big[\, A \odot B \;\|\; A \odot (1 - B)\,\big] \in [0,1]^{64},
  \label{eq:set-attn}
\end{equation}
and $\tilde{v}$ takes the place of $v$ in the attention average. The query, the key, the softmax over
positions and the output projection are unchanged, and no parameter is added: the head's
$64$ value channels are reinterpreted as $32$ intersections and $32$ differences. What attention
delivers at a position is then an average of bounded quantities, which is the property
Section~\ref{sec:naming-attn} reads and Section~\ref{sec:mono-comp} pays for --- a channel gathers
nothing far from its resting level unless a key row is scaled for it.

Everything else is the baseline of Section~\ref{sec:circuit-model}: twelve layers of width
$d = 768$, twelve heads, learned positional embeddings, pre-norm blocks, no bias anywhere, and tied
input and output embeddings. The parameter count is the baseline's but for the one scalar $g_\ell$ per layer.

\subsection{The pressure}
\label{sec:setmodel-pressure}

Left to the language-modeling loss alone, the pure construction did not train: above roughly half
conversion the earlier papers found training diverging late, sooner the larger the converted
fraction. What lets it run everywhere is an auxiliary objective from \citet{oskin2026offaxis}, where
it was introduced for a different purpose --- to prescribe where in the stack a conventional
transformer holds its intermediate content relative to the readout.

Write $h^{(\ell)}_t$ for the residual after layer $\ell$ at position $t$, and decode it through the
tied readout as the logit lens does,
\begin{equation}
  z^{(\ell)}_t = \mathrm{LN}\big(h^{(\ell)}_t\big)\,E^{\top}, \qquad
  \tilde{z}^{(\ell)}_t = z^{(\ell)}_t - \operatorname{mean}_{v}\, z^{(\ell)}_{t,v},
  \label{eq:press-decode}
\end{equation}
centering over the vocabulary. Let $\tilde{z}_t$ be the same quantity for the model's final logits.
Nothing is held fixed: the term's gradient reaches both the layer's decode and the final prediction,
so the pressure can meet its target by moving either. The objective asks each layer's decode to sit
at a prescribed angle to the final prediction,
\begin{equation}
  \mathcal{L}_{\angle} \;=\; \frac{\alpha}{L}\sum_{\ell=0}^{L-1}\;\mathbb{E}_{t}\Big[
  \big(\cos\angle(\tilde{z}^{(\ell)}_t,\,\tilde{z}_t) - \cos\theta_\ell\big)^{2}\Big],
  \qquad \cos\angle(a,b) = \frac{\langle a,b\rangle}{\|a\|\,\|b\|},
  \label{eq:press-cos}
\end{equation}
with the target angle a step over depth,
\begin{equation}
  \theta_\ell \;=\; \begin{cases} 90^{\circ}, & \ell = 0,\dots,5,\\[2pt] 0^{\circ}, & \ell = 6,\dots,11. \end{cases}
  \label{eq:press-sched}
\end{equation}
The first half of the stack is asked to keep everything it computes orthogonal to the answer, and
the second half to point at it. That is the two-phase structure \citet{oskin2026offaxis} measured
in the conventional baseline --- a concept phase held off the readout axis, a token phase committed
to it --- written down as a target rather than left to emerge. The expectation runs over $128$ positions drawn at random from each batch, the same positions for
every layer; the coefficient is $\alpha = 3$ and is held constant
through training, the term is added to the cross-entropy, and it is switched off at evaluation. It
touches no parameter and inserts nothing into the forward pass.

Two facts about this pressure are established elsewhere and matter here. On the conventional model it
is a lottery: prescribing the step through the loss alone, \citet{oskin2026offaxis} report six of
eight seeds diverging outright, and the survivors reaching the geometry under strain. That paper's
remedy is a fixed orthogonal rotation inserted at the boundary layer, which supplies the frame the
loss asks for instead of making the model grow one. The set-operator arm uses no such device. Under
the loss alone, all eight seeds converge, inside $0.2$ nats of one another, at a quality inside the
baseline's converged range (Table~\ref{tab:mono-quality}). The constraint that a conventional
transformer meets only by luck, and pays for in quality when it does, this construction meets every
time for nothing. No run of this arm without the schedule is part of this paper, so the comparison is with the earlier
papers' runs rather than a paired ablation. What those runs suggest is that a bounded stack left to
itself has to find the concept--token boundary on its own and to commit on the readout axis at every
depth at once, and saturates trying; the schedule tells it where the boundary is. Why a bounded
two-operand unit is at ease with a geometry a GELU unit is not, and whether that reading is the whole
of it, are not settled in this paper;
Section~\ref{sec:mono-what} finds the construction's legibility comes from its order-preserving
operands rather than from the set operations, and the trainability may have the same source.

\subsection{Two shaping terms}

Two small auxiliary terms act on the operands during training, both at coefficient $0.003$. Writing $v$ for an operand's value and the means over
the operands of a layer and the tokens of a batch,
\begin{equation}
  \mathcal{L}_{\mathrm{crisp}} = 0.003\;\operatorname{mean}\big[\,v\,(1 - v)\,\big], \qquad
  \mathcal{L}_{\mathrm{var}} = 0.003\;\operatorname{mean}_{u}\Big[\max\Big(0,\; 1 - \frac{\operatorname{Var}_t\, v_u}{\tau}\Big)\Big],
  \quad \tau = 0.003.
  \label{eq:set-aux}
\end{equation}
The first rewards an operand for sitting near $0$ or $1$ rather than in the middle of its range. The
second is a floor on how much an operand varies across the tokens of a batch: an operand that has
settled at a constant, and so responds to nothing, pays until its variance clears $\tau$. Both are
applied to the feed-forward operands and to the attention memberships of
Equation~\ref{eq:set-attn}. The crispness term is also the ``shaping'' of Table~\ref{tab:models}, and it is the
one of the two that is defined only for a bounded activation, since $v(1-v)$ needs $v \in [0,1]$ to
be a penalty at all; that is why the softplus and ReLU arms carry it at zero. The sigmoid arm of
Section~\ref{sec:mono} was trained with and without it, and the two agree on the measurements that
comparison rests on, so the term is not what produces those results. They
do not agree everywhere: the two part company on where a model delivers context over depth, which
Section~\ref{sec:naming-attn} reports and leaves open.

\subsection{The recipe}

Training is the baseline's (Section~\ref{sec:circuit-model}): one epoch of OpenWebText at batch
size $16$ and context length $2048$, AdamW with $\beta = (0.9, 0.95)$ and weight decay $0.1$, peak
learning rate $3 \times 10^{-4}$ with $2000$ warmup steps and cosine decay to a tenth of peak,
$272{,}687$ steps, gradient clipping, and \textsc{bf16} on a single GPU. The auxiliary terms are
added to the cross-entropy from the first step with their coefficients fixed. Nothing was tuned for
this arm: it inherits the baseline's learning rate and warmup unchanged. Eight seeds were trained
under the screen of Section~\ref{sec:circuit-seeds}, and all eight pass it.

What the rest of the paper reads from the result is the same as for any other arm, with one
substitution. A unit's read side is operand $A$'s row $w^{a}_u$, and the agreement figures of
Section~\ref{sec:mono-legibility} are taken on it; Section~\ref{sec:mono-legibility} also reads
$w^{b}_u$ and finds both operands nameable at roughly twice the baseline's single row. A unit's write
side is two columns rather than one, and an edit to one of them lands on one branch of the unit
(Section~\ref{sec:mono-edit}). The second operand is the one thing this construction has that the
others do not, and Section~\ref{sec:mono-gate} is what it turns out to be worth.

\secbarrier

%% file: pitfalls.tex
\section{Ash heap}
\label{sec:baseline-pitfalls}

Every method below is the sensible thing to do --- the obvious statistic, the standard method, the natural correction --- and each either does
not work or, worse, returns a confident number pointing the wrong way. The failure that matters is the one that returns 16.9 times the null on a model that is measurably rank-one, or that ranks a diverged checkpoint as the most legible thing in the study.
They are set out at length because anyone repeating this measurement will reach for the same ones. Maybe they
can find a way to make them work.  Maybe the text here will spare them the trouble.  They fall into five families: the
models can rig the comparison, the estimators are biased, the cuts and summaries are arbitrary, the
ablation searches fail their own checks, and the standard reading methods do not survive contact
with a conventional unit. After those come the failures specific to naming a component --- which of
Section~\ref{sec:naming}'s negative results are the instrument's fault, the conditions under which
its numbers hold, fitting a unit's drivers from upstream, reading a write column against the
vocabulary, and naming a unit by where it fires --- and, last, what the failures leave: the tests
that say what a unit's unnameable remainder is. The dependency graph has an appendix of its own
(Appendix~\ref{sec:appgraph}).

\subsection{The models can rig the comparison}

\paragraph{A worse model reads as more legible.} The largest failure is the one Table~\ref{tab:baseline-seeds} records. Define the effective dimensionality of a layer's
token table as the participation ratio of its eigenvalue spectrum,
\begin{equation}
  \mathrm{PR}\big(\tilde{E}[l]\big) = \frac{\big(\sum_i \lambda_i\big)^2}{\sum_i \lambda_i^2},
  \qquad \lambda_i \text{ the eigenvalues of } \mathrm{Cov}_t\big(\tilde{E}[l,t]\big),
  \label{eq:pr}
\end{equation}
which reads as a count of independent directions. Averaged over the first five layers a converged
seed carries 98.4 to 106.7 of them, and never falls below 75.8 at any depth. The excluded seeds
carry 30.3 to 45.7, and two of them collapse almost completely at a single layer: 2.2 and 1.1
independent directions at layer 1, out of 768. A collapsed representation is easier to predict
because there is less variety left to predict, and the excluded seeds accordingly score 40.7 to 49.2
percent on $\mathrm{ov}@8$ against 33.6 to 38.4 for the five that converged. \emph{Every excluded
seed reads as more legible than every converged one.} Including a single one would shrink the gap
this paper later reports by about a third; the effect is large enough, and in the wrong direction
often enough, that quality is reported beside every legibility figure in this paper.

\paragraph{A fairness correction that rewards silence.} A unit that no
token clearly drives has a noisy causal ranking and therefore scores low, which suggests dropping
such units before comparing models. The correction backfires, because the model with more of them
gains more from their removal: dropping units with no clear trigger lifts the baseline by about six
points. The correction rewards whichever model has more such units, so applying it in the interest of fairness silently favors one side of any comparison it is used in. The uncorrected
number is reported.

\subsection{The estimators are biased}

\paragraph{The dimensionality metric is biased both ways.}
Equation~\ref{eq:pr} looks safer than it is for being scale-free and parameter-free. Each row
of $\tilde{E}$ is a mean over $|\mathcal{C}|$ contexts and carries estimation noise, and that noise is
approximately isotropic in $d$ dimensions, so the measured covariance is
\begin{equation}
  \mathrm{Cov}_{\text{meas}} \;\approx\; \mathrm{Cov}_{\text{true}}
  \;+\; \frac{\sigma^2_{\text{within}}}{|\mathcal{C}|}\, I ,
  \label{eq:ridge}
\end{equation}
with $\sigma^2_{\text{within}}$ the variance of a token's state across contexts. An isotropic ridge
pulls the participation ratio toward the ambient dimension, so Equation~\ref{eq:pr}
\emph{over}states dimensionality. Worse, the ridge is not constant across the models being compared:
$\sigma^2_{\text{within}}$ measures how much a token's representation moves with context, and a
better model has more of it, so the correction needed is largest exactly where the measurement is
being used to argue. Across the seeds here $\sigma^2_{\text{within}}$ varies by a factor of seven hundred.
A second bias runs the other way, since sample eigenspectra are over-dispersed relative to the
population, which makes Equation~\ref{eq:pr} \emph{under}state dimensionality, most severely for the
seeds with the highest true value.

Neither bias can be argued away, so both are removed by measurement. Splitting the contexts into two
disjoint halves and taking the participation ratio of the symmetrized cross-covariance
\begin{equation}
  C = \tfrac{1}{2}\big(M + M^{\top}\big), \qquad
  M = \frac{1}{|V|}\sum_{t \in V}
      \big(\tilde{E}_1[l,t]-\bar{\tilde{E}}_1\big)\big(\tilde{E}_2[l,t]-\bar{\tilde{E}}_2\big)^{\!\top}
  \label{eq:prsplit}
\end{equation}
removes the ridge, because noise independent between the halves contributes nothing in expectation.
Doing so moves every seed by less than one unit of $\mathrm{PR}$. In this case the objection does not bite.

\paragraph{A small table understates.} Agreement depends on
$|\mathcal{C}|$ in Equation~\ref{eq:etil} and is still rising at 32 contexts; going from 4 to 32
raises baseline $\mathrm{ov}@8$ from 34.6 to 40.2 percent. The estimand is a property of the model
rather than of the sample, so a larger table is a better estimate rather than a thumb on the scale.
Two consequences follow. Absolute figures must be quoted with the table size that produced them. And
a paper that splits its contexts evenly between building the table and testing against it, as the
obvious protocol does, is deliberately halving the quality of its own instrument for a hygiene
requirement that costs nothing to satisfy another way.

\paragraph{Scoring on the table's own contexts.} If
$\mathcal{C}$ and $\mathcal{C}'$ overlap, then $s_u(t)$ in Equation~\ref{eq:decode} is, up to
centering, the mean pre-activation over the same contexts whose mean activation defines
$\bar{a}_u(t)$. The two sides then share their sampling noise, and for a monotone $\phi$ the
comparison reduces to a question about the curvature of $\phi$ rather than about the parameters.
Overlapping the two sets is cheap. The flag
that shares the contexts also doubles the table the naming side is built from, and it is the table
that moves the number: holding the table at sixteen contexts, overlapping against disjoint is $39.2$
against $38.9$ under the ranked summary and $59.7$ against $59.5$ under the mean. The leakage is about a third of a point. Every number in this paper uses disjoint
sets regardless, which costs nothing and settles the question rather than arguing it.

\paragraph{The reproducibility floor.} Recomputed from an independent sample of contexts, a
baseline unit's top-$K$ list agrees with itself, at $K=8$, 64 percent of the time at eight contexts
per half and 86 percent at ninety-six. Any statement about how well two \emph{different} procedures agree is
meaningless without that floor, since a procedure that cannot reproduce itself cannot be expected to
match anything else. Measured against it, several apparently large disagreements fall inside the
range the ground truth varies over on its own.

\subsection{The cuts and summaries are arbitrary}

\begin{figure}[htbp]
\centering
\includegraphics[width=0.72\textwidth]{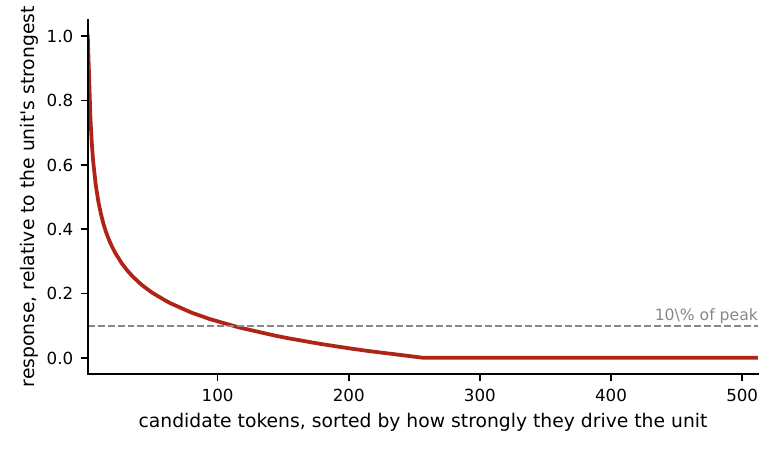}
\caption{Causal response of a baseline unit across candidate tokens, sorted from strongest, median
over units. The response spreads over roughly 115 of 512 candidates and does not fall below a tenth
of its peak until rank 111. A list of any fixed length samples the leftmost sliver: the top eight
capture 18 percent of the total, the top thirty-two 42 percent. Context length 2048, 32 contexts.}
\label{fig:setprofile}
\end{figure}

\paragraph{Any single $K$ is an arbitrary cut.} Define a unit's effective set size from its causal
profile, as the participation ratio of the response in excess of the unit's own median:
\begin{equation}
  e_u(t) = \max\!\big(0,\; \bar{a}_u(t) - \operatorname{med}_t \bar{a}_u(t)\big),
  \qquad
  n_u = \frac{\big(\sum_t e_u(t)\big)^2}{\sum_t e_u(t)^2}.
  \label{eq:neff}
\end{equation}
A baseline unit has $n_u = 115$ at the median, quartiles 82 and 148
(Figure~\ref{fig:setprofile}). Which token lands at rank $K$ rather than $K+1$ is close to a coin
flip, so a measurement read at one $K$ spends most of its resolution on boundary churn. Reporting
$\widehat{\mathrm{ov}}@K$ over a sweep of $K$ rather than at a chosen cut is the only version of this
measurement reported here.

\paragraph{Thresholds chosen in the wrong coordinates.} Almost every quantity here needs a cutoff
somewhere, and a cutoff is a claim about where the interesting part of a distribution lies. Four
separate measurements in this work failed for that reason alone. A threshold placed at the midpoint
of an activation's nominal range measured silence rather than selectivity, because the units in
question operate in the lowest seventh of that range. A range set by the first and ninety-ninth
percentiles discarded the top one percent of a unit's response, which for a selective unit is the
entire signal. Quantile bins put resolution where the probability mass is rather than where the
signal is, and were non-monotone in the number of bins as a result. And a criterion counting tokens
more than three standard deviations \emph{above} a unit's mean is correct for a right-skewed quantity
and silently discards half the evidence for a symmetric one. The general form is the same each time:
a cutoff was chosen in coordinates that suited one of the things being compared. The defense that
works is to plot the distribution before choosing, and to build in a check the measurement must pass,
such as requiring quantization error to fall monotonically as bins are added.

\paragraph{The summary statistic moves the answer.} Equation~\ref{eq:causal} averages activation
across contexts, asking what a unit responds to most \emph{strongly}. Ranking tokens within each
context and averaging those ranks asks what it responds to most \emph{reliably}. Two defensible summaries of the same measurements disagree, moving baseline $\mathrm{ov}@8$ from 34.6 to 52.4 percent and changing the shape of the curve over $K$ as well as its level. No external ground truth settles the choice: any measure of a unit's downstream effect
is itself monotone in $a_u$, which returns the question to how activations should be aggregated
across contexts.

\paragraph{Why two methods give one answer.} In the face of that
disagreement, the tempting inference is that a unit has two token sets and the two statistics find
one each. It does not follow. Two methods producing two sets establishes only that the set is not
unique; there could be three, or many, or none, with each statistic slicing an unstructured surface
differently. Running six summary statistics rather than two shows the actual structure on the
baseline: four magnitude-based measures agree with one another, and the single ordering-based measure
sits outside them. That is a consensus and an outlier, invisible if the two statistics one happens to run are the two most different.

\subsection{Which summary the causal set is taken under}
\label{sec:pit-convention}

The outlier is the one this paper's per-seed tables are computed with, and that has to be said.
Equation~\ref{eq:causal} averages a unit's activation over
contexts and takes the top $K$: the tokens that drive it hardest. The alternative ranks the tokens
within each context and averages those positions: the tokens that drive it most consistently. The
first is what the model integrates, since a unit's write is linear in its activation and a token that
spikes once does contribute a lot once. The second is what a \emph{description} ought to mean, since
a name that holds in one context out of eight is not a name.

The two disagree unequally. Moving from the mean to the ranked summary costs
the baseline $17.8$ points of agreement at $K=8$ and the set-operator model $1.7$. Compared directly,
with no decoding involved --- the overlap between the two conventions' own top-$8$ lists --- they
agree on $63$ percent of the baseline's units and on $75$ to $79$ percent of the bounded models',
against a chance rate of $1.5$.

That is a better result than either number alone. For a bounded unit, what it
responds to most strongly and what it responds to most reliably are close to the same list, so the
question ``what does this unit respond to'' has an answer that does not depend on how it is asked.
For a conventional unit the two lists come apart. The naming question is well posed for one
construction and under-determined for the other, which is a statement about the constructions rather
than about the instrument, and it survives whichever summary a reader prefers.

Two consequences follow. The first is that a legibility \emph{ratio} between two architectures is
not a portable quantity. The ratio here moves from about twice to about a third again depending on a
convention neither architecture chose, and the movement is architecture-specific, so comparing that
ratio against a number computed elsewhere under an unstated convention is not a comparison. That
applies to this paper's numbers as much as to anyone's, which is why the ratio is not the form any
claim here is put in.

The second is narrower and sharper. Ranks are invariant under any strictly increasing transform, so
under the ranked summary a strictly monotone unit's causal ordering \emph{is} its pre-activation
ordering, exactly. An order-preserving arm's agreement with its own ceiling is then $1$ by construction rather than by measurement, and the set-operator and sigmoid models duly return identical ceiling and realized figures. The empirical content of that comparison lives entirely on the
baseline's side, which is why Section~\ref{sec:mono-legibility} reports it under the mean summary,
where both sides can move: there the baseline reaches $0.81$ of its ceiling and the sigmoid model
$0.98$.

\subsection{How the standard reading methods fail on a conventional unit}

\paragraph{Max-activating examples look fine.} The default method in the field is to rank
inputs by how strongly they drive a unit and read the top of that list. Run on the baseline it gives
every one of 36{,}864 units an apparently individual answer: across two converged seeds the strongest
candidate is distributed over all 512 candidates, and the most popular single winner accounts for
only 1.1 and 2.8 percent of units. The problem is that nothing about the output looks degenerate. The top of the list is the part the parameters can name --- it is the same $K=8$ regime
where agreement is 59 percent above chance --- and the list continues into
Table~\ref{tab:gelu-units}. A method that returns a distinct-looking name for every unit while the
set behind that name is a grab-bag will not report its own failure, and reading only the top is how
the failure goes unnoticed.

\begin{table}[htbp]
\centering
\small
\caption{What six baseline units respond to: the top-$K$ tokens of their causal profile at $K=40$,
for units at the median of the effective-size distribution. Unselected --- these are the units the
median picks out, not ones chosen to make a point. Each row is one unit. The definition of
Section~\ref{sec:whatitmeans} asks that a reader be able to say what drives the component; this is the material that reading has to work with, and
Section~\ref{sec:naming-provenance} is what sorting it by source recovers. Context length 2048, 32 contexts, 512 candidate tokens.}
\label{tab:gelu-units}
\begin{tabular}{@{}lp{0.88\textwidth}@{}}
\toprule
unit & the tokens that drive it \\
\midrule
A & \texttt{\ about}, \texttt{\ him}, \texttt{\ them}, \texttt{\ who}, \texttt{\ whether}, \texttt{\ number}, \texttt{\ how}, \texttt{\ seen}, \texttt{\ name}, \texttt{\ B}, \texttt{\ that}, \texttt{\ E}, \texttt{\ U}, \texttt{\ \$}, \texttt{\ 3}, \texttt{\ In}, \texttt{\ go}, \texttt{\ F}, \texttt{\ 6}, \texttt{\ 7}, \texttt{\ T}, \texttt{\ New}, \texttt{\ 9}, \texttt{\ each}, \texttt{\ So}, \texttt{\ law}, \texttt{\ C}, \texttt{\ S}, \texttt{\ this}, \texttt{\ 5}, \texttt{\ then}, \texttt{re}, \texttt{\ went}, \texttt{\ A}, \texttt{\ change}, \texttt{B}, \texttt{\ 8}, \texttt{\ M}, \texttt{\ But}, \texttt{\ including} \\
\addlinespace
B & \texttt{000}, \texttt{\ United}, \texttt{\ This}, \texttt{\ million}, \texttt{\ 20}, \texttt{\ American}, \texttt{\ 0}, \texttt{\ That}, \texttt{\ 2}, \texttt{\ If}, \texttt{\ Trump}, \texttt{\ 10}, \texttt{\%}, \texttt{10}, \texttt{\ 1}, \texttt{\ The}, \texttt{\ It}, \texttt{0}, \texttt{\ But}, \texttt{\ 12}, \texttt{7}, \texttt{If}, \texttt{It}, \texttt{\ \&}, \texttt{6}, \texttt{This}, \texttt{\ percent}, \texttt{\ For}, \texttt{\ We}, \texttt{8}, \texttt{5}, \texttt{But}, \texttt{\ 5}, \texttt{\ 6}, \texttt{\ 4}, \texttt{2}, \texttt{'re}, \texttt{\ 9}, \texttt{\ 7}, \texttt{\ 8} \\
\addlinespace
C & \texttt{al}, \texttt{h}, \texttt{\ call}, \texttt{'}, \texttt{an}, \texttt{year}, \texttt{ar}, \texttt{\ public}, \texttt{man}, \texttt{I}, \texttt{\ If}, \texttt{\ called}, \texttt{\ both}, \texttt{\ open}, \texttt{a}, \texttt{,"}, \texttt{\ [}, \texttt{ll}, \texttt{\ more}, \texttt{\ As}, \texttt{i}, \texttt{ed}, \texttt{\ long}, \texttt{\ I}, \texttt{\ water}, \texttt{\ game}, \texttt{\ US}, \texttt{\ free}, \texttt{\ F}, \texttt{F}, \texttt{\ hard}, \texttt{\ much}, \texttt{C}, \texttt{\ once}, \texttt{as}, \texttt{\ part}, \texttt{\ system}, \texttt{\ per} \\
\addlinespace
D & \texttt{\ power}, \texttt{\ on}, \texttt{\ So}, \texttt{\ business}, \texttt{on}, \texttt{\ help}, \texttt{\ York}, \texttt{\ money}, \texttt{\ working}, \texttt{\ company}, \texttt{com}, \texttt{\ off}, \texttt{\ support}, \texttt{\ now}, \texttt{x}, \texttt{\ work}, \texttt{He}, \texttt{This}, \texttt{\ He}, \texttt{\ here}, \texttt{\ who}, \texttt{\ As}, \texttt{\ one}, \texttt{\ part}, \texttt{\ give}, \texttt{\ They}, \texttt{man}, \texttt{\ data}, \texttt{\ know}, \texttt{\ him}, \texttt{\ \&}, \texttt{z}, \texttt{\ say}, \texttt{\ might}, \texttt{\ he}, \texttt{\ today}, \texttt{\ This}, \texttt{u}, \texttt{\ very}, \texttt{c} \\
\addlinespace
E & \texttt{\ from}, \texttt{\ by}, \texttt{\ your}, \texttt{\ these}, \texttt{\ each}, \texttt{\ them}, \texttt{as}, \texttt{\ with}, \texttt{\ In}, \texttt{),}, \texttt{\ says}, \texttt{\ those}, \texttt{\ in}, \texttt{)}, \texttt{In}, \texttt{\ using}, \texttt{?}, \texttt{\ used}, \texttt{As}, \texttt{\ this}, \texttt{\ ]}, \texttt{\ as}, \texttt{).}, \texttt{\ said}, \texttt{'s}, \texttt{6}, \texttt{ing}, \texttt{\ K}, \texttt{\ --}, \texttt{\ him}, \texttt{\ were}, \texttt{9}, \texttt{in}, \texttt{the}, \texttt{]}, \texttt{,}, \texttt{\ This}, \texttt{\ his}, \texttt{en} \\
\addlinespace
F & \texttt{<|endoftext|>}, \texttt{\textbackslash n}, \texttt{\ police}, \texttt{\ R}, \texttt{as}, \texttt{\ four}, \texttt{\ where}, \texttt{\ all}, \texttt{\ women}, \texttt{\ L}, \texttt{\ State}, \texttt{\ five}, \texttt{\ p}, \texttt{\ She}, \texttt{\ must}, \texttt{\ political}, \texttt{\ government}, \texttt{\ 9}, \texttt{\ 8}, \texttt{\ much}, \texttt{\ money}, \texttt{\ And}, \texttt{\ M}, \texttt{If}, \texttt{\ later}, \texttt{\ just}, \texttt{(}, \texttt{9}, \texttt{\ first}, \texttt{\ possible}, \texttt{\ T}, \texttt{),}, \texttt{\ When}, \texttt{or}, \texttt{\ left}, \texttt{L}, \texttt{But}, \texttt{-}, \texttt{\ D}, \texttt{There} \\
\bottomrule
\end{tabular}
\end{table}

\begin{table}[htbp]
\centering
\small
\caption{The same presentation as Table~\ref{tab:gelu-units}, for the sigmoid model: top-$K$ token
sets at $K=40$ for units at the median of the effective-size distribution, unselected. Offered for
comparison with the baseline's rows above. Section~\ref{sec:mono-legibility} reads these through the
dependency graph instead, which is the measurement the paper relies on.}
\label{tab:mono-units}
\begin{tabular}{@{}lp{0.87\textwidth}@{}}
\toprule
unit & the tokens that drive it \\
\midrule
1 & \texttt{\ city}, \texttt{\ place}, \texttt{\ here}, \texttt{\ home}, \texttt{\ country}, \texttt{\ States}, \texttt{\ school}, \texttt{\ York}, \texttt{\ local}, \texttt{\ state}, \texttt{\ world}, \texttt{\ there}, \texttt{\ across}, \texttt{\ around}, \texttt{\ found}, \texttt{\ point}, \texttt{\ and}, \texttt{\ community}, \texttt{\ State}, \texttt{\ find}, \texttt{\ US}, \texttt{\ American}, \texttt{\ left}, \texttt{\ family}, \texttt{\ see}, \texttt{\ in}, \texttt{\ enough}, \texttt{\ course}, \texttt{\ run}, \texttt{\ away}, \texttt{\ asked}, \texttt{\ able}, \texttt{\ least}, \texttt{\textbackslash n}, \texttt{\ or}, \texttt{\ that}, \texttt{\ within}, \texttt{\ where}, \texttt{,}, \texttt{\ same} \\
\addlinespace
2 & \texttt{\ As}, \texttt{As}, \texttt{as}, \texttt{\ as}, \texttt{\ what}, \texttt{'re}, \texttt{<0xEF>}, \texttt{\ were}, \texttt{\ be}, \texttt{ic}, \texttt{\ was}, \texttt{\ place}, \texttt{\ are}, \texttt{is}, \texttt{es}, \texttt{\ part}, \texttt{\ being}, \texttt{\ example}, \texttt{al}, \texttt{e}, \texttt{\ is}, \texttt{ar}, \texttt{a}, \texttt{'s}, \texttt{\ how}, \texttt{\ \texttt{<0xEF>}}, \texttt{\ also}, \texttt{\ so}, \texttt{s}, \texttt{\ So}, \texttt{\ than}, \texttt{re}, \texttt{\ those}, \texttt{an}, \texttt{m}, \texttt{\ where}, \texttt{\ that}, \texttt{\ case}, \texttt{\ That}, \texttt{\ am} \\
\addlinespace
3 & \texttt{\ M}, \texttt{M}, \texttt{\textbackslash n}, \texttt{\ K}, \texttt{\ T}, \texttt{\ didn}, \texttt{\ L}, \texttt{\ don}, \texttt{\ D}, \texttt{T}, \texttt{m}, \texttt{\ an}, \texttt{\ from}, \texttt{\ H}, \texttt{B}, \texttt{\ P}, \texttt{D}, \texttt{p}, \texttt{P}, \texttt{\ a}, \texttt{t}, \texttt{\ doesn}, \texttt{\ A}, \texttt{\ like}, \texttt{A}, \texttt{\ went}, \texttt{\ think}, \texttt{\ B}, \texttt{\ go}, \texttt{\ high}, \texttt{\ C}, \texttt{\ F}, \texttt{\ against}, \texttt{\ the}, \texttt{\ fact}, \texttt{\ something}, \texttt{\ if}, \texttt{d}, \texttt{\ his}, \texttt{L} \\
\bottomrule
\end{tabular}
\end{table}

\paragraph{What concentration measures.} The natural way to put a number on
``is this unit nameable'' is to ask how concentrated its response is on its best single name, scored
against a null. This ranks a collapsed model as the most nameable thing in the study. A model whose
code is nearly rank-one scored 16.9 times its null, and under a metric that also counts how many
\emph{distinct} names the units receive, the same model produced nine of them across 1536 units,
scoring below the random null. When most units share one dominant component they share their top
inputs, so each scores a high concentration while being individually meaningless. Any nameability score needs a distinctness measure read beside it.

\paragraph{Rotating the model to fit the lens.} If mid-stack writes are illegible because
their frame disagrees with the readout, the direct fix is to train the model to keep them in the same
frame. We tried this as an auxiliary objective pulling each layer's lens distribution toward the
final one, over a range of strengths. Two things go wrong. The angle saturates: past a certain point
more pressure buys almost nothing, and the frames stop near fourteen degrees rather than at zero,
because the early layers cannot both decode to the output and still compute. And quality falls
sharply well before that, by up to forty-eight percent on LAMBADA at the strongest setting. The only
setting that costs nothing leaves the frame essentially where it started. Imposing the alignment
architecturally instead, with a learned orthogonal transform after every block trained to cancel each
layer's rotation, fails differently and more informatively: perplexity degrades by roughly an order
of magnitude \emph{and} the measured rotation increases. The geometry resists being removed, consistent with its being load-bearing rather than incidental~\citep{oskin2026offaxis}.

\paragraph{Why the lens is a derivation.} A sparse autoencoder
trained on these activations does recover features a person can name, at good reconstruction and with
few dead units. We reproduce that, the strongest positive result any method in this appendix produces. It is set aside here by choice rather than by definition. Nothing about legibility forbids a working fitted explainer of this kind. But an autoencoder is a second model,
trained on activations harvested from a corpus, and it explains the network in the same sense that a
second network trained to predict a network's outputs explains
it~\citep{bricken2023monosemanticity,cunningham2023sparse,bills2023neurons}: understanding the
explainer becomes part of the job.
A sparse autoencoder is a real and useful thing to have, and the gap between it and reading the weights is what this paper is about. One methodological note: an
$\ell_1$ penalty tuned by hand gave a solution with more than two hundred active units out of
two hundred and fifty-six, which is a dense rotation wearing a dictionary's clothes. Fixing the
active count by construction, with a top-$k$ activation, avoids the blind coefficient search
entirely.

\subsection{The ablation searches fail their own checks}
\label{sec:pit-ablate}

Two of the measurements in Section~\ref{sec:circuit} are searches over ablations, and both of the
obvious ways to run one are wrong. Neither failure shows up in a median.

\paragraph{Bisection lacks its invariant.} The required graph is the smallest prefix of the one-hop
set whose removal changes the prediction, and the cheap way to find it is to bisect on prefix
length: $\log n_{90}$ ablations instead of a scan. That is valid only if $\mathrm{flip}(k)$ is
monotone --- if, once removing the top $k$ changes the prediction, removing the top $k{+}1$ changes
it too. On $11$ to $28$ percent of predictions it is not. Some larger prefix restores the prediction
that a smaller one broke, and the same failure appears in all four architectures checked
(Table~\ref{tab:mono}).

The reason is the structure Appendix~\ref{sec:circuit-branch} measures. The dependency graph is not a
tree. Its nodes share sources heavily --- that overlap is exactly why the closure grows so much more
slowly than the branching factor would imply --- so a prefix of the ranking is not a set of
independent contributions that can be removed one at a time. Removing the top $k$ deletes components
that the components ranked below $k$ also draw on, which changes what those components compute. The $k{+}1$-th ablation is therefore a different perturbation of a different graph, with the remaining nodes moved. Nothing about that guarantees a monotone
sequence, and the surviving prediction after a larger ablation is the downstream compensation
familiar as self-repair \citep{mcgrath2023hydra}, here inside this instrument rather than in a
model under study.

Bisection, with nothing to stand on under those conditions, fails in a predictable direction.
It agrees with the true minimum on $84$ to $100$ percent of predictions and otherwise lands high, by
as much as $25$ components on Llama-3.2-1B. The individual predictions behind a correct-looking table of medians built by bisection are wrong.

\begin{table}[htbp]
\centering\small
\caption{Sweeping every prefix length instead of bisecting. ``Flip at all'' counts predictions that
change when some prefix is removed; the required graph is undefined for the rest, and they are
excluded from the medians reported elsewhere. ``Worst overshoot'' is the largest amount by which
bisection exceeded the true minimum on a single prediction.}
\label{tab:mono}
\input{T_mono}
\end{table}

\begin{figure}[htbp]
\centering
\includegraphics[width=\textwidth]{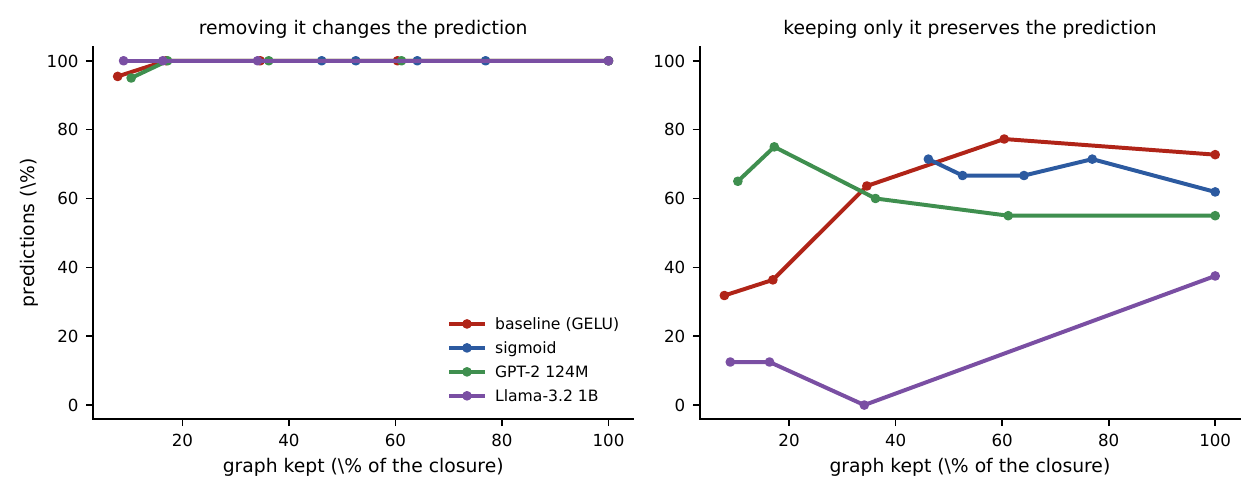}
\caption{Both directions on the same pruned graphs. Left: removing the graph changes the prediction
almost as often at a tenth of the closure as at all of it. Right: keeping \emph{only} the graph,
which is the sufficiency direction --- the curves are not monotone, and on three of four models
keeping more preserves the prediction \emph{less} often. That is a failure of the instrument rather than a property of the circuit.}
\label{fig:prune}
\end{figure}

\paragraph{Keeping more can preserve less.} Appendix~\ref{sec:circuit-closure} reports the
sufficiency direction failing on the baseline. It fails the same way on models this paper did not
train, which is what makes it a property of the measurement. Keeping only a pruned graph preserves
the prediction between $0$ and $77$ percent of the time, and on three of the four models below
(Table~\ref{tab:prune-all}, Figure~\ref{fig:prune}) the curve is not monotone in how much is kept: GPT-2 peaks at $\theta = 0.3$ and falls thereafter, and
Llama-3.2-1B reads $12$, $12$, $0$, $38$. A set that genuinely produced a prediction could not do
worse when enlarged.

The cause is that keeping a subset ablates everything else, so the model is evaluated far off its
own distribution, and how far depends on the subset in a way that has nothing to do with whether the
subset is the circuit. The necessity direction in the same table is unaffected: it removes a small
set from an otherwise intact model, which is a mild perturbation of a model still on its
distribution. That asymmetry is why a set cannot be certified sufficient by enlarging it until it works. The
instrument that does work runs the other way: Section~\ref{sec:circuit-suf} searches directly for a
small set and tests it by running the model, scoring its controls against the floor the ablation
itself creates.

\begin{table}[htbp]
\centering\small
\caption{Pruning by the per-node threshold $\theta$, across four models. Columns as in
Table~\ref{tab:prune}. The same predictions are used at every $\theta$, so rows within a model are
paired.}
\label{tab:prune-all}
\input{T_prune}
\end{table}

\subsection{Whether the instrument or the model is at fault}
\label{sec:naming-verdict}

Every negative result in Section~\ref{sec:naming} admits two readings --- the model has no structure
of the kind sought, or the measurement could not have found it --- and they are separated here rather
than left to the reader, because three of the four turn out to be the instrument.

The frame is the clearest case. Scoring a writer through the unembedding and its reader in that
reader's own basis returns a correlation of $-0.02$ against a random control, which reads as a
component passing nothing to the one that consumes it. The same edges scored in one frame correlate $+0.28$, with nothing about the model changed between the two numbers.

The fit is the second. Reconstructing a unit's drivers from upstream components returns negative
held-out variance under every pool tried, which invites the conclusion that no such relation exists.
Selecting three atoms from a dictionary of $1{,}440$ costs about twenty-nine bits against thirty
observations, so the search exhausts the sample before a coefficient is estimated; and the profile
being reconstructed reproduces across disjoint contexts at $0.648$, so the signal was there
throughout. An ordering, which selects nothing, recovers it (Section~\ref{sec:naming-provenance}).

Attention is the third. Its components rank a unit's drivers close to chance, and attention supplies
$23$ percent of the components in a trace and is frequently among those a prediction cannot lose.
Driver sets are built by substituting a token at one position and reading a unit at the same
position; attention assembles its output at a position from \emph{other} positions. A term that does
not move when the probe moves cannot order the probe's results, whatever it contributes to the
prediction.

The fourth is the model. A component operating on a direction no token induces has no vocabulary name at any resolution. The write column comes no closer to a token than a random
direction through the early and middle stack. A fitted rotation recovers nothing over not rotating.
And a shared basis fitted across units explains the \emph{nameable} response better than the rest.
Where the reading stops there, it stops because the object is not lexical.

\subsection{The opposing side of the accounting}
\label{sec:pit-opposing}

Section~\ref{sec:circuit-accounting} sets the opposing mass aside as traffic. Whether that is right
can be tested at the readout directly, by asking whether the mass
pushing a token \emph{down} is any different from what an arbitrary token receives. The test uses
the contrastive pairs of Section~\ref{sec:circuit-suf}, $A \rightarrow X$ and $B \rightarrow Y$,
where what separates the two positions is what attention carried in from earlier context. At $B$,
where $X$ lost, every feed-forward unit and attention channel has a signed push on $X$, and
the negative side of that accounting is set against three floors: the same accounting for a
competitor the model merely considered, a random token from its top fifty; for a random vocabulary
token; and for $X$'s own contributions with their signs shuffled and their magnitudes kept, which is
what picking the most negative of tens of thousands of terms produces on its own.
Table~\ref{tab:supp} reports the comparison. The token that lost carries the same negative side as a
token nobody was deciding about. The ratio of its negative to its positive mass matches the
competitor's to within $0.02$ in every model, and a random token's is one; reaching ninety percent
of its negative mass takes the same share of the model as it takes with the signs shuffled, to
within a point; and removing its eight most negative components lifts it by about as much as the
same operation lifts the competitor, within a logit in every model and by less than it lifts a
random token on two of them. Attention channels carry the negative mass in proportion to their
number, and no single component switches on to push $X$ down where it loses: the most negative push
at $B$ among components that were not pushing $X$ down at $A$, scored against the spread of its kind,
is within a unit of the same statistic for the control tokens in every model, and on the bounded
arms a unit cannot switch at all, since a nonnegative activation fixes the sign of its push on a
token. ``Not this'' is written onto the readout by thousands of components in every context and
for every token, and it is the mirror of the positive side rather than a signal about the token.

Much of that mass is a constant per component rather than anything the position decides: split
each component's activation into its mean over positions and the deviation from it, and the mean
carries about three quarters of the mass on both sides of the baseline's accounting, effectively all
of it on the two bounded arms, half on GPT-2 and a quarter on Gemma-3 and Qwen. Rerunning the
comparison with that constant part removed and with ablation to the mean rather than to zero
changes none of its conclusions: every entry of Table~\ref{tab:supp} moves by a few hundredths or
a few points and the circuits of Table~\ref{tab:pair} stay at tens of components. The one case where
the constant part matters is the seed of the last row of Table~\ref{tab:reading}: four attention
channels writing large constants at every position dominate its signed sum, and with the constant
part removed its net denominator returns from one or two components to $94$, the net is negative on
$3$ percent of pairs instead of $39$, and its negative side sits at the floor like every other
model's.

\begin{table}[htbp]
\centering\small
\caption{The negative side of a token that lost, against its floor. On the contrastive pairs of
Section~\ref{sec:circuit-suf}, at the position where $X$ lost, the mass pushing $X$ down over the mass pushing it up, for $X$, for a
competitor from the model's top fifty and for a random vocabulary token; the share of the model's
components it takes to reach ninety percent of $X$'s negative mass, against the same count with the
signs of $X$'s contributions shuffled; and the logit lift from removing the eight components
pushing each token down hardest. Medians over pairs. Feed-forward units and attention channels
alike.}
\label{tab:supp}
\input{T_supp}
\end{table}

\subsection{Conditions for trusting these numbers}
\label{sec:naming-conditions}

Every measurement in Section~\ref{sec:naming} can be made to return a confident wrong answer.
Table~\ref{tab:reading} states the conditions that separate the two outcomes, each with the number
that moves when it is dropped, because a condition offered without its consequence is easy to read as
fastidiousness and skip.

\begin{table}[htbp]
\centering\small
\caption{Conditions for reading a component. The last column gives the measurement with the
condition and without it. None of these is a matter of
taste, and a reading that drops one is not a weaker version of this one --- it is capable of
returning the opposite sign.}
\label{tab:reading}
\begin{tabular}{@{}p{0.24\textwidth}p{0.40\textwidth}p{0.28\textwidth}@{}}
\toprule
condition & why it is not optional & what it cost \\
\midrule
Read at the readout, or in the layer's own frame. Never mix the two.
& The residual stream is not in token coordinates until the last few layers, so a mid-stack
  quantity decoded through the unembedding is being asked what it would mean somewhere it is not.
& Writer-to-reader agreement measured $-0.02$ across frames and $+0.28$ within one; the first reads
  as ``the graph does not link''. \\
\addlinespace
Give every concentration or overlap statistic a matched random control.
& Random directions concentrate, and random sets overlap. Neither statistic has a meaningful zero.
& A top-eight response share of $0.105$ looks like selectivity until the random control returns
  $0.097$. \\
\addlinespace
Check that the null could have been rejected before believing it.
& A test whose null sits near the ceiling cannot discriminate, so its null result carries no evidence of absence.
& Projecting onto the span of all candidate tokens put the null at $0.665$ of a $768$-dimensional
  space; the excess was zero everywhere because the test had no power. \\
\addlinespace
Separate a component's direction from its behavior.
& Selectivity can occupy a small part of a direction whose bulk does something unrelated, so the two
  questions have different answers.
& Reconstruction of the direction sits at chance while $21$ to $73$ percent of units are
  selective beyond the random $95$th percentile. \\
\addlinespace
Filter the candidate token set to words.
& Frequent fragments and punctuation dominate a raw frequency list and appear at the top of every
  component's reading, which says more about the corpus than about the component.
& Unfiltered, \texttt{'\^{}\^{}\^{}\^{}'} and bare subword fragments headed the detection list of
  most units in the gallery. \\
\addlinespace
Probe at the context length the model was trained at.
& Statistics that depend on a variance floor are computed against the wrong floor at a shorter
  context, and channels that operate at range are scored as dead.
& Probes run at $256$ tokens on models trained at $2048$ reported a population of dead channels
  that does not exist. \\
\addlinespace
Remove a component's constant write before reading a signed accounting.
& A channel that writes nearly the same large vector at every position contributes a constant to
  every token's accounting, of either sign. That is a bias, not a decision, and left in, it can
  dominate a signed sum.
& On one of the baseline's five converged seeds, four layer-1 attention channels output between
  $115$ and $143$ in magnitude at every position with a spread of $0.002$. The net denominator
  returned one or two components and the net went negative on $39$ percent of contrastive pairs,
  on a model whose predictions were as confident as the other seeds'. With each component's mean
  write removed, the count returned to $94$ (Appendix~\ref{sec:pit-opposing}). \\
\bottomrule
\end{tabular}
\end{table}

\subsection{Fitting a unit's drivers from its upstream components}
\label{sec:naming-fitfail}

Section~\ref{sec:naming-provenance} ranks upstream components rather than fitting them, because
every fitted version failed, all of them in the same way, and the failure mode is invisible in the
statistic normally used to guard against it.

A unit's pre-activation is a sum of upstream write columns carried forward and read through its row,
so the natural estimator selects a few upstream components and regresses the unit's response on their
activations, scoring on tokens held out from the selection. Four candidate pools were tried. The co-members of a unit's own sufficient set; the union of every
component appearing in any sufficient set; the components that share sufficient sets with the target,
against a control matched on how often a component is reused; and the twelve upstream components
whose carried write columns align most strongly with the unit's read row. Every candidate was carried through the fitted layer maps, the
carried embedding was forced into the design so that what was scored was explanation beyond the
starting residual, and attention channels were included alongside feed-forward units.

All four returned \emph{negative} held-out variance --- worse than predicting the mean --- and the
strongest pool returned the worst result, $-0.306$ against $-0.009$ for the carried embedding alone.

Observations per parameter, the usual guard, says the design was adequate: three atoms fitted on
thirty tokens is ten observations each. It is the wrong count. The atoms are selected as well as
fitted, and selecting three columns from a dictionary of $1{,}440$ costs $\log_2\binom{1440}{3}
\approx 29$ bits against thirty observations. The search consumes the sample before a coefficient is
estimated, so the columns that survive are the ones correlated with noise in the training half, and
those anti-correlate out of sample. That is why the scores are negative rather than small, and it is
why raising the sample and lowering the atom count --- which moved observations per parameter from
$1.8$ to $10$ --- changed nothing.

The reading that the model simply has no such structure is ruled out separately. Split the sixteen
contexts in half and the two halves agree on the anonymous profile at $0.648$ (CI $[0.643, 0.654]$,
$n=14{,}123$), so roughly two thirds of the variance being fitted is reproducible signal. A ranking, which selects nothing and fits nothing, reaches the always-predictable target that the estimator never could.

\subsection{Reading a write column against the vocabulary}
\label{sec:geometry-tried}

The naming of Section~\ref{sec:naming-vocab} works from predictions and never reads a column. That was
not the first approach tried, and the measurements that failed are collected here, because their
failure has a single cause.

A component's contribution to a prediction is a product of two terms, how hard it fired and how well
its column aligns with what the readout needs. Measured on the baseline, the column puts the tokens it
is credited for at the $86$th percentile of its own ordering over the vocabulary, and a third of them
inside its own top one percent --- so the column is far from silent. But the firing carries $52$
percent of the variance in the contribution. The column narrows the field to roughly the top seventh
of the vocabulary and the context chooses from inside it, so no measurement of the column alone can
recover which token a component is credited for. Reversing the question confirms it: given components
already labeled by what they write, retrieving a label from column geometry alone beats chance by
only $1.23$ times on the baseline and $1.35$ on the sigmoid model, real but far too weak to invert.

What was tried, and what each returned. The best cosine between a column and any single centered
unembedding row sits at chance through two thirds of every model (Table~\ref{tab:write}). Splitting a
layer's write into the part a fixed map of the incoming state accounts for and the part left over does not rescue it (Table~\ref{tab:split}). The set a
column points at is no tighter than the set an arbitrary direction picks out, once the comparison is
made against a random \emph{direction} rather than a random set of tokens. Morphological axes exist in
the embedding --- plural, capitalization, tense, all consistent at $0.29$ to $0.55$ when each pair is
scored against an axis built from the others --- but write columns lie along them at chance, and a
component's contribution to a plural prediction is separable from a singular one only in the aggregate
and not per column. The readout's own geometry is no help either: one direction holds most of its
variance and none of the residual stream's energy, and removing it costs nothing while removing the
two behind it costs eleven points of top-one accuracy, so there is no free correction to apply.

\begin{table}[htbp]
\centering\small
\caption{Where a write column sits relative to the readout, by depth. Each entry is the median
column's best cosine with any single centered unembedding row, in excess of the same statistic for a
random direction, so $0$ is chance and the chance level itself is given for scale. Feed-forward and
attention are the same kind of object here --- one column of an output projection, scaled by a scalar
the model computes --- and are measured identically. Bold marks the last third, the only band in
which any model clears chance. As a check that the statistic can see readout-directed writing at all,
the state entering the final norm scores $0.249$ on it against these chance levels, and an
unembedding row scores $1.000$.}
\label{tab:write}
\input{T_write}
\end{table}

\begin{table}[htbp]
\centering\small
\caption{Where the token alignment of a layer's write sits, once the write is split into the part a
fixed map of the incoming state accounts for and the part left over
(Section~\ref{sec:circuit-bookkeeping}). Both are scored with the same best cosine against any
centered unembedding row used in Table~\ref{tab:write}. ``First layers'' is the largest value over
the opening quarter of the stack and ``rest'' the median over the remainder. The alignment is in the
fitted part, which is what carrying the embedding forward looks like; the leftover is at chance
throughout. Layers where the leftover is under a quarter of the write are excluded from its range,
since the direction of a vector that small carries no information --- this drops the sigmoid model's
first four layers and none of the baseline's.}
\label{tab:split}
\input{T_split}
\end{table}

This is why the naming had to come from predictions.

\subsection{Naming a unit by where it fires}
\label{sec:geometry-firing}

The move that names the write side does not transfer, and the failure is about the architecture
rather than about the measurement. One could collect the positions in
ordinary text where a unit fires hardest and name it by the token sitting there, and two units in
three do acquire a name that way. But the token at a position is present in the residual only at the
very bottom: it starts at $+0.45$ and is down to $+0.003$ one layer later. Above that a unit is reading what the model has built rather than the token that seeded it, and any agreement between the two is running through whatever that token caused. Measured by depth, the agreement behaves accordingly ---
it is three times stronger at layer zero than at layer eleven, and falls the whole way. Naming by firing position therefore describes the contexts a unit turns on in, a real fact about the unit rather than what it reads.

The same objection disposes of naming a unit by the token some fixed distance \emph{before} its
firing positions. That token is not in the residual either, and the instrument cannot be completed by
sweeping the distance: whatever a unit keys on may lie hundreds of positions back, so any offset is a
truncation. The parameters read against the layer's own table are the instrument that answers what a
unit reads, and they answer it sharply for the head of the list and not below.

\begin{table}[htbp]
\centering\small
\caption{Attention alignment split by head, at the two ends of the stack. Each entry is the range
across the model's twelve heads of that head's write columns weighted by how hard they fire, scored
by the same best cosine with any centered unembedding row but reported as an absolute cosine rather
than as an excess, with the chance level beside it for scale. Layer zero is aligned in every model
because the residual there is the token; the last layer separates the baseline from the rest.}
\label{tab:heads}
\input{T_heads}
\end{table}

Attention was measured the same way and reads the same, with one difference of its own: it leans on
few of its columns, the top one percent carrying up to a fifth of a block's firing where a
feed-forward block's carry at most a fortieth. A median over all columns therefore describes a typical
column of a block that mostly uses a few, which is why a channel's contribution sits further above its
median column than a unit's does. Whether the columns a block uses are also the readable ones is a
separate question and the answer is mostly no: weight each column by how hard it fires and compare
against the same weights permuted across columns, and only the first attention block shows a large
gap, where the residual is still the token itself.

One cross-model difference does survive the measure. At the last layer the baseline's heads straddle chance
while every other model's sit above it (Table~\ref{tab:heads}) --- read through the typical column,
the baseline holds its computation off the readout axis until later than the rest.

\subsection{What the tokens outside the concept are}
\label{sec:naming-remainder}

Section~\ref{sec:naming-about} splits a unit's driver set into the part forming an embedding-space
concept and a remainder the cosine cannot speak to, and reports the second as $72$ percent of the
response on the baseline, and $61$ to $72$ percent across the four models measured here. Three explanations of that remainder can be tested without consulting the embedding for the
verdict, and none of the three survives. What the tests leave is a description rather than a name, and
the description is what Section~\ref{sec:naming-provenance} then works from: they establish that the
remainder is real and single before it is read in another basis. They run over $1{,}444$ units drawn
at random across four models and sixteen contexts, and every number is quoted beside the same
statistic on the same unit's coherent tokens (Table~\ref{tab:remainder}).

The first possibility is that the remainder is nothing, its tokens there by sampling accident. The
test is \emph{reliability}: build the response profile on one half of the contexts, build it on the
disjoint other half, and ask whether the remainder's ordering reproduces. Noise does not reproduce.
It reproduces. The rank correlation between halves runs from $0.52$ to $0.64$ across the four models,
against $0.59$ to $0.75$ for the coherent tokens of the same units. For a minority the answer is the
other way, $11$ to $20$ percent of units falling below $0.2$, and for those the remainder is sampling
noise. For the rest it is response the model reproduces on contexts it has not seen.

The second is that the remainder is several concepts at once, one unit carrying three paradigm
classes with no single coherent direction among them. The test is \emph{substructure}: count the
eigen-directions of the remainder's own similarity matrix that exceed the largest eigenvalue of a
size-matched random set. It carries none. Not one unit of the $1{,}444$ shows more than a single
direction above the null, and $15$ to $40$ percent show none at all.

The third is that the remainder is relational, and a syntagmatic class has a signature: a paradigm
unit fires on its token wherever the token appears, while a relational unit fires only in the right
structure. The test is \emph{context-dependence}, the coefficient of variation of the response across
contexts. The signature is absent. The remainder varies slightly \emph{less} across contexts than the
coherent part on every model, and the share of units running the other way is $31$ to $50$ percent.

What the three leave is a remainder that is real, single, and no more context-bound than the tokens a
concept does explain. That is a sharper object than a blind spot: it is not an artifact to be
explained away, and it is not several things that better tools would separate. It is one thing, read
in a basis it is not written in --- which is what Section~\ref{sec:naming-provenance} measures by
changing the basis to the components upstream of the unit.

One candidate can be ruled out, and it is the obvious one. Most of what a transformer does between
layers is carriage (Section~\ref{sec:circuit-bookkeeping}), so the natural guess is that a unit's
unnameable response is that unit taking part in carriage rather than responding to anything. Fitting
the layer-to-layer map gives the subspace carriage works in, and each write column can be scored by
how much of itself lies there. Across $326$ baseline units and $278$ sigmoid units measured both
ways, that score correlates with the nameable share at $+0.14$ and $+0.22$: the wrong sign, and the
wrong sign after controlling for depth and for membership in the residual's own high-variance
subspace, with which the carriage span shares no more than a random overlap. Units that write more
into carriage are slightly \emph{more} nameable. There is also a reason the guess was unlikely to
hold as stated. A unit writes one direction whatever drives it, so how much of it is carriage is a
property of its write, while how much of it is nameable is a property of its read. The two need not
have anything to say to each other.

The read side admits the same test and fails it too. A component that carries the stream has to
\emph{read} what is being carried, so its read row should lie in the carriage span; unlike the write
column, the read row varies with what drives the unit, so this comparison is not ruled out by
construction. Read rows sit in that span at $1.26$ and $1.34$ times a matched random direction, about
what write columns manage, and the lift predicts nothing about nameability: $r = +0.007$ on the
baseline and $+0.177$ on the sigmoid model, the second again the wrong sign. Reading the carried
stream is not what the unnameable response is.

\paragraph{The remainder against an untrained weight.} A read row that never received a useful gradient would
still look like its initialization, and a random direction has no token that names it, which would
make a unit unnameable by default. Comparing the checkpoint at $50{,}000$ steps against the end of
training settles it without recourse to an initialization seed: not one read row of $36{,}864$ in
either model is static. The median cosine between the two checkpoints is $0.58$ and $0.64$, each row
traveling about four fifths of its own norm over that stretch, and no unit moves less than a tenth
of it. Nameability is flat across the range --- $r = +0.011$ on the baseline, and the least-moved and
most-moved halves of the population are both at $28$ percent nameable to the point.
\begin{table}[htbp]
\centering\small
\caption{What the tokens outside the concept are. ``Coherent'' is the part of a unit's top forty that
clears a cut calibrated against a random set of the same size, and the share is of the unit's
response. Reliability is the rank correlation of a token ordering built on one half of the contexts
against the disjoint other half. Substructure counts eigen-directions of the remainder above a
size-matched null. Context variation is the coefficient of variation of the response across contexts.
Units are sampled uniformly at random per layer, and a unit is examined only where both sides have at
least four tokens.}
\label{tab:remainder}
\input{T_remainder}
\end{table}

\secbarrier

%% file: T_mono.tex
\begin{tabular}{lrrrrr}
\toprule
model & predictions & flip at all & $\mathrm{flip}(k)$ monotone & bisection exact & worst overshoot \\
\midrule
baseline (GELU) & 39 & 25 & 72\% & 84\% & 4 \\
sigmoid & 38 & 37 & 89\% & 100\% & 0 \\
GPT-2 124M & 33 & 31 & 84\% & 94\% & 16 \\
Llama-3.2 1B & 32 & 29 & 83\% & 90\% & 25 \\
\bottomrule
\end{tabular}

%% file: T_prune.tex
\begin{tabular}{llrrrr}
\toprule
model & $\theta$ & components & \% of model & \% of closure & prediction changes \\
\midrule
baseline (GELU) & 0.1 & 56 & 0.12\% & 8\% & 95\% \\
 & 0.3 & 122 & 0.26\% & 17\% & 100\% \\
 & 0.5 & 249 & 0.54\% & 35\% & 100\% \\
 & 0.7 & 434 & 0.94\% & 60\% & 100\% \\
 & 0.9 & 720 & 1.56\% & 100\% & 100\% \\
\addlinespace
sigmoid & 0.1 & 36 & 0.08\% & 46\% & 100\% \\
 & 0.3 & 41 & 0.09\% & 53\% & 100\% \\
 & 0.5 & 50 & 0.11\% & 64\% & 100\% \\
 & 0.7 & 60 & 0.13\% & 77\% & 100\% \\
 & 0.9 & 78 & 0.17\% & 100\% & 100\% \\
\addlinespace
GPT-2 124M & 0.1 & 102 & 0.22\% & 10\% & 95\% \\
 & 0.3 & 170 & 0.37\% & 17\% & 100\% \\
 & 0.5 & 358 & 0.78\% & 36\% & 100\% \\
 & 0.7 & 604 & 1.31\% & 61\% & 100\% \\
 & 0.9 & 987 & 2.14\% & 100\% & 100\% \\
\addlinespace
Llama-3.2 1B & 0.1 & 154 & 0.09\% & 9\% & 100\% \\
 & 0.3 & 282 & 0.17\% & 16\% & 100\% \\
 & 0.5 & 588 & 0.36\% & 34\% & 100\% \\
 & 0.9 & 1724 & 1.05\% & 100\% & 100\% \\
\addlinespace
\bottomrule
\end{tabular}

%% file: T_supp.tex
\footnotesize\setlength{\tabcolsep}{5pt}
\begin{tabular}{@{}lrrrrrrr@{}}
\toprule
& \multicolumn{3}{c}{negative / positive mass} & \multicolumn{2}{c}{$n_{90}$, negative side} & \multicolumn{2}{c}{lift, 8 removed} \\
\cmidrule(lr){2-4}\cmidrule(lr){5-6}\cmidrule(lr){7-8}
model & loser & competitor & random & loser & shuffled & loser & competitor \\
\midrule
baseline (GELU) & 0.96 & 0.97 & 1.02 & 26\% & 26\% & 0.3 & 0.0 \\
sigmoid & 0.89 & 0.90 & 0.99 & 18\% & 18\% & 1.9 & 1.8 \\
set operators & 0.89 & 0.90 & 0.99 & 19\% & 18\% & 0.7 & 0.2 \\
GPT-2 124M & 0.86 & 0.88 & 1.01 & 24\% & 23\% & 1.6 & 1.8 \\
Gemma-3 1B & 0.86 & 0.87 & 1.00 & 16\% & 16\% & 2.3 & 2.2 \\
Qwen2.5 1.5B & 0.91 & 0.92 & 1.00 & 20\% & 20\% & 1.3 & 1.4 \\
\bottomrule
\end{tabular}

%% file: T_write.tex
\setlength{\tabcolsep}{5pt}
\begin{tabular}{lrrrrrrr}
\toprule
& & \multicolumn{3}{c}{feed-forward} & \multicolumn{3}{c}{attention} \\
\cmidrule(lr){3-5}\cmidrule(lr){6-8}
model & chance & early & middle & late & early & middle & late \\
\midrule
baseline (GELU) & 0.143 & -0.002 & -0.001 & \textbf{+0.006} & -0.001 & -0.002 & \textbf{+0.005} \\
sigmoid & 0.143 & +0.003 & -0.001 & \textbf{+0.031} & -0.007 & -0.001 & \textbf{+0.026} \\
softplus & 0.141 & +0.001 & -0.001 & \textbf{+0.042} & +0.001 & -0.000 & \textbf{+0.066} \\
ReLU & 0.141 & +0.002 & -0.002 & \textbf{+0.022} & +0.021 & -0.001 & \textbf{+0.027} \\
set operators & 0.136 & -0.008 & -0.006 & \textbf{+0.024} & -0.005 & -0.002 & \textbf{+0.058} \\
unshaped sigmoid & 0.143 & -0.001 & -0.002 & \textbf{+0.032} & +0.001 & +0.001 & \textbf{+0.054} \\
\bottomrule
\end{tabular}

%% file: T_split.tex
\setlength{\tabcolsep}{6pt}
\begin{tabular}{lrrrr}
\toprule
& & \multicolumn{2}{c}{fitted part} & leftover \\
\cmidrule(lr){3-4}\cmidrule(lr){5-5}
model & chance & first layers & rest & range over depth \\
\midrule
baseline (GELU) & 0.143 & \textbf{0.202} & 0.142 & 0.139--0.146 \\
sigmoid & 0.143 & \textbf{0.157} & 0.151 & 0.139--0.142 \\
\bottomrule
\end{tabular}

%% file: T_heads.tex
\setlength{\tabcolsep}{5pt}
\begin{tabular}{lrrrrr}
\toprule
& & \multicolumn{2}{c}{first layer} & \multicolumn{2}{c}{last layer} \\
\cmidrule(lr){3-4}\cmidrule(lr){5-6}
model & chance & range & median head & range & median head \\
\midrule
baseline (GELU) & 0.143 & 0.160--0.602 & 0.374 & 0.141--0.173 & 0.162 \\
sigmoid & 0.143 & 0.159--0.665 & 0.358 & 0.185--0.267 & 0.229 \\
softplus & 0.141 & 0.148--0.550 & 0.242 & 0.222--0.300 & 0.261 \\
ReLU & 0.142 & 0.136--0.299 & 0.216 & 0.184--0.233 & 0.203 \\
set operators & 0.136 & 0.134--0.526 & 0.281 & 0.182--0.247 & 0.202 \\
unshaped sigmoid & 0.142 & 0.146--0.475 & 0.240 & 0.209--0.284 & 0.253 \\
\bottomrule
\end{tabular}

%% file: T_remainder.tex
\begin{tabular}{lrrrrrrr}
\toprule
& & \multicolumn{2}{c}{coherent} & \multicolumn{3}{c}{remainder} \\
\cmidrule(lr){3-4}\cmidrule(lr){5-7}
model & units & tokens & \% of response & reliability & directions & context var. \\
\midrule
baseline (GELU) & 338 & 11 & 28\% & 0.52 \small{(0.59)} & 1 & 1.11 \small{(1.26)} \\
sigmoid & 324 & 11 & 28\% & 0.64 \small{(0.75)} & 1 & 0.65 \small{(0.68)} \\
softplus & 347 & 14 & 39\% & 0.58 \small{(0.67)} & 1 & 0.73 \small{(0.79)} \\
set operators & 435 & 15 & 39\% & 0.60 \small{(0.69)} & 1 & 0.68 \small{(0.68)} \\
\bottomrule
\end{tabular}

%% file: appgraph.tex
\section{The dependency graph}
\label{sec:appgraph}

Section~\ref{sec:circuit} identifies the components that carry a prediction and the smallest set
that builds it, and neither measurement needs to know where a component's own input came from.
That question has an answer, and this appendix gives it: the edge relation between two components,
the recursion that closes it into a graph, and what the resulting object is good for.

Two results here are used elsewhere in the paper. The edges are causal rather than merely
attributed, which is what licenses Section~\ref{sec:naming} to describe a component by what its
inputs carry. And a graph's second level cannot be estimated by multiplying its first level by a
branching factor --- the product overstates the measured count, by a factor that grows with the
branching itself --- which anyone measuring these graphs is likely to need.

The rest is a route that does not arrive. The closure, built to answer the sufficiency question, answers it worse than a set of the same size taken straight off the contribution ranking.
Section~\ref{sec:circuit-suf} reports what does work.

\subsection{Edges, branching, and the closure}
\label{sec:circuit-branch}

If $n_{90}$ is one hop, the natural next question is how quickly the structure widens below it. For
a node $x$ in the one-hop set, the share of its incoming drive owed to an upstream component $v$ is
\begin{equation}
  e_{v \to x} \;=\; a_v \, \big\langle w_v,\; \gamma_{\ell} \odot r_x \big\rangle \big/ \rho_{\ell},
  \label{eq:edge}
\end{equation}
with $r_x$ the read row of $x$ and $\gamma_\ell$ the gain of the norm feeding its sublayer. The
scale $\rho_\ell$ is a single number for a given $x$, so it cancels from every share and the
recursion needs no normalization of its own.

The ninety percent here is a different quantity from the one of Section~\ref{sec:circuit-accounting}. That bar was on contribution to the readout, and there is one such total per
prediction. This one is on contribution to a single node's pre-activation, and there is a different
total for every node in the graph. What the same fraction is a fraction \emph{of} differs.
With that fixed, define the \emph{branching factor} as the number of upstream components needed to
reach ninety percent of a node's own incoming drive.

Equation~\ref{eq:contrib} is already exact, so it is fair to ask what a second level adds. The
contributions sum to the logit, so the one-hop set is not an approximation of the answer that a
second level improves --- it is a complete account of \emph{the logit, given the activations}. Every
component in it writes onto the readout direction; that is what selects it; and no member depends on
any other. The one-hop set is a star around the readout rather than a graph.

What the accounting takes as given is $a_u$. The write projection in Equation~\ref{eq:contrib} is a
parameter and is fixed, but the activation is an observed number, and nothing so far says where it
came from. Equation~\ref{eq:edge} is the first statement in this paper that relates two components
to each other rather than each to the readout, and what it decomposes is not the logit but the
pre-activation behind $a_u$. That is also the influence the exact decomposition structurally
omits: a component in an early layer is credited for its own write landing on the readout, never for
the fact that later layers read it and act on what they find.

Nothing stops that step from repeating. Every source it returns is itself a component with a read
row, so the same question can be put to it, and to whatever it returns, until no node remains that
has not been asked. Call the fixed point the \emph{closure}: the one-hop set together with
everything reachable from it by following ninety percent of each node's own incoming drive. It is the
backward dependency graph of a single prediction, and every member of it is counted rather than
estimated.

Two properties keep the recursion from running away. A component can draw only on components
strictly below it, so the graph is acyclic and ordered by depth. And the sets overlap heavily:
different nodes draw on many of the same sources, so each level adds far less than the previous
level's size times the branching factor. On the baseline the counts run $53$, $336$, $736$ over six
levels, after which the frontier is empty --- $1.6$ percent of the model's components. Every model
in this paper closes this way, at four to ten levels, and the closure runs three to six times the
one-hop set on the shaped arms, fourteen times on the baseline, and twelve to sixty on the models
trained by other people.

That overlap is why the product of $n_{90}$ and the branching factor is never reported here as a
stand-in for the second level. The product counts every component shared by two nodes once per node; across
the eighteen models here it overstates the measured two-level closure by $1.6$ to $9.8\times$,
median $3.6\times$, by a factor that grows with the branching factor itself and so distorts
comparisons between models as well as the size of any one.

A graph built this way can also be thinned, by lowering the share of a node's incoming drive its
sources have to account for. Writing that share $\theta$, the closure above is $\theta = 0.9$, and
smaller values keep only the strongest edges out of each node. Both measurements that follow are read
at several values of $\theta$, and they do not agree about what happens as it falls.

Only feed-forward nodes are expanded. An attention component here is a named location --- a layer, a head, and a channel
within that head. It is scored by Equation~\ref{eq:edge} like any other component, it is ranked
against them, it enters the one-hop set and the closure on the same footing, and when a graph is
ablated it is ablated by name. Nothing about its contribution is unknown.

What is not done is to ask what produced the value that channel read. That question is about a different graph. A feed-forward unit at position $p$ draws on the
residual stream at $p$, so its sources sit at the same position and lower layers, and the recursion
stays inside a depth-ordered acyclic graph. An attention channel at $p$ draws on values written at
earlier positions, so following it leaves the token whose prediction is being explained and opens a
graph over the sequence. This paper measures the graph at one position and treats attention as its
boundary: between a third and three fifths of a closure is attention, identified and ablated but not
followed. What the graph is
worth is an activation-function question, taken up in Sections~\ref{sec:mono-branch}
and~\ref{sec:stock-branch}; here the point is only that these edges are real.

These edges are causal rather than merely attributed. Removing a node's eight highest-ranked sources
leaves it at $0.34$ of its original activation; removing eight components drawn at random from the
same upstream pool leaves it at $1.0004$. The separation holds at every count tested, and the
random control never moved a node by more than a few percent.

\begin{figure}[htbp]
\centering
\includegraphics[width=\textwidth]{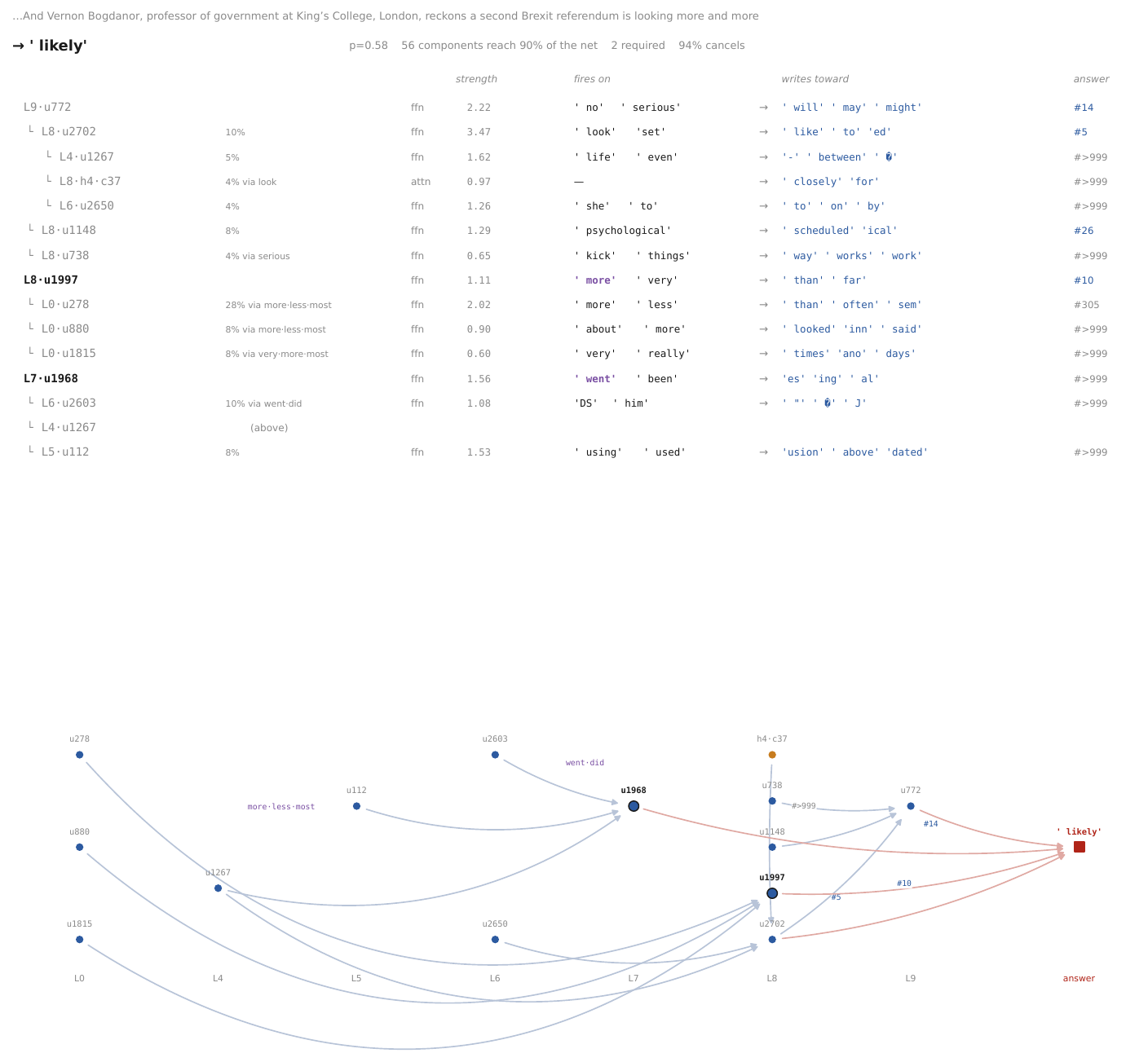}
\caption{One prediction's edges, read one hop. Bold marks the components whose removal changes the
prediction; indented rows are what such a component draws on, with the share of its incoming drive
they supply and, where writer and reader both rank a token in their first eight, that token. The
graph beneath draws the same components, depth running left to right, each joined to the answer with
the rank the predicted token holds in what it promotes.}
\label{fig:edgeexample}
\end{figure}

Figure~\ref{fig:edgeexample} is one prediction read this way, drawn from the same random sample as the
gallery and under the same filters. The context ends \emph{a second Brexit referendum is looking more
and more} and the model predicts \emph{ likely} at $p = 0.58$; $56$ components reach ninety percent
of the net, $94$ percent of the contribution mass cancels, and two components are required.

The four required and contributing components in it do not all relate to the answer the same way.
Unit $1997$ of layer $8$ fires on \emph{ more}, \emph{ very} and \emph{ most}, and the three units
feeding it hardest all sit in layer $0$ and write exactly those degree words: the three edges are
labeled with degree words, \emph{ more}, \emph{ less}, \emph{ most} and \emph{ very}, each a token
that writer and this reader both rank in their first eight. It promotes the predicted token tenth. Unit $772$ of layer $9$ answers to \emph{ no} and
\emph{ serious} and writes modal verbs, promoting \emph{ likely} fourteenth. A comparative detector
fed by degree words, and a modal detector, are a reasonable account of why a sentence ending
\emph{more and more} continues with \emph{ likely}.

Unit $1968$ of layer $7$ is the other case. It fires on
past-tense verbs and auxiliaries --- \emph{ went}, \emph{ been}, \emph{ did}, \emph{ could} --- and
writes word-continuation fragments --- \emph{es}, \emph{ing}, \emph{er}. Both sides are legible.
What it does not do is promote the answer: it ranks \emph{ likely} past three thousand, so the
predicted token is nowhere near what this unit specializes in. It is required anyway. Its
contribution to that particular logit is large enough that removing it changes the prediction, which
is a reminder that promoting a token and contributing to it are different things, and that the
components a prediction cannot lose need not be the ones pushing toward it.

\subsection{Aggregating influence along paths}

The overlap half of that problem has a standard treatment. \citet{abnar2020attentionflow} face it
directly when aggregating attention across layers --- paths through a transformer share edges, so
summing path weights double-counts --- and answer it by treating the attention graph as a flow
network with attention weights as capacities and taking a maximum flow. That construction does not
transfer here unaltered, for two reasons. Capacities must be non-negative, and
attention weights are; the contributions of Equation~\ref{eq:contrib} are signed, and replacing them
with magnitudes or with their positive parts is exactly the accounting this appendix argues against.
And a maximum flow scores a path by its bottleneck edge rather than by what it delivers, which
answers how much could be routed rather than how much arrives.

For a signed and locally linear quantity the overlap is in fact benign: contributions along
different paths add, so a path-sum over the graph is well defined and shared edges need no special
handling. What is \emph{not} linear is the intervention --- ablation removes a component and the
survivors then compute something else --- and the intervention rather than the attribution is what makes $\mathrm{flip}(k)$ non-monotone. A ranking by total path-summed influence rather than by direct
contribution is the natural next version of this measurement, and choosing a set by cut rather than
by prefix is the version after that. The prefix is reported because it is defined without further
assumptions and can be checked directly by ablation, and it is stated as a bound rather than as a
minimum.

\subsection{A pruned graph is still the reason}

Thinning the closure barely weakens it. A tenth of
it --- $56$ components, about one part in a thousand of the model --- still changes the prediction
$95$ percent of the time, and a fifth of it changes every prediction tested
(Table~\ref{tab:prune}). Necessity survives pruning that removes most of the graph.

\begin{table}[htbp]
\centering\small
\caption{Pruning the graph by lowering the per-node threshold $\theta$. ``Prediction changes''
ablates the pruned graph; ``preserved'' keeps only it and replaces everything else. The same
predictions are used at every $\theta$, so the rows are paired.}
\label{tab:prune}
\input{T_prune_base}
\end{table}

\subsection{Why sufficiency needs a search rather than a closure}
\label{sec:circuit-closure}

The sufficiency test keeps a set, replaces every other component in the model, and asks whether the
same token still comes out. Put to the closure it is a fair question. Put to the one-hop set it is
not, because replacing everything outside that set replaces the very drive that makes its members
fire, so the graph is what the test needs.

Two controls decide whether the recursion earns its place. A size-matched random set says whether
the closure beats an arbitrary set of the same size, and the top-$|C|$ components by direct
contribution say whether it beats simply taking more of the one-hop ranking.

\begin{table}[htbp]
\centering\small
\caption{Sufficiency against the one-hop set and against its closure, with both controls at matched
size. ``Keep-more'' is the validity check described below: how much preservation rises as the kept set
grows from nothing to $2{,}048$ components. Where that curve is flat the test
discriminates nothing and no entry should be read as a measurement; Section~\ref{sec:stock-instrument}
reports a model where it is flat.}
\label{tab:closure}
\input{T_closure}
\end{table}

This measurement needs more predictions than it looks like it needs. At $28$ of them
(Table~\ref{tab:closure}) the closure preserves the prediction on $54$ percent against $32$ for a
size-matched set taken further down the contribution ranking, which reads as twenty-two points in
the recursion's favor. At $150$ the sign reverses: $60$ percent against $65$ on the baseline, $63$
against $78$ on the sigmoid model, and $65$ against $87$ on GPT-2. A $95$ percent interval on a proportion near a half is $\pm 19$ points at $n=28$ and
$\pm 8$ at $n=150$, so twenty-two points at the smaller sample sits inside its own noise. What it buys on any model depends on that model's branching factor: where the factor
is small the closure is barely larger than the one-hop set and there is little for the recursion to
add, which is the case for the constructions of Section~\ref{sec:mono}.

A keep-only number means nothing unless keeping more helps, and that has to be checked rather than
assumed. Raising the kept set from nothing to $2{,}048$ components moves preservation by sixty-three
points, so the test has room to discriminate and the closure's advantage over the controls is a real
difference rather than noise around a ceiling. Every model measured behaves this way but one
(Section~\ref{sec:stock-instrument}). Tightening the recursion does not repair it. Raising the per-node threshold from $0.9$ to $0.99$
and on to the point where the cut no longer binds leaves preservation flat or slightly lower on
every model tested, while the set grows: $59$ to $56$ percent on the baseline as the closure goes
from $640$ to $786$ components. Ancestry is the wrong currency for this test: what preserves a prediction is components that write toward it, and the closure spends its budget on ancestry.

Keeping \emph{only} a pruned graph is not even monotone in how much is kept --- preservation peaks at
$\theta = 0.7$ and falls at $\theta = 0.9$, where strictly more of the graph is retained
(Table~\ref{tab:prune}) --- and Appendix~\ref{sec:pit-ablate} shows that to be a property of the
instrument rather than of this model.

Both numbers in this appendix are necessity statements, and neither shows that the components in it
can produce the answer on their own. The search of Section~\ref{sec:circuit-suf} reaches eight
components, two orders of magnitude below the closure, and a different set.

\subsection{What the replacement choice does to a keep-only number}
\label{sec:circuit-asym}

Removing a component is not free of choices. A transformer has no null value for a component's
output, so an ablation must write something in its place, and three are in common use: zero, the
component's mean output sampled over inputs, and a value resampled from a different input. Which of
these is chosen barely moves the measurement above; the three agree to within seven points. The lens identifies the \emph{necessary} parts of the transformer for that
prediction, and replacing them quickly breaks the prediction.

Run in reverse --- keep only the components the lens found and replace everything else --- the
measurement changes character (Table~\ref{tab:asymmetry}). All three choices drastically
impact the correctness: the prediction survives in $33$ percent of cases under zero replacement and
$3$ percent under resampling. The components the lens identifies are thus necessary but \emph{are not
sufficient} on their own to make the prediction. Section~\ref{sec:circuit-suf} reports what does reproduce a prediction; the question here is the
choice of zero against the mean and resample alternatives.
Zero replacement, the most flattering of the three, flatters for a reason that has nothing to do with the circuit. Section~\ref{sec:circuit-accounting} showed that most of what a model writes
at a position pushes \emph{away} from the predicted token. Zeroing every component outside the
circuit deletes that opposing mass along with everything else, so the target logit can come out
\emph{higher} than the intact model produces.

\begin{table}[htbp]
\centering\small
\caption{Sufficiency under the three standard replacement baselines: the share of predictions
preserved when only the circuit is kept and everything else is replaced. The choice of replacement
moves the answer by thirty points on the same circuit. The necessity measurement, on the same
components and the same three choices, moves by seven. Zero, the most favorable replacement, is the least informative, for the reason given in the text.}
\label{tab:asymmetry}
\input{T_asymmetry}
\end{table}

\secbarrier

%% file: T_prune_base.tex
\begin{tabular}{llrrrr}
\toprule
model & $\theta$ & components & \% of model & \% of closure & prediction changes \\
\midrule
baseline (GELU) & 0.1 & 56 & 0.12\% & 8\% & 95\% \\
 & 0.3 & 122 & 0.26\% & 17\% & 100\% \\
 & 0.5 & 249 & 0.54\% & 35\% & 100\% \\
 & 0.7 & 434 & 0.94\% & 60\% & 100\% \\
 & 0.9 & 720 & 1.56\% & 100\% & 100\% \\
\addlinespace
\bottomrule
\end{tabular}

%% file: T_closure.tex
\begin{tabular}{lr}
\toprule
 & baseline (GELU) \\
\midrule
one hop & 38 \\
\addlinespace
\multicolumn{2}{l}{\emph{set size}} \\
\quad closure & 147 \\
\quad one hop & 32\% \\
\addlinespace
\multicolumn{2}{l}{\emph{prediction preserved keeping only}} \\
\quad closure & \textbf{54\%} \\
\quad top-$|C|$ & 32\% \\
\quad random & 0\% \\
\quad keep-more & $+63$ \\
\bottomrule
\end{tabular}

%% file: T_asymmetry.tex
\begin{tabular}{lrrrrrr}
\toprule
& \multicolumn{3}{c}{necessity: prediction changed} & \multicolumn{3}{c}{sufficiency: prediction preserved} \\
\cmidrule(lr){2-4}\cmidrule(lr){5-7}
model & zero & mean & resample & zero & mean & resample \\
\midrule
baseline (GELU) & 81\% & 74\% & 75\% & 33\% & 5\% & 3\% \\
\bottomrule
\end{tabular}